\documentclass{article}
\usepackage{arxiv}
\usepackage[utf8]{inputenc} % allow utf-8 input
\usepackage[T1]{fontenc}    % use 8-bit T1 fonts
\usepackage{hyperref}       % hyperlinks
\usepackage{url}            % simple URL typesetting
\usepackage{booktabs}       % professional-quality tables
\usepackage{amsfonts}       % blackboard math symbols
\usepackage{nicefrac}       % compact symbols for 1/2, etc.
\usepackage{microtype}      % microtypography

\usepackage{comment}
\usepackage{dsfont}
\usepackage{amssymb}
\usepackage{amsmath}
\usepackage{amsthm}  % FOR ARXIV
\DeclareMathOperator*{\argmax}{\arg\!\max}

\usepackage{graphicx}
\usepackage{mathtools}
\usepackage[algoruled,boxed,linesnumbered]{algorithm2e}
\usepackage{parskip}
\usepackage{textcomp}
\usepackage{subcaption}
\usepackage{physics}
\usepackage[tableposition=top]{caption}
\usepackage[table,dvipsnames]{xcolor}
\usepackage{float}
\usepackage{breakcites}

\hypersetup{bookmarksnumbered}

\definecolor{MyLightGray}{gray}{0.9}
\definecolor{MyDarkGray}{gray}{0.6}

\SetCommentSty{mycommfont}

\newtheorem{definition}{Definition}[section]
\newtheorem{theorem}{Theorem}[section]
\newtheorem{corollary}{Corollary}[theorem]
\newtheorem{lemma}[theorem]{Lemma}
\newtheorem{remark}{Remark}

\date{}

\title{Lipschitzness Is All You Need To Tame Off-policy
Generative Adversarial Imitation Learning}

\author{ % FOR ARXIV
 Lionel~Blond\'e\thanks{Correspondence to Lionel Blond\'e: \texttt{lionel.blonde@unige.ch}.} \\
 University of Geneva, \\
 HES-SO, Switzerland \\
 \And
 Pablo~Strasser \\
 University of Geneva, \\
 HES-SO, Switzerland \\
 \And
 Alexandros~Kalousis \\
 University of Geneva, \\
 HES-SO, Switzerland \\
}

\begin{document}

\maketitle

\begin{abstract}
Despite the recent success of reinforcement learning in various domains,
these approaches remain, for the most part, deterringly sensitive to hyper-parameters
and are often riddled with essential engineering feats allowing their success.
We consider the case of off-policy generative adversarial imitation learning,
and perform an in-depth review, qualitative and quantitative, of the method.
We show that forcing the learned reward function to be local Lipschitz-continuous
is a \textit{sine qua non} condition for the method to perform well.
We then study the effects of this necessary condition and provide several theoretical results
involving the local Lipschitzness of the state-value function.
We complement these guarantees with empirical evidence attesting to the strong
positive effect that the consistent satisfaction of the Lipschitzness constraint on the reward has
on imitation performance.
Finally, we tackle a generic pessimistic reward preconditioning add-on
spawning a large class of reward shaping methods, which
makes the base method it is plugged into provably more robust, as shown in several
additional theoretical guarantees.
We then discuss these through a fine-grained lens and share our insights.
Crucially, the guarantees derived and
reported in this work are valid for \emph{any} reward
satisfying the Lipschitzness condition, nothing is specific to imitation.
As such, these may be of independent interest.
% \keywords{ % FOR ML
%     Imitation Learning
%     \and
%     Reinforcement Learning
%     \and
%     Lipschitz-continuity
%     \and
%     Generative Adversarial Networks
%     \and
%     Deep Learning
% }
\end{abstract}

\section{Introduction}

Imitation learning (IL) \cite{Bagnell2015-ni} sets out to design artificial agents able to adopt
a behavior demonstrated via a set of expert-generated trajectories.
Also referred to as \textit{``teaching by showing''} \cite{Schaal1997-vi},
IL can replace tedious tasks such as manual hard-coded agent programming, or
hand-crafted reward design \textit{``reward shaping''} \cite{Ng1999-lv}
for the agent to be trained via reinforcement learning (RL) \cite{Sutton1998-ow}.
Besides, in contrast with the latter, imitation learning does not necessarily involve
agent-environment interactions.
This feature is particularly appealing in real-world domains such as robotics
\cite{Atkeson1997-db,Schaal1997-vi,Ratliff2007-fc,Billard2008-jb},
where the artificial agent is physically implemented with expensive hardware,
and the environment contains enough external entities
(\textit{e.g.} humans, other artificial agents, other costly devices)
to raise safety concerns \cite{Ha2020-vb,Kahn2016-cq,Ray2019-xf,Held2017-wo}.
When controls are provided in the demonstrations
(or recovered via inverse dynamics from the available kinematics \cite{Hanna2017-iz}),
we can treat said controls as regression targets,
and learn a mimicking policy with a simple, supervised approach.
This interaction-free approach (simulated or physical, real-world interactions),
called \emph{behavioral cloning} (BC),
has enabled the success of various endeavors
in robotic manipulation and locomotion \cite{Ratliff2007-fc,Wang2017-eq},
in autonomous driving ---
with the first self-driving vehicle \cite{Pomerleau1989-nh,Pomerleau1990-lm} thirty years ago
and more recently with \cite{Gu2020-fo} using Waymo's open dataset \cite{Sun2019-qc} ---
and also in grand challenges like \textsc{AlphaGo} \cite{Silver2016-my}
and \textsc{AlphaStar} \cite{Vinyals2019-vx}.
Due to its conceptual simplicity, we expect BC to still be a part of the pipeline
for the most ambitious enterprises going forward,
especially as open datasets get slowly released.

Despite its practical advantages, BC is extremely data-hungry \textit{w.r.t.} the amount
of expert demonstrations it needs to yield robust, high-fidelity policies.
Besides, unless corrective behavior is present in the dataset
(\textit{e.g.} in autonomous driving, how to drive back onto the road),
the policy learned via BC will not be able to internalize this behavior.
Once in a situation from which it can not recover,
there will be a permanent \emph{covariate shift} between its current observations and
the demonstrated ones.
The controls learned in a supervised manner on the expert dataset are therefore useless,
due to the distributional shift.
As a result, the agent's errors will compound, a phenomenon coined by \cite{Ross2010-eb}
as \emph{compounding errors}.
In \textsc{Section}~\ref{compoundvars},
we stress how the latter echoes the \emph{compounding variations} phenomenon,
exhibited as part of the theoretical contributions of this work.
To address the shortcomings of BC, \cite{Abbeel2004-rb} proposes to harness
the innate credit assignment \cite{Sutton1998-ow} capabilities of RL,
by first trying to learn the cost function underlying the demonstrated behavior
(inverse RL \cite{Ng2000-qd}),
before using this cost to optimize a policy via RL.
The succession of inverse RL and RL is called apprenticeship learning (AL) \cite{Abbeel2004-rb},
and can, by design,
yield policies that can recover from out-of-distribution situations thanks to RL's
built-in temporal abstraction mechanisms.
Cost learning however is incredibly tedious, and successful approaches end up requiring
coarse relaxations to avoid being deterringly computationally-expensive
\cite{Abbeel2004-rb,Syed2008-su,Syed2008-zo,Ho2016-xn}.
Ultimately, as noted by \cite{Ziebart2008-fe},
setting out to recovering the cost signal under which
the expert demonstrations are optimal (base assumption of inverse RL) is an ill-posed objective
--- echoing the reward shaping considerations from \cite{Ng1999-lv}.
In line with this statement, generative adversarial imitation learning (GAIL) \cite{Ho2016-bv}
departs from the typical AL pipeline, and replaces learning
the optimal cost (``optimal'' in the inverse RL sense)
by learning a \emph{surrogate} cost function.
GAIL does so by leveraging generative adversarial networks \cite{Goodfellow2014-yk},
as the name hints.
The method is described in greater detail in \textsc{Section}~\ref{gail}.
Due to the RL step it involves (like any AL method),
GAIL suffers from poor sample-efficiency \textit{w.r.t.} the amount
of interactions it needs to perform with the environment.
This caveat has since been addressed, notably by transposition to the off-policy setting,
concurrently in SAM \cite{Blonde2019-vc} and DAC \cite{Kostrikov2019-jo}
(\textit{cf.} \textsc{Section}~\ref{bridge}).
Both adversarial IL methods
leverage actor-critic architectures,
consequently suffering from a greater exposure to instabilities.
These weaknesses are mitigated with various complementary techniques,
and cautious hyper-parameter tuning.

In this work, we set out to first conduct a thorough
theoretical and empirical
investigation into off-policy
generative adversarial imitation learning, to pinpoint which are the techniques
that are instrumental in performing well,
and shed light over which are ones that can be discarded or disregarded
without decrease in performance.
Ultimately, we would like to exhibit the techniques that are \emph{sufficient} for
the method to achieve peak performance.
Virtually every algorithmic design choice made in this work is supported
by an ablation study reported in the \textsc{Appendix}.
We start by describing the base off-policy adversarial imitation learning method
at the core of this work in \textsc{Section}~\ref{bridge}.
We then undertake diagnoses of the various issues that arise from the combination of
bilevel optimization problems at the core of the investigated model
in \textsc{Section}~\ref{lipalluneed}.
A key contribution of our work consists in showing that enforcing a Lipschitzness
constraint on the learned surrogate reward is a \emph{necessary} condition
for the method to even learn anything
--- in our consumer-grade, computationally affordable hardware setting.
We study it closely, providing empirical evidence of the importance of this constraint
through detailed ablation results in \textsc{Section}~\ref{empres1}.
We follow up on this empirical evidence with theoretical results in \textsc{Section}~\ref{theory},
characterizing the
Lipschitzness of the state-action value function under said reward Lipschitzness condition,
and discuss the obtained variation bounds subsequently.
Crucially, we show that without variation bounds on the reward,
a phenomenon we call \emph{compounding variations}
can cause the variations of the state-action value to explode.
As such, the theoretical results reported in \textsc{Section}~\ref{theory}
--- and discussed in \textsc{Section}~\ref{discussion} ---
corroborate the empirical evidence exhibited in \textsc{Section}~\ref{empres1}.
\textbf{\emph{Note, the theoretical results
reported in this work are valid for any reward
satisfying the condition, they readily transfer to the general RL setting and are not specific to imitation.}}
The theoretically-grounded Lipschitzness condition,
implemented as a gradient penalty,
is in practice a \emph{local} Lipschitzness condition.
We therefore investigate \emph{where}
(\textit{i.e.} on which samples, on which input distribution)
the local Lipschitzness regularization should be enforced.
We propose a new interpretation of the regularization scheme
through an RL perspective,
make an intuitively grounded claim on where to enforce the constraint to get the best results,
and corroborate our claim empirically (\textit{cf.} \textsc{Section}~\ref{gradpenrl}).
Crucially, we show that
the consistent satisfaction of the Lipschitzness constraint on the reward
is a strong predictor of how well the mimicking agent performs empirically
(\textit{cf.} \textsc{Section}~\ref{understand}).
Finally, we introduce a generic pessimistic reward preconditioner which
makes the base method it is plugged into provably more robust,
as attested by its companion guarantees
(\textit{cf.} \textsc{Section}~\ref{purpleandres}).
\textbf{\emph{Again, these guarantees
are not not specific to imitation and can be of independent interest for the RL community.}}
Among the reported insights, we give an illustrative example of how the simple technique can further increase
the robustness of the method it is plugged into.
We release the code as an open-source\footnote{Code made available at the URL:
\texttt{https://github.com/lionelblonde/liayn-pytorch}.} project.

\section{Related work}
\label{related}

Off-policy generative adversarial imitation learning, which is the object of this work,
involves learning a parametric surrogate reward function, from expert demonstrations.
By design \cite{Ho2016-bv,Blonde2019-vc,Kostrikov2019-jo},
this signal is learned at the same time as the policy, and is therefore subject to
non-stationarities (\textit{cf.} \textsc{Section}~\ref{nonstat}).
This reward regime is reminiscent of the \emph{reward corruption} phenomenon
\cite{Everitt2017-ql,Romoff2018-if}, which posits that the real-world rewards are imperfect
(\textit{e.g.} uncontrolled task specification change, sensor defects, reward hacking)
and must therefore be treated as such, \textit{i.e.} non-stationary at the very least.
Despite being learned and therefore liable to non-stationary behavior,
our reward is internal
--- as opposed to outside the agent's and practitioner's scope ---
and is therefore fully observable, as well as controllable via the practitioner-specified
algorithmic design.
The reward corruption can consequently be acted upon, and more easily mitigated than if
it originated from a \emph{black box} reward
originating from the unknown environment.

The demonstrations on the other hand are available from the very beginning,
and do not change as the policy learns.
In that respect, our approach differs from \textit{observational learning} \cite{Borsa2017-ab},
where the policy learns to imitate another by \emph{observing} it itself learn in the
environment --- and therefore does not strictly qualify as an expert at the task.
Observational learning draws clear parallels
with the teacher-student scheme in policy distillation \cite{Rusu2015-sr}.
While our reward is changing since the policy changes and due to the inherent learning
dynamics of function approximators, in observational learning, the reward would be changing
also due to the expert still learning, causing a distributional drift.

Multi-armed bandits \cite{Robbins1952-sp}
have received a lot of attention in recent years
to formalize and model problems of sequential decision making under uncertainty.
In the context of this work, the most appropriate variants of bandits
are \emph{stateful} contextual multi-armed bandits.
As the name hints, such models formalize decision making specific to given situations
(\textit{i.e.} contexts, states), in which the situations are \textit{i.i.d.}-sampled.
We consider the case of reinforcement learning, where the situations are
entangled, along with the decisions themselves, in a Markov decision process
(\textit{cf.} \textsc{Section}~\ref{prelim}).
In particular, non-stationary reward channels in Markov decision processes
have been studied extensively
(\textit{cf.} \textsc{Section}~\ref{nonstat}).
Among these, adversarial bandits \cite{Auer1995-mm} can be seen as the archetype or worst-case
reward corruption scenario, in which an adversary
--- possibly driven by malevolent intents ---
decides on the reward given to the agent.
In these models, the common way to deal with non-stationary reward processes is to
assume the reward variations in time are upper-bounded, either per-decision or
over longer time periods.
We give a comprehensive account of sequential decision making under uncertainty
in non-stationary Markov decision processes
in \textsc{Appendix}~\ref{nsmdps}.
By contrast, our theoretical guarantees are built on the premise that
the reward function's variations are bounded \emph{over the input space} by
assuming that the reward function is locally Lipschitz-continuous over it.
We make the same assumption on the dynamics of the multi-stage decision process,
as well as on the control policy.
While our theoretical results ultimately characterize the value function's robustness
in terms of Lipschitz-continuity,
\cite{Fonteneau2010-hu,Fonteneau2013-jw}
start from the same assumptions,
propose an estimator of the expected return, and derive bounds on its bias and variance.
Derived in the offline RL setting,
their bounds increase as the \textit{``dispersion''} of the offline dataset increases.
As such, our findings and dicussions carried out in \textsc{Section}~\ref{discussion} echo their work.

Several works have recently attempted to address the overfitting problem GAIL suffers from.
This is due to the discriminator being able to trivially distinguish agent-generated samples
from expert-generated ones, which occurs when the learning dynamics of the adversarial
game are not properly balanced.
As such, the gist of said techniques is to either weaken the discriminator directly or
make its classification task harder,
which unsurprisingly exactly coincides with the typical techniques used to cope with
overfitting in (binary) classification.
These techniques are, in no particular order:
reducing the discriminator's capacity
--- by plugging the classifier on top of an
independent perception stack
(\textit{e.g.} random features, state-action value convolutional layers) \cite{Reed2018-ga},
smoothing the positive labels with uniform random noise \cite{Blonde2019-vc},
adopting a positive-unlabeled classification objective
(instead of the traditional positive-negative one) \cite{Xu2019-uo},
using a gradient penalty (originally from \cite{Gulrajani2017-mr}) regularizer
\cite{Blonde2019-vc,Kostrikov2019-jo},
leveraging an adaptive information bottleneck in the discriminator network
\cite{Peng2018-mo},
enriching the expert dataset via task-specific data augmentation \cite{Zolna2019-wj}.
In this work,
we do not propose a new regularization technique.
Instead, we perform an in-depth analysis of the simplest
techniques
--- in terms of conceptual simplicity, implementation time,
number of parameters, and computational cost \cite{Hernandez2020-dv} ---
and ultimately find that the gradient penalty regularizer
achieves the best trade-off.

A large-scale empirical study of adversarial imitation learning \cite{Orsini2021-fv}, released very recently,
considers a wide range of hyper-parameter settings, reporting results for more than
500k trained agents.
The authors conclude that their study adds nuances to ours (this work).
In particular, they argue that
while the regularization techniques that urge the reward to be Lipschitz-continuous indeed do
improve the performance (hence corroborating what we show in the first investigation of our work;
\textit{cf.} \textsc{Section}~\ref{empres1}),
more traditional regularizers (\textit{e.g.} weight decay, dropout) can often perform similarly.
In this work, we align the notion of smoothness with the Lipschitz-continuity of a function approximator,
and are therefore focusing, from \textsc{Section}~\ref{empres1} onward, on gradient penalization because it
\emph{explicitly} enforces the reward to be smooth.
More importantly, reward Lipschitzness is among the premises of our theoretical guarantees.
In the results reported in \cite{Orsini2021-fv}, the discriminator regularization schemes that can perform on par
with schemes enforcing Lipschitz-continuity explicitly
(gradient penalization \cite{Gulrajani2017-mr},
and spectral normalization \cite{Miyato2018-wc}), which are always the top performers, are:
dropout \cite{Srivastava2014-oa},
weight decay \cite{Loshchilov2017-hn},
and mixup \cite{Zhang2017-as} (performing data augmentation).
Regularization schemes such as dropout, weight decay, and data augmentation
are less often seen through the lens of smoothness regularization than through the lens of generalization,
despite generalization being among the beneficial effects of smoothness \cite{Rosca2020-yg}.
Used in the last layer, weight decay \cite{Loshchilov2017-hn} punishes spikes in elements of the weight matrix
by limiting its norm, hence not allowing the output of the network to change too much.
Dropout \cite{Srivastava2014-oa} applies masks over hidden activations, making the network return similar
outputs when inputs only differ slighly.
When using data augmentation (\textit{e.g.} in mixup \cite{Zhang2017-as}),
the network is forced to be close-to-invariant to purposely crafted variations of the input.
These regularizers do not enforce Lipschitzness over the input space as explicitly
as gradient penalties and spectral normalization do;
nevertheless, they do encourage Lipschitzness implicitly, making the predictor more robust as a result.
Specifically, as noted in \cite{Gouk2021-dz},
when a neural function approximator is trained with dropout, the Lipschitz constant of each layer is
multiplied by $1-r$,
where $r$ is the dropout rate.
It is also noted in \cite{Cisse2017-tl} that using weight decay regularization at the last layer
controls the Lipschitz constant of the network.
All in all, the methods reported by \cite{Orsini2021-fv} as performing the best are the ones
enforcing Lipschitz-continuity over the input space explicitly, and these can be matched by
regularization schemes that encourage Lipschitzness over the input space implicitly.
As such, these results are complementary to the ones we report in our first investigation
in \textsc{Section}~\ref{empres1}, where we found that direct, explicit gradient penalization
exceeds the performance of other evaluated regularizers.
As we report, not constraining the Lipschitzness of the discriminator
yields the worst results among the evaluated alternatives.
Keeping the Lipschitz constant of the discriminator in check seems essential.
Perhaps more importantly, the empirical investigation we conduct in \textsc{Section}~\ref{empres1},
and that is complemented by \cite{Orsini2021-fv},
motivates the derivation of our novel theoretical guarantees.
Through these, we provide insights as to \emph{why} keeping the Lipschitz constant of the reward in check seems
to play such an important role in the stability of the value in off-policy adversarial IL.
The considerable computational budget spent in \cite{Orsini2021-fv} attests to how challenging
the tackled problem is.

In \cite{Hafner2011-rv}, Hafner and Riedmiller
advocate for the use of a \emph{smooth} reward signal in RL.
\cite{Lange2012-cc} presents it as one key method to make learning values in offline RL less tedious.
Sharp changes in reward value are hard
to represent and internalize by the action-value neural function approximator.
Using a smooth reward surrogate
derived from the original \textit{``jumpy’’} reward signal such that the trends are preserved
but the crispness is attenuated proved instrumental empirically.
Our observation about reward Lipschitz-continuity being a crucial component of our
off-policy imitation learning pipeline is in line with the suggestion of \cite{Hafner2011-rv}.
On top of providing empirical evidence of its benefits,
we also provide a number of theoretical results characterizing
what the reward smoothness does on the value function smoothness.

Finally, we point out that local Lipschitz-continuity conditions are
also found in the adversarial robustness literature.
Notably, \cite{Finlay2018-is} encourages Lipschitzness via gradient regularization,
as is done in our work.
Similarly, \cite{Hardt2015-qf}
derives bounds under a Lipschitz-continuity assumption on the loss.

\section{Background}

\paragraph{Setting.}
In this work, we address the problem of an agent whose goal is,
in the absence of extrinsic reinforcement signal \cite{Singh2009-cm},
to \emph{imitate} the behavior demonstrated by an expert \cite{Bagnell2015-ni},
expressed to the agent via a pool of trajectories.
The agent is never told how well she performs or what the optimal actions are,
and is not allowed to query the expert for feedback.

\paragraph{Preliminaries.}
\label{prelim}
The intrinsic behavior of the decision maker is represented by the \emph{policy} $\pi_\theta$,
modeled by a neural network with parameter $\theta$,
mapping states to probability distributions over actions.
Formally, the conditional probability density over actions that the agent
concentrates at action $a_t$ in state $s_t$ is denoted by $\pi_\theta(a_t | s_t)$,
for all discrete timestep $t \geq 0$.
We model the environment the agent interacts with as an infinite-horizon, memoryless, and stationary
\emph{Markov Decision Process} (MDP) \cite{Puterman1994-pf}
formalized as the tuple
$\mathbb{M} \vcentcolon= (\mathcal{S}, \mathcal{A}, p, \rho_0, u, \gamma)$.
$\mathcal{S} \subseteq \mathbb{R}^n$ and $\mathcal{A} \subseteq \mathbb{R}^m$
are respectively the state space and action space.
$p$ and $\rho_0$ define the \emph{dynamics} of the world,
where $p(s_{t+1} | s_t, a_t)$ denotes the stationary conditional probability density
concentrated at the next state $s_{t+1}$
when stochastically transitioning from state $s_t$ upon executing action $a_t$,
and $\rho_0$ denotes the initial state probability density.
$u$ denotes a stationary \emph{reward process}
that assigns, to any state-actions pairs,
a real-valued reward $r_t$ distributed as $r_t \sim u(\cdot | s_t, a_t)$.
Finally, $\gamma \in [0, 1)$ is the discount factor.
We make the MDP \emph{episodic}
by positing the existence of an absorbing state in every trace of interaction and
enforcing $\gamma = 0$ to formally trigger episode termination
once the absorbing state is reached.
Since our agent does not receive rewards from the environment,
she is in effect interacting with an MDP lacking a reward process $r$.
Our method however encompasses learning a surrogate reward parameterized
by a deterministic function approximator
such as a neural network with parameter $\varphi$, denoted by $r_\varphi$, and
whose learning procedure will be reported subsequently.
Consequently, our agent effectively interacts with the augmentation of the previous MDP defined as
$\mathbb{M}^* \vcentcolon= (\mathcal{S}, \mathcal{A}, p, \rho_0, r_\varphi, \gamma)$.
A \emph{trajectory} $\tau_\theta$ is a trace of $\pi_\theta$ in $\mathbb{M}^*$, succession of
consecutive \emph{transitions} $(s_t, a_t, r_t, s_{t+1})$,
where $r_t \coloneqq r_\varphi(s_t, a_t)$.
A \emph{demonstration} is the set of state-actions pairs $(s_t, a_t)$
extracted from a trajectory collected by the expert policy $\pi_e$
in $\mathbb{M}$.
The \emph{demonstration dataset} $\mathcal{D}$ is a set of demonstrations.

\paragraph{Objective.}
Building on the reward hypothesis at the core of reinforcement learning (any task
can be defined as the maximization of a reward), to act optimally, our agents
must be able to deal with delayed signals and maximize the long-term cumulative reward.
To address credit assignment, we use the concept of \emph{return},
the discounted sum of rewards from timestep $t$ onwards, defined as
$R_t^\gamma
\coloneqq \sum_{k=0}^{+\infty} \gamma^k r_{t+k}
\coloneqq \sum_{k=0}^{+\infty} \gamma^k r_\varphi(s_{t+k}, a_{t+k})$
in the infinite-horizon regime.
By taking the expectation of the return
with respect to all the future states and actions in $\mathbb{M}^*$,
after selecting $a_t$ in $s_t$ and following $\pi_\theta$ thereafter,
we obtain the state-action value ($Q$-value) of the policy $\pi_\theta$ at $(s_t, a_t)$:
$Q^{\pi_\theta}(s_t, a_t) \coloneqq
\mathbb{E}_{
s_{t+1} \sim p(\cdot | s_t, a_t),
a_{t+1} \sim \pi_\theta(\cdot | s_{t+1}), \ldots}
[R_t^\gamma]$
(\textit{abbrv.} $\mathbb{E}_{\pi_\theta}^{>t}[R_t^\gamma]$).
At state $s_t$, a policy $\pi_\theta$ that picks $a_t$ verifying:
\[a_t = \argmax_{a \in \mathcal{A}} Q^{\pi_\theta} (s_t, a)\]
therefore acts optimally looking onwards from $s_t$.
Ultimately, an agent acting optimally at all times maximizes
$V^{\pi_\theta} (s_0)
\coloneqq \mathbb{E}_{a_0 \sim \pi_\theta(\cdot | s_0)}[Q^{\pi_\theta}(s_0, a_0)]$
for any given start state $s_0 \sim \rho_0$.
\textit{In fine}, we can now define the \emph{utility function}
(also called \emph{performance objective} \cite{Silver2014-dk}) to which
our agent's policy $\pi_\theta$ must be solution of:
$\pi_\theta = \argmax_{\pi \in \Pi} U_0(\pi)$ where $U_t(\pi) \coloneqq V^\pi(s_t)$
and $\Pi$ is the search space of parametric function approximators,
\textit{i.e.} deep neural networks.

\paragraph{Generative Adversarial Imitation Learning.}
\label{gail}
GAIL \cite{Ho2016-bv}
trains a binary classifier $D_\varphi$, called \emph{discriminator},
where samples from $\pi_e$ are positive-labeled, and those from $\pi_\theta$ are negative-labeled.
It borrows its name from \textit{Generative Adversarial Networks} \cite{Goodfellow2014-yk}:
the policy $\pi_\theta$ plays the role of generator and is optimized to fool
the discriminator $D_\varphi$ into classifying its generated samples (negatives),
as positives.
As such, the prediction value indicates to what extent $D_\varphi$ believes $\pi_\theta$'s
generations are coming from the expert, and therefore constitutes a good measure of
mimicking success.
GAIL does not try to recover the reward function that underlies the expert's behavior.
Rather, it learns a similarity measure between $\pi_e$ and $\pi_\theta$,
and uses it as a \emph{surrogate} reward function.
We say that $\pi_\theta$ and $D_\varphi$ are \textit{``trained adversarially''} to denote
the two-player game they are intricately tied in:
$D_\varphi$ is trained to assert with confidence
whether a sample has been generated by $\pi_\theta$,
while $\pi_\theta$ receives increasingly greater rewards as $D_\varphi$'s confidence in
said assertion lowers.
\textit{In fine}, the surrogate reward measures the confusion of $D_\varphi$.
In this work, the neural network function approximator modeling $D_\varphi$
uses a sigmoid as output layer activation, \textit{i.e.} $D_\varphi \in [0, 1]$.
The exact zero case is bypassed numerically for $\log \circ D_\varphi$ to always exist,
by adding an infinitesimal value $\epsilon > 0$ to $D_\varphi$ inside the logarithm.
The same numerical stability trick is used for $\log \circ (1 - D_\varphi)$ to avoid the exact one case
(\textit{cf.} reward formulations in \textsc{Section}~\ref{bridge}).

\section{Comprehensive refresher on the sample-efficient adversarial mimic}
\label{bridge}

Building on TRPO \cite{Schulman2015-jt}, GAIL \cite{Ho2016-bv} inherits its
policy evaluation subroutine,
consisting in learning a parametric estimate of the state-value function
$V_\omega \approx V^{\pi_\theta}$ via Monte-Carlo estimation
over samples collected by $\pi_\theta$.
While it uses function approximation
to estimate $V^{\pi_\theta}$, hoping it generalizes better
than a straight-forward non-parametric Monte-Carlo estimate (discounted sum),
we will reserve the term \emph{actor-critic} for architectures in which the
state-value $V^{\pi_\theta}(\cdot)$ or Q-value $Q^{\pi_\theta}(\cdot, \cdot)$ is learned
via Temporal-Difference (TD) \cite{Sutton1988-to}.
This terminology choice is adopted from \cite{Sutton1998-ow}
(\textit{cf.} \textsc{Chapter} 13.5).
A \emph{critic} is used for bootstrapping, as in the TD update rule
(whatever the bootstrapping degree is).
As such, TRPO is not an actor-critic, while algorithms learning their value via TD,
such as DDPG \cite{Silver2014-dk,Lillicrap2016-xa}, are actor-critic architectures.
Albeit hindered from various weaknesses (\textit{cf.} \textsc{Section}~\ref{triad}),
and forgetting for a moment
that it is combined with function approximation \cite{Sutton1999-ii,Silver2014-dk},
the TD update is able to propagate information quicker as the backups are shorter and
therefore do not need to reach episode termination to learn, in contrast with
Monte-Carlo estimation.
That is without even involving fictitious, memory,
or experience replay mechanisms \cite{Lin1992-pp}.
By design, TD learning is less data-hungry
(\textit{w.r.t.} interactions in the environment),
and involving replay mechanisms \cite{Lin1992-pp,Lillicrap2016-xa,Wang2016-mp}
significantly adds on to its inherent sample-efficiency.
Based on this line of reasoning, SAM \cite{Blonde2019-vc} and DAC \cite{Kostrikov2019-jo}
addressed the deterring sample-complexity of GAIL by,
among other improvements (\textit{cf.} \cite{Blonde2019-vc,Kostrikov2019-jo}),
using an actor-critic architecture to replace TRPO for policy evaluation and improvement.
SAM \cite{Blonde2019-vc} uses DDPG \cite{Lillicrap2016-xa},
whereas DAC \cite{Kostrikov2019-jo} uses TD3 \cite{Fujimoto2018-pe}.
Both were released concurrently, and both report significant improvements in sample-efficiency
(up to two orders of magnitude).  % depending on the environment
Standing as the stripped-down model that brought sample-efficiency to GAIL,
we take SAM as base.
Albeit described momentarily in the body of this work,
we urge the reader eager to understand every single aspect of the laid out algorithm to also refer to
the section in which we describe the experimental setting, \textit{cf.}~\ref{empres1}.

We now lay out the constituents of SAM \cite{Blonde2019-vc}, and how their learning
procedures are orchestrated.
The agent's behavior is dictated by a \emph{deterministic} policy $\mu_\theta$,
the critic $Q_\omega$ assigns $Q$-values to actions picked by the agent,
and the reward $r_\varphi$ assesses to what degree the agent behaves like the expert.
As usual, $\theta$, $\omega$, and $\varphi$ denote the respective parameters of
these neural function approximatiors.
To explore when carrying out rollouts in the environment, $\mu_\theta$ is perturbed
both in parameter space by adaptive noise injection in $\theta$
\cite{Plappert2018-rl, Fortunato2017-af},
and action space by adding the temporally-correlated response
of an Ornstein-Uhlenbeck noise process \cite{Uhlenbeck1930-an,Lillicrap2016-xa}
to the action returned by $\mu_\theta$.
Formally, in state $s_t$, action $a_t$ is sampled from
$\pi_\theta(\cdot | s_t) \coloneqq \mu_{\theta + \epsilon}(s_t) + \eta_t$,
where $\epsilon \sim \mathcal{N}(0, \sigma_a^2)$ ($\sigma_a$ adapts conservatively such that
$|\mu_{\theta + \epsilon}(s_t) - \mu_\theta(s_t)|$ remains below a certain threshold),
and where $\eta_t$ is the response of the Ornstein-Uhlenbeck process \cite{Uhlenbeck1930-an}
$\mathfrak{N}_{OU}$ at timestep $t$ in the episode,
such that $\eta_t \coloneqq \mathfrak{N}_{OU}(t, \sigma_b)$.
Note, $\mathfrak{N}_{OU}$ is reset upon episode termination.
As a first minor contribution, we carried out an ablation study
on exploration strategies, and report the results in
\textsc{Appendix}~\ref{ablationexplo}.
While the utility of temporally-correlated noise is somewhat limited to dynamical systems,
both parameter noise and input noise injections
have proved beneficial in generative modeling with GANs
(\cite{Zhao2017-bs} and \cite{Arjovsky2017-ne}, respectively).
As in GAIL \cite{Ho2016-bv} (described earlier in \textsc{Section}~\ref{gail}),
the discriminator $D_\varphi$ is trained via an adversarial
training procedure \cite{Goodfellow2014-yk} against the policy $\pi_\theta$.
The surrogate reward $r_\varphi$ used to augment MDP $\mathbb{M}$ into $\mathbb{M}^*$
is derived from $D_\varphi$ to reflect the incentive that the agent needs to complete
the task at hand.
In the tasks we consider in this work
(simulated robotics environments \cite{Brockman2016-un},
based on the \textsc{MuJoCo} \cite{Todorov2012-gc} physics engine,
and described in \textsc{Table}~\ref{envtable})
an episode terminates either
\textit{a)} when the agent \textit{fails} to complete
the task according to an task-specific criterion hard-coded in the environment,
or \textit{b)} when the agent has performed a number of steps in the
environments that exceeds a predefined hard-coded \emph{timeout},
which we left to its default value
--- with the exception of \texttt{HalfCheetah}, in which \textit{a)} does not apply.
Due to \textit{a)}, the agent can decide to truncate its return by triggering its own
failure, and decide to ``cut its losses'' when it is penalized too heavily for not succeeding
according to the task criterion.
Always-negative rewards (\textit{e.g.} per-step ``$-1$'' reward
to urge to agent to complete the task quickly \cite{Kaelbling1993-dv})
can therefore make the agent give up and trigger termination the earliest possible,
as this would maximize its return.
On the other hand, always-positive rewards can make the agent content with its
sub-optimal actions which would prevent it from pursuing higher rewards,
as long as it remains alive.
This phenomenon has been dubbed \textit{survival bias} in \cite{Kostrikov2019-jo}.
Notably, this discussion highlights the tedious challenge
that reward shaping \cite{Ng1999-lv} usually represents to practitioners
when designing a new task.
Stemming from their generator loss counterparts in the GAN literature,
the \emph{minimax (saturating)} reward variant is $r_\varphi \coloneqq -\log(1-D_\varphi)$,
and the \emph{non-saturating} reward variant is $\log(D_\varphi)$.
The minimax reward is always positive, the non-saturating reward is always negative,
and the sum of the two can take positive and negative values.
We found empirically that using the minimax reward, despite being always positive,
yielded by far the best results compared to the sum of the two variants.
The performance gap is reduced in the \texttt{HalfCheetah} task which was expected
since it is the only task in which the agent can not trigger an early termination.
We report these comparative results in \textsc{Appendix}~\ref{ablationreward}.
Crucially, these results show that the base method considered in this work
can already successfully mitigate survival bias, without requiring
additional reward shaping.
In summary, we use the formulation $r_\varphi \coloneqq - \log(1 - D_\varphi)$,
unless stated otherwise explicitly.

We also adopt the mechanism introduced in \cite{Kostrikov2019-jo}
that wraps the absorbing transitions
(agent-generated \emph{and} expert-generated)
to enable the discriminator to distinguish between
terminations caused by failure
and terminations triggered by the artificially hard-coded timeout.
The method enables the discriminator to penalize the agent for terminating by failure
when the expert would, with the same action and in the same state, terminate by
reaching the episode timeout without failing.
In such a scenario,
without wrapping the absorbing transitions, the agent perfectly imitates the expert in the
eyes of the discriminator, which is not the case.
We use the wrapping mechanism in every experiment.
Nonetheless, we omit it from the equations and algorithms for legibility.
Giving the agent the ability to differentiate between terminations that are due to time limits
and those caused by the environment had proved crucial
for the decision maker to continue beyond the time limit.
The significant role played by the explicit inclusion of the notion of time in RL
has been established by Harada in \cite{Harada1997-id}, yet without much follow-up,
until being revived in \cite{Pardo2018-xi} where the authors demonstrate that
a careful inclusion of the notion of time in RL can meaningfully impact performance.

\begin{figure}
\centering
\begin{subfigure}[t]{0.24\textwidth}
\centering
\includegraphics[width=\linewidth]{%
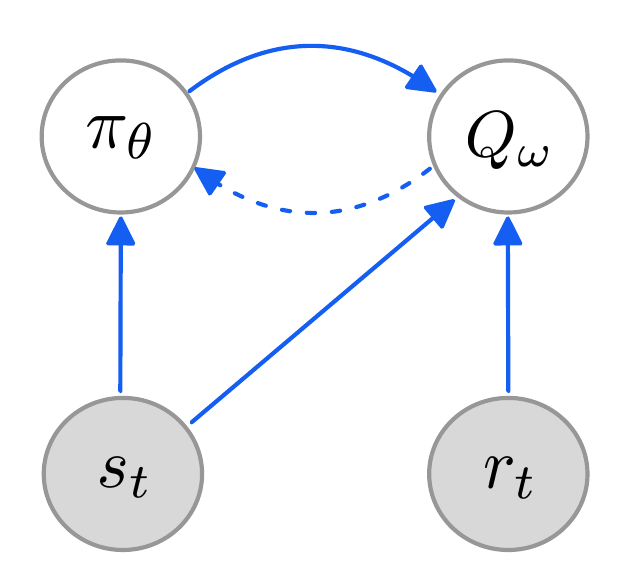}
\caption{Actor-Critic \cite{Sutton1999-ii}}
\label{diag:ac}
\end{subfigure}
\begin{subfigure}[t]{0.24\textwidth}
\centering
\includegraphics[width=\linewidth]{%
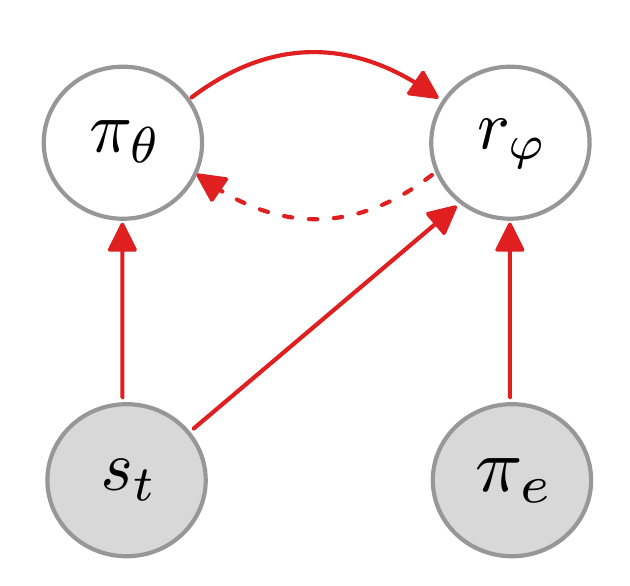}
\caption{GAIL \cite{Ho2016-bv}}
\label{diag:gail}
\end{subfigure}
\begin{subfigure}[t]{0.33\textwidth}
\centering
\includegraphics[width=\linewidth]{%
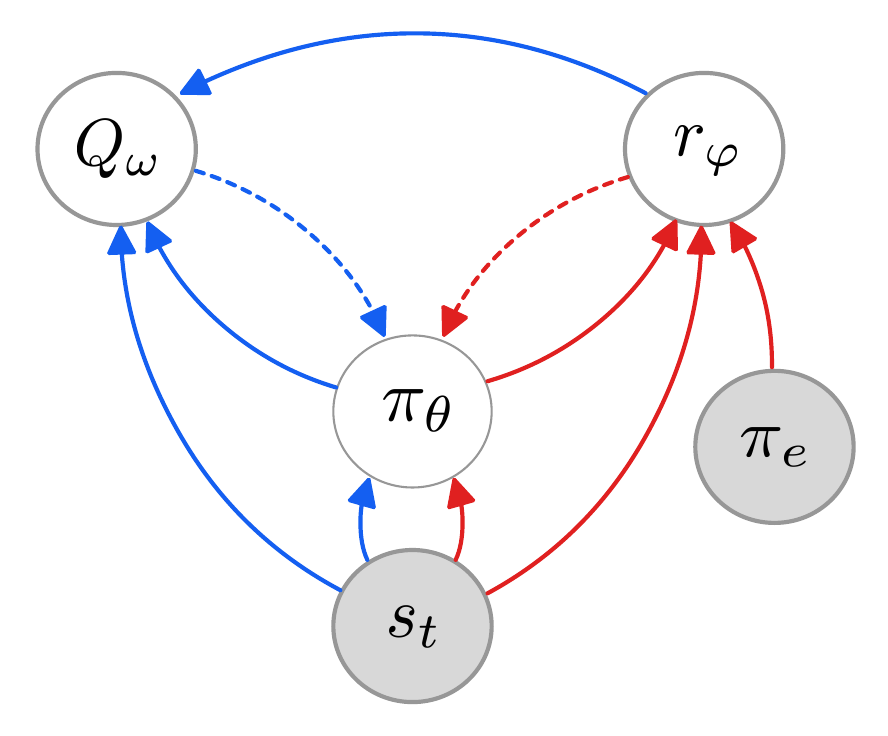}
\caption{SAM \cite{Blonde2019-vc}}
\label{diag:sam}
\end{subfigure}
\caption{Information flows (plain arrows)
and gradient flows (dotted arrows) between modules.
Best seen in color.}
\label{infogradflow}
\end{figure}

By assuming the roles of opponents in a GAN,
$\theta$ and $\varphi$ are tied in a \emph{bilevel} optimization problem
(as highlighted in \cite{Pfau2016-ft}).
Similarly, by defining an actor-critic architecture,
$\theta$ and $\omega$ are also tied in a bilevel optimization problem.
We notice the dual role of $\theta$, which is intricately tied in both bilevel problems.
As such, what SAM \cite{Blonde2019-vc} sets out to solve
can be dubbed a \emph{$\theta$-coupled twin bilevel}
optimization problem.
Note, $Q_\omega$ uses the parametric reward $r_\varphi$ as a scalar detached from the
computational graph of the $(\theta, \omega)$ bilevel problem,
as having gradients flow back from $Q_\omega$ to $\varphi$ would
prevent $D_\varphi$ from being learned as intended,
\textit{i.e.} adversarially in the $(\theta, \varphi)$ bilevel problem.
The information and gradient flows occurring between the components are
illustrated in \textsc{Figure}~\ref{infogradflow}.
As we show via numerous ablation studies in this work,
training this \emph{$\theta$-coupled twin bilevel} system to completion
is severely prone to instabilities and highly sensitive to hyper-parameters.
Ultimately, we show that $r_\varphi$'s Lipschitzness is a \textit{sine qua non} condition
for the method to perform well, and study the effects of
this necessary condition in several theoretical results
in \textsc{Section}~\ref{theory}.

Sample-efficiency is achieved through the use of a replay mechanism \cite{Lin1992-pp}:
every component (every neural network, $\theta$, $\omega$, and $\varphi$)
is trained using samples from the replay buffer $\mathcal{R}$ \cite{Mnih2013-rb,Mnih2015-iy},
a \textit{``first in, first out''} queue of fixed retention window,
to which new rollout samples (transitions) are sequentially added,
and from which old rollout samples are sequentially removed.
Note however that when a transition is sampled from $\mathcal{R}$,
its reward component is re-computed using the most recent $r_\varphi$ update.
\cite{Blonde2019-vc} and \cite{Kostrikov2019-jo} were the first to train $D_\varphi$
with experience replay, in a non-\textit{i.i.d.} context (Markovian),
for increased learning stability.
Borrowing the common terminology,
the reward is therefore effectively \textit{``learned off-policy''}.
Let $\beta$ be the off-policy distribution that corresponds to uniform sampling
over $\mathcal{R}$. $\beta$ is therefore effectively a mixture of past policy updates
$[\theta_{i - \Delta + 1}, \ldots, \theta_{i - 1}, \theta_i]$,
where the mixing depends on $\mathcal{R}$'s retention window, and the number of collected
samples per iteration.

We first introduce $\rho^\pi_{\mathbb{M}^*}$, which denotes
the discounted state visitation frequency
of an arbitrary policy $\pi$ in $\mathbb{M}^*$.
Formally,
$\rho^\pi_{\mathbb{M}^*} (s)
\coloneqq
\sum_{t=0}^{+\infty} \gamma^t \mathbb{P}^\pi_{\mathbb{M}^*} [S_t=s]$,
where $\mathbb{P}^\pi_{\mathbb{M}^*} [S_t=s]$ is the probability of reaching state $s$
at timestep $t$ when interacting with the MDP $\mathbb{M}^*$
by acting according to $\pi$.
Since $\sum_{s \in \mathcal{S}} \rho^\pi_\mathbb{M}(s) = 1 / (1-\gamma)$,
$\rho^\pi_\mathbb{M}$ can be seen as a probability distribution over states up to a constant factor.
Due to the presence of the discount factor $\gamma$, $\rho^\pi_{\mathbb{M}^*} (s)$ has higher value
if $s$ is visited earlier than later in the infinite-horizon trajectory.
In practice, we relax the definition to its non-discounted counterpart and to
the episodic regime case, as is usually done.
Plus, since every interaction is done in MDP $\mathbb{M}^*$, we use the shorthand $\rho^\pi$.
From this point forward, when states $s_t$ are sampled uniformly
from the replay buffer $\mathcal{R}$ --- in effect, following policy $\beta$ ---
the expectation over said samples will be denoted as $\mathbb{E}_{s_t \sim \rho^\beta} [\cdot]$.

We now go over how each module ($\theta$, $\omega$, and $\varphi$) is optimized in this work.
We optimize $\varphi$ with the binary cross-entropy loss, where positive-labeled samples are
from $\pi_e$, and negative-labeled samples are from $\beta$:
\begin{align}
\ell_\varphi \coloneqq
\mathbb{E}_{s_t \sim \rho^{\pi_e}, a_t \sim \pi_e}[-\log(1 - D_\varphi(s_t, a_t))]
+ \mathbb{E}_{s_t \sim \rho^\beta, a_t \sim \beta}[-\log(D_\varphi(s_t, a_t))]
\label{varphiloss}
\end{align}
In this work, unless stated otherwise, $\varphi$ is regularized with
gradient penalization $\mathfrak{R}_\varphi^\zeta (k)$,
subsuming the original formulation proposed in \cite{Gulrajani2017-mr},
which was used in SAM \cite{Blonde2019-vc} and DAC \cite{Kostrikov2019-jo}:
\begin{align}
\ell_\varphi^\textsc{GP}
&\coloneqq \ell_\varphi + \lambda \, \mathfrak{R}_\varphi^\zeta (k)
\coloneqq \ell_\varphi + \lambda \,
\mathbb{E}_{s_t \sim \rho^{\zeta}, a_t \sim \zeta}
[(\lVert  \nabla _{s_t,a_t} \, D_\varphi(s_t,a_t) \rVert - k )^2]
\label{varphilossgp}
\end{align}
The regularizer will be the object of several downstream analyses and discussions
(\textit{cf.} \textsc{Sections}~\ref{gradpen} and~\ref{gradpenrl}).
The meaning of $\lambda$, $k$ and $\zeta$ will be given in \textsc{Section}~\ref{gradpen}.

The critic's parameters $\omega$ are updated by gradient decent on the
TD loss \cite{Sutton1988-to}, using the
multi-step version \cite{Peng1996-xn} (\textit{``$n$-step''})
of the Bellman target (R.H.S. of the expected Bellman equation),
which has proven beneficial for policy evaluation
\cite{Hessel2017-ns,Fernando_Hernandez-Garcia2019-bk}.
The loss optimized by the critic is:
\begin{align}
\ell_\omega \coloneqq \mathbb{E}_{s_t \sim \rho^\beta, a_t \sim \beta}[
(Q_\omega(s_t, a_t)
- Q^\text{targ})^2]
\label{omegaloss}
\end{align}
where the target $Q^\text{targ}$ uses \textit{softly}-updated \cite{Lillicrap2016-xa}
target networks \cite{Mnih2013-rb,Mnih2015-iy},
$\theta'$ and $\omega'$, and is defined as:
\begin{align}
Q^\text{targ} &\coloneqq
\sum_{k=0}^{n-1} \gamma^{k} r_\varphi(s_{t+k}, a_{t+k})
+ \gamma^n Q_{\omega'} (s_{t+n}, \mu_{\theta'}(s_{t+n}))
\qquad
\blacktriangleright\text{\small{\textit{Bellman target}}} \\
(\theta',\omega') &\leftarrow (1 - \tau) (\theta',\omega') + \tau (\theta,\omega)
\quad 0 \leq \tau \leq 1
\qquad \qquad
\blacktriangleright\text{\small{\textit{target networks update}}}
\end{align}

Finally, since $\mu_\theta$ is deterministic, its utility value at timestep $t$ is
$U_t(\mu_\theta) = V^{\mu_\theta}(s_t)
= Q^{\mu_\theta}(s_t, \mu_\theta(s_t))
\approx \mathbb{E}_{s_t \sim \rho^\beta}[Q_\omega(s_t, \mu_\theta(s_t))]
\eqqcolon \mathcal{U}_\theta$,
where the approximation is
due to the actor-critic design involving the use of function approximators.
To maximize its utility at $t$, $\theta$ must take a gradient step in the ascending direction,
derived according to the
\emph{deterministic policy gradient theorem} \cite{Silver2014-dk}:
\begin{align}
\nabla_\theta \, U_t(\mu_\theta)
&\approx \nabla_\theta \, \mathcal{U}_\theta \label{thetautil} \\
&=\nabla_\theta \, \mathbb{E}_{s_t \sim \rho^\beta}
[Q_\omega(s_t, \mu_\theta(s_t))] \\
&=\mathbb{E}_{s_t \sim \rho^\beta}[
\nabla_\theta \mu_\theta(s_t) \nabla_a Q_\omega (s_t, a) \rvert_{a = \mu_\theta(s_t)}]
\label{policygrad}
\end{align}
This last step (\textsc{eq}~\ref{policygrad}) emerges from the natural
assumption that  $\forall s \; \nabla_\theta \, s = 0$,
since the analytical form of $\mathbb{M}$'s dynamics, $p$, is unknown.
To overcome the inherent \emph{overestimation bias} \cite{Thrun1993-or}
hindering Q-Learning and actor-critic methods based on greedy action selection
(\textit{e.g.}~DDPG \cite{Lillicrap2016-xa}),
and therefore suffered by our critic $Q_\omega$,
we apply the actor-critic counterpart of double-Q learning \cite{Van_Hasselt2010-qk}
--- analogously, Double-DQN \cite{Van_Hasselt2015-uc} for DQN ---
proposed in Twin-Delayed DDPG (\textit{abbrv.} TD3) \cite{Fujimoto2018-pe}.
This add-on method, simply called \emph{clipped double-Q learning} (\textit{abbrv.} CD),
consists in learning an additional (or \textit{``twin''}) critic,
and using the smaller of the two associated Q-values in the Bellman target,
used in the temporal-difference error of both critics.
For its reported benefits at minimal cost,
we also use the other main add-on proposed in TD3 \cite{Fujimoto2018-pe}
called \emph{target policy smoothing}.
The latter adds noise to the target action
in order for the deterministic policy not to pick actions with erroneously high Q-values,
as such input noise injection effectively smooths out the Q landscape along changes in action.
Target policy smoothing (or target smoothing, \textit{abbrv.} TS)
draws strong inspiration from the SARSA \cite{Sutton1998-ow} learning update
since it uses a perturbation of the greedy next-action
in the learning update rule, which makes the method more robust against noisy inputs
and therefore potentially safer in a safety-critical scenario.
Note, while value overfitting primarily impedes policies that are deterministic by design,
stochastic policies that prematurely collapse to their mode \cite{Schulman2015-jt}
are deterministic in effect and as such are impeded too.
In particular, fitting the value estimate against an expectation
of \emph{similar} bootstrapped target value estimates forces similar actions
to have similar values, which corresponds --- by definition ---
to making the Q-function locally Lipschitz-continuous.
As such, the induced smoothness over Q is to be understood
in terms of \emph{local Lipschitz-continuity}
(or equivalently, \emph{local Lipschitzness}), which we define in \textsc{Definition}~\ref{lipdef}.
More generally, the concept of smoothness that is at the core of the analyses
laid out in this work is the concept of Lipschitz-continuity.
Interestingly, we show later in \textsc{Section}~\ref{notenough},
formally and from first principles,
that target policy smoothing is equivalent to applying a regularizer on Q that induces
Lipschitz-continuity \textit{w.r.t.} the action input.
In addition, we align the notion of \emph{robustness} of a function approximator
with the value of its \emph{Lipschitz constant} (\textit{cf.}~\textsc{Definition}~\ref{lipdef}): a
$k_1$-Lipschitz-continuous
function approximator will be characterized as
\emph{more robust} than another $k_2$-Lipschitz-continuous function approximator if and only if $k_1 \leq k_2$.
As such, in this work, the notions of smoothness and robustness are both aligned with the notion of
Lipschitz-continuity.

\begin{definition}[local $k$-Lipschitz-continuity]
\label{lipdef}
Let $f$ be a function
$\mathcal{X} \subseteq \mathbb{R}^{n} \rightarrow \mathcal{Y} \subseteq \mathbb{R}^{m}$,
$x \mapsto f(x)$,
and $C^0$ (continuous) over $\mathcal{X}$.
We denote the euclidean norms of $\mathcal{X}$ and $\mathcal{Y}$ by
$\lVert \cdot \rVert_\mathcal{X}$ and $\lVert \cdot \rVert_\mathcal{Y}$ respectively,
and the Frobenius norm of the $\mathbb{R}^{m \times n}$ matrix space by $\lVert \cdot \rVert_F$.
Lastly, let $k$ be a non-negative real, $k \geq 0$.

\textit{(a)} $f$ is $k$-Lipschitz-continuous over $\mathcal{X}$ iff,
$\forall x, x' \in \mathcal{X}$,
\[\lVert f(x) - f(x') \rVert_\mathcal{Y} \leq k \, \lVert x - x' \rVert_\mathcal{X}\]

\textit{(b)} If $f$ is also differentiable,
then $f$ is $k$-Lipschitz-continuous over $\mathcal{X}$ iff,
$\forall x, x' \in \mathcal{X}$,
\[\lVert  \nabla \, f(x) \rVert_F \leq k\]
In either case, if the inequality is verified, $k$ is called the \textit{Lipschitz constant} of $f$.
The symbol $\nabla$, historically reserved to denote the gradient operator,
is here used to denote the Jacobian operator of the vector function $f$, to maintain symmetry with
the notations and appellations used in previous works.

\textit{(c)} Let $X$ be a subspace of $\mathcal{X}$, $X \subseteq \mathcal{X}$.
$f$ is said \emph{locally} $k$-Lipschitz-continuous over
$X \subseteq \mathcal{X}$ iff, for all $x \in X$,
there exists a neighborhood $U_x$ of $x$
such that $f$ is $k$-Lipschitz-continuous over $U_x$.
\end{definition}

Based on \textsc{Definition}~\ref{lipdef}~\textit{(b)}
the gradient penalty in \textsc{eq}~\ref{varphilossgp},
effectively enforces local Lipschitz-continuity over the support of the
$\zeta$ distribution (described later in \textit{cf.} \textsc{Section}~\ref{gradpen}),
a subspace of the state-action joint space.

Unless specified otherwise, we use both the clipped double-Q learning and target policy smoothing
add-on techniques in all the experiments reported in this work.
We ran an ablation study on both techniques to illustrate their respective benefits,
and support our algorithmic design choice to use them.
We report said ablations in \textsc{Appendix}~\ref{ablationtd3}.

We describe the inner workings of SAM in \textsc{Algorithm}~\ref{algosam}
\footnote{The symbols~``$\textcolor{blue}{\diamond}$''
and ``$\textcolor{red}{\diamond}$'' appearing
in front of line numbers in \textsc{Algorithm}~\ref{algosam}
are related to the distributed learning scheme
used in this work, which we describe in section \ref{empres1}.}.

Since our agent learns a parametric reward --- differentiable by design ---
along with a deterministic policy, we \emph{could}, in principle,
use the gradient
$\mathbb{E}_{s_t \sim \rho^\beta}[\nabla_\theta \mu_\theta(s_t)
\nabla_a r_\varphi (s_t, a) \rvert_{a = \mu_\theta(s_t)}]$
(constructed by analogy with \textsc{eq}~\ref{policygrad})
to update the policy.
\cite{Blonde2019-vc} raised the question of whether one \emph{should} use this gradient
and answered in the negative:
while the gradient in \textsc{eq}~\ref{policygrad}
guides the policy towards behaviors that maximize the long-term return of the agent,
effectively trying to address the credit assignment problem,
the gradient involving $r_\varphi$ in place of $Q_\omega$ is myopic,
and does not encourage the policy to think more than one step ahead.
It is obvious that back-propagating through $Q_\omega$,
literally designed to enable the policy to reason across longer time ranges,
will be more helpful to the policy towards solving the task.
The authors therefore discard the gradient involving $r_\varphi$.
Nonetheless, we set out to investigate whether the latter can favorably
assist the gradient in \textsc{eq}~\ref{policygrad} in solving the task,
when both gradients are used \emph{in conjunction}.
Drawing a parallel with the line of work using unsupervised auxiliary tasks
to improve representation learning in visual tasks
\cite{Jaderberg2016-sf,Shelhamer2016-kd,Mirowski2016-fm,Doersch2015-fy},
we define the gradient
$\mathbb{E}_{s_t \sim \rho^\beta}[\nabla_\theta \mu_\theta(s_t)
\nabla_a Q_\omega (s_t, a) \rvert_{a = \mu_\theta(s_t)}]$
as the \emph{main} gradient,
and
$\mathbb{E}_{s_t \sim \rho^\beta}[\nabla_\theta \mu_\theta(s_t)
\nabla_a r_\varphi (s_t, a) \rvert_{a = \mu_\theta(s_t)}]$
as the \emph{auxiliary} gradient,
which we denote by $g_m$ and $g_a$ respectively.
Based on our previous argumentation, allowing the myopic $g_a$
to take the upper hand over $g_m$
could have a disastrous impact on solving the task:
combining the $g_m$ and $g_a$ must be done conservatively.
As such, we use the auxiliary gradient only if it amplifies the
main gradient.
We measure the complementarity of the main and auxiliary tasks
by the cosine similarity between their respective gradients, $\mathfrak{S}(g_m, g_a))$,
as done in \cite{Du2018-xc},
and assemble the new composite gradient
$g_c \coloneqq g_m + \max(0, \mathfrak{S}(g_m, g_a)) \, g_a$.
By design, $g_a$ is added to $g_m$ only if the cosine
similarity between them, $\mathfrak{S}(g_m, g_a))$, is positive,
and will, in that case, be scaled by said cosine similarity.
If the gradients are collinear, they are summed: $g_c = g_m + g_a$.
If they are orthogonal or if the similarity is negative, $g_a$ is discarded: $g_c = g_m$.
Our experiments comparing the usage of $g_c$ and $g_m$
(\textit{cf.} \textsc{Figure}~\ref{cosimplots} in \textsc{Appendix}~\ref{cossim})
show that using the composite gradient $g_c$
does not yield any improvement over using only $g_m$.
By monitoring the values taken by $\mathfrak{S}(g_m, g_a))$,
we noticed that the cosine similarity was almost always negative, yet close to $0$,
hence $g_c = g_m$, which trivially explains why the results are almost identical.

\section{Lipschitzness is all you need}
\label{lipalluneed}

This section aims to
put the emphasis on what makes off-policy generative adversarial imitation learning
challenging.
When applicable, we propose solutions to these challenges,
supported by intuitive and empirical evidence.
\textit{In fine}, as the section name hints, we found that
--- in our experimental and computational setting,
described at the beginning of \textsc{Section}~\ref{empres1} ---
forcing the local Lipschitzness of the reward
is a \emph{sine qua non} condition for good performance,
while also being \emph{sufficient} to achieve peak performance.

\IncMargin{1em}
\begin{algorithm}
\SetAlgoLined
\SetNlSty{}{}{}
\SetKwInput{KwInit}{init}
\KwInit{initialize
the random seeds of each framework used for sampling,
the random seed of the environment $\mathbb{M}$,
the neural function approximators' parameters ($\theta$, $\varphi$, $\omega$),
their target networks as exact frozen copies,
the rollout cache $\mathcal{C}$,
the replay buffer $\mathcal{R}$.}
\While{no stopping criterion is met}{
    \tcc{Interact with the world to collect new samples}
    \Repeat{the rollout cache $\mathcal{C}$ is full}{
        Perform action $a_t \sim \pi_\theta(\cdot|s_t)$ in state $s_t$
        and receive the next state $s_{t+1}$ and termination indicator $d$
        returned by the environment $\mathbb{M}^*-\{r_\varphi\}$\;
        Store the reward-less transition $(s_t, a_t, s_{t+1})$
        in the rollout cache $\mathcal{C}$\;
    }
    Dump the content of the rollout cache $\mathcal{C}$ into the replay buffer $\mathcal{R}$,
    then flush $\mathcal{C}$\;
    \tcc{Train every modules}
    \ForEach{training step per iteration}{
        \ForEach{reward training step per iteration}{
            Get a mini-batch of samples from the replay buffer $\mathcal{R}$\;
            Get a mini-batch of samples from the expert demonstration dataset $\mathcal{D}$\;
            \SetNlSty{}{\textcolor{red}{$\diamond$}}{}
            Perform a gradient \emph{descent} step along
                $\nabla_\varphi \, \ell_\varphi^\textsc{GP}$
                (\textit{cf.} \textsc{eq}~\ref{varphiloss})
                using \emph{both} mini-batches:
                $$
                \ell_\varphi^\textsc{GP}
                \coloneqq \mathbb{E}_{s_t \sim \rho^{\pi_e}, a_t \sim \pi_e}[-\log(1 - D_\varphi(s_t, a_t))]
                + \mathbb{E}_{s_t \sim \rho^\beta, a_t \sim \beta}[-\log(D_\varphi(s_t, a_t))]
                + \lambda \, \mathfrak{R}_\varphi^\zeta (k)
                $$
                where $\mathfrak{R}_\varphi^\zeta (k) \coloneqq
                \mathbb{E}_{s_t \sim \rho^{\zeta}, a_t \sim \zeta}
                [(\lVert  \nabla _{s_t,a_t} \, D_\varphi(s_t,a_t) \rVert - k )^2]$
                is a gradient penalty regularizer\;
        }
        \ForEach{agent training step per iteration}{
            Get a mini-batch of samples from the replay buffer $\mathcal{R}$\;
            Augment every reward-less transition sampled from $\mathcal{R}$
                with the learned reward surrogate $r_\varphi$:
                $(s_t, a_t, s_{t+1}) \to (s_t, a_t, r_\varphi (s_t,a_t), s_{t+1})$
                (omitting here the use of $n$-step returns for simplicity)\;
            \SetNlSty{}{\textcolor{red}{$\diamond$}}{}
            Perform a gradient \emph{descent} step along
                $\nabla_\omega \, \ell_\omega$
                (\textit{cf.} \textsc{eq}~\ref{omegaloss})
                using the mini-batch:
                $$
                \ell_\omega \coloneqq \mathbb{E}_{s_t \sim \rho^\beta, a_t \sim \beta}[
                (Q_\omega(s_t, a_t)
                - Q^\text{targ})^2]
                $$
                where
                $Q^\text{targ} \coloneqq \sum_{k=0}^{n-1} \gamma^{k} r_\varphi(s_{t+k}, a_{t+k})
                + \gamma^n Q_{\omega'} (s_{t+n}, \mu_{\theta'}(s_{t+n}))$ is the $n$-step Bellman target\;
            Perform a gradient \emph{ascent} step along
                $\nabla_\theta \, \mathcal{U}_\theta$
                (\textit{cf.} \textsc{eq}~\ref{thetautil})
                using the mini-batch:
                $$
                \mathcal{U}_\theta \coloneqq
                \mathbb{E}_{s_t \sim \rho^\beta}
                [Q_\omega(s_t, \mu_\theta(s_t))]
                $$
                \;
            \SetNlSty{}{}{}
            Update the target networks using the new $\omega$ and $\theta$\;
        }
    }
    Adapt parameter noise standard deviation $\sigma$
    used to define $\pi_\theta$ from $\mu_\theta$ (\textit{cf.} \textsc{Section}~\ref{bridge})\;
    \tcc{Evaluate the trained policy}
    \ForEach{evaluation step per iteration}{
        \SetNlSty{}{\textcolor{blue}{$\diamond$}}{}
        Evaluate the empirical return of $\mu_\theta$
            in $\mathbb{M}$, using the task reward $r$
            (\textit{cf.} \textsc{Section}~\ref{bridge})\;
        \SetNlSty{}{}{}
    }
}
\caption{SAM: Sample-efficient Adversarial Mimic}
\label{algosam}
\end{algorithm}
\DecMargin{1em}

\subsection{A Deadlier Triad}
\label{triad}
In recent years, several works
\cite{Fujimoto2018-pe,Fu2019-kb,Achiam2019-os}
have carried out in-depth diagnoses
of the inherent problems of Q-learning \cite{Watkins1989-ir,Watkins1992-gl}
--- and bootstrapping-based actor-critic architectures
by extension --- in the function approximation regime.
Note, while the following issues directly apply to DQN \cite{Mnih2013-rb,Mnih2015-iy},
which even introduces additional difficulties
(\textit{e.g.} target networks, replay buffer),
we limit the scope of this section to Q-learning, to eventually make our point.
Q-learning under function approximation possesses properties that,
when used in conjunction, make the algorithm brittle, prone to unstable behavior,
as well as tedious to bring to convergence.
Without caution, the algorithm is bound to diverge.
These properties constitute the \emph{deadly triad} \cite{Sutton1998-ow,Van_Hasselt2018-ql}:
function approximation, bootstrapping, and off-policy learning.

Since the method we consider in this work \textit{per se} follows an
actor-critic architecture,
it possesses all three properties, and is therefore inclined to diverge
and suffer from instabilities.
Additionally, since the learned reward $r_\varphi$ is:
\emph{a)} defined from binary classifier predictions
--- discriminator's predicted probabilities of being expert-generated ---
estimated via function approximation,
\emph{b)} learned at the same time as the policy,
and \emph{c)} learned off-policy
--- with the negative samples coming from the replay distribution $\beta$,
the method we study consequently introduces an extra layer of complication in the deadly triad.
We now go over the three points and explain to what extent they each exacerbate
the divergence-inducing properties that form the deadly triad.

To tackle point \emph{a)}, we introduce explicit residuals
to represent the various sources of error involved in temporal-difference learning,
and illustrate how these residuals accumulate over the course of an episode.
We will use the shorthand $\mathbb{E}[\cdot]$ for expectations for the sake of legibility.
We take inspiration from \textsc{eq} $(12)$ in \cite{Fujimoto2018-pe},
where a bias term is introduced in the TD error due to the function approximation of the Q-value,
as the Bellman equation is never exactly satisfied in this regime.
Borrowing the terminology from the statistical risk minimization literature,
while the original bias suffered by the TD error was due to the \emph{estimation error}
caused by bootstrapping,
function approximation is responsible for an extra \emph{approximation error} contribution.
The sum of these two errors is represented with the residual $\delta_\omega$.
Let us now consider $D_\varphi (s,a)$, the estimated probability that a sample
$(s, a)$ is coming from expert demonstrations.
Formally,
$D_\varphi (s,a)
= \mathbb{P}_\varphi[\textsc{Expert}(s,a)]$, where the event is
defined as $\textsc{Expert}(s,a) \coloneqq
\text{``} s \sim \rho^{\pi_e} \, \land \, a \sim \pi_e \text{''}$,
and where $\mathbb{P}_\varphi$ denotes the probability estimated
with the approximator $\varphi$.
In the same vein,
we distinguish the error contributions:
the approximation error is caused by the choice of function approximatior class
(\textit{e.g.} two-layer neural networks with hyperbolic tangent activations),
and the estimation error is due to the gap between the estimations of our classifier
and the predictions of the \emph{Bayes classifier}
--- the classifier with the lowest misclassification rate in the chosen class.
This gap can be written as
$\lvert D_\varphi(s_t,a_t) - \textsc{Bayes} (s_t,a_t) \rvert$,
where $\textsc{Bayes} (s,a)
= \mathbb{P}_{\textsc{Bayes}}[\textsc{Expert}(s,a)]$,
by analogy with the previous notations.
\textit{In fine}, we introduce the residual $\delta_\varphi$
that represents the contribution of both errors in the learned reward $r_\varphi$, hence:
\begin{align}
Q_\omega(s_t,a_t)
&= r_\varphi(s_t,a_t) - \delta_\varphi(s_t, a_t)
+ \gamma \mathbb{E}[Q_\omega(s_{t+1},a_{t+1})] - \delta_\omega(s_t, a_t) \\
&= [r_\varphi(s_t,a_t) - \delta_\varphi(s_t, a_t) - \delta_\omega(s_t, a_t)]
+ \gamma \mathbb{E}[Q_\omega(s_{t+1},a_{t+1})] \\
&= \Delta_{\varphi,\omega}(s_t, a_t) + \gamma \mathbb{E}[Q_\omega(s_{t+1},a_{t+1})] \\
&= \Delta_{\varphi,\omega}(s_t, a_t) + \gamma \mathbb{E}
[
\Delta_{\varphi,\omega}(s_{t+1}, a_{t+1})
+ \gamma \mathbb{E}[Q_\omega(s_{t+2},a_{t+2})]
] \\
&= \mathbb{E}\Bigg[\sum_{k=0}^{+\infty} \gamma^k \, \Delta_{\varphi,\omega}(s_{t+k}, a_{t+k})\Bigg]
\end{align}
where
$\Delta_{\varphi,\omega}(s_t, a_t)
\coloneqq r_\varphi(s_t,a_t) - \delta_\varphi(s_t, a_t) - \delta_\omega(s_t, a_t)$.

As observed in \cite{Fujimoto2018-pe} when estimating the accumulation of error due to
function approximation in the standard RL setting, the variance of
the state-action value is proportional to the variance of both the return
and the Bellman residual $\delta_\omega$.
Crucially, in our setting involving the learned imitation reward $r_\varphi$,
it is \emph{also} proportional to the
variance of the residual $\delta_\varphi$, containing contributions of
both the approximation error \emph{and} estimation error of $r_\varphi$.
As a result, the variance of the estimate also suffers from
a critically stronger dependence on $\gamma$
(\textit{cf.} ablation study in \textsc{Appendix}~\ref{ablationdiscount}).
Intuitively, as we propagate rewards further (higher $\gamma^k$ value),
their induced residual error triggers a greater increase
in the variance of the Q-value estimate.
In addition to its effect on the variance, the additional residual also clearly
impacts the overestimation bias \cite{Thrun1993-or}
it is afflicted by, which further advocates the use
of dedicated techniques such as Double Q-learning \cite{Fujimoto2018-pe,Van_Hasselt2010-qk},
as we do in this work (\textit{cf.} \textsc{Section}~\ref{bridge}).
All in all, by introducing an extra source of approximation and estimation error,
we further burden TD-learning.

Moving on to points \textit{b)}
--- the reward is learned at the same time as the policy ---
and \textit{c)}
--- the reward is learned off-policy using samples from the replay policy $\beta$ ---
we see that each statement allow us to qualify the reward $r_\varphi$ as
a \emph{non-stationary} process.
Conceptually, by considering a additive decomposition of the reward $r_\varphi$
into a stationary $r_\varphi^\textsc{stat}$
and a non-stationary contribution $r_\varphi^\textsc{non-stat}$,
we see that following an accumulation analysis similar to the previous one
shows that the variance of the state-action value is proportional to
the variances of each contribution.
While the variance of $r_\varphi^\textsc{stat}$ can be important and therefore can
have a considerable impact on the variance of the Q-value estimate,
it can usually be somewhat tamed with online normalization techniques and mitigated with
techniques enabling the agent to cope with rewards of vastly different scales
(\textit{e.g.} \textsc{Pop-art} \cite{Van_Hasselt2016-bh}).
We show later that such methods do not help when the underlying reward is non-stationary
(\textit{cf.} \textsc{Section}~\ref{nonstat} for empirical results).
The variance of the non-stationary contribution $r_\varphi^\textsc{non-stat}$,
indeed is, due to its continually-changing nature, untameable with these regular techniques
relying on the usual stationarity assumption ---
unless additional dedicated mechanisms are integrated
(\textit{e.g.} change point detection techniques).
Naturally, the non-stationary contribution also has an effect on the bias of the
estimation, and \textit{a fortiori} on its overestimation bias
(as with \textit{a)}).
We note that the argument made in the context of Q-learning by \cite{Fu2019-kb}
naturally transfers to the TD-learning objective optimized in this work:
the objective is non-stationary, due to
\textit{i)} the \emph{moving target} problem
--- caused by using bootstrapping to learn an estimate that is updated every iteration
and \textit{ii)} the \emph{distribution shift} problem
--- caused by learning the Q-value estimate off-policy using $\beta$,
effectively being a mixture of past policies,
which changes every iteration.
Point \textit{i)} is a source of non-stationarity since the target of the supervised
objective is moving with the prediction as iterations go by,
due to using bootstrapping.
Fitting the current estimate against the target defined from this very estimate is an ordeal,
and \textit{b)} makes the task even harder by having the reward move too,
given it is also learned, at the same time.
The target of the TD objective therefore now has two moving pieces,
one from bootstrapping (\textit{i)}),
one from reward learning (\textit{b)}).
The distribution shift problem \textit{ii)}, stemming from the Q-value being learned off-policy,
is naturally worsened by the reward being estimated off-policy \textit{c)}.
Note, although both the reward and Q-value are learned with
samples from $\beta$, the actual mini-batches used to perform the gradient update of each estimate
might be different in practice.
As such, the TD error would be optimized using samples from a mixture of past policies that
is different from the mixture under which the reward is learned,
and then use this reward trained under a different effective distribution in the Bellman target.
All in all, by introducing a extra sources of non-stationarity (\textit{b)} and \textit{c)}),
we further burden the non-stationarity of TD-learning (\textit{i)} and \textit{ii)}).

\subsection{Continually changing rewards}
\label{nonstat}

In a non-stationary MDP,
the non-stationarities can manifest in the dynamics
\cite{Nilim2005-rq,Da_Silva2006-gj,Xu2007-iq,Lim2013-ml,Abdallah2016-pd},
in the reward process
\cite{Even-dar2005-rg,Dick2014-br},
or in both conjointly \cite{Yu2009-xp,Yu2009-yo,Abbasi-Yadkori2013-vd,Gajane2018-ee,
Padakandla2019-mb,Yu2019-sc,Lecarpentier2019-av}
(\textit{cf.} \textsc{Appendix}~\ref{nsmdps} for a review of
sequential decision making under uncertainty in non-stationary MDPs).
In this work, we focus on the MDP $\mathbb{M}^*$
whose transition distribution $p$ is stationary \textit{i.e.} not changing over time.
As discussed in \textsc{Section}~\ref{triad}, the reward process defined by $r_\varphi$ is
however non-stationary.
In particular, $r_\varphi$ is \emph{drifting},
\textit{i.e.} gradually changes at an unknown rate,
due to the reward being learned at the same time as the policy,
but also due to it being estimated off-policy.
While the former reason is true in the on-policy setting as well,
the latter is specific to the off-policy setting, on which we focus in this work.
Indeed, in \emph{on-policy} generative adversarial imitation learning,
the parameter sets $\varphi$ and $\theta$ are involved in a bilevel optimization
problem (\textit{cf.} \textsc{Section}~\ref{gail}) and consequently are intricately tied.
$\varphi$ is trained via an adversarial procedure opposing it to $\theta$ in
a zero-sum two-player game.
At the same time, $\theta$ is trained by policy gradients to optimize $\pi_\theta$'s
episodic accumulation of rewards generated by $r_\varphi$.
The synthetically generated rewards perceived by the agent
are, in effect, sampled from a stochastic process
that incrementally changes over the course of the policy updates,
effectively qualifying $r_\varphi$ as a drifting non-stationary reward process.

By moving to the off-policy setting
--- for reasons laid out earlier in \textsc{Section}~\ref{bridge} ---
the zero-sum two-player game is not opposing $r_\varphi$ and $\pi_\theta$,
but $r_\varphi$ and $\beta$, where $\beta$ is the off-policy distribution
stemming from experience replay.
As the parameter set $\theta$ go through gradient updates,
the new policies $\pi_\theta$ are
added to the mixture of past policies $\beta$.
Crucially, to perform its parameter update at a given iteration, the policy $\pi_\theta$
uses transitions augmented with rewards generated by $r_\varphi$,
whose latest update was trying to distinguish between samples from $\pi_e$ and $\beta$
(as opposed to $\pi_e$ and $\pi_\theta$ in the on-policy setting).
Since $\pi_\theta$ is drifting, $\beta$ is also drifting based on how experience replay operates.
Nevertheless, by being a mixture of previous policy updates, $\beta$ potentially drifts less
that $\pi_\theta$, since, in effect, two consecutive $\beta$ distributions
are mixing over a wide overlap of the same past policies.
In reality however, $\beta$ corresponds to uniformly sampling a mini-batch from
the replay buffer. Consecutive $\beta$ can therefore
be uncontrollably distant from each other in practice, making the distributional drift
of the reward more tedious to deal with than in the on-policy setting.
Using large mini-batches and distributed multi-core architectures
somewhat levels the playing field though.

The adversarial bilevel optimization problem guiding the adaptive tuning of $r_\varphi$
for every $\pi_\theta$ update
is reminiscent of the stream of research pioneered by \cite{Auer1995-mm}
in which the reward is generated by an omniscient \emph{adversary},
either arbitrarily or adaptively with potentially malevolent drive
\cite{Yu2009-xp,Yu2009-yo,Lim2013-ml,Gajane2018-ee,Yu2019-sc}.
Non-stationary environments are almost exclusively tackled from a theoretical perspective
in the literature (\textit{cf.} previous references).
Specifically, in the \emph{drifting} case, the non-stationarities are traditionally dealt with
via the use of sliding windows.
The accompanying (dynamic) regret analyses all rely on strict assumptions.
In the switching case, one needs to know the number of occurring switches beforehand,
while in the drifting case, the change variation need be upper-bounded.
Specifically, \cite{Besbes2014-hm,Cheung2019-yq} assume the total change to be
upper-bounded by some preset variation budget,
while \cite{Cheung2019-cu} assumes the variations are uniformly bounded in time.
\cite{Ortner2019-pw} assumes that the \textit{incremental}
variation (as opposed to \textit{total} in \cite{Besbes2014-hm,Cheung2019-yq}) is upper-bounded
by a \textit{per-change} threshold.
Finally, in the same vein, \cite{Lecarpentier2019-av} posits \emph{regular evolution},
by making the assumption that both the transition and reward functions are Lipschitz-continuous
\textit{w.r.t.} time.
By contrast, our approach relies on imposing local Lipschitz-continuity of the reward
over the input space, which will be described later in \textsc{Section}~\ref{gradpen}.

Online return normalization methods ---
using statistics computed over the entire return history
(reminiscent of sliding window methods)
to whiten the current return estimate ---
are the usual go-to solution to deal with rewards (and \textit{a fortiori} returns)
whose scale can vary a lot, albeit still under stationarity assumption.
We investigate whether online return normalization methods
and \textsc{Pop-Art} \cite{Van_Hasselt2016-bh} can have a positive impact on learning performance,
when the process underlying the reward is learned at the same time as the policy,
via experience replay.
Given that the reward distribution can drift at an unknown rate
(although influenced by the learning rate used to train $\varphi$),
it is fair to assume that we might benefit from such methods,
especially considering how unstable a twin bilevel
optimization problem can be.
On the other hand, as learning progresses, older rewards are -- especially in early training ---
\emph{stale}, which can potentially pollute the running statistics accumulated
by these normalization techniques.
The results obtained in this ablation study are reported
in \textsc{Appendix}~\ref{ablationretnorm}.

We observe that neither return normalization nor \textsc{Pop-Art} provide an improvement
over the baseline.
On the contrary, in \texttt{Hopper} and \texttt{Walker2d},
we see that they even yield significantly
poorer performance within the allowed runtime,
compared to the base method using neither return normalization nor \textsc{Pop-Art}
(\textit{cf.}~\textsc{Figure}~\ref{ablationretnorm}).
We propose an explanation of this phenomenon
based on the \emph{stability-plasticity dilemma} \cite{Carpenter1987-wd}.
In early training, the policy $\pi_\theta$ changes at a fast rate and with a high amplitude
when going through gradient updates,
due to being a randomly initialized neural function approximator.
The reward $r_\varphi$ is in a symmetric situation, but is also influenced by the rate
of change of $\theta$, being trained in an adversarial game.
In order to keep up with this fast pace of change in early training,
the critic $Q_\omega$ --- using the reward $r_\varphi$ in its own learning objective ---
needs to be sufficiently flexible to accommodate and adapt quickly to these frequent changes.
In other words, the critic's \emph{plasticity} must be high.
Since reward estimates from $r_\varphi$ become stale after a few $\varphi$ updates,
we also want our critic to avoid using stale reward to prevent
the degradation of $\omega$.
This property is referred to as \emph{stability} in \cite{Carpenter1987-wd}.
\textit{In fine}, the critic must be plastic and stable.
Note, using the current reward update to augment the sample transitions with their reward,
as done in this work, provides the critic with such stability.
However, return normalization and \textsc{Pop-Art} use stale running statistics
estimates to whiten the state-action values returned by the critic,
which prevents both plasticity (values need to change fast with the reward,
normalization slows down this process)
and harms stability due to the staleness of the obsolete reward that are
\textit{``baked in''} the running statistics.
The obtained results corroborate the previous analysis
(\textit{cf.} \textsc{Appendix}~\ref{ablationretnorm}).

We conclude this section by  discussing the reward learning dynamics.
While in the transient regime, the reward process is effectively non-stationary,
it gradually becomes stationary
as it reaches a steady-state regime.
Nonetheless, the presence of such stabilization does not guarantee that the desired
equilibrium has been reached.
Indeed, as we will discuss in the next section,
adversarial imitation learning has proved to be prone to overfitting.
We now address it.

\subsection{Overfitting cascade}
\label{overfitting}

Being based on a binary classifier, the synthetic reward process $r_\varphi$ is
inherently susceptible to overfitting, and it has been shown (\textit{cf.} subsequent references)
that it indeed does.
As exhibited in \textsc{Section}~\ref{related},
several endeavors have proposed techniques to prevent the learned reward from overfitting,
individually building on traditional regularization methods aimed to address overfitting
in classification.
These techniques either make the discriminator model weaker
\cite{Reed2018-ga,Blonde2019-vc,Kostrikov2019-jo,Peng2018-mo},
or make the classification task harder
\cite{Blonde2019-vc,Xu2019-uo,Zolna2019-wj},
to deter the discriminator from relying on non-salient features
to trivially distinguish between samples from $\pi_e$ and $\pi_\theta$
($\pi_e$ and $\beta$ in our off-policy setting, \textit{cf.} \textsc{Section}~\ref{nonstat}).

On a more fundamental level,
the ability of deep neural networks
to generalize (and \textit{a fortiori} to circumvent overfitting)
had been attributed to the flatness of the loss landscape in
the neighborhoods of minima of the loss function \cite{Hochreiter1997-ft,Keskar2017-dg}
--- provided the optimization method is a variant of stochastic gradient descent.
While it has more recently been shown that sharp minima \textit{can} generalize \cite{Dinh2017-iz},
we argue and show both empirically and analytically that,
in the off-policy setting tackled in this work,
flatness of the reward function around the maxima
--- corresponding to the positive samples, \textit{i.e.} the expert data ---
is paramount for good empirical performance.
In other words, we argue that the presence of peaks in the reward function
caused by the discriminator overfitting on the expert data
(non-salient features in the worst case)
is the major source of optimization issues occuring in off-policy GAIL.
As such, we focus on methods that address overfitting by inducing flatness in the
learned reward function around expert samples, subject to being peaked on the
reward landscape.
An obvious candidate to enforce this desired flatness property is
gradient penalty regularization,
inducing Lipschitz-continuity on the reward function $r_\varphi$,
over its input space $\mathcal{S} \times \mathcal{A}$,
which has been described earlier in \textsc{Section}~\ref{bridge},
and will be the object of \textsc{Sections}~\ref{gradpen} and~\ref{gradpenrl}.

Simply put, reward overfitting translates to the presence of peaks on the reward landscape.
Even in the case where these peaks exactly coincide with the expert data
(perfect classification, the discriminator coincides with the Bayes classifier
of the function class),
peaked reward landscapes (\textit{i.e.} sparse reward setting) can be tedious to optimize over.
Crucially, peaks in $r_\varphi$ \emph{can potentially}
cause peaks in the state-action value landscape $Q_\omega$.
When policy evaluation is done via Monte-Carlo estimation,
the length of the rollouts likely attenuates the contribution of individual peaked rewards
aggregated during the rollout into a discounted sum.
If the peaks were not predominant in the rollout, the associated empirical estimate of the value
will not be peaked (relative to its neighboring values).
By contrast, the TD's bootstrapping-based objective does not attenuate peaks in $r_\varphi$,
which consequently causes peaks in $Q_\omega$.
Note, using multi-steps returns \cite{Peng1996-xn}
can help mitigate the phenomenon and benefit from the
attenuation effect witnessed in the Monte-Carlo estimation described above,
hence our usage of multi-step returns in this work (\textit{cf.} \textsc{Section}~\ref{bridge}).

Narrow peaks in the state-action value estimate $Q_\omega$
can cause the deterministic policy $\mu_\theta$
to itself overfit to these peaks on the $Q_\omega$ landscape.
As such overfitting \emph{cascades} from rewards to the policy,
and hampers policy optimization (\textit{cf.} \textsc{eq}~\ref{policygrad}).
Furthermore, peaks in Q-values can severely hinder temporal-difference optimization since,
by design, these outlying values can appear in either the predicted Q-value or
the target Q-value.
As such, echoing the observations and analyses made in
\textsc{Sections}~\ref{triad} and~\ref{nonstat},
bootstrapping makes the optimization more tedious,
when bringing sampled-efficiency to GAIL.
These irregularities naturally transfer to the loss landscape,
exacerbating the innate irregularity of loss landscapes when using
neural networks as function approximators \cite{Li2018-ko},
making it harder to optimize over \textsc{eq}~\ref{omegaloss}.
\textit{In fine}, peaks on the reward landscape can cascade
and impede both policy improvement and evaluation.

In the next section (\textsc{Section}~\ref{gradpen}),
we discuss how to enforce Lipschitz-continuity in usual neural architectures,
before going over empirical results corroborating our previous analyses
(\textsc{Section}~\ref{empres1}).
Ultimately, we show that \textit{not} forcing Lipschitz-continuity on the learned surrogate reward
yields poor results, making it a \textit{sine qua non} condition for success.

\subsection{Enforcing Lipschitz-continuity in deep neural networks}
\label{gradpen}

Designed to address the shortcomings of the original GAN \cite{Goodfellow2014-yk},
whose training effectively minimizes a Jensen-Shannon divergence between
generated and real distributions,
the Wasserstein GAN (WGAN) \cite{Arjovsky2017-la} leverages
the Wasserstein metric.
Specifically, the authors of \cite{Arjovsky2017-la}
use the dual representation of the \emph{Wasserstein-1} metric
under a \emph{1-Lipschitz-continuity}
(\textit{cf.} \textsc{Definition}~\ref{lipdef})
assumption over the discriminator,
which allow them to employ the
Kantorovich-Rubinstein duality theorem,
to eventually arrive at a tractable loss
one can optimize over.

In the Wasserstein GAN \cite{Arjovsky2017-la},
the weights of the discriminator
--- called \emph{critic} to emphasize that it is no longer a classifier ---
are \emph{clipped}.
While not equivalent to enforcing the $1$-Lipschitz constraint their
model is theoretically built on,
clipping the weights \emph{does} loosely enforce Lipschitz-continuity, with
a Lipschitz constant depending on the clipping boundaries.
This simple technique however disrupts, by its design, the optimization dynamics.
As emphasized in \cite{Gulrajani2017-mr}, clipping the weights of the Wasserstein critic
can result in a pathological optimization landscape,
echoing the analysis carried out in \textsc{Section}~\ref{overfitting}.

In an attempt to address this issue,
the authors of \cite{Gulrajani2017-mr} propose to impose the underlying
$1$-Lipschitz constraint via another method,
fully integrated into the bilevel optimization problem
as a gradient penalty regularization.
When augmented with this gradient penalization technique,
WGAN --- dubbed WGAN\textit{-GP} --- is shown to yield consistently better
results, enjoys more stable learning dynamics,
and displays a smoother loss landscape \cite{Gulrajani2017-mr}.
Interestingly, the regularization technique has proved
to yield better results even in the original GAN \cite{Lucic2017-nz},
despite it not being grounded on the Lipschitzness footing like WGAN \cite{Arjovsky2017-la}.
In addition, following in the footsteps of the comprehensive study proposed in \cite{Lucic2017-nz},
\cite{Kurach2018-cs} shows empirically that the WGAN loss does not outperform
the original GAN consistently across various hyper-parameter settings,
and advocates for the use of the original GAN loss,
along with the use of spectral normalization \cite{Miyato2018-wc},
and gradient penalty regularization \cite{Gulrajani2017-mr}
to achieve the best results
(albeit at an increased cost in computation in visual domains).
In line with these works (\cite{Lucic2017-nz,Kurach2018-cs}),
we therefore commit to the archetype GAN loss formulation \cite{Goodfellow2014-yk},
as has been laid out earlier in \textsc{Section}~\ref{bridge}
when describing the discriminator objective in \textsc{eq}~\ref{varphiloss}.
We now remind the objective optimized by the discriminator
(\textit{cf.} \textsc{eq}~\ref{varphilossgp}),
where the generalized form of the gradient penalty, $\mathfrak{R}_\varphi^\zeta (k)$,
subsumes the original penalty \cite{Gulrajani2017-mr} as well as variants
that will be studied later in \textsc{Section}~\ref{gradpenrl}:
\begin{align}
\ell_\varphi^\textsc{GP}
&\coloneqq \ell_\varphi + \lambda \, \mathfrak{R}_\varphi^\zeta (k)
\coloneqq \ell_\varphi + \lambda \,
\mathbb{E}_{s_t \sim \rho^{\zeta}, a_t \sim \zeta}
[(\lVert  \nabla _{s_t,a_t} \, D_\varphi(s_t,a_t) \rVert - k )^2]
\label{eqgp}
\end{align}
In \textsc{eq}~\ref{eqgp}, $\lambda$ corresponds to the weight attributed to the
regularizer in the objective (\textit{cf.} ablation in \textsc{Section}~\ref{gradpenrl}),
and $\lVert \cdot \rVert$ depicts the euclidean norm in the appropriate vector space.
$\zeta$ is the distribution defining \emph{where}
in the input space $\mathcal{S} \times \mathcal{A}$ the Lipschitzness constraint
should be enforced.
$\zeta$ is defined from $\pi_e$ and $\beta$.
In the original gradient penalty formulation \cite{Gulrajani2017-mr},
$\zeta$ corresponds to sampling points uniformly in segments
\footnote{The segment joining the arbitrary points $x$ and $y$ in $\mathbb{R}^d$
is the set of points defined as
$S \coloneqq \{(1 - \alpha)x + \alpha y \; | \; \alpha \in [0, 1] \}$.
Sampling a point $z \in \mathbb{R}^d$ uniformly from $S$ corresponds to
sampling $\alpha \sim \operatorname{unif}(0,1)$,
before assembling $z \coloneqq (1 - \alpha)x + \alpha y$.}
joining points from the generated data and real data,
grounded on the derived theoretical results (\textit{cf.} Proposition 1 in \cite{Gulrajani2017-mr})
that the optimal discriminator is
$1$-Lipschitz along these segments.
While it does not mean that enforcing such constraint will make the discriminator optimal,
it yields good results in practice.
We discuss several formulations of $\zeta$ in \textsc{Section}~\ref{gradpenrl},
evaluate them empirically and propose intuitive arguments explaining the obtained results.
In particular, we adopt an \emph{RL viewpoint} and propose an alternate
ground as to why the regularizer has enabled successes in
control and search tasks, as reported in \cite{Blonde2019-vc,Kostrikov2019-jo}.
In particular, in \cite{Gulrajani2017-mr},
the $1$-Lipschitz-continuity is encouraged
by using $\mathfrak{R}_\varphi^\zeta (1)$ as regularizer.

Additionally, in line with the observations done in \cite{Gulrajani2017-mr},
we investigated with
\textit{a)} replacing $\mathfrak{R}_\varphi^\zeta (k)$ with a \emph{one-sided} alternative
defined as $\mathbb{E}_{s_t \sim \rho^{\zeta}, a_t \sim \zeta}
[\max(0, \lVert  \nabla _{s_t,a_t} \, D_\varphi(s_t,a_t) \rVert - k )^2]$, and
\textit{b)} ablating online batch normalization of the state input from the discriminator.
The alternative regularizer of \textit{a)} encourages the norm to be \emph{lower} than $k$
(formally, $\lVert \nabla _{s_t,a_t} \, D_\varphi(s_t,a_t) \rVert \leq k$)
in contrast to the original regularizer that enforces it to be \emph{close} to $k$.
While the one-sided version describes the notion of $k$-Lipschitzness more accurately
(\textit{cf.} \textsc{Definition}~\ref{lipdef}),
it yields similar results overall,
as shown in \textsc{Appendix}~\ref{ablationonesided}.
Crucially, we conclude from these experiments that it is \emph{sufficient} to have
the norm remain upper-bounded by $k$, or equivalently, to have $D_\varphi$ be Lipschitz-continuous.
In other words, we do not need to impose a stronger constraint
than $k$-Lipschitz-continuity
on the discriminator to achieve peak performance,
in the context of this ablation study.
As for \textit{b)}, online batch normalization of the state input is mostly hurting performance.
as reported in \textsc{Appendix}~\ref{ablationbn}.
We therefore arrive at the same conclusions as \cite{Gulrajani2017-mr}:
\textit{a)} we use the \textit{two-sided} formulation of $\mathfrak{R}_\varphi^\zeta (k)$ described in
\textsc{eq}~\ref{eqgp} since using the once-sided variant yields no improvement,
and \textit{b)} we omit the online batch normalization of the state input in the
discriminator since it hurts performance, while still using this normalization scheme in
the policy and critic (more details about the technique will be given when we describe our
experimental setting in the next section, \textsc{Section}~\ref{empres1}).

\subsection{Diagnosing the importance of Lipschitzness empirically
in off-policy adversarial imitation learning}
\label{empres1}

Before going over the empirical results reported in this section,
we describe our experimental setting.
Unless explicitly stated otherwise,
every experiment --- reported in both this section and \textsc{Section}~\ref{purpleandres} ---
is run in the same base setting.
In addition, the used hyper-parameters are made available in
\textsc{Appendix}~\ref{hps}.

\subsubsection{Environments}

In this work, we consider the simulated robotics, continuous control environments
built with the \textsc{MuJoCo} \cite{Todorov2012-gc} physics engine,
and provided to the community through the OpenAI Gym API \cite{Brockman2016-un}.
We use the following versions of the environments: \texttt{v3} for
\texttt{Hopper}, \texttt{Walker2d},
\texttt{HalfCheetah}, \texttt{Ant}, \texttt{Humanoid},
and \texttt{v2} for \texttt{InvertedDoublePendulum}.
For each of these, the dimension $n$ of a given state $s \in \mathcal{S} \subseteq \mathbb{R}^n$
and the dimension $m$ of a given action $a \in \mathcal{A} \subseteq \mathbb{R}^m$
scale as the degrees of freedom (DoFs) associated with the environment's underlying
\textsc{MuJoCo} model.
As a rule of thumb, the more complex the articulated physics-bound model is
(\textit{i.e.} more limbs, joints with greater DoFs),
the larger both $n$ and $m$ are.
The intrinsic difficulty of the simulated robotics task scales super-linearly with $n$ and $m$,
albeit considerably faster with $m$ (policy's output) than with $n$ (policy's input).

Omitting their respective versions, \textsc{Table}~\ref{envtable} reports the
state and action dimensions ($n$ and $m$ respectively) for all the environments tackled
in this work, and are ordered, from left to right, by increasing
state and action dimensions, \texttt{Humanoid-v3} being the most challenging.
Since we consider, in our experiments, expert datasets composed of at most $10$ demonstrations
($10$ is the default number; when we use $5$, we specify it in the caption),
we report return statistics (mean $\mu$ and standard deviation $\sigma$,
formatted as $\mu(\sigma)$ in \textsc{Table}~\ref{envtable}) aggregated over
the set of $10$ deterministically-selected demonstrations (the $10$ first in our fixed pool)
that every method requesting for $10$ demonstrations will receive.
To reiterate: in this work, every single method and variant will receive exactly the same
demonstrations, due to an explicit seeding mechanism in every experiment.
The reported statistics therefore identically apply to every method or variant using
$10$ demonstrations.
By design, this reproducibility asset naturally extends to settings
requesting fewer.

\subsubsection{Demonstrations}

As in \cite{Ho2016-bv}, we subsampled every demonstration with a $1/u$ ratio ---
an operation called \textit{temporal dropout} in \cite{Duan2017-tg}.
For a given demonstration, we sample an index $i_0$ from the discrete uniform distribution
$\operatorname{unif}\{0, u - 1\}$ to determine the first subsampled transition.
We then take one transition every $u$ transition from the initial index $i_0$.
\textit{In fine}, the subsampled demonstration is extracted from the original one
of length $l$ by only preserving the transitions of indices
$\{i_0 + ku \: | \: 0 \leq k < \lfloor l / u \rfloor\}$.
Since the experts achieve very high performance in the \textsc{MuJoCo} benchmark
(\textit{cf.} last column of \textsc{Table}~\ref{envtable})
they never fail their task and live until the \textit{``timeout''} episode termination
triggered by OpenAI Gym API, triggered once the horizon of $1000$ timesteps is reached,
in every environments considered in this work.
As such, most demonstrations have a length $l \approx 1000$ transitions
(sometimes less but always above $950$).
Since we use the sub-sampling rate $u=20$, as in \cite{Ho2016-bv},
the subsampled demonstrations have a length of
$|\{i_0 + ku \: | \: 0 \leq k < \lfloor l / u \rfloor\}| = \lfloor l / u \rfloor \approx 50$
transitions.

We wrap the absorbing states
in both the expert trajectories beforehand and
agent-generated trajectories at training time,
as introduced in \cite{Kostrikov2019-jo}.
Note, this assumes knowledge about the nature
--- organic (\textit{e.g.} falling down)
and triggered (\textit{e.g.} timeout flag set at a fixed episode horizon) ---
of the episode terminations (if any) occurring in the expert trajectories.
Considering the benchmark,
it is trivial to individually determine their natures in our work,
which makes said assumption of knowledge weak.
We trained the experts from which the demonstrations were then extracted
using the on-policy state-of-the-art PPO \cite{Schulman2017-ou} algorithm.
We used early stopping to halt the expert training processes when
a phenomenon of diminishing returns is observed in its empirical return,
typically attained by the $20$ million interactions mark.
We used our own parallel PPO implementation,
written in PyTorch \cite{Paszke2019-zf}, and will share the code upon acceptance.
The IL endeavors presented in this work have also been implemented
with this framework.

\subsubsection{Distributed training}

The distributed training scheme employed to obtain every empirical imitation learning result
exhibited in this work uses the MPI message-passing standard.
Upon launch, an experiment spins $n$ workers,
each assigned with an identifying unique rank $0 \leq r < n$.
They all have symmetric roles, except the rank $0$ worker,
which will be referred to as the \textit{``zero-rank''} worker.
The role of each worker is to follow the studied algorithm ---
SAM (\textit{cf.} \textsc{Algorithm}~\ref{algosam})
in the experiments reported in this section, and the proposed extension
PURPLE
in the experiments reported later
in \textsc{Section}~\ref{purpleandres}.
The zero-rank worker exactly follows the algorithm, while
the $n-1$ other workers omit
the evaluation phase (denoted by the symbol~``$\textcolor{blue}{\diamond}$'' appearing
in front of the line number).
The random seed of each worker is defined deterministically from its rank
and the \textit{base} random seed given as a hyper-parameter by the practitioner,
and is used to \textit{a)} determine the behavior of every stochastic entity involved in
the worker's training process,
and \textit{b)} determine the stochasticity of the environment it interacts with.

Before every gradient-based parameter update step
--- denoted in \textsc{Algorithm}~\ref{algosam}
by the symbol~``$\textcolor{red}{\diamond}$'' appearing
in front of the line number ---
the zero-rank worker gathers the gradients across the $n-1$ other workers,
and aggregates them via an averaging operation,
and sends the aggregate to every worker.
Upon receipt, every worker of the pool then uses the aggregated gradient
in its own learning update.
Since the parameters are synced across workers before the learning process kicks off,
this \emph{synchronous} gradient-averaging scheme
ensures that the workers all have the same parameters
throughout the entire learning process (same initial parameters, then same updates).
This distributed training scheme leverages learners seeded differently in their own
environments, also seeded differently, to accelerate exploration, and above all
provide the model with greater robustness.

Every imitation learning experiment whose results are reported in this work
has been run for a fixed wall-clock duration
--- 12 or 48 hours, as indicated in their respective captions ---
due to hardware and computational infrastructure constraints.
While the effective running time appears in the caption of every plot,
the latter still depict the temporal progression of the methods in terms of \emph{timesteps},
the number of interactions carried out with the environment.
The reported performance corresponds to the undiscounted empirical return,
computed using the reward returned by the environment (available at evaluation time),
gathered by the non-perturbed policy $\mu_\theta$ (deterministic) of the zero-rank worker.
Every experiment uses $16$ workers,
and can therefore be executed on most desktop consumer-grade computers.
Lastly, we monitored every experiment with the Weights \& Biases \cite{Biewald2020-wb} tracking and visualization tool.

Additionally, we run each experiment with $5$ different \textit{base} random seeds ($0$ to $4$),
raising the effective seed count per experiment to $80$.
Each presented plot depicts the mean across them with a solid line,
and the standard deviation envelope
(half a standard deviation on either side of the mean) with a shaded area.

Finally, we use an \emph{online} observation normalization scheme,
instrumental in performing well in continuous control tasks.
The running mean and standard deviation used to standardize the observations are
computed using an online method to represent the statistics of the entire history of observation.
These statistics are updated with the mean and standard deviation computed over
the concatenation of latest rollouts collected by each parallel worker,
making is effectively an \emph{online distributed}
batch normalization \cite{Ioffe2015-ls} variant.

\begin{table}
\centering
\begin{tabular}{c|cc|c}
\hline
Environment&State dim. $n$&Action dim. $m$&Expert Return $\mu(\sigma)$ \\
\hline
\rowcolor{MyLightGray}
\texttt{IDP}&$11$&$1$&$9339.966(1.041)$ \\
\texttt{Hopper}&$11$&$3$&$4111.823(56.81)$ \\
\rowcolor{MyLightGray}
\texttt{Walker2d}&$17$&$6$&$6046.116(13.76)$ \\
\texttt{HalfCheetah}&$17$&$6$&$7613.154(36.25)$ \\
\rowcolor{MyLightGray}
\texttt{Ant}&$111$&$8$&$6688.696(48.83)$ \\
\texttt{Humanoid}&$376$&$17$&$9175.152(98.94)$ \\
\hline
\end{tabular}
\caption{State and action dimensions, $n$ and $m$,
of the studied environments from the \textsc{MuJoCo}
\cite{Todorov2012-gc} simulated robotics benchmark from
OpenAI Gym \cite{Brockman2016-un}.
(\textit{abbrv.} \texttt{IDP} for \texttt{InvertedDoublePendulum}, the continuous control
counterpart of \texttt{Acrobot}.)
In the last column, we report both the mean $\mu$ and standard deviation $\sigma$
(formatted as $\mu(\sigma)$ in the table)
of the expert's returns, aggregated across the set of $10$ demonstrations used in this work.}
\label{envtable}
\end{table}

\subsubsection{Empirical results}

\begin{figure}
  \center
  \begin{subfigure}[t]{0.49\textwidth}
    \center\scalebox{0.18}[0.18]{\includegraphics{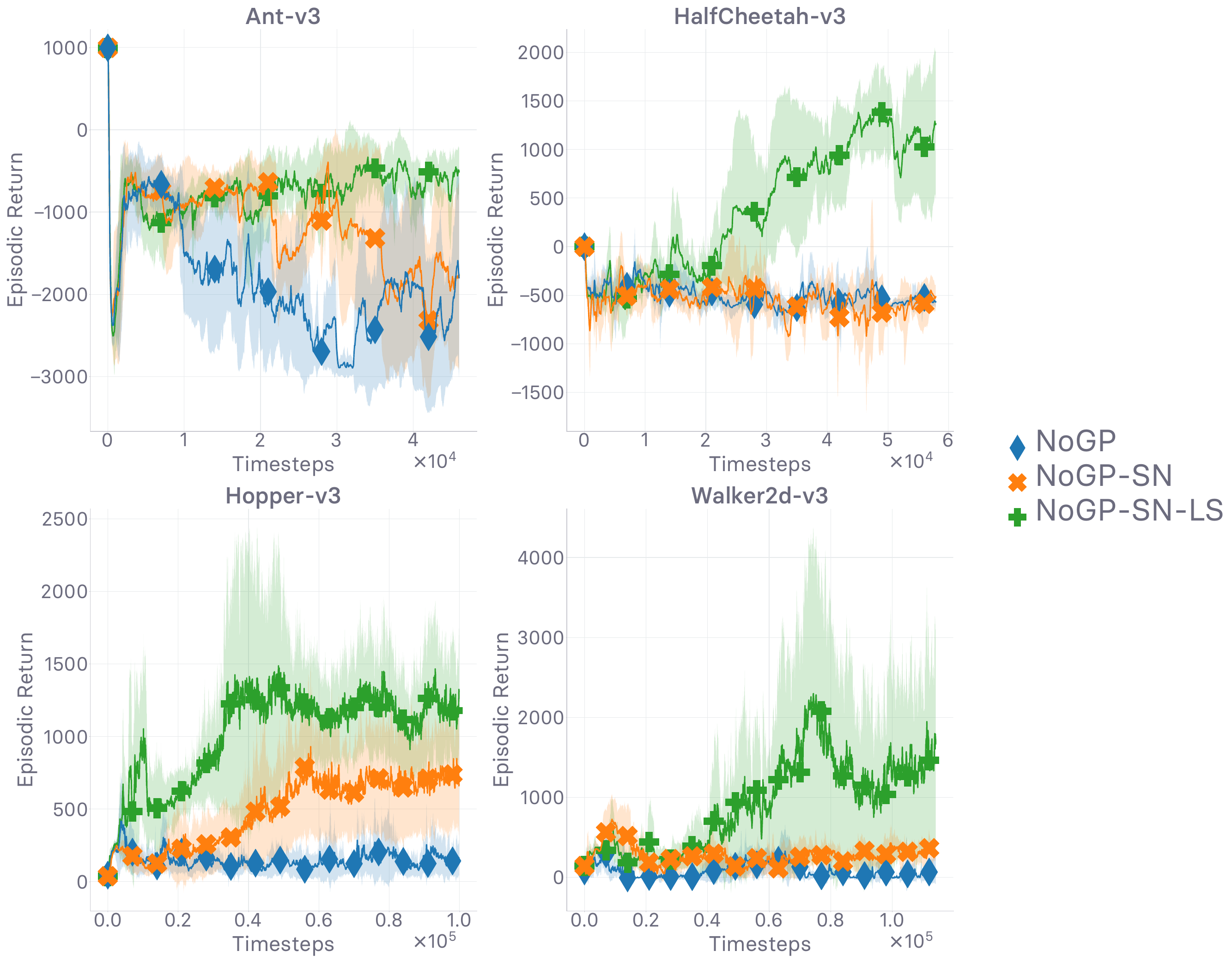}}
    \caption{Evolution of return values \textit{(higher is better)}}
  \end{subfigure}
  \begin{subfigure}[t]{0.49\textwidth}
    \center\scalebox{0.18}[0.18]{\includegraphics{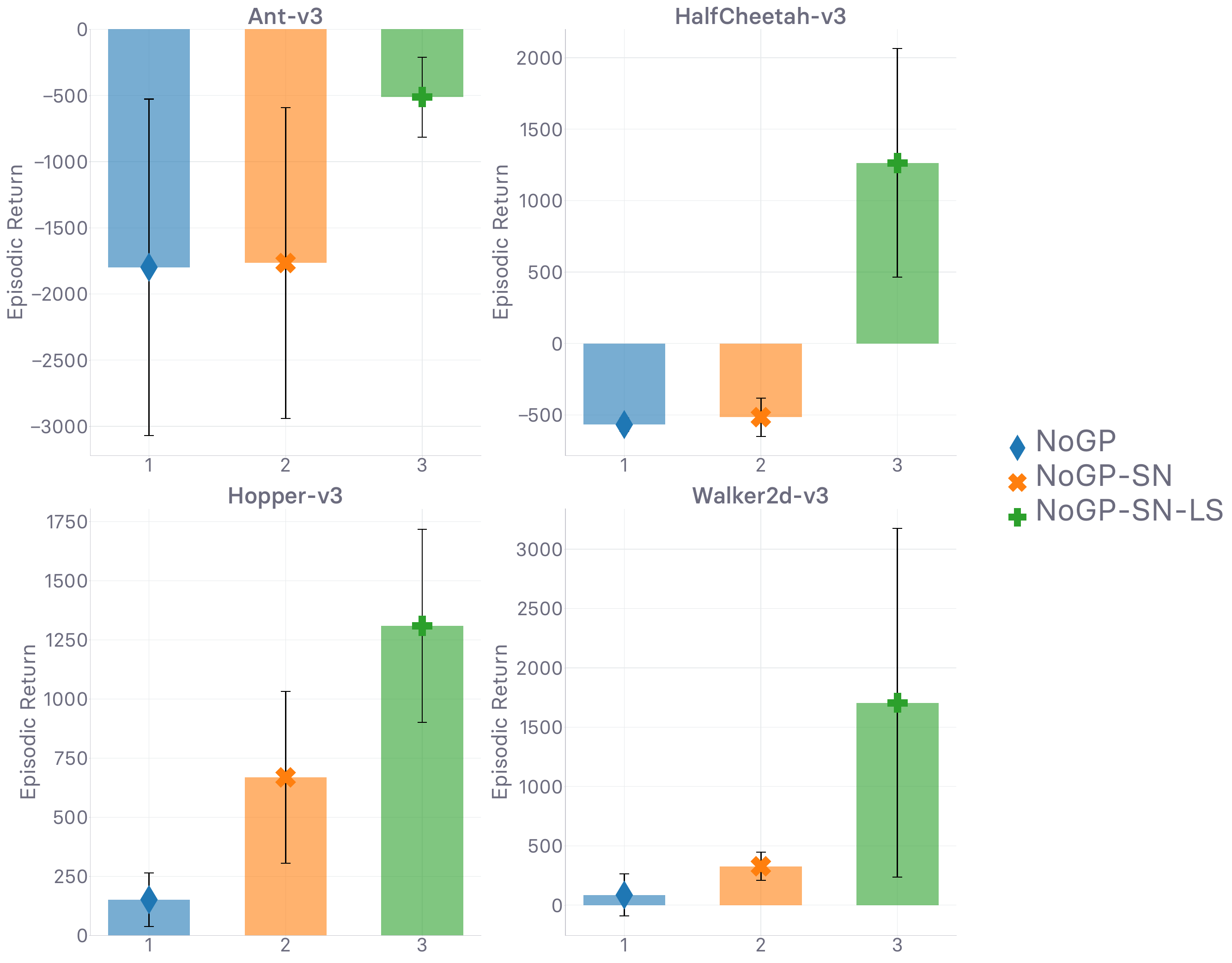}}
    \caption{Final return values at timeout \textit{(higher is better)}}
  \end{subfigure}
  \caption{
  Evaluation of several methods while \emph{not} using GP.
  Legend described in text.
  Runtime is 12 hours.}
  \label{resplotsnogp}
\end{figure}

\begin{figure}
  \center
  \begin{subfigure}[t]{0.49\textwidth}
    \center\scalebox{0.18}[0.18]{\includegraphics{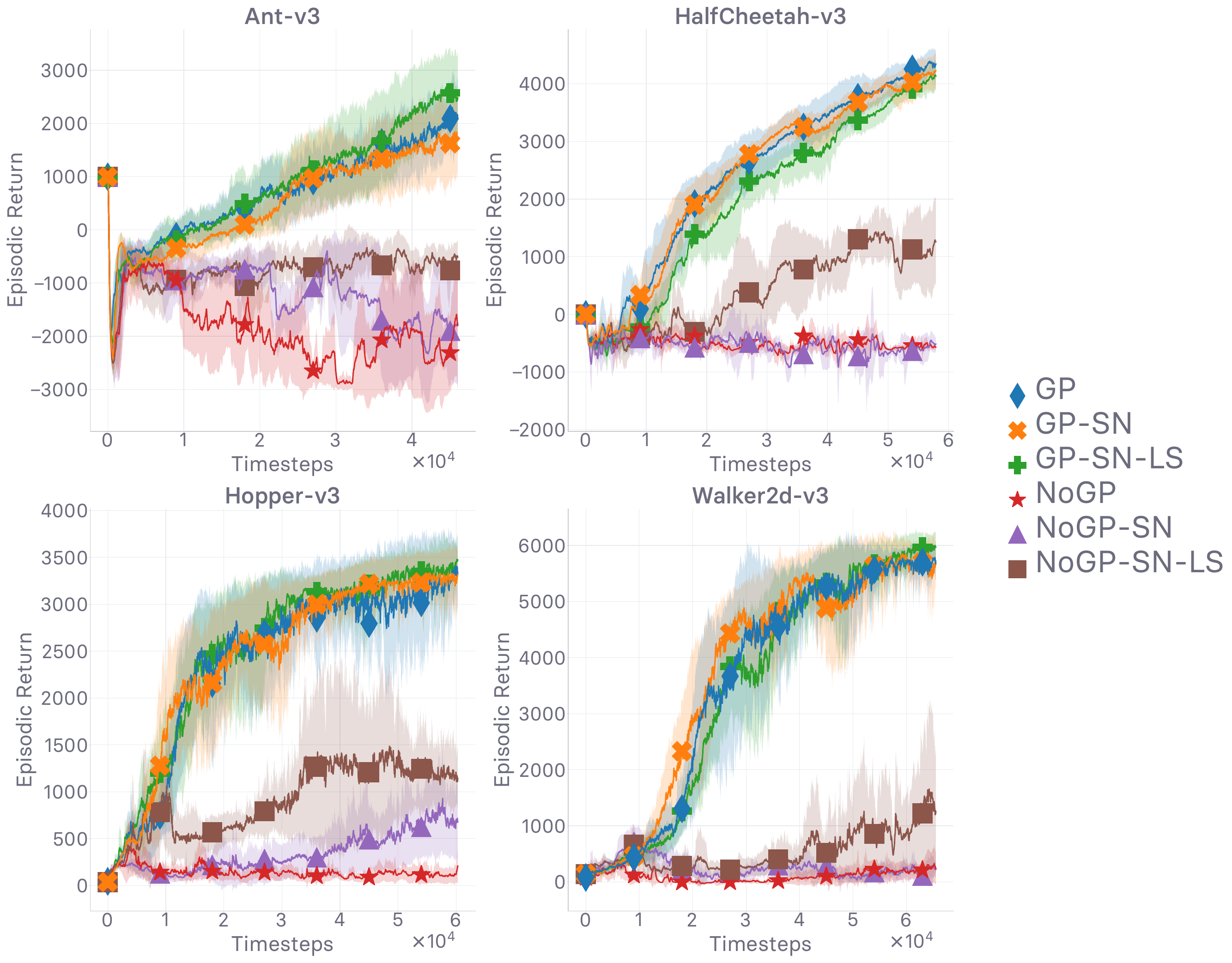}}
    \caption{Evolution of return values \textit{(higher is better)}}
  \end{subfigure}
  \begin{subfigure}[t]{0.49\textwidth}
    \center\scalebox{0.18}[0.18]{\includegraphics{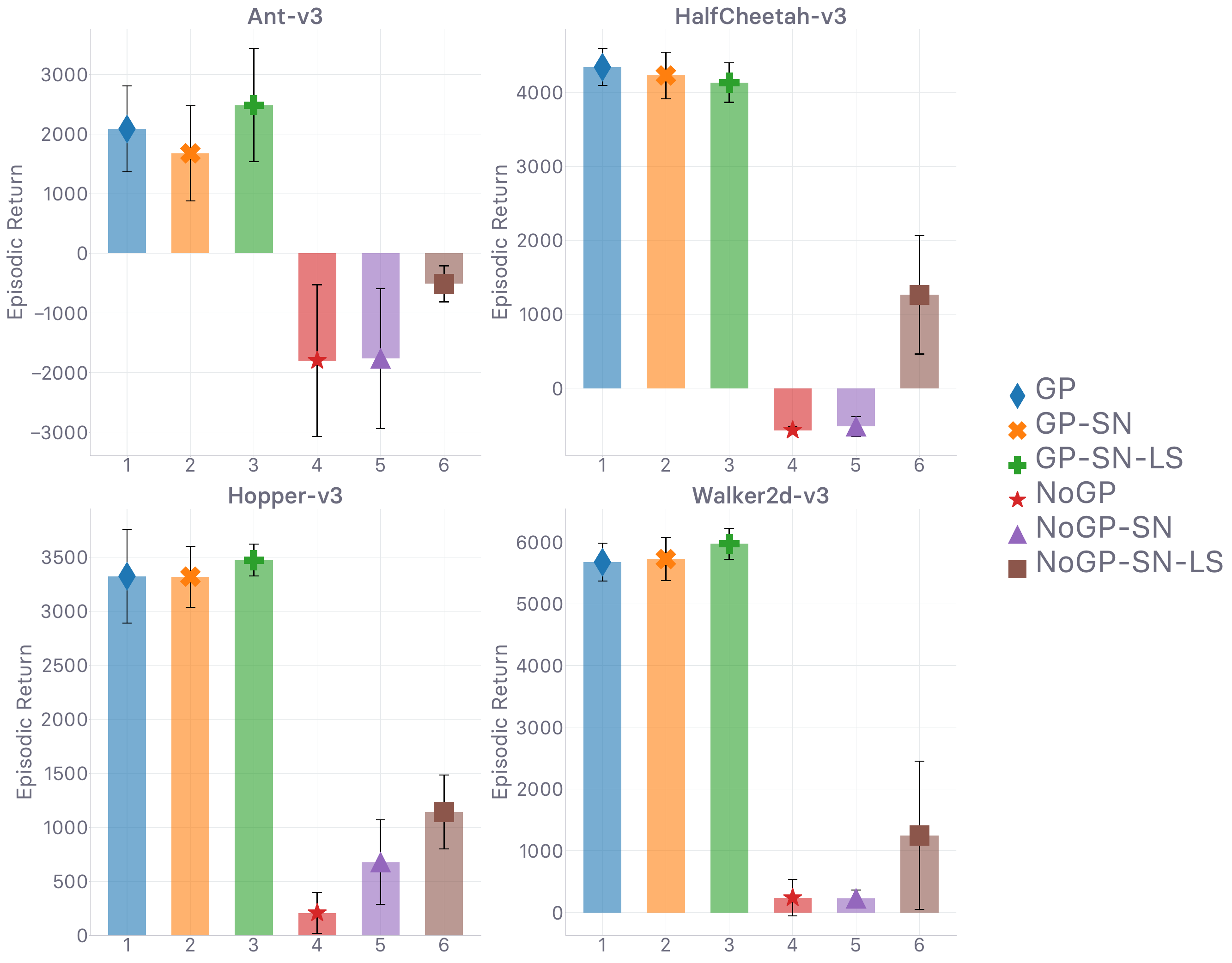}}
    \caption{Final return values at timeout \textit{(higher is better)}}
  \end{subfigure}
  \caption{
  Evaluation of several methods showing the necessity of GP.
  Legend described in text.
  Runtime is \textbf{\emph{12 hours}}.}
  \label{resplotsgp}
\end{figure}

\begin{figure}
  \center
  \begin{subfigure}[t]{0.99\textwidth}
    \center\scalebox{0.18}[0.18]{\includegraphics{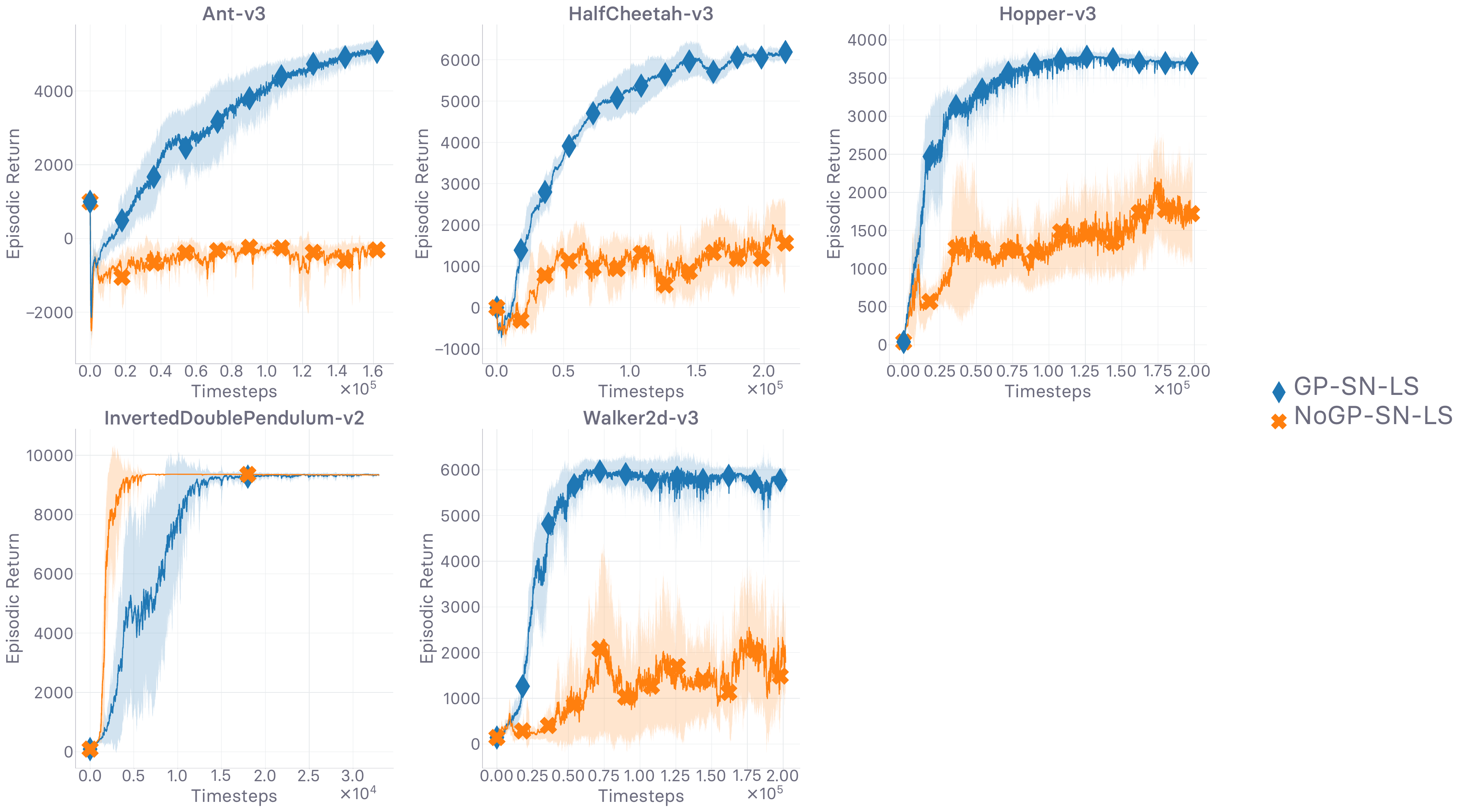}}
  \end{subfigure}
  \caption{
  Evaluation of several methods showing the necessity of GP.
  Legend described in text.
  Runtime is \textbf{\emph{48 hours}}.}
  \label{resplotsgp2}
\end{figure}

We now go over our first set of empirical results, whose goal is to show
to what extent gradient penalty regularization is needed.
The compared methods all use SAM (\textit{cf.} \textsc{Section}~\ref{bridge})
as base.

First, \textsc{Figure}~\ref{resplotsnogp} compares several modular configurations,
which are described using the following handles in the legend.
\texttt{GP} means that gradient penalization (GP)
(\textit{cf.} \textsc{Section}~\ref{gradpen}) is used.
\texttt{NoGP} means that GP is not used
(using $\ell_\varphi$ instead of $\ell_\varphi^\textsc{GP}$).
Note, \texttt{NoGP} is the only negative handle that we use, since it it central to our analyses.
When any other technique is not in use, it is simply absent from the handle in the legend.
\texttt{SN} means that spectral normalization (SN) \cite{Miyato2018-wc} is used.
SN normalizes the discriminator's weights to have a norm close to $1$,
drawing a direct parallel with GP.
In line with what the large-scale ablation studies on GAN add-ons advocate
\cite{Lucic2017-nz,Kurach2018-cs}, SN is used in most modern GAN architectures for its simplicity.
We here investigate if SN is enough to keep the gradient in check, or if GP is necessary.
\texttt{LS} denotes one-sided uniform label smoothing, consisting in replacing
the positive labels only (hence \textit{one-sided}), which are normally equal to $1$ (expert, real),
by a \textit{soft label} $u$, distributed as $u \sim \operatorname{unif}(0.7,1.2)$.
We do not consider Variational Discriminator Bottleneck (VDB) \cite{Peng2018-mo}
in our comparisons since
\textit{a)} we prefer to focus on stripped-down canonical methods,
and \textit{b)} the information bottleneck forced on the discriminator's hidden representation
boils down to smoothing the labels anyway, as shown recently in \cite{Muller2019-rr}.

In \textsc{Figure}~\ref{resplotsnogp}, we see that \emph{not using GP} (\texttt{NoGP})
prevents the agent
from learning anything valuable: the agent barely collects \emph{any reward at all}.
While using SN can improve performance slightly (\texttt{NoGP-SN}),
the addition of LS (\texttt{NoGP-SN-LS})
\emph{considerably} improves performance over the two previous candidates.
Nonetheless, despite the sizable runtime, all three perform poorly
and are a far cry from achieving the same empirical return as the expert
(\textit{cf.} \textsc{Table}~\ref{envtable}).
In contrast with \textsc{Figure}~\ref{resplotsnogp},
\textsc{Figure}~\ref{resplotsgp} and \textsc{Figure}~\ref{resplotsgp2}
show to what extent introducing GP in the off-policy imitation learning algorithm
considered in this work impacts performance positively.
The performance gap is \emph{substantial}
--- in every environment except the easiest one considered, \texttt{InvertedDoublePendulum-v2},
as described in \textsc{Table}~\ref{envtable}.
As soon as GP is in use, the agent achieves near-expert performance
(\textit{cf.} \textsc{Table}~\ref{envtable}).
\textit{In fine}, \textsc{Figure}~\ref{resplotsnogp} shows that
\emph{without} GP, neither SN nor LS are enough to enable the agent
to mimic the expert with high fidelity,
while \textsc{Figure}~\ref{resplotsgp}  and \textsc{Figure}~\ref{resplotsgp2}
show that \emph{with} GP,
extra methods such as LS barely improve performance.
These results support our claim: gradient penalty is, (\emph{empirically})
\emph{necessary} and \emph{sufficient} to ensure near-expert performance in
\emph{off-policy} generative adversarial imitation learning,
in our computational setting.

\begin{figure}
  \center
  \begin{subfigure}[t]{0.99\textwidth}
    \center\scalebox{0.18}[0.18]{\includegraphics{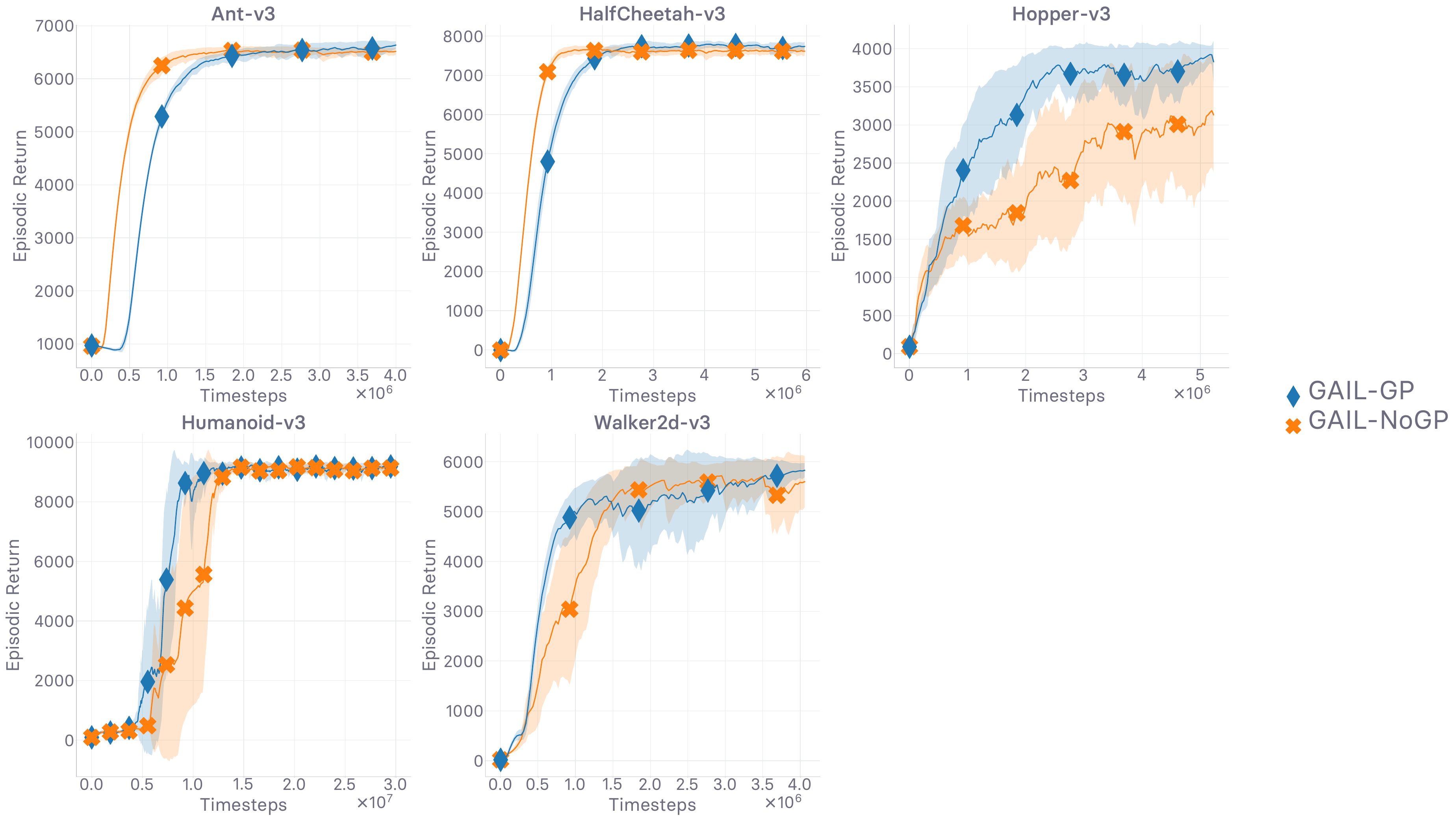}}
  \end{subfigure}
  \caption{
  Ablation study on GP
  in \emph{on-policy} GAIL.
  We see that the agent is still able to learn policies achieving peak performance
  even without GP, in contrast to the off-policy version of the algorithm.
  In the most difficult environment of the \textsc{MuJoCo} suite
  (\textit{cf.} \textsc{Table}~\ref{envtable}),
  \texttt{Humanoid}, GP achieves best performance.
  Runtime is 12 hours.}
  \label{resplotsgail}
\end{figure}

We also conducted an ablation of GP in the \emph{on-policy} setting,
reported in \textsc{Figure}~\ref{resplotsgail}.
We see that across the range of environments,
GP does not assume the same decisive role as in the off-policy setting.
In fact, the agent reaches peak performance earlier \emph{without} GP in
two challenging environments, \texttt{Ant} and \texttt{HalfCheetah},
out of the five considered.
Nevertheless, it still allows the agent to attain peak empirical return faster in
\texttt{Hopper}, \texttt{Walker2d}, and perhaps most strikingly,
in the extremely complex \texttt{Humanoid} environment.
All in all, while GP can help in the on-policy setting, in is not
\emph{necessary} as in the off-policy setting studied in this work.
In line with the analyses led in \textsc{Sections}
\ref{triad},~\ref{nonstat}, and~\ref{overfitting},
the results of \textsc{Figure}~\ref{resplotsgail} somewhat
corroborate our claim that the presence of bootstrapping
in the policy evaluation objective creates a \emph{bottleneck}, that can be
addressed by enforcing a Lipschitz-continuity constraint --- GP ---
on the reward learned for imitation.

\begin{figure}
  \center
  \begin{subfigure}[t]{0.49\textwidth}
    \center\scalebox{0.16}[0.16]{\includegraphics{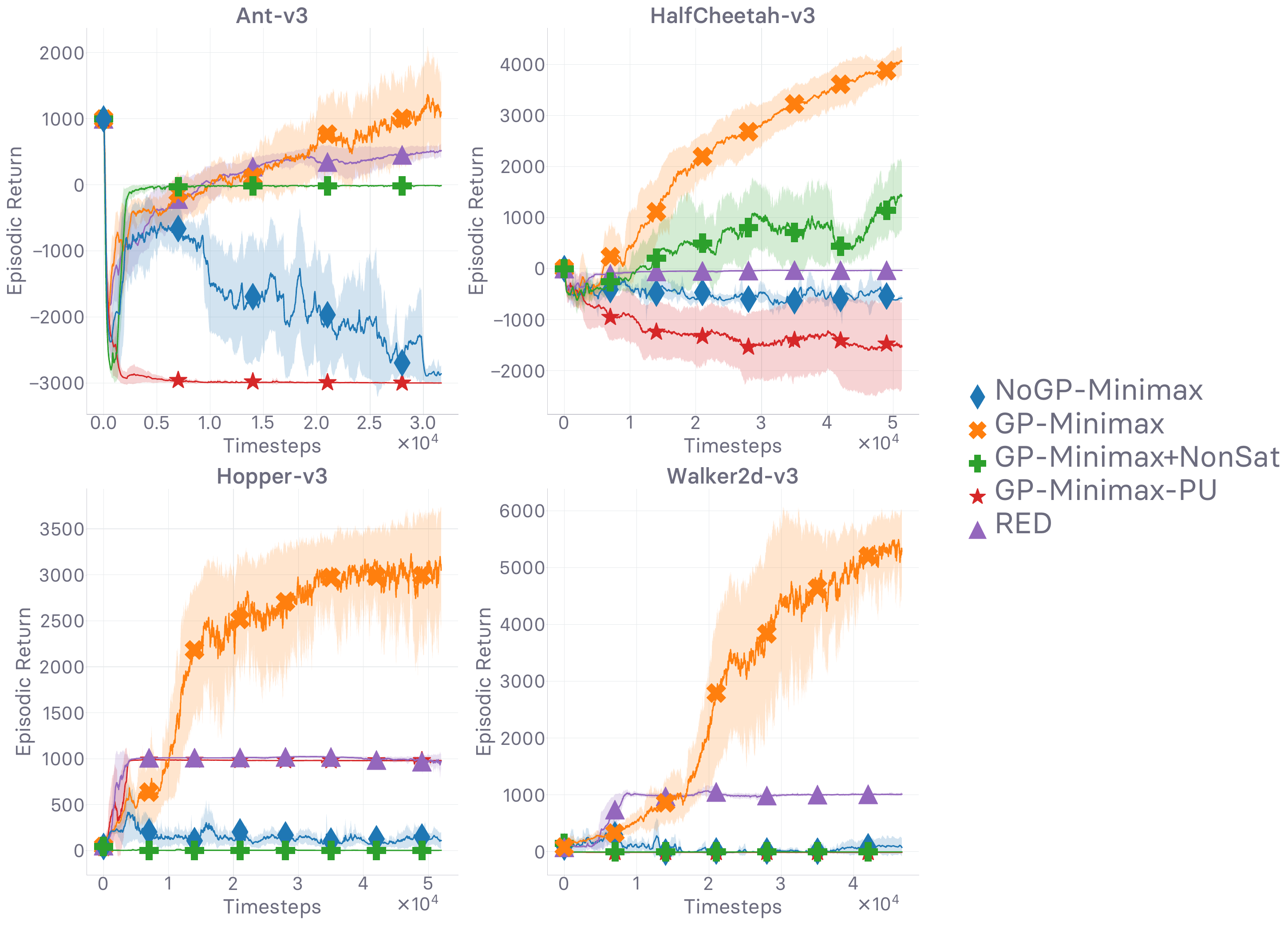}}
    \caption{Evolution of return values \textit{(higher is better)}}
  \end{subfigure}
  \begin{subfigure}[t]{0.49\textwidth}
    \center\scalebox{0.16}[0.16]{\includegraphics{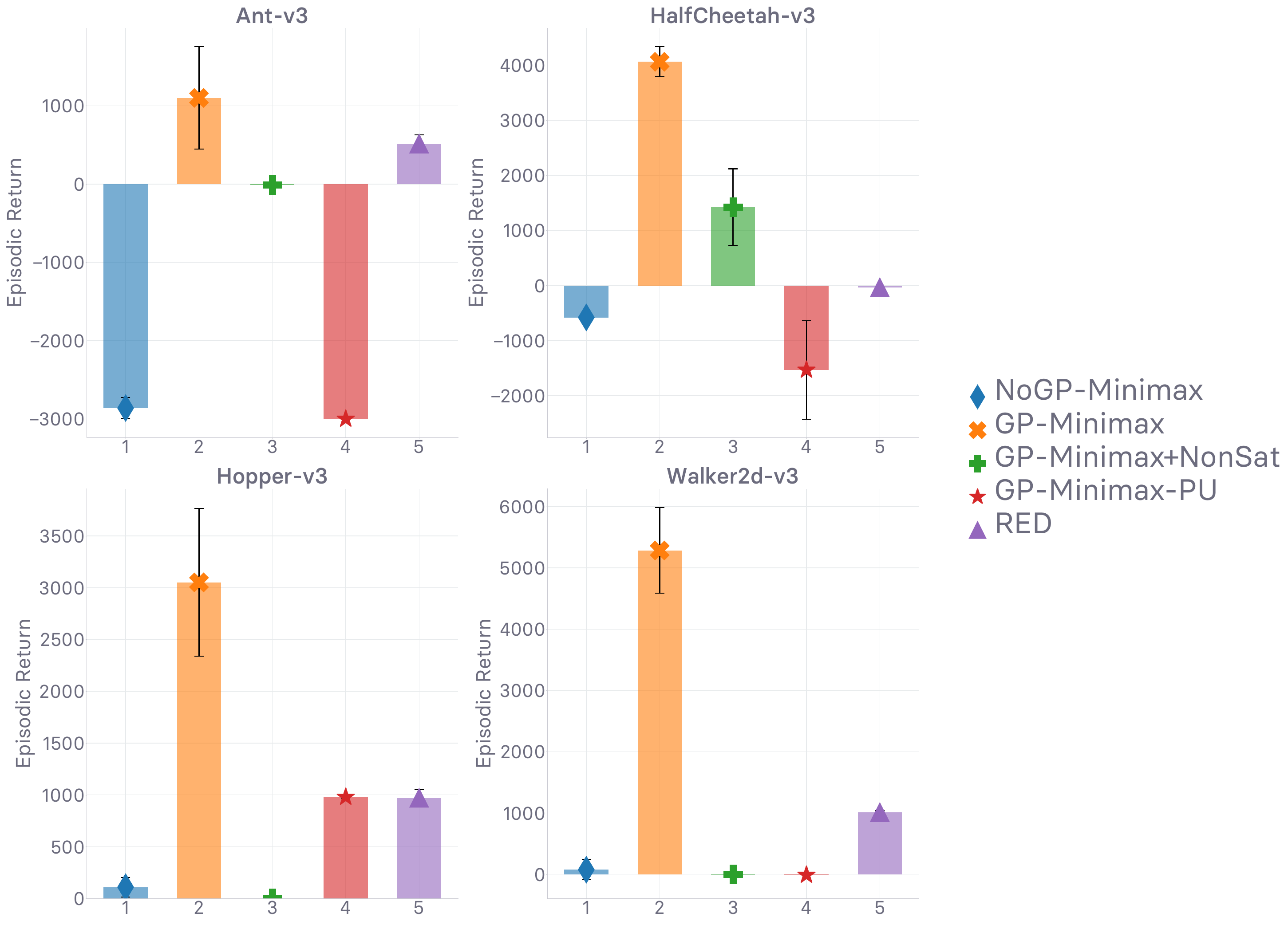}}
    \caption{Final return values at timeout \textit{(higher is better)}}
  \end{subfigure}
  \caption{
  Evaluation of several alternate reward formulations.
  Legend described in text.
  Runtime is 12 hours.}
  \label{resplotsredpu}
\end{figure}

\textsc{Figure}~\ref{resplotsredpu} compares SAM,
with and without GP,
against several alternate versions of the objective used
to train the surrogate reward for imitation.
We introduce the following new handles to denote these methods.
\textit{``RED''} means that the random expert distillation (RED) \cite{Wang2019-pd}
method is used to learn the imitation reward, replacing the adversarial one in SAM.
RED is based on random network distillation (RND) \cite{Burda2018-vl},
an exploration method using the prediction error of a learned network against a random fixed target
as a measure of novelty, and use it to craft a reward bonus.
Instead of updating the network while training to keep the novelty estimate tuned to
the current exploration level of the agent, RED trains the RND predictor network
to predict the random fixed target on the expert dataset \emph{before} training the policy.
RED then uses the prediction error to assemble a reward signal for the imitation agent,
who is rewarded \emph{more} if the actions it picks are deemed \emph{not novel},
as that means the agent's occupancy measure matches the occupancy of what has been seen before,
\textit{i.e.} the expert dataset. As such, RED is a technique that rewards the agent for
matching the distribution support of the expert policy $\pi_e$.
Note, as opposed to adversarial imitation, the RED reward is not updated during training,
which technically protects it from overfitting.
\textit{``PU''} means that we learn the reward via adversarial
imitation, but using the discriminator objective
recently proposed in positive-unlabeled (PU) GAIL \cite{Xu2019-uo}.
Briefly, the method considers that while the expert-generated samples
are positive-labels,
the agent-generated ones are unlabeled (as opposed to negative-labeled).
Intuitively, it should prevent the discriminator overfitting on irrelevant features
when it becomes difficult for the discriminator to tell agent and expert apart.

The wrapping mechanism
--- consisting in wrapping the \emph{absorbing} transitions,
which we described in \textsc{Section}~\ref{bridge} ---
is used in every experiment reported in \textsc{Figure}~\ref{resplotsredpu},
\emph{including RED}.
In addition, note, we only use GP in the adversarial context we introduced it in.
We do not use GP with RED.
Each technique is re-implemented based on the associated paper,
with the same hyper-parameters, with the exception of RED:
instead of using the per-environment scale for the prediction loss on which the
RED reward is built, we keep a running estimate of the standard deviation of this
prediction loss and rescale said prediction loss with its running standard deviation.
This modification is consistent with the rescaling done in
the paper RED is based on RND.
By contrast, the per-environment scales in RED's official implementation
span several orders of magnitude (four).
We here opt for environment-agnostic methods.

The results in \textsc{Figure}~\ref{resplotsredpu}
show that the wrapping techniques introduced in \cite{Kostrikov2019-jo}
and described in \textsc{Section}~\ref{bridge}
increases performance overall.
Like we have shown before in \textsc{Figures}~\ref{resplotsnogp}, \ref{resplotsgp}, and \ref{resplotsgp2},
not using GP causes a considerable drop in performance.
PU prevents the agent to learn an expert-like policy, in every environment.
Note, while the comparison is fair, PU was introduced in \emph{visual} tasks.
In particular, we see that, in \texttt{Hopper}, PU's empirical return hits a plateau
at about $1000$ reward units (\textit{abbrv.} r.u.).
We observe the exact same phenomenon with RED, for which it occurs
in \emph{every} environment.
This is caused by the agent being stuck performing the same sub-optimal actions,
accumulating sub-optimal outcomes until episode termination artificially triggered by timeout.
The agent exploits the fact that it has a lifetime upper-bounded by said timeout
and is therefore \emph{biased} by its \emph{survival}
(survival bias, \textit{cf.} \textsc{Section}~\ref{bridge}).
The RED agents are in effect staying alive until termination,
and therefore avoid falling down (organic trigger) until the timeout (artificial trigger)
is reached.
While the reward used in RED is not negative, the agent quickly reaches a performance level
at which all the rewards are almost identical
--- since the RED reward is trained \emph{beforehand}, with no chance of adaptive tuning like
training the reward \emph{at the same time} allows in this work, and since
RED's score is based on how the agent and expert distribution match.
Once the agent is similar enough to the expert, it always gets the same rewards
and has therefore no incentive to resemble the expert with higher fidelity.
Instead, it is content and just tries to live through the episode.
This propensity to survival bias explains why such care was taken to hand-tune
its scale.
Finally, even though wrapping absorbing transitions generally improves performance,
\textsc{Figure}~\ref{resplotsredpu} shows that survival bias is avoided
even \emph{without} it (occurrence in \texttt{Hopper} has been overcome).

The results in \textsc{Figure}~\ref{resplotsgp}
provide empirical evidence
that enforcing Lipschitz-continuity on $D_\varphi$
over the input space via the gradient regularization (\textit{cf.} \textsc{eq}~\ref{eqgp})
is \emph{necessary} and \emph{sufficient}
for the agent to achieve expert performance in the considered off-policy setting.
We therefore ask the question:
is the positive impact that GP has on training imitation policies via bootstrapping explained
\textit{a)} by its \emph{direct} effect on the reward smoothness,
or \textit{b)} by its \emph{indirect} effect on the state-action value smoothness?
We argue that \emph{both} contribute to the stability and performance of the studied method.
While point \textit{a)} is intuitive from the analyses laid out in \textsc{Section}
\ref{triad},~\ref{nonstat}, and~\ref{overfitting}, we believe that point
\textit{b)} deserves further analysis and discussion.
As such, we derive theoretical results to qualify, both qualitatively and quantitatively,
the Lipschitz-continuity that is potentially \emph{implicitly enforced} on the state-action value
when assuming the Lipschitz-continuity of the reward.
These results are reported in \textsc{Section}~\ref{theory},
and will hopefully help us answer the previous question.
A discussion of the \emph{indirect} effect and how it compares to the direct effect
implemented by target smoothing is carried out in \textsc{Section}~\ref{notenough}.

\section{Pushing the analysis further: robustness guarantees and provably more robust extension}

\subsection{Robustness guarantees: state-action value Lipschitzness}
\label{theory}

In this section, we ultimately show that enforcing a Lipschitzness constraint on
the reward $r_\varphi$ has the effect of enforcing a Lipschitzness
constraint on the associated state-action value $Q_\varphi$.
Note, $Q_\varphi$ is the \textit{real} Q-value derived from $r_\varphi$,
while $Q_\omega$ is a function approximation of it.
We discuss this point in more detail in \textsc{Section}~\ref{discussion}.
We characterize and discuss the conditions under which such result is satisfied,
as well as how the exhibited Lipschitz constant for $Q_\varphi$ relates to
the one enforced on $r_\varphi$.
We work in the \emph{episodic} setting, \textit{i.e.} with a finite-horizon $T$,
which is achieved by assuming that $\gamma=0$ once an absorbing state is reached.
Note, since we optimize over mini-batches in practice,
nothing guarantees that the Lipschitz constraint is satisfied
by the learned function approximation
\emph{globally} across the whole joint space $\mathcal{S} \times \mathcal{A}$,
at every training iteration.
In such setting, we are therefore reduced to \emph{local} Lipschitzness,
defined as Lipschitzness in neighborhoods around samples at which the constraint is applied.
The provenance of these samples is not the focus of this theoretical section and
assume they are agent-generated.
We study the effect of enforcing Lipschitzness constraints
on other data distributions in \textsc{Section}~\ref{gradpenrl}.

\paragraph{Notations.}
Given a function
$f: \mathbb{R}^n \times \mathbb{R}^m \rightarrow \mathbb{R}^d$,
taking the pair of vectors $(x, y)$ as inputs, we denote by
$\nabla _{x, y} \, f$ the pair of Jacobians associated with $x$ and $y$,
$\nabla _x \, f$ and $\nabla _y \, f$ respectively,
which are rectangular matrices in
$\mathbb{R}^{d \times n}$ and $\mathbb{R}^{d \times n}$ respectively.
Now that the stable concepts and notations have been laid out,
we introduce the variables $x_i$ and $y_i$,
indexed by $i \in \mathcal{I} \subseteq \mathbb{N}$.
Note, indices $i$'s' do not depict different occurrences of the $x$ variable: the
$x_i$'s and $y_i$'s are distinct variables.
These families of variables will enable us to formalize
the Jacobian of $f$ with respect to $(x_i, y_i)$ evaluated at $(x_{i'}, x_{i'})$,
defined as $(\dv*{f (x_{i'}, y_{i'})}{x_i}, \dv*{f (x_{i'}, y_{i'})}{y_i})$,
where $i' \in \mathcal{I}, i' \geq i$.
To lighten the notations, we overload the symbol $\nabla$ and introduce the shorthands
$\nabla_x^i[f]_{i'} \coloneqq \dv*{f (x_{i'}, y_{i'})}{x_i}$ and
$\nabla_y^i[f]_{i'} \coloneqq \dv*{f (x_{i'}, y_{i'})}{y_i}$.
By analogy, the shorthand $\nabla_{x,y}^i[f]_{i'}$ denotes the pair
$(\nabla_x^i[f]_{i'}, \nabla_y^i[f]_{i'})$.
In this work, the difference between the index of derivation $i$
and the index of evaluation $i'$, $i - i' \leq 0$ will be referred to
as \emph{gap}.
We use $\lVert \cdot \rVert _F$ to denote the Frobenius norm, which
a) is naturally defined over rectangular matrices in $\mathbb{R}^{m \times n}$
and b) is \textit{sub-multiplicative}:
$\lVert UV \rVert _F \leq \lVert U \rVert _F \, \lVert V \rVert _F$,
for $U$ and $V$ rectangular with compatible sizes
(provable via Cauchy-Schwarz inequality).
In proofs, we use ``$\otimes$'' for matrix multiplication,
to avoid collisions with the scalar product.

\begin{lemma}[recursive inequality --- induction step]
\label{lemma}
Let the MDP with which the agent interacts be deterministic,
with the dynamics of the environment determined
by the function $f: \mathcal{S} \times \mathcal{A} \rightarrow \mathcal{S}$.
The agent follows a deterministic policy $\mu: \mathcal{S} \rightarrow \mathcal{A}$
to map states to actions,
and receives rewards from
$r_\varphi: \mathcal{S} \times \mathcal{A} \rightarrow \mathbb{R}$
upon interaction.
The functions $f$, $\mu$ and $r_\varphi$ need be $C^0$ and differentiable
over their respective input spaces.
This property is satisfied by the usual neural network function approximators.
The ``almost-everywhere'' case can be derived from this lemma without major changes
(relevant when at least one activation function is only differentiable almost-everywhere, ReLU).
\textbf{\emph{(a)}} Under the previous assumptions,
for $k \in [0, T-t-1] \cap \mathbb{N}$ the following \textbf{recursive inequality} is verified:
\begin{align}
\lVert \nabla_{s,a}^t[r_\varphi]_{t+k+1} \rVert ^2_F
&\leq
C_t
\, \lVert \nabla_{s,a}^{t+1}[r_\varphi]_{t+k+1} \rVert ^2_F
\end{align}
where $C_t \coloneqq A_t^2 \max(1, B_{t+1}^2)$,
$A_t$ and $B_t$ being defined as the supremum norms associated with the Jacobians of $f$
and $\mu$ respectively, with values in $\mathbb{R} \cup \{+\infty\}$:
\begin{align}
\forall t \in [0, T] \cap \mathbb{N} \text{,} \quad
\begin{cases}
A_t \coloneqq \lVert\nabla_{s,a}^t[f]_t\rVert _\infty
= \sup \big\{\lVert\nabla_{s,a}^t[f]_t\rVert _F \, : \, (s_t, a_t) \in
\mathcal{S} \times \mathcal{A} \big\} \\
B_t \coloneqq \lVert\nabla_s^t[\mu]_t\rVert _\infty
= \sup \big\{\lVert\nabla_s^t[\mu]_t\rVert _F \, : \, s_t \in
\mathcal{S} \big\}
\end{cases}
\end{align}
\textbf{\emph{(b)}} Additionally, by introducing \textbf{time-independent} upper bounds
$A, B \in \mathbb{R} \cup \{+\infty\}$
such that $\forall t \in [0, T] \cap \mathbb{N}$,
$A_t \leq A$ and $B_t \leq B$, the recursive inequality becomes:
\begin{align}
\lVert \nabla_{s,a}^t[r_\varphi]_{t+k+1} \rVert ^2_F
&\leq
C
\, \lVert \nabla_{s,a}^{t+1}[r_\varphi]_{t+k+1} \rVert ^2_F
\end{align}
where $C \coloneqq A^2 \max(1, B^2)$ is the time-independent counterpart of $C_t$.
\end{lemma}

\emph{Proof of \textsc{Lemma}~\ref{lemma}~\emph{(a)}.}
First, we take the derivative with respect to each variable separately:
\begin{align}
\nabla_s^t[r_\varphi]_{t+k+1}
&= \dv*{r_\varphi (s_{t+k+1}, a_{t+k+1})}{s_t} \\
&= \dv*{r_\varphi \big(f(s_{t+k}, a_{t+k}), \mu(f(s_{t+k}, a_{t+k}))\big)}{s_t} \\
&= \dv{r_\varphi (s_{t+k+1}, a_{t+k+1})}{s_{t+1}}
\otimes \dv{f(s_t, a_t)}{s_t} \\
&\qquad + \dv{r_\varphi (s_{t+k+1}, a_{t+k+1})}{a_{t+1}}
\otimes \dv{\mu(s_{t+1})}{s_{t+1}}
\otimes \dv{f(s_t, a_t)}{s_t} \nonumber \\
&= \nabla_s^{t+1}[r_\varphi]_{t+k+1} \otimes \nabla_s^t[f]_t
+ \nabla_a^{t+1}[r_\varphi]_{t+k+1} \otimes \nabla_s^{t+1}[\mu]_{t+1} \otimes \nabla_s^t[f]_t
\end{align}
\begin{align}
\nabla_a^t[r_\varphi]_{t+k+1}
&= \dv*{r_\varphi (s_{t+k+1}, a_{t+k+1})}{a_t} \\
&= \dv*{r_\varphi \big(f(s_{t+k}, a_{t+k}), \mu(f(s_{t+k}, a_{t+k}))\big)}{a_t} \\
&= \dv{r_\varphi (s_{t+k+1}, a_{t+k+1})}{s_{t+1}}
\otimes \dv{f(s_t, a_t)}{a_t} \\
&\qquad + \dv{r_\varphi (s_{t+k+1}, a_{t+k+1})}{a_{t+1}}
\otimes\dv{\mu(s_{t+1})}{s_{t+1}}
\otimes\dv{f(s_t, a_t)}{a_t} \nonumber \\
&= \nabla_s^{t+1}[r_\varphi]_{t+k+1} \otimes \nabla_a^t[f]_t
+ \nabla_a^{t+1}[r_\varphi]_{t+k+1} \otimes \nabla_s^{t+1}[\mu]_{t+1} \otimes \nabla_a^t[f]_t
\end{align}

By assembling the norm with respect to both input variables, we get:
\begin{align}
\lVert \nabla_{s,a}^t&[r_\varphi]_{t+k+1} \rVert ^2_F \nonumber \\
&= \lVert \nabla_s^t[r_\varphi]_{t+k+1} \rVert ^2_F
+ \lVert \nabla_a^t[r_\varphi]_{t+k+1} \rVert ^2_F \\
&=
\lVert
\nabla_s^{t+1}[r_\varphi]_{t+k+1} \otimes \nabla_s^t[f]_t
+ \nabla_a^{t+1}[r_\varphi]_{t+k+1} \otimes \nabla_s^{t+1}[\mu]_{t+1} \otimes \nabla_s^t[f]_t
\rVert ^2_F \\
&\qquad +
\lVert
\nabla_s^{t+1}[r_\varphi]_{t+k+1} \otimes \nabla_a^t[f]_t
+ \nabla_a^{t+1}[r_\varphi]_{t+k+1} \otimes \nabla_s^{t+1}[\mu]_{t+1} \otimes \nabla_a^t[f]_t
\rVert ^2_F \nonumber \\
%%%%
&\leq
\lVert
\nabla_s^{t+1}[r_\varphi]_{t+k+1} \otimes \nabla_s^t[f]_t
\rVert ^2_F
\qquad
\blacktriangleright\text{\small{\textit{triangular inequality}}} \\
&\qquad +
\lVert
\nabla_a^{t+1}[r_\varphi]_{t+k+1} \otimes \nabla_s^{t+1}[\mu]_{t+1} \otimes \nabla_s^t[f]_t
\rVert ^2_F \nonumber \\
&\qquad +
\lVert
\nabla_s^{t+1}[r_\varphi]_{t+k+1} \otimes \nabla_a^t[f]_t
\rVert ^2_F \nonumber \\
&\qquad +
\lVert
\nabla_a^{t+1}[r_\varphi]_{t+k+1} \otimes \nabla_s^{t+1}[\mu]_{t+1} \otimes \nabla_a^t[f]_t
\rVert ^2_F \nonumber \\
%%%%
&\leq
\lVert\nabla_s^{t+1}[r_\varphi]_{t+k+1}\rVert ^2_F
\,
\lVert\nabla_s^t[f]_t\rVert ^2_F
\qquad
\blacktriangleright\text{\small{\textit{sub-multiplicativity}}} \\
&\qquad +
\lVert\nabla_a^{t+1}[r_\varphi]_{t+k+1}\rVert ^2_F
\,
\lVert\nabla_s^{t+1}[\mu]_{t+1}\rVert ^2_F
\,
\lVert\nabla_s^t[f]_t\rVert ^2_F \nonumber \\
&\qquad +
\lVert\nabla_s^{t+1}[r_\varphi]_{t+k+1}\rVert ^2_F
\,
\lVert\nabla_a^t[f]_t\rVert ^2_F \nonumber \\
&\qquad +
\lVert\nabla_a^{t+1}[r_\varphi]_{t+k+1}\rVert ^2_F
\,
\lVert\nabla_s^{t+1}[\mu]_{t+1}\rVert ^2_F
\,
\lVert\nabla_a^t[f]_t\rVert ^2_F \nonumber \\
%%%%
&=
\lVert\nabla_s^{t+1}[r_\varphi]_{t+k+1}\rVert ^2_F
\,
\big(
\lVert\nabla_s^t[f]_t\rVert ^2_F +
\lVert\nabla_a^t[f]_t\rVert ^2_F
\big)
\qquad
\blacktriangleright\text{\small{\textit{factorization}}} \\
&\qquad +
\lVert\nabla_a^{t+1}[r_\varphi]_{t+k+1}\rVert ^2_F
\,
\lVert\nabla_s^{t+1}[\mu]_{t+1}\rVert ^2_F
\,
\big(
\lVert\nabla_s^t[f]_t\rVert ^2_F +
\lVert\nabla_a^t[f]_t\rVert ^2_F
\big) \nonumber \\
%%%%
&=
\lVert\nabla_s^{t+1}[r_\varphi]_{t+k+1}\rVert ^2_F
\,
\lVert\nabla_{s,a}^t[f]_t\rVert ^2_F
\qquad
\blacktriangleright\text{\small{\textit{total norm}}} \\
&\qquad +
\lVert\nabla_a^{t+1}[r_\varphi]_{t+k+1}\rVert ^2_F
\,
\lVert\nabla_s^{t+1}[\mu]_{t+1}\rVert ^2_F
\,
\lVert\nabla_{s,a}^t[f]_t\rVert ^2_F \nonumber
\end{align}

Let $A_t$, $B_t$ and $C_t$ be time-dependent quantities defined as:
\begin{align}
\forall t \in [0, T] \cap \mathbb{N} \text{,} \quad
\begin{cases}
A_t \coloneqq \lVert\nabla_{s,a}^t[f]_t\rVert _\infty
= \sup \big\{\lVert\nabla_{s,a}^t[f]_t\rVert _F \, : \, (s_t, a_t) \in
\mathcal{S} \times \mathcal{A} \big\} \\
B_t \coloneqq \lVert\nabla_s^t[\mu]_t\rVert _\infty
= \sup \big\{\lVert\nabla_s^t[\mu]_t\rVert _F \, : \, s_t \in
\mathcal{S} \big\} \\
C_t \coloneqq A_t^2 \max(1, B_{t+1}^2)
\end{cases}
\label{aandb}
\end{align}

Finally, by substitution, we obtain:
\begin{align}
\lVert \nabla_{s,a}^t[r_\varphi]_{t+k+1} \rVert ^2_F
&\leq
A_t^2 \lVert\nabla_s^{t+1}[r_\varphi]_{t+k+1}\rVert ^2_F +
A_t^2 B_{t+1}^2 \lVert\nabla_a^{t+1}[r_\varphi]_{t+k+1}\rVert ^2_F \\
&\leq
A_t^2 \max(1, B_{t+1}^2) \big(\lVert\nabla_s^{t+1}[r_\varphi]_{t+k+1}\rVert ^2_F +
\lVert\nabla_a^{t+1}[r_\varphi]_{t+k+1}\rVert ^2_F\big) \\
&=
A_t^2 \max(1, B_{t+1}^2) \, \lVert \nabla_{s,a}^{t+1}[r_\varphi]_{t+k+1} \rVert ^2_F
\qquad
\blacktriangleright\text{\small{\textit{total norm}}}
\label{lasteqlemma} \\
&=
C_t \, \lVert \nabla_{s,a}^{t+1}[r_\varphi]_{t+k+1} \rVert ^2_F
\qquad
\blacktriangleright\text{\small{$C_t$ \textit{definition}}}
\end{align}
which concludes the proof of \textsc{Lemma}~\ref{lemma} \emph{(a)}. \qed

\emph{Proof of \textsc{Lemma}~\ref{lemma}~\emph{(b)}.}
By introducing time-independent upper bounds $A$ and $B$ such that
$A_t \leq A$ and $B_t \leq B$
$\, \forall t \in [0, T] \cap \mathbb{N}$,
as well as $C \coloneqq A^2 \max(1, B^2)$,
we obtain, by substitution in \textsc{eq}~\ref{lasteqlemma}:
\begin{align}
\lVert \nabla_{s,a}^t[r_\varphi]_{t+k+1} \rVert ^2_F
&\leq
A^2 \max(1, B^2) \, \lVert \nabla_{s,a}^{t+1}[r_\varphi]_{t+k+1} \rVert ^2_F \\
&=
C \, \lVert \nabla_{s,a}^{t+1}[r_\varphi]_{t+k+1} \rVert ^2_F
\end{align}
which concludes the proof of \textsc{Lemma}~\ref{lemma} \emph{(b)}. \qed

\textsc{Lemma}~\ref{lemma} tells us how the norm of the Jacobian associated with a gap
between derivation and evaluation indices equal to $t+1$ relate to the
norm of the Jacobian associated with a gap equal to $t$.
We will use this recursive property to prove our first theorem, \textsc{Theorem}~\ref{theorem1}.
Additionally, from this point forward, we will use the time-independent upper-bounds exclusively,
\textit{i.e.} \textsc{Lemma}~\ref{lemma} \textit{(b)}.

\begin{theorem}[gap-dependent reward Lipschitzness]
\label{theorem1}
In addition to the assumptions laid out in lemma~\ref{lemma},
we assume that the function $r_\varphi$ is $\delta$-Lipschitz
over $\mathcal{S} \times \mathcal{A}$.
Since $r_\varphi$ is $C^0$ and differentiable over $\mathcal{S} \times \mathcal{A}$,
this assumption can be written as
$\lVert \nabla_{s,a}^u[r_\varphi]_u \rVert _F \leq \delta$,
where $u \in [0, T] \cap \mathbb{N}$.
\textbf{\emph{(a)}} Then, under these assumptions, the following is verified:
\begin{align}
\lVert \nabla_{s,a}^t[r_\varphi]_{t+k} \rVert ^2_F
&\leq
\delta ^2 \, \prod_{u=0}^{k-1} C_{t+u}
\end{align}
where $k \in [0, T] \cap \mathbb{N}$ and $C_v$ is defined as
in \textsc{Lemma}~\ref{lemma}~\textit{(a)}, $\forall v \in [0, T] \cap \mathbb{N}$.
\textbf{\emph{(b)}} Additionally, by involving the time-independent upper bounds
introduced in \textsc{Lemma}~\ref{lemma}~\textit{(b)}, we have the following:
\begin{align}
\lVert \nabla_{s,a}^t[r_\varphi]_{t+k} \rVert ^2_F
&\leq
C^k \, \delta ^2
\end{align}
where $k \in [0, T] \cap \mathbb{N}$ and $C$ is defined as
in \textsc{Lemma}~\ref{lemma} \textit{(b)}.
\end{theorem}

\emph{Proof of \textsc{Theorem}~\ref{theorem1}~\emph{(a)}.}
We will prove \textsc{Theorem}~\ref{theorem1}~\emph{(a)} by induction.

Let us introduce the dummy variable $v$,
along with the induction hypothesis for $v$:
\begin{align}
\lVert \nabla_{s,a}^t[r_\varphi]_{t+v} \rVert ^2_F
&\leq
\delta ^2 \, \prod_{u=0}^{v-1} C_{t+u}
\qquad
\blacktriangleright\text{\small{\textit{induction hypothesis}}}
\label{indhyp1}
\end{align}
where $v$ represents the gap between the derivation timestep and the evaluation timestep.

\emph{Step 1: initialization.} When the gap $v=0$, \textsc{eq}~\ref{indhyp1} becomes
$\lVert \nabla_{s,a}^t[r_\varphi]_t \rVert ^2_F \leq \delta ^2$,
$\forall t \in [0, T] \cap \mathbb{N}$,
which is trivially verified since it exactly corresponds to \textsc{Theorem}~\ref{theorem1}'s
main assumption.

\emph{Step 2: induction.} Let us assume that \textsc{eq}~\ref{indhyp1} is verified for $v$ fixed,
and show that \textsc{eq}~\ref{indhyp1} is satisfied when the gap is equal to $v+1$.
\begin{align}
\lVert \nabla_{s,a}^t[r_\varphi]_{t+v+1} \rVert ^2_F
&\leq
C_t \, \lVert \nabla_{s,a}^{t+1}[r_\varphi]_{t+v+1} \rVert ^2_F
\qquad
\blacktriangleright\text{\small{\textit{\textsc{Lemma}~\ref{lemma} (a)}}} \\
&\leq
C_t \, \delta^2 \, \prod_{u=0}^{v-1} C_{t+1+u}
\qquad
\blacktriangleright\text{\small{\textit{\textsc{eq}~\ref{indhyp1} since gap is $v$, at $t+1$}}} \\
&=
C_t \, \delta^2 \, \prod_{u=1}^{v} C_{t+u}
\qquad
\blacktriangleright\text{\small{\textit{index shift}}} \\
&=
\delta^2 \, \prod_{u=0}^{v} C_{t+u}
\qquad
\blacktriangleright\text{\small{\textit{repack product}}}
\end{align}
\textsc{eq}~\ref{indhyp1} is therefore satisfied for $v+1$ when assumed at $v$,
which proves the induction step.

\emph{Step 3: conclusion.} Since \textsc{eq}~\ref{indhyp1} has been verified for both the
initialization and induction steps,
the hypothesis is valid
$\forall v \in [0, T] \cap \mathbb{N}$, which concludes the proof of
\textsc{Theorem}~\ref{theorem1}~\emph{(a)}. \qed

\emph{Proof of \textsc{Theorem}~\ref{theorem1}~\emph{(b)}.}
We will prove \textsc{Theorem}~\ref{theorem1}~\emph{(b)} by induction.

Let us introduce the dummy variable $v$,
along with the induction hypothesis for $v$:
\begin{align}
\lVert \nabla_{s,a}^t[r_\varphi]_{t+v} \rVert ^2_F
&\leq
C^v \, \delta ^2
\qquad
\blacktriangleright\text{\small{\textit{induction hypothesis}}}
\label{indhyp2}
\end{align}
where $v$ represents the gap between the derivation timestep and the evaluation timestep.

\emph{Step 1: initialization.} When the gap $v=0$, \textsc{eq}~\ref{indhyp2} becomes
$\lVert \nabla_{s,a}^t[r_\varphi]_t \rVert ^2_F \leq \delta ^2$,
$\forall t \in [0, T] \cap \mathbb{N}$,
which is trivially verified since it exactly corresponds to \textsc{Theorem}~\ref{theorem1}'s
main assumption.

\emph{Step 2: induction.} Let us assume that \textsc{eq}~\ref{indhyp2} is verified for $v$ fixed,
and show that \textsc{eq}~\ref{indhyp2} is satisfied when the gap is equal to $v+1$.
\begin{align}
\lVert \nabla_{s,a}^t[r_\varphi]_{t+v+1} \rVert ^2_F
&\leq
C \, \lVert \nabla_{s,a}^{t+1}[r_\varphi]_{t+v+1} \rVert ^2_F
\qquad
\blacktriangleright\text{\small{\textit{\textsc{Lemma}~\ref{lemma} (b)}}} \\
&\leq
C \, C^v \, \delta ^2
\qquad
\blacktriangleright\text{\small{\textit{\textsc{eq}~\ref{indhyp2} since gap is $v$}}} \\
&=
C^{v+1} \, \delta ^2
\end{align}
\textsc{eq}~\ref{indhyp2} is therefore satisfied for $v+1$ when assumed at $v$,
which proves the induction step.

\emph{Step 3: conclusion.} Since \textsc{eq}~\ref{indhyp2} has been verified for both the
initialization and induction steps,
the hypothesis is valid
$\forall v \in [0, T] \cap \mathbb{N}$, which concludes the proof of
\textsc{Theorem}~\ref{theorem1}~\emph{(b)}. \qed

This result shows that when there is a gap $k$ between the derivation and evaluation indices,
the norm of the Jacobian of $r_\varphi$ is upper-bounded by a \emph{gap-dependent}
quantity equal to $\sqrt{C^k} \delta$, over the entire input space.
Crucially, this property applies if and only if the gap between the
timestep of the derivation variable and the timestep of the evaluation variable is
equal to $0$, hence the use of the same letter $u$ in the assumption formulation.

\begin{theorem}[state-action value Lipschitzness]
\label{theorem2}
We work under the assumptions laid out in
both \textsc{Lemma}~\ref{lemma} and \textsc{Theorem}~\ref{theorem1}, and repeat
the main lines here
for \textsc{Theorem}~\ref{theorem2} to be self-contained:
a) The functions $f$, $\mu$ and $r_\varphi$ are $C^0$ and differentiable
over their respective input spaces,
and b) the function $r_\varphi$ is $\delta$-Lipschitz
over $\mathcal{S} \times \mathcal{A}$, i.e.
$\lVert \nabla_{s,a}^u[r_\varphi]_u \rVert _F \leq \delta$,
where $u \in [0, T] \cap \mathbb{N}$.
Then the quantity $\nabla_{s,a}^u[Q_\varphi]_u$ exists
$\forall u \in [0, T] \cap \mathbb{N}$,
and verifies:
\begin{align}
\lVert \nabla_{s,a}^t[Q_\varphi]_t \rVert _F
\leq
\left\{
\begin{aligned}
&\delta \, \sqrt{\frac{1 - \big( \gamma^2 C \big)^{T - t}}{1 - \gamma^2 C}},
&\qquad  &\text{if $\gamma^2 C \neq 1$} \\
&\delta\sqrt{T - t},
&\qquad &\text{if $\gamma^2 C = 1$}
\end{aligned}
\right.
\end{align}
$\forall t \in [0, T] \cap \mathbb{N}$,
where
$C \coloneqq A^2 \max(1, B^2)$, with $A$ and $B$ time-independent upper bounds of
$\lVert\nabla_{s,a}^t[f]_t\rVert _\infty$ and $\lVert\nabla_s^t[\mu]_t\rVert _\infty$
respectively
(see \textsc{eq}~\ref{aandb} for definitions of the supremum norms).
\end{theorem}

\emph{Proof of \textsc{Theorem}~\ref{theorem2}.}
With finite horizon $T$, we have
$Q_\varphi (s_t, a_t) \coloneqq \sum_{k=0}^{T-t-1} \gamma^k \, r_\varphi (s_{t+k}, a_{t+k})$,
$\forall t \in [0, T] \cap \mathbb{N}$,
since $f$, $\mu$, and $r_\varphi$ are all deterministic (no expectation).
Additionally, since $r_\varphi$ is assumes to be $C^0$ and differentiable
over $\mathcal{S} \times \mathcal{A}$,
$Q_\varphi$ is by construction also $C^0$ and differentiable over $\mathcal{S} \times \mathcal{A}$.
Consequently, $\nabla_{s,a}^u[Q_\varphi]_u$ exists, $\forall u \in [0, T] \cap \mathbb{N}$.
Since both $r_\varphi$ and $Q_\varphi$ are scalar-valued (their output space is $\mathbb{R}$),
their Jacobians are the same as their gradients.
We can therefore use the linearity of the gradient operator:
$\nabla_{s,a}^t[Q_\varphi]_t = \sum_{k=0}^{T-t-1} \gamma^k \, \nabla_{s,a}^t[r_\varphi]_{t+k}$,
$\forall t \in [0, T] \cap \mathbb{N}$.
\begin{align}
\lVert \nabla_{s,a}^t[Q_\varphi]_t \rVert _F^2
&= \Bigg\lVert \sum_{k=0}^{T-t-1} \gamma^k \, \nabla_{s,a}^t[r_\varphi]_{t+k} \Bigg\rVert _F^2
\qquad
\blacktriangleright\text{\small{\textit{operator's linearity}}} \\
&\leq \sum_{k=0}^{T-t-1} \gamma^{2k} \, \lVert \nabla_{s,a}^t[r_\varphi]_{t+k} \rVert _F^2
\qquad
\blacktriangleright\text{\small{\textit{triangular inequality}}} \\
&\leq \sum_{k=0}^{T-t-1} \gamma^{2k} \, C^k\, \delta ^2
\qquad
\blacktriangleright\text{\small{\textit{\textsc{Theorem}~\ref{theorem1}}}} \\
&= \delta^2 \sum_{k=0}^{T-t-1} \big( \gamma^2 C \big)^k
\label{geomsum}
\end{align}
When $\gamma^2 C = 1$, we obtain $\lVert \nabla_{s,a}^t[Q_\varphi]_t \rVert _F^2
= \delta^2 (T - t)$.
On the other hand, when $\gamma^2 C \neq 1$:
\begin{align}
\lVert \nabla_{s,a}^t[Q_\varphi]_t \rVert _F^2
&\leq \delta^2 \, \frac{1 - \big( \gamma^2 C \big)^{T - t}}{1 - \gamma^2 C}
\qquad
\blacktriangleright\text{\small{\textit{finite sum of geometric series}}}
\end{align}
\begin{align}
\implies \quad
\lVert \nabla_{s,a}^t[Q_\varphi]_t \rVert _F^2
\leq
\left\{
\begin{aligned}
& \delta^2 \, \frac{1 - \big( \gamma^2 C \big)^{T - t}}{1 - \gamma^2 C},
&\qquad  &\text{if $\gamma^2 C \neq 1$} \\
& \delta^2 (T - t),
&\qquad &\text{if $\gamma^2 C = 1$}
\end{aligned}
\right.
\end{align}
By applying $\sqrt{\cdot}$ (monotonically increasing) to the inequality,
we obtain the claimed result. \qed

Finally, we derive a corollary from \textsc{Theorem}~\ref{theorem2}
corresponding to the infinite-horizon regime.

\begin{corollary}[infinite-horizon regime]
\label{corollary1}
Under the assumptions of \textsc{Theorem}~\ref{theorem2},
including that $r_\varphi$ is $\delta$-Lipschitz over $\mathcal{S} \times \mathcal{A}$,
and assuming that $\gamma^2 C < 1$, we have, in the infinite-horizon regime:
\begin{align}
\lVert \nabla_{s,a}^t[Q_\varphi]_t \rVert _F
&\leq \frac{\delta}{\sqrt{1 - \gamma^2 C}}
\end{align}
which translates into $Q_\varphi$ being $\frac{\delta}{\sqrt{1 - \gamma^2 C}}$-Lipschitz
over $\mathcal{S} \times \mathcal{A}$.
\end{corollary}

\emph{Proof of \textsc{Corollary}~\ref{corollary1}.}
We now have
$Q_\varphi (s_t, a_t) \coloneqq \sum_{k=0}^{+\infty} \gamma^k \, r_\varphi (s_{t+k}, a_{t+k})$,
$\forall t \in [0, T] \cap \mathbb{N}$,
since $f$, $\mu$, and $r_\varphi$ are all deterministic
and are now working working under the infinite-horizon regime.
Considering the changes in $Q_\varphi$'s definition, the first part of the proof
can be done by analogy with the proof of \textsc{Theorem}~\ref{theorem2},
until \textsc{eq}~\ref{geomsum}, which is our starting point.
In this regime, $\gamma^2 C \geq 1$ yields an infinite sum in \textsc{eq}~\ref{geomsum},
which results in an uninformative (because infinite) upper-bound on
$\lVert \nabla_{s,a}^t[Q_\varphi]_t \rVert _F$.
On the other hand, when $\gamma^2 C < 1$
(note, we always have $\gamma^2 C \geq 0$ by definition),
the infinite sum in \textsc{eq}~\ref{geomsum} is defined.
Since we have shown that $\gamma^2 C < 1$ is the only setting in which the sum is defined,
we continue from the infinite-horizon version of \textsc{eq}~\ref{geomsum}
with $\gamma^2 C < 1$ onwards. Hence,
\begin{align}
\lVert \nabla_{s,a}^t[Q_\varphi]_t \rVert _F^2
\leq \delta^2 \sum_{k=0}^{+\infty} \big( \gamma^2 C \big)^k
= \frac{\delta^2}{1 - \gamma^2 C}
\qquad
\blacktriangleright\text{\small{\textit{infinite sum of geometric series}}}
\end{align}
Using $\sqrt{\cdot}$ (monotonically increasing) on both sides
concludes the proof of \textsc{Corollary}~\ref{corollary1}. \qed

To conclude the section, we now give interpretations of the derived theoretical results,
discuss the implications of our results, and also exhibit to what extent they
transfer to the practical setting.

\subsection{Discussion I: implications and limitations of the theoretical guarantees}
\label{discussion}

\subsubsection{Function approximation bias}

\textsc{Theorem}~\ref{theorem2} exhibits the Lipschitz constant of $Q_\varphi$
when $r_\varphi$ is $\delta$-Lipschitz.
In practice however, the state-action value (or value function) is usually modeled
by a neural network, and learned via gradient descent
either by using a Monte-Carlo estimate of the collected return as regression target,
or by bootstrapping using a subsequent model estimate \cite{Sutton1988-to}.
We therefore have access to a learned estimate $Q_\omega$, as opposed to the real
state-action value $Q_\varphi$.
As such, the results derived in \textsc{Theorem}~\ref{theorem2} will transfer favorably into
the function approximation setting as $Q_\omega$ becomes
a better parametric estimate of $Q_\varphi$.
Note, the reward is denoted by $r_\varphi$ for the reader to easily distinguish it
from the \emph{black-box} reward traditionally returned by the environment.
Albeit arbitrary, the notation $r_\varphi$ allows for the reward to be modeled by
a neural network parameterized by the weights $\varphi$, and learned via gradient descent,
as is indeed the case in this work.
Crucially, having control over $r_\varphi$ in practice allows for the enforcement of constraints,
making the $\delta$-Lipschitzness assumption in \textsc{Theorem}~\ref{theorem1},
\textsc{Theorem}~\ref{theorem2} and \textsc{Corollary}~\ref{corollary1}
practically satisfiable via gradient penalization~\ref{gradpen}.
It is crucial to note that, while function approximation creates a gap between
theory and practice for the $Q$-value (\emph{worse} when bootstrapping),
there is a meaningfully lesser gap for the reward as
the $\delta$-Lipschitzness constraint is directly enforced on the parametric reward $r_\varphi$.

\subsubsection{Value Lipschitzness}

In \textsc{Corollary}~\ref{corollary1} we showed that
$\lVert \nabla_{s,a}^t[Q_\varphi]_t \rVert _F \leq \delta / \sqrt{1 - \gamma^2 C}$,
in the infinite-horizon regime,
when $r_\varphi$ is assumed $\delta$-Lipschitz over $\mathcal{S} \times \mathcal{A}$,
and assuming $\gamma^2 C < 1$.
In other words, in this setting,
enforcing $r_\varphi$ to be $\delta$-Lipschitz
causes $Q_\varphi$ to be $\Delta_\infty$-Lipschitz,
where $\Delta_\infty \coloneqq \delta / \sqrt{1 - \gamma^2 C}$,
$C \coloneqq A^2 \max(1, B^2)$,
and $A$, $B$ are upper-bounds of
$\lVert\nabla_{s,a}^t[f]_t\rVert _\infty$, $\lVert\nabla_s^t[\mu]_t\rVert _\infty$.
Starting from the assumption that $\gamma^2 C < 1$,
we trivially arrive at $\sqrt{1 - \gamma^2 C} < 1$, then $1 / \sqrt{1 - \gamma^2 C} > 1$,
and since $\delta \geq 0$ by definition (\textit{cf.} \textsc{Section}~\ref{gradpen}),
we finally get $\Delta_\infty > \delta$.
Without loss of generality,
consider the case in which $r_\varphi$ is \emph{not} a contraction, \textit{i.e.} $r_\varphi$
is $\delta$-Lipschitz $C^0$ over $\mathcal{S} \times \mathcal{A}$, with $\delta \geq 1$.
As a result, $\Delta_\infty > \delta \geq 1$, \textit{i.e.} $\Delta_\infty > 1$,
which means that, under the considered conditions,
$Q_\varphi$ is \emph{not} a contraction over $\mathcal{S} \times \mathcal{A}$ either.
The latter naturally extends to any $u \in \mathbb{R}_+$ that lower-bounds $\delta$:
if $\delta > u$, then $\Delta_\infty > u$, $\forall u \in \mathbb{R}_+$.
Lipschitz functions and especially contractions
are at the core of many fundamental
results in dynamics programming, hence also in reinforcement learning.
Crucially, the Bellman operator being a contraction
causes a fixed point iterative process, such as value iteration \cite{Sutton1998-ow},
to converge to a unique fixed point whatever the starting iterate of $Q$.
Since we learn $Q_\varphi$ with temporal-difference learning \cite{Sutton1988-to} via
a bootstrapped objective, the convergence of our method is a direct consequence
of the contractant nature of the Bellman operator.
As such the Lipschitzness-centric analysis laid out in this section is complementary to the latter.
It provides a characterization of $Q_\varphi$'s Lipschitzness
over the input space $\mathcal{S} \times \mathcal{A}$
as opposed to over iterates, \textit{i.e.} time.
As such, our analysis therefore does not give convergence guarantees of an iterative process,
which are already carried over from temporal-difference learning at the core of our algorithm.
Rather, we provide \emph{variation upper-bounds} for $Q_\varphi$
when $r_\varphi$ has upper-bounded variations:
if $r_\varphi$ is $\delta$-Lipschitz,
then $Q_\varphi$ is $\Delta_\infty$-Lipschitz.
\textit{In fine}, this result has an immediate corollary, derived previously in this block:
if the variations of $r_\varphi$ are lower-bounded by $\delta$,
then the variations of $Q_\varphi$ are lower-bounded by $\Delta_\infty>\delta$.

\subsubsection{Compounding variations}
\label{compoundvars}

The relative position of $\gamma^2 C$ with respect to $1$ is instrumental in the
behavior of the exhibited variation bounds, in both the finite- and infinite-horizon settings.
In the latter, we see that the upper-bound gets to infinity when $\gamma^2 C$
(non-negative by definition, and lower than $1$ as necessary condition
for the infinite sum to exist) gets closer to $1$ from below.
In the former, we focus on the $\gamma^2 C \neq 1$ case, as in the other case, the
bound does not even depend on $\gamma^2 C$.
As such, we study the value of $\lVert \nabla_{s,a}^t[Q_\varphi]_t \rVert _F$'s upper-bound
in the finite-horizon setting when $\gamma^2 C \neq 1$,
dubbed $\Delta_t \coloneqq \delta\sqrt{1 - (\gamma^2 C)^{T-t} / 1 - \gamma^2 C}$.
Beforehand, we would remind the reader how the bounded quantity should behave
throughout an episode.
Since $Q_\varphi$ is defined as the expected sum of \emph{future} rewards $r_\varphi$,
predicting such value should get increasingly tainted with uncertainty as it tries to
predict across long time ranges.
As such, predicting $Q_\varphi$ at time $t=0$ is the most challenging,
as it corresponds to the value of an entire trajectory,
whereas predicting $Q_\varphi$ at time $t=T$ is the easiest
(equal to last reward $r_\varphi$).
Higher horizons $T$ consequently make the prediction task more difficult,
as do discount factors $\gamma$ closer to $1$.
We now discuss $\Delta_t$.
As long as $\gamma^2 C \neq 1$, $\Delta_t$ gets to $0$ as $t$ gets to $T$.
This is consistent with the previous reminder:
as $t$ gets to $T$, the $Q_\varphi$ estimation task becomes easier,
hence the variation bound ($\Delta_t$) due to prediction uncertainty should decrease to $0$.
As $t$ gets to $0$ however, the behavior of $\Delta_t$ depends on the value of $\gamma^2 C$:
if $\gamma^2 C \gg 1$, $\Delta_t$ explodes to infinity, whereas
for reasonable values of $\gamma^2 C$, $\Delta_t$ does not.
Since $C \coloneqq A^2 \max(1, B^2)$, $\gamma^2 C \gg 1$ translates to
$((\exists u > 1): A \gg u) \lor ((\exists v > 1): B \gg v)$.
Let us assume that $A$ ($B$) not only upper-bounds every $A_t$ ($B_t$) but is also the tightest
time-independent bound: $A \coloneqq A_{t'}$ ($B \coloneqq B_{t''}$)
where $t' = \argmax_t{A_t}$ ($t'' = \argmax_t{B_t}$).
We then have
$((\exists u > 1)(\exists t'): A_{t'} \gg u)
\lor ((\exists v > 1)(\exists t''): B_{t''} \gg v)$,
\textit{i.e.}
$((\exists u > 1)(\exists t'): \lVert\nabla_{s,a}^{t'}[f]_{t'}\rVert _\infty \gg u)
\lor ((\exists v > 1)(\exists t''): \lVert\nabla_s^{t''}[\mu]_{t''}\rVert _\infty \gg v)$
over $\mathcal{S} \times \mathcal{A}$.
Note, the ``or'' is inclusive.
In other words, if the variations (in space) of the policy or the dynamics are large
in the early stage of an episode ($0 \leq t \ll T$),
then $\Delta_t$ (variation bound on $Q_\varphi$) explodes.
The exhibited phenomenon is somewhat reminiscent of the compounding of
errors isolated in \cite{Ross2010-eb}.

\subsubsection{Is value Lipschitzness enough?}
\label{notenough}

We showed that under mild conditions, and in finite- and infinite- horizon regimes,
$r_\varphi$ Lipschitzness implies $Q_\varphi$ Lipschitzness, \textit{i.e.}
that if similar state-action are mapped to similar rewards by $r_\varphi$,
then $Q_\varphi$ also maps then to similar state-action values.
This regularization desideratum is evocative of the
\emph{target policy smoothing} add-on introduced in \cite{Fujimoto2018-pe},
already presented earlier in \textsc{Section}~\ref{bridge}.
In short, target policy smoothing perturbs the target action slightly.
In effect, the temporal-difference optimization now fits the value estimate against an expectation
of \emph{similar} bootstrapped target value estimates.
Forcing similar action to have similar values naturally smooths out the value estimate,
which by definition emulates the enforcement of a Lipschitzness constraint on the value,
and as such mitigates value overfitting which
deterministic policies are prone to.
While its smoothing effect on the value function is somewhat intuitive,
we set out to investigate formally how target policy smoothing
affects the optimization dynamics, and particularly to what extent it smooths out
the state-action value landscape.
Since the function approximator $Q_\omega$ is optimized as a supervised learning problem
using the traditional squared loss criterion, we first study how perturbing the inputs with
additive random noise, denoted by $\xi$, impacts the optimized criterion,
and what kind of behavior it encourages in the predictive function.
As such, to lighten the expressions, we consider the supervised criterion
$C(x) \coloneqq (y - f(x))^2$,
where $f(x)$ is the predicted vector
at the input vector $x$, and $y$ is the supervised target vector.
We also consider, in line with \cite{Fujimoto2018-pe}, that the noise is
sampled from a spherical zero-centered Gaussian distribution,
omitting here that the noise is truncated for legibility,
hence $\xi \sim \mathcal{N}(0, \sigma^2 I)$.
The criterion injected with input noise is
$C_\xi (x) \coloneqq C (x + \xi) = (y - f(x + \xi))^2$.
Assuming the noise has small amplitude (further supporting the original truncation),
we can write the second-order Taylor series expansion of the perturbed criterion near $\xi=0$,
as a polynomial of $\xi$:
\begin{align}
C_\xi (x) = C(x)
+ \sum_i \pdv{C}{x_i} \bigg\rvert_{x} \xi_i
+ \frac{1}{2} \sum_i\sum_j \pdv{C}{x_i}{x_j} \bigg\rvert_{x} \xi_i \xi_j
+ \mathcal{O}(\lVert \xi \rVert ^3)
\end{align}
where $\lVert \cdot \rVert$ denotes the euclidean norm in the appropriate vector space.
From this point forward, we assume the noise has a small enough norm to allow the
third term, $\mathcal{O}(\lVert \xi \rVert ^3)$, to be neglected.
By integrating over the noise distribution, we obtain:
\begin{align}
\int C_\xi (x) p(\xi)d\xi = C(x)
+ \sum_i \pdv{C}{x_i} \bigg\rvert_{x} \int \xi_i p(\xi)d\xi
+ \frac{1}{2} \sum_i\sum_j \pdv{C}{x_i}{x_j} \bigg\rvert_{x} \int \xi_i \xi_j p(\xi)d\xi
\label{intnoise}
\end{align}
Since the noise is sampled from
the zero-centered and spherical distribution $\mathcal{N}(0, \sigma^2I)$,
we have respectively that
$\int \xi_i p(\xi)d\xi = 0$ and
\[\int \xi_i \xi_j p(\xi)d\xi
= \int \xi_i^2 \delta_{ij} p(\xi)d\xi
= \delta_{ij} \int \xi_i^2  p(\xi)d\xi
= \delta_{ij} \sigma^2\],
where $\delta_{ij}$ is the Kronecker symbol.
By injecting these expressions in \textsc{eq}~\ref{intnoise}, we get:
\begin{align}
\int C_\xi (x) p(\xi)d\xi
= C(x)
+ \frac{\sigma^2}{2} \sum_i \pdv[2]{C}{x_i} \bigg\rvert_{x}
= C(x)
+ \frac{\sigma^2}{2} \operatorname{Tr}(H_x \, C)
\end{align}
where $\operatorname{Tr}(H_x \, C)$ is the trace of the Hessian of the criterion $C$,
\textit{w.r.t.} the input variable $x$.
We now want to express the exhibited regularizer $\operatorname{Tr}(H_x \, C))$
as a function of the derivatives of the prediction function $f$,
and therefore calculate the consecutive derivative sums:
\begin{align}
\sum_i \pdv{C}{x_i} \bigg\rvert_{x}
&= -2 \sum_i \big ( y - f(x) \big ) \pdv{f}{x_i} \bigg\rvert_{x} \\
\sum_i \pdv[2]{C}{x_i} \bigg\rvert_{x}
&= 2 \sum_i \Bigg [ \bigg ( \pdv{f}{x_i} \bigg\rvert_{x} \bigg ) ^2
- \big ( y - f(x) \big ) \pdv[2]{f}{x_i} \bigg\rvert_{x} \Bigg ]
\end{align}
hence,
\begin{align}
\int C_\xi (x) p(\xi)d\xi
&= C(x)
+ \sigma^2
\sum_i \Bigg [ \bigg ( \pdv{f}{x_i} \bigg\rvert_{x} \bigg ) ^2
- \big ( y - f(x) \big ) \pdv[2]{f}{x_i} \bigg\rvert_{x} \Bigg ]
\end{align}
\textit{In fine}, we can write, in a more condensed form:
\begin{align}
\mathbb{E}_\xi [C(x + \xi)]
= C(x) + \sigma^2 \Big [ \lVert \nabla_x \, f \rVert^2
- \operatorname{Tr}\big(C(x) H_x \, f \big) \Big ]
\end{align}
The previous derivations
--- derived somewhat similarly in \cite{Webb1994-ar} and \cite{Bishop1995-ea} ---
show that minimizing the criterion with
noise injected in the input is equivalent to minimizing the criterion
without any noise \emph{and} a regularizer containing norms of both
the Jacobian and Hessian of the prediction function $f$.
As raised in \cite{Bishop1995-ea},
the second term of the regularizer is unsuitable for the design of
a practically viable learning algorithm, since
\textit{a)} it involves prohibitively costly second-order derivatives,
and \textit{b)} it is not positive definite, and consequently not lower-bounded,
which overall makes the regularizer a bad candidate for an optimization problem loss.
Nevertheless, \cite{Bishop1995-ea} further shows that this regularization is equivalent
to the use of a standard Tikhonov-like positive-definite regularization scheme
involving \emph{only} first-order derivatives,
provided the noise has small amplitude --- ensured here with a small $\sigma$
and noise clipping.
As such, the regularizer induced by the input noise $\xi$ is equivalent to
$\sigma^2 \big [ \lVert \nabla_x \, f \rVert^2 \big ]$,
and by direct analogy, we can say that
target policy smoothing induces an implicit regularizer on the TD objective,
of the form $\sigma^2 \big [ \lVert \nabla_a \, Q_{\omega'} \rVert^2 \big ]$,
Note, $\omega'$ are the target critic parameters, given that
target policy smoothing adds noise to the target action,
an input of target critic value $Q_{\omega'}$.
By construction,
the target parameters $\omega'$ slowly follow the online parameters $\omega$
(\textit{cf.} \textsc{Section}~\ref{bridge}).
In addition, temporal-difference learning urges $Q_\omega$ to move closer to
$Q_{\omega'}$ by design
(\textit{cf.} \textsc{eq}~\ref{omegaloss}).
Consequently, properties enforced on one set of parameters
should \emph{eventually} be transfered to the other,
such that \textit{in fine} both $\omega$ and $\omega'$ possess the given property
only explicitly enforced on one (albeit delayed).
Based on this line of reasoning,
the temporal-difference learning dynamics and soft target updates should
make the theoretically equivalent $\sigma^2 \big [ \lVert \nabla_a \, Q_{\omega'} \rVert^2 \big ]$
regularizer enforce smoothness on the online parameters $\omega$ too,
even if it explicitly only constrains the target weights $\omega'$.
All in all, we have shown that target smoothing is equivalent to
adding a regularizer to the temporal-difference error to minimize when learning $Q_\omega$,
where said regularizer is reminiscent of the gradient penalty regularizer,
presented earlier in \textsc{eq}~\ref{eqgp}.
As such, target smoothing \emph{does} implement a gradient penalty regularization,
but on $Q_\omega$.
Crucially, the gradient in the penalty is only taken
\textit{w.r.t.} the action dimension, but not \textit{w.r.t.} the state dimension.
In spite of the use of target policy smoothing in our method,
it was not enough to yield stable learning behaviors,
as shown in \textsc{Section}~\ref{empres1}.
Gradient penalization was an absolute necessity.
Even though both methods encourage $Q_\omega$ to
be smoother (directly in \cite{Fujimoto2018-pe},
and indirectly via reward Lipschitzness in this work),
on its own, learning a smooth $Q_\omega$ estimate seems not to be \emph{sufficient}
for our method to work:
learning a smooth $r_\varphi$ estimate to serve as basis for $Q_\omega$
seems to be a \emph{necessary} condition.

\subsubsection{Indirect reward regularization}
\label{indirectreg}

The theoretical guarantees we have derived
(\textit{cf.}~\textsc{Theorem}~\ref{theorem1},
\textsc{Theorem}~\ref{theorem2} and \textsc{Corollary}~\ref{corollary1})
all build on the premise that the reward $r_\varphi$
is $\delta$-Lipschitz over the joint input space $\mathcal{S} \times \mathcal{A}$,
\textit{i.e.} that $\lVert \nabla_{s,a}^t[r_\varphi]_t \rVert _F \leq \delta$.
Crucially, we do \emph{not} enforce this regularity property \emph{directly} is practice, but instead
urge the discriminator $D_\varphi$ to be $k$-Lipschitz by restricting the norm of the Jacobian of the latter
via regularization (\textit{cf.}~\textsc{eq}~\ref{varphilossgp}).
We here set out to figure out to what extent the $k$-Lipschitzness enforced onto $D_\varphi$
propagates and transfers to $r_\varphi$; in particular, whether it results in
the \emph{indicrectly}-urged $\delta$-Lipschitzness of $r_\varphi$,
with $\delta \neq k$ outside of edge cases.
While $k$ is fixed throughout the lifetime of the agent, $\delta$ need not be.
As such, discussing the behavior of this evolving Lipschitz constant \textit{w.r.t.} the learning dynamics
is crucial to better understand \emph{when} the guarantees we have just derived
(whose main premise is $\lVert \nabla_{s,a}^t[r_\varphi]_t \rVert _F \leq \delta$) apply in practice.
As laid out ealier in \textsc{Section}~\ref{bridge}, in this work, we consider two forms of reward,
crafted purely from the scores returned by $D_\varphi$:
the minimax (saturating) one $r_\varphi^\textsc{mm} \coloneqq -\log(1-D_\varphi)$
and the non-saturating one $r_\varphi^\textsc{ns} \coloneqq \log(D_\varphi)$
(names purposely chosen to echo their counterpart GAN generator loss).
Although we opted for the minimax form (based on the ablation study we carried out on the matter,
\textit{cf.}~\textsc{Appendix}~\ref{ablationreward}), we here tackle and discuss both forms,
as we suspect there could be more to it than just zero-order numerics.
Analyzing first-order behavior is the crux of most GAN design breakthroughs, which is far from surprising, considering
how intertwined the inner networks are (generator $G$, and discriminator $D$). Yet, in adversarial IL,
the policy (playing the role of $G$) does not receive gradients flowing back from $D$ like in GANs. Instead, it gets
a reward signal crafted from $D$'s returned scalar value, detached from the computational graph,
and try to maximize it over time via \emph{policy-}gradient optimization.
The discussion in adversarial IL has thus always limited to the numerics of the reward signal and
how to shape it in a way that faciliates the resolution of the task at hand (similarly to how we discuss
the impact of its shape when reporting our last empirical findings of \textsc{Section}~\ref{empres1}).

By constrast, we here are interested in the gradients of these rewards ($r_{\varphi, \textsc{mm}}$
and $r_{\varphi, \textsc{ns}}$)
in this studied adversarial IL context,
with the end-goal of characterizing their Lipschitz-continuity (or absence thereof).
Their respective Jacobians' norms, under the setting laid out earlier in \textsc{Section}~\ref{theory}, are
$\lVert \nabla_{s,a}^t[r_\varphi^\textsc{mm}]_t \rVert _F
= \lVert \nabla_{s,a}^t[D_\varphi]_t \rVert _F \, \big/ \, (1 - D_\varphi(s_t,a_t))$ and
$\lVert \nabla_{s,a}^t[r_\varphi^\textsc{ns}]_t \rVert _F
= \lVert \nabla_{s,a}^t[D_\varphi]_t \rVert _F \, \big/ \, D_\varphi(s_t,a_t)$, with $D_\varphi(s_t,a_t) \in (0,1)$
($D_\varphi$'s score is wrapped with a sigmoid).
As laid out above, we here posit that $D_\varphi$ is $k$-Lipschitz-continuous as founding assumption ---
$\lVert \nabla_{s,a}^t[D_\varphi]_t \rVert _F \leq k$. We can now upper-bound the Jacobians' norms
unpacked above with the Lipschitz constant of $D_\varphi$:
$\lVert \nabla_{s,a}^t[r_\varphi^\textsc{mm}]_t \rVert _F
\leq k \, \big/ \, (1 - D_\varphi(s_t,a_t))$ and
$\lVert \nabla_{s,a}^t[r_\varphi^\textsc{ns}]_t \rVert _F
\leq k \, \big/ \, D_\varphi(s_t,a_t)$.
Since $D_\varphi(s_t,a_t) \in (0,1)$, both denominators (for either reward form) are in $(0,1)$,
which makes the Jacobian's norm of either reward form unbounded
over its domain (due to $D_\varphi \to 0$ from above for $r_\varphi^\textsc{ns}$;
due to $D_\varphi \to 1$ from below for $r_\varphi^\textsc{mm}$),
despite the $D_\varphi$'s $k$-Lipschitzness.
Since treating the entire range of values that \emph{can} be taken by $D_\varphi(s_t,a_t)$, $(0,1)$,
lead us to a dead end, and leaving us unable to upper-bound neither
$\lVert \nabla_{s,a}^t[r_\varphi^\textsc{mm}]_t \rVert _F$ nor
$\lVert \nabla_{s,a}^t[r_\varphi^\textsc{ns}]_t \rVert _F$,
we now adopt a more granular approach and procede by dichotomy.
As such, $\exists \, \ell \in (0,1)$ verifying $0 < \ell \ll 1$
such that $1 \, \big/ \, D_\varphi(s_t,a_t)$
\big(and as a result also $\lVert \nabla_{s,a}^t[r_\varphi^\textsc{ns}]_t \rVert _F
\leq k \, \big/ \, D_\varphi(s_t,a_t)$\big)
is unbounded when $D_\varphi(s_t,a_t) \in (0,\ell]$ and bounded when $D_\varphi(s_t,a_t) \in (\ell,1)$.
Similarly, $\exists \, L \in (0,1)$ verifying $0 \ll L < 1$
such that $1 \, \big/ \, (1 - D_\varphi(s_t,a_t))$
\big(and as a result also $\lVert \nabla_{s,a}^t[r_\varphi^\textsc{mm}]_t \rVert _F
\leq k \, \big/ \, (1 - D_\varphi(s_t,a_t))$\big)
is bounded when $D_\varphi(s_t,a_t) \in (0,L]$ and unbounded when $D_\varphi(s_t,a_t) \in (L,1)$.
If we were to figure out the \emph{effective} range covered by $D_\varphi$'s values throughout the learning process,
we would maybe be able to exploit the dichotomy.

In practice, the untrained agent initially performs poorly at the imitation task,
and is therefore assigned low scores by $D_\varphi$ (near $0$,
as ``$0$'' is the label assigned to samples from the agent in the classification update $D_\varphi$ goes through
every iteration).
As learning progresses, the agent's scores gradually shift towards $1$
--- the label used for expert samples in $D_\varphi$'s update,
and \emph{optimally} converge to the central value of $0.5$ in the $(0,1)$ range that $D_\varphi$ can describe.
Indeed, the \emph{perfect} discriminator consistently predicts scores equal to $0.5$ for the agent's actions
\cite{Goodfellow2017-pv}: the agent has managed to perfectly confuse $D_\varphi$ as to where the data it is fed
comes from (both sources, expert and agent, are perceived as equiprobable).
What matters for $\lVert \nabla_{s,a}^t[r_\varphi]_t \rVert _F$ (either form) to be bounded \emph{in practice} is
for it to be bounded for values of $D_\varphi$ in $(0,M]$, where $0.5 \leq M < 1$
(the values \emph{realistically} taken by $D_\varphi$ throughout the learning process).
Since $M < L$ in effect (for $L$, \textit{cf.} dichotomy above),
we can conclude that $\lVert \nabla_{s,a}^t[r_\varphi^\textsc{mm}]_t \rVert _F$
is effectively bounded: $\exists \, \delta$, $0 \leq \delta < +\infty$, such that
$\lVert \nabla_{s,a}^t[r_\varphi^\textsc{mm}]_t \rVert _F \leq \delta$.
We however can not conclude as such for $\lVert \nabla_{s,a}^t[r_\varphi^\textsc{ns}]_t \rVert _F$,
however close to zero $\ell$ might be (for $\ell$, \textit{cf.} dichotomy above).
It is not rare for $D_\varphi$ to take $0$ as value early in training,
which makes $\lVert \nabla_{s,a}^t[r_\varphi^\textsc{ns}]_t \rVert _F$ unbounded
in the interval described by the values taken by D in practice: $(0, M]$.
Interestingly, when $D_\varphi$ is near $0$ early in training,
$\lVert \nabla_{s,a}^t[r_\varphi^\textsc{mm}]_t \rVert _F
\leq k \, \big/ \, (1 - D_\varphi(s_t,a_t)) \approx k$.
The lowest upper-bound for $\lVert \nabla_{s,a}^t[r_\varphi^\textsc{mm}]_t \rVert _F$
is $\delta \approx k$, and can only happen early in the training process, when $D_\varphi$ correctly
classifies the agent's actions as coming from the agent.
In other words, the Lipschitz constant of $r_\varphi^\textsc{mm}$ is at its lowest early in training.
Besides, as the agent becomes more proficient at mimicking the expert and therefore collects higher scores
from $D_\varphi$, $\delta$ increases monotonically and grows aways from its initial value $k$.
Compared to the alternative (highest Lipschitz constant early in training and then monotonically decreasing
as the scores increase when the agent gets better at the task, nearing the lowest value of $k$ when
$D_\varphi \to 1$),
which as it turns out is exactly the behavior adopted by $r_\varphi^\textsc{ns}$,
the behavior of $r_\varphi^\textsc{mm}$ is far more desirable.

Crucially, to sum up, $r_\varphi^\textsc{ns}$ is not Lipschitz early in training when the agent would benefit most
from regularity in the reward landscape. $r_\varphi^\textsc{mm}$ however \emph{is} Lipschitz-continuous
early in training, with the lowest Lipschitz constant of its lifetime, which aligns with the Lipschitz constant
enforced on $D_\varphi$ ($\delta \approx k$).
As such, $r_\varphi^\textsc{mm}$ is at its most regular when the agent needs it most (early, when it knows nothing),
and then becomes less and less restrictive (the Lipschitz constant $\delta$ increases) as the
agent collects higher similarity scores with the expert from $D_\varphi$.
One could therefore see $r_\varphi^\textsc{mm}$ as having built-in \textit{``training wheels''},
which gradually phase out as the agent becomes better, providing less safety as the agent becomes more
proficient at the imitation task.
To conclude this discussion point, with the minimax reward form $r_\varphi \coloneqq r_\varphi^\textsc{mm}$,
we have $\lVert \nabla_{s,a}^t[D_\varphi]_t \rVert _F \leq k
\implies \lVert \nabla_{s,a}^t[r_\varphi]_t \rVert _F \leq \delta$ in practice.
This means that the premise of our theoretical guarantees consisting in positing that
the reward is $\delta$-Lipschitz-continuous \emph{can} be satisfied in practice by enforcing
$k$-Lipschitz-continuity on $D_\varphi$ via gradient penalty regularization (\textit{cf.}~\textsc{eq}~\ref{eqgp}).
This is \emph{not} the case when $r_\varphi \coloneqq r_\varphi^\textsc{ns}$.
We propose this analytical observation as an explanation as to why using $r_\varphi^\textsc{ns}$
yields such poor results in our reported ablation, \textit{cf.}~\textsc{Appendix}~\ref{ablationreward}.
Our discussion detaches itself from the one adopting a zero-order numerics scope, laid out in
\textsc{Section}~\ref{empres1}, by discussing first-order numerics instead, which blends into our Lipschitzness
narrative.

\subsubsection{Local smoothness}

The local Lipschitzness assumption is reminiscent of many theoretical results
in the study of robustness to adversarial examples.
Notably, \cite{Yang2020-bz} shows that local Lipschitzness is correlated with
empirical robustness and accuracy in various benchmark datasets.
As mentioned when we justified the \emph{local} nature of the Lipschitz-continuity notion tackled in this work
(\textit{cf.}~\textsc{Definition}~\ref{lipdef}),
we optimize the different modules over mini-batches of samples.
While forcing the constraint to be satisfied globally might be feasible in some
low-dimensional supervised or unsupervised learning problems,
the notion of fixed dataset does not exist \textit{a priori} in reinforcement learning.
\textsc{Section}~\ref{gradpenrl} describes, compares and discusses
the effect of \emph{where} the local Lipschitzness constraint is enforced
(\textit{e.g.} expert demonstration manifold, fictitious replay experiences).
Wherever the regularizer is applied,
the constraint is local nonetheless.
One can therefore not guarantee that the $\delta$-Lipschitz-continuity of $r_\varphi$,
formalized as $\lVert \nabla_{s,a}^t[r_\varphi]_t \rVert _F \leq \delta$,
and urged by enforcing
$\lVert \nabla_{s,a}^t[D_\varphi]_t \rVert _F \leq k$ via gradient penalization
(\textit{cf.} our previous discussion on indirect reward regularization in \textsc{Section}~\ref{indirectreg}),
will be satisfied \emph{everywhere}
in $\mathcal{S} \times \mathcal{A}$.
Plus, considering that \textsc{Theorem}~\ref{theorem2} and \textsc{Corollary}~\ref{corollary1}
rely on the satisfaction of the constraint on $r_\varphi$ along every trajectory,
which is likely not to be verified in practice,
we can say with high confidence that the constraint on $Q_\varphi$,
$\lVert \nabla_{s,a}^t[Q_\varphi]_t \rVert _F \leq \Delta_\infty$,
will not be satisfied over the whole joint input space either.
Still, we can hope to enhance the coverage of the subspace on which the constraint
$\lVert \nabla_{s,a}^t[r_\varphi]_t \rVert _F \leq \delta$ is satisfied,
dubbed $\mathfrak{C}$, by doing more $r_\varphi$ learning updates with the regularizer
--- technically, $D_\varphi$ learning updates encouraging $D_\varphi$ to satisfy
$\lVert \nabla_{s,a}^t[D_\varphi]_t \rVert _F \leq k$ via gradient penalization,
\textit{cf.}~\textsc{eq}~\ref{eqgp}.
From this point onward,
we will qualify a state-action pair $(s_t, a_t)$
--- equivalently, an action $a_t$ in a given state $s_t$ ---
as \textit{``$\mathfrak{C}$-valid''} if it belongs to $\mathfrak{C} \ni (s_t, a_t)$,
\textit{i.e.}~if $r_\varphi$ is $\delta$-Lipschitz, verifying
$\lVert \nabla_{s,a}^t[r_\varphi]_t \rVert _F \leq \delta$.
Note, the notion of $\mathfrak{C}$-validity is inherently local,
since we have defined the notion for a single given input pair $(s_t, a_t)$.
As such, future statements about $\mathfrak{C}$-validity will all be local ones by essence.
In addition,
despite having $\lVert \nabla_{s,a}^t[D_\varphi]_t \rVert _F \leq k
\implies \lVert \nabla_{s,a}^t[r_\varphi]_t \rVert _F \leq \delta$
\emph{in practice} for the minimax reward form
(\textit{cf.} our previous discussion on indirect reward regularization in \textsc{Section}~\ref{indirectreg}),
there is not an exact equivalence between $r_\varphi$ being $\delta$-Lipschitz
and $D_\varphi$ being $k$-Lipschitz in theory.
Therefore, we will qualify a state-action pair $(s_t, a_t)$
--- equivalently, an action $a_t$ in a given state $s_t$ ---
as \textit{``approximately $\mathfrak{C}$-valid''} if
$D_\varphi$ is $k$-Lipschitz, verifying
$\lVert \nabla_{s,a}^t[D_\varphi]_t \rVert _F \leq k$.
As it has been made clear by now, $D_\varphi$'s $k$-Lipschitzness is encouraged by
plugging a gradient penalty regularizer $\mathfrak{R}_\varphi^\zeta (k)$ into $D_\varphi$'s loss
(\textit{cf.}~\textsc{eq}~\ref{eqgp}).
Despite being encouraged, $\lVert \nabla_{s,a}^t[D_\varphi]_t \rVert _F \leq k$ can nonetheless not be guaranteed
solely from the application of the regularizer at $(s_t, a_t)$.
As such, to cover all bases,
we will qualify a state-action pair $(s_t, a_t)$
--- equivalently, an action $a_t$ in a given state $s_t$ ---
as \textit{``probably approximately $\mathfrak{C}$-valid''} if
$(s_t, a_t)$ is in the support of the distribution $\zeta$ that determines where the
gradient penalty regularizer $\mathfrak{R}_\varphi^\zeta (k)$ of $\ell_\varphi^\textsc{GP}$ is applied
in $\mathcal{S} \times \mathcal{A}$,
\textit{i.e.}~if $(\operatorname{supp} \; \zeta) \ni (s_t, a_t)$.
A probably approximately $\mathfrak{C}$-valid point is supported by the distribution
that describes where $\lVert \nabla_{s,a}^t[D_\varphi]_t \rVert _F \leq k$ is enforced, and as such,
$\mathfrak{R}_\varphi^\zeta (k)$ may be applied at this point.

Importantly, the policy might, due to its exploratory motivations,
pick an action $a_t$ in state $s_t$ that is not $\mathfrak{C}$-valid.
Depending on where the constraint will then be enforced, the sample might then
be $\mathfrak{C}$-valid after $r_\varphi$'s update
(technically, indirectly via $D_\varphi$'s update; \textit{cf.}~\textsc{Section}~\ref{gradpenrl}).
This observation motivates the investigation we carry out in \textsc{Section}~\ref{understand},
in which we define a soft $\mathfrak{C}$-validity pseudo-indicator of $\mathfrak{C}$
(\textit{cf.}~\textsc{eq}~\ref{pseudoindicator})
that enables us to assess whether the
agent consistently performs approximately $\mathfrak{C}$-valid actions
when it interacts with the MDP $\mathbb{M}^*$ following $\mu_\theta$.

\subsection{A new reinforcement learning perspective on gradient penalty}
\label{gradpenrl}

We begin by considering a few variants of the original gradient penalty regularizer
\cite{Gulrajani2017-mr}
introduced in \textsc{Section}~\ref{gradpen}.
Each variant corresponds to a particular case of the \emph{generalized} version of the
regularizer, described in \textsc{eq}~\ref{eqgp}.
Subsuming all versions, we remind \textsc{eq}~\ref{eqgp} here for didactic purposes:
\begin{align}
\ell_\varphi^\textsc{GP}
\coloneqq \ell_\varphi + \lambda \, \mathfrak{R}_\varphi^\zeta (k)
\coloneqq \ell_\varphi + \lambda \,
\mathbb{E}_{s_t \sim \rho^{\zeta}, a_t \sim \zeta}
[(\lVert  \nabla _{s_t,a_t} \, D_\varphi(s_t,a_t) \rVert - k )^2]
\end{align}
where $\zeta$ is the distribution that describes \emph{where} the regularizer is applied
--- where the Lipschitz-continuity constraint is enforced in the input space
$\mathcal{S} \times \mathcal{A}$.
In \cite{Gulrajani2017-mr}, $\zeta$ corresponds to sampling point uniformly along segments
joining samples generated by the agent following its policy and samples generated by
the expert policy, \textit{i.e.} samples from the expert demonstrations $\mathcal{D}$.
Formally, focusing on the action only for legibility --- the counterpart formalism
for the state is derived easily by using the visitation distribution instead of the policy ---
$a \sim \zeta$ means $a = u \, a' + (1-u) \, a''$, where $a' \sim \pi_\theta$, $a'' \sim \pi_e$,
and $u \sim \operatorname{unif}(0,1)$.
The distribution $\zeta$ we have just described corresponds to the transposition of the
GAN formulation to the GAIL setting, which is an \emph{on}-policy setting.
Therefore, in this work, we amend the $\zeta$ previously described,
and replace it with its \emph{off}-policy counterpart, where
$a' \sim \beta$ (\textit{cf.} \textsc{Section}~\ref{bridge}).
As for the penalty target, \cite{Gulrajani2017-mr} use $k=1$, in line with the theoretical
result derived by the authors.
By contrast, DRAGAN \cite{Kodali2017-xt} use a $\zeta$ such that
$a \sim \zeta$ means $a = a'' + \epsilon$, where $a'' \sim \pi_e$,
and $\epsilon \sim \mathcal{N}(0, 10)$.
Like WGAN-GP \cite{Gulrajani2017-mr}, DRAGAN uses the penalty target $k=1$.
Finally, for the sake of symmetry, we introduce a reversed version of DRAGAN,
dubbed NAGARD (name reversed).
To the best of our knowledge, the method has not been explored in the literature.
NAGARD also uses $k=1$ as penalty target, but perturbs the policy-generated samples
as opposed to the expert ones:
$a \sim \zeta$ means $a = a' + \epsilon$, where $a' \sim \beta$
(\emph{off}-policy setting),
and $\epsilon \sim \mathcal{N}(0, 10)$.
We use $\lambda=10$ in all the variants, in line with the original
hyper-parameter settings in \cite{Gulrajani2017-mr} and \cite{Kodali2017-xt}.

\begin{figure}
\centering
\begin{subfigure}[t]{0.38\textwidth}
\centering
\includegraphics[width=\linewidth]{%
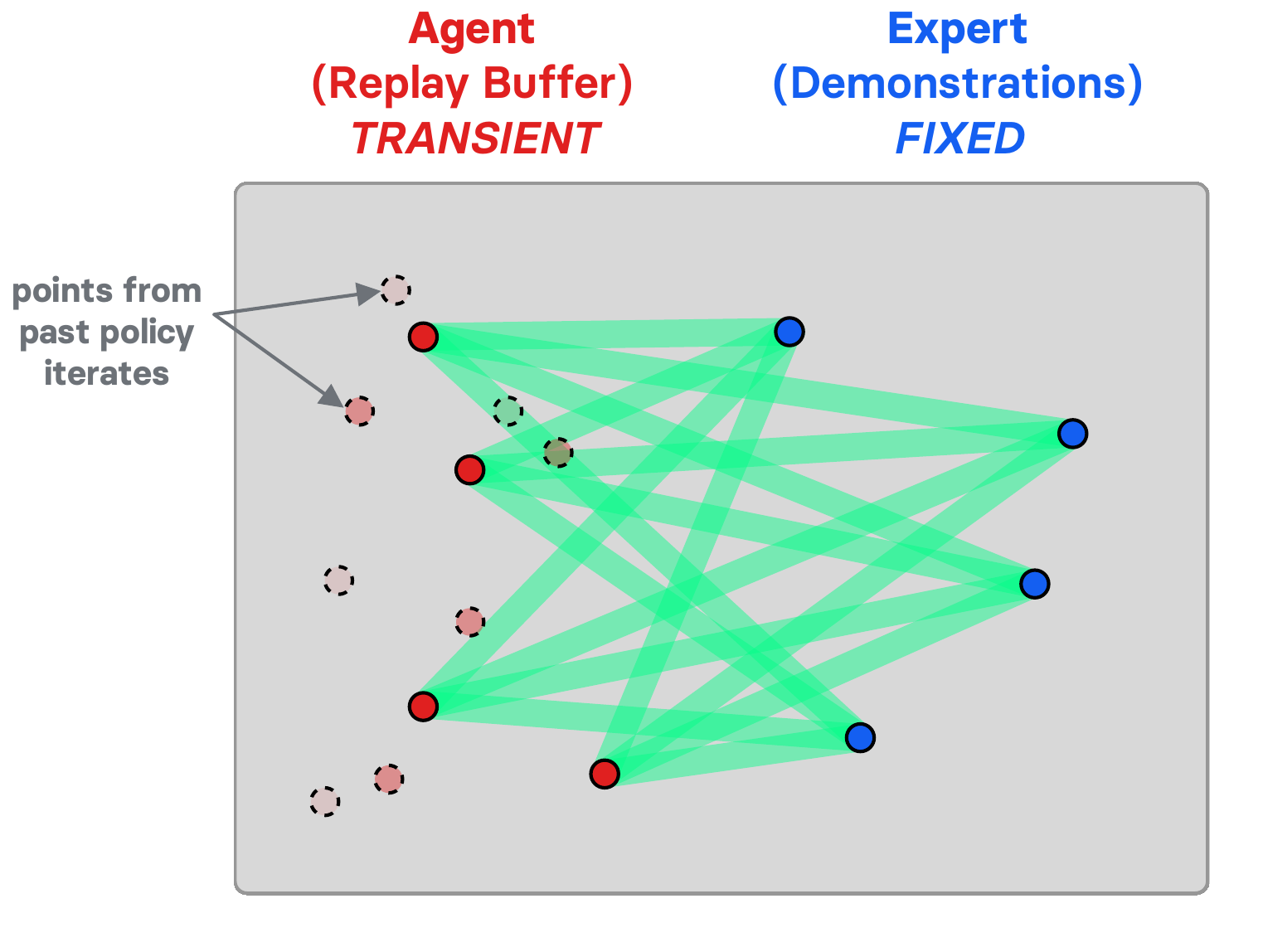}
\caption{WGAN-GP \cite{Gulrajani2017-mr}}
\label{wgangp}
\end{subfigure}
\begin{subfigure}[t]{0.30\textwidth}
\centering
\includegraphics[width=\linewidth]{%
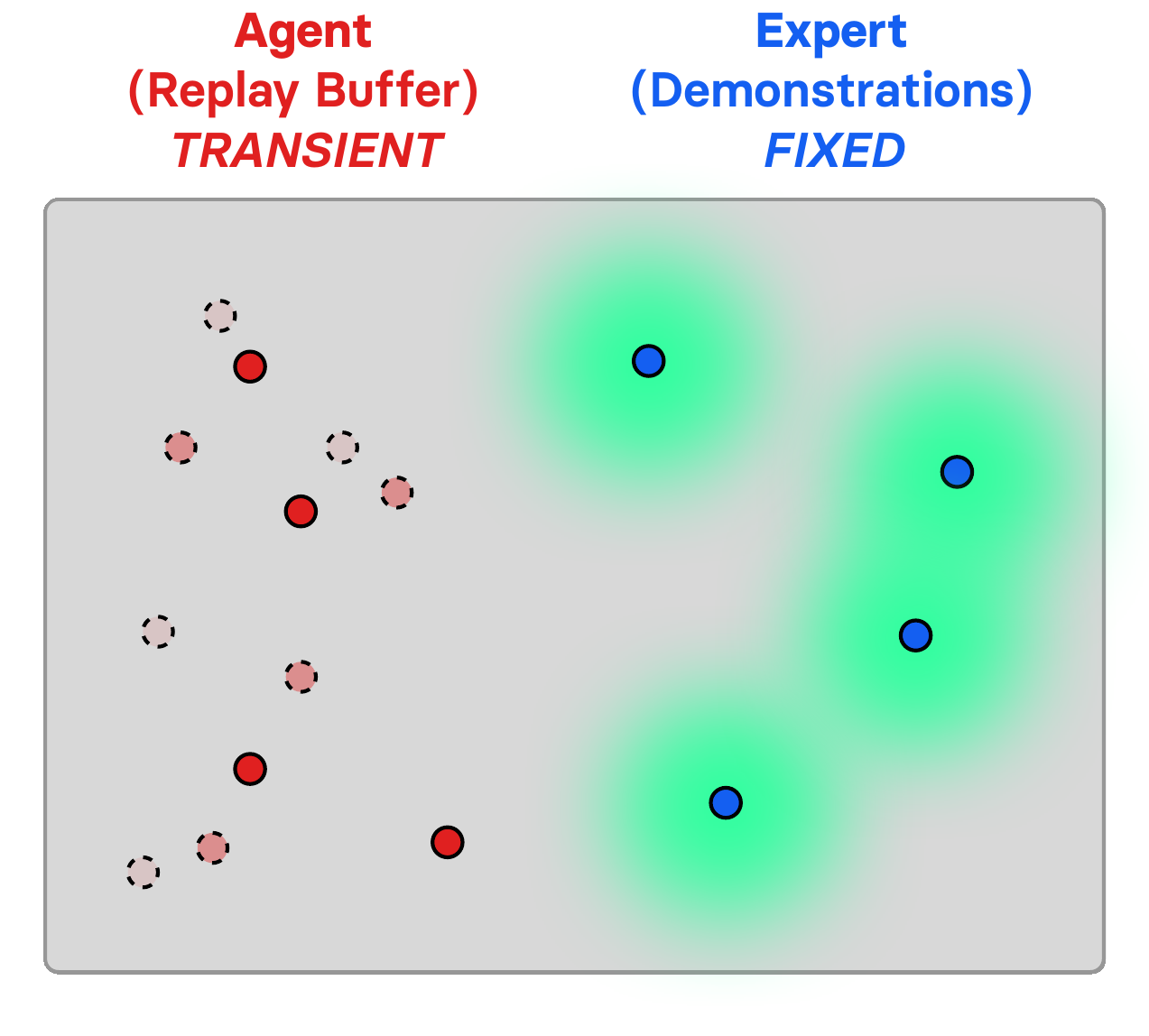}
\caption{DRAGAN \cite{Kodali2017-xt}}
\label{dragan}
\end{subfigure}
\begin{subfigure}[t]{0.30\textwidth}
\centering
\includegraphics[width=\linewidth]{%
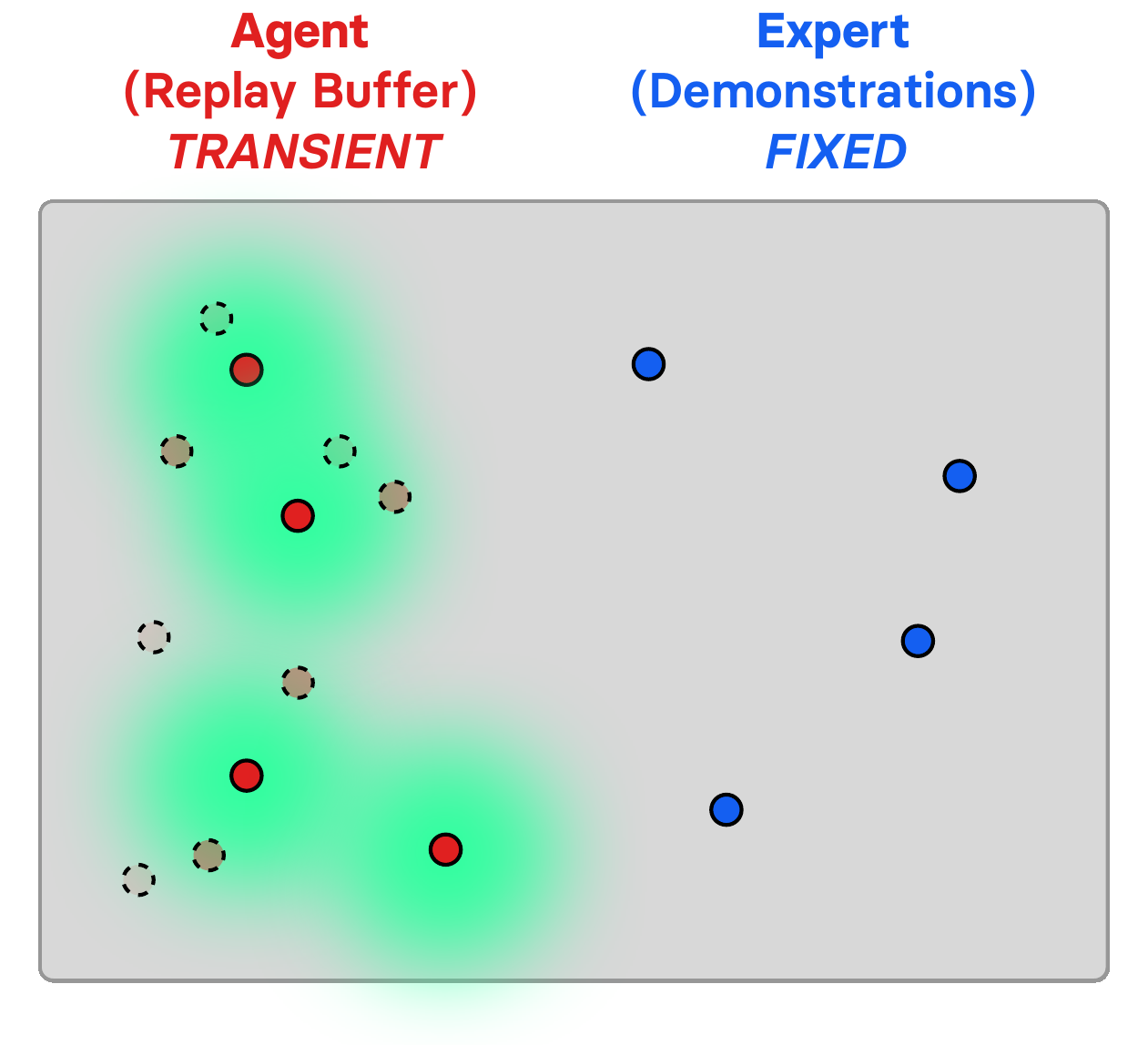}
\caption{NAGARD (illustrative \emph{only})}
\label{nagard}
\end{subfigure}
\caption{Schematic representation (in green) of the support of the $\zeta$ distribution,
depicting \emph{where} the gradient penalty regularizer is enforced, at a given iteration,
and for all iterations throughout the lifetime of the learning agent.
It corresponds to the subspace of $\mathcal{S} \times \mathcal{A}$
on which the Lipschitz-continuity constraint is applied:
where the state-action pairs are \emph{likely} $\mathfrak{C}$-valid.
The intensity of the green color indicates the probability assigned by the distribution
$\zeta$ on the state-action pair. The more opaque the coloration, the higher the probability.
Best seen in color.}
\label{gpvariants}
\end{figure}

\textsc{Figure}~\ref{gpvariants} depicts in green the subspace of the input space
$\mathcal{S} \times \mathcal{A}$ where the $k$-Lipschitz-continuity constraint,
formalized as $\lVert \nabla_{s,a}^t[D_\varphi]_t \rVert _F \leq k$, and enouraged
in $\ell_\varphi^\textsc{GP}$ by $\mathfrak{R}_\varphi^\zeta (k)$,
is applied.
In other words, \textsc{Figure}~\ref{gpvariants} highlights the support of the distribution $\zeta$ for each variant,
which have just been described above.
As such, the green areas in \textsc{Figures}~\ref{dragan},~\ref{nagard}, and~\ref{wgangp}
are schematic depictions of where the state-actions pairs are \emph{probably approximately $\mathfrak{C}$-valid}.

One conceptual difference between the DRAGAN penalty and the two others is that
the support of the distribution $\zeta$ does not change throughout the entire training process
for the former, while is does for the latter.
Borrowing the intuitive terminology used in \cite{Kodali2017-xt},
WGAN-GP proposes a \textit{coupled penalty},
while DRAGAN (like NAGARD) propose a \textit{local} penalty.
In \cite{Kodali2017-xt}, the authors perform a comprehensive empirical study
of mode collapse, and diagnose that the generator collapsing to single modes is
often coupled with the discriminator displaying sharp gradients around the samples from
the real distribution.
In model-free generative adversarial imitation learning,
the generator does not have access to the gradient of the discriminator with respect
to its actions in the backward pass, although it could be somewhat accessed using
a model-based approach \cite{Baram2017-es}.
In spite of not being accessible \textit{per se},
the sharpness of the discriminator's gradients near real samples observed in \cite{Kodali2017-xt}
translates, in the setting considered in this work,
to sharp rewards, which we referred to as reward overfitting and
was discussed thoroughly in \textsc{Section}~\ref{overfitting}.
As such, mode collapse mitigation in the GAN setting translates to
a problem of credit assignment in our setting, caused by the peaked reward landscape
(\textit{cf.} \textsc{Appendix}~\ref{ablationdiscount} to witness the sensitivity
\textit{w.r.t.} the discount factor $\gamma$, controlling how far ahead in the episode
the agent looks).
The stability issues the methods incur in either settings are on par.
Both gradient penalty regularizers aim to address these stability weaknesses,
and do so by enforcing a Lipschitz-continuity constraint,
albeit on a different support $\operatorname{supp} \; \zeta$
(\textit{cf.} \textsc{Figure}~\ref{gpvariants}).

\begin{figure}
  \center
  \begin{subfigure}[t]{0.99\textwidth}
    \center\scalebox{0.18}[0.18]{\includegraphics{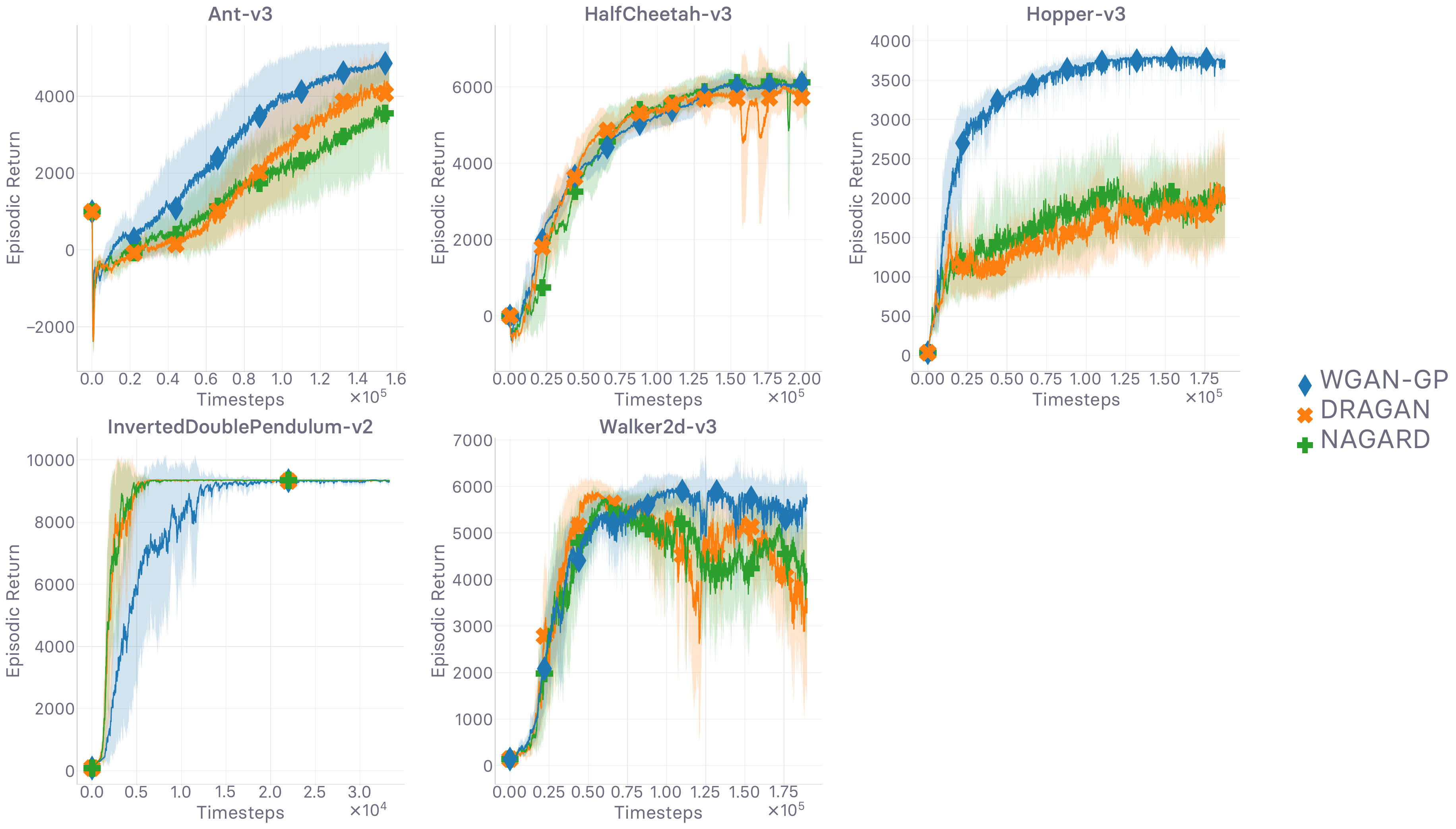}}
    \caption{Evolution of return values \textit{(higher is better)}}
  \end{subfigure}
  \begin{subfigure}[t]{0.99\textwidth}
    \center\scalebox{0.18}[0.18]{\includegraphics{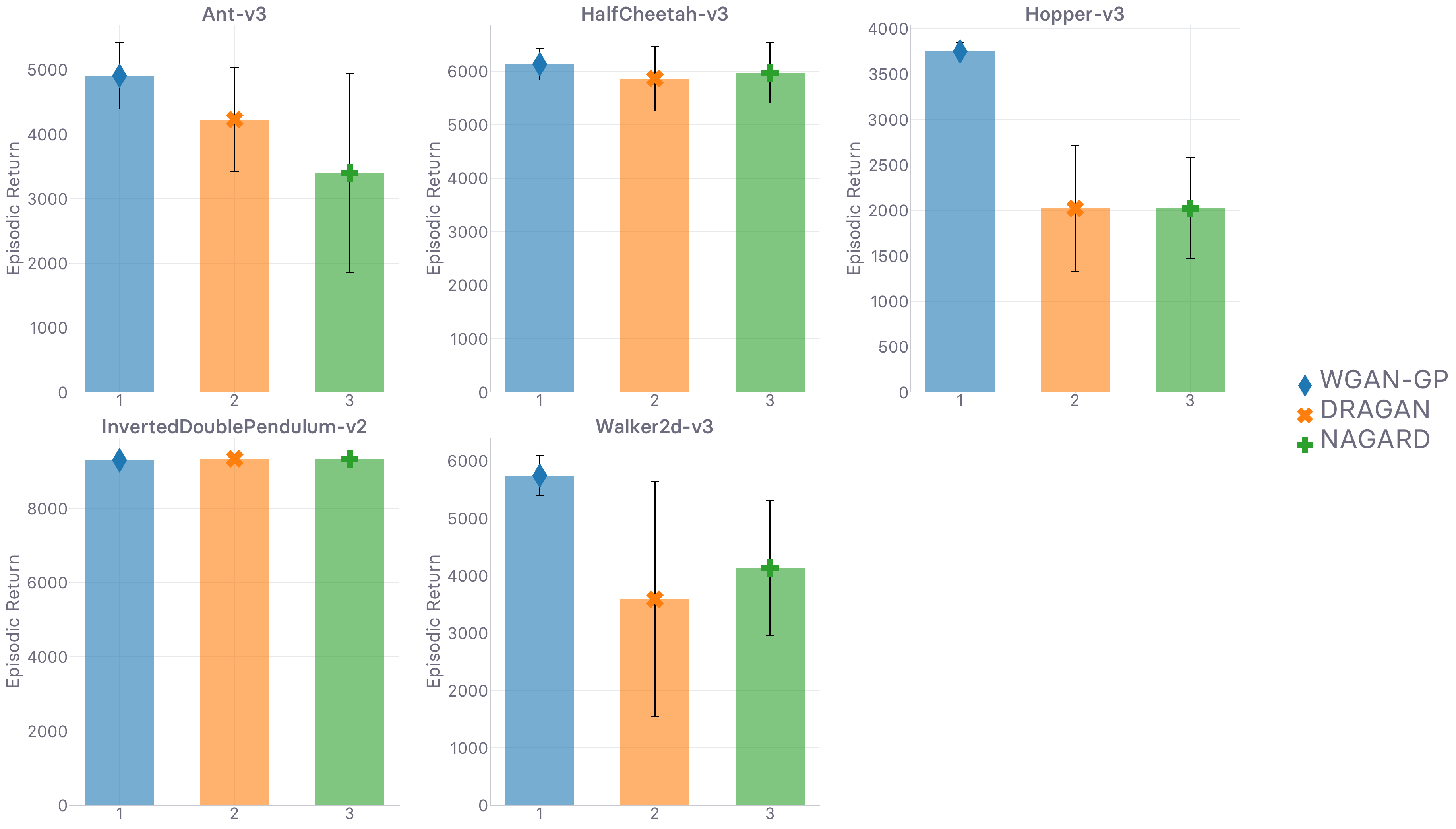}}
    \caption{Final return values at timeout \textit{(higher is better)}}
  \end{subfigure}
  \caption{
  Evaluation of gradient penalty variants.
  Explanation in text.
  Runtime is 48 hours.}
  \label{penvariantsplots}
\end{figure}

As mentioned earlier in \textsc{Section}~\ref{gradpen},
the distribution $\zeta$ used in WGAN-GP \cite{Gulrajani2017-mr}
is motivated by the fact that --- as they show in their work ---
the \emph{optimal} discriminator is $1$-Lipschitz along lines joining
real and fake samples.
The authors of \cite{Kodali2017-xt} deem the assumptions underlying this result
to be unrealistic, which naturally weakens the ensuing method derived from this line
of reasoning.
They instead propose DRAGAN, whose justification is straightforward and unarguable:
since they witness sharp discriminator gradients around real samples,
they introduce a \emph{local} penalty that
aims to smooth out the gradients of the discriminator \emph{around} the real data points.
Formally, as described above when defining the distribution $\zeta$ associated with the approach,
it tries to ensure Lipschitz-continuity of the discriminator in the neighborhoods
(additive Gaussian noise perturbations)
of the real samples.
The generator or policy is more likely to escape the narrow peaks of the optimization
landscape --- corresponding to the real data points ---
with this extra stochasticity.
\textit{In fine}, in our setting, DRAGAN can dial down the sharpness of the reward landscape
at expert samples the discriminator overfits on.
This technique should therefore fully address the shortcomings
raised and discussed in \textsc{Section}~\ref{gradpen}.
While the method seem to yield better results than WGAN-GP
in generative modeling with generative adversarial nets,
the empirical results we report in \textsc{Figure}~\ref{penvariantsplots} show otherwise.
All the considered penalties help close the significant performance gap
reported in \textsc{Figure}~\ref{resplotsgp}, in almost every environment,
but the penalty from WGAN-GP generally pulls ahead.
Additionally, not only does is display higher empirical return,
it also crucially exhibits more stable and less jittery behavior.

Despite the apparent disadvantage of \emph{local} penalties
(DRAGAN \cite{Kodali2017-xt} and NAGARD) compared to WGAN-GP
in terms of their schematically-depicted $\operatorname{supp} \; \zeta$ sizes
(\textit{cf.} \textsc{Figure}~\ref{gpvariants}),
it is important to remember that the additive Gaussian perturbation is
distributed as $\mathcal{N}(0,10)$.
For these local methods, $\zeta$ is therefore covering a
\emph{large} \footnote{Considering the observations are clipped to be in $[-5.0, 5.0]$,
as is customary in the \textsc{MuJoCo} \cite{Todorov2012-gc} benchmark \cite{Brockman2016-un},
an additive Gaussian perturbation with $\sigma^2=10$ can,
in all fairness, be qualified as \emph{large}.}
area around the central sample, including with high probability samples that are,
according to the discriminator, from both categories
--- fake samples (predicted as from $\beta$),
and real samples (predicted as from $\pi_e$).
As such, the perceived diameter of the green disks in the schematic representations in
\textsc{Figures}~\ref{dragan} and~\ref{nagard} maybe smaller than it would be in reality.
It is crucial to consider the coverage of the different $\zeta$ distributions
as they determine how strongly the Lipschitz-continuity property is potentially
enforced at a given state-action pair, for a fixed number of discriminator updates.
Consequently, for a given optimization step,
while the \emph{local} penalties are --- somewhat ironically ---
applying the Lipschitz-continuity constraint on data points \emph{scattered} around the
agent- (NAGARD) or expert-generated (DRAGAN) samples,
the $\operatorname{supp} \; \zeta$ for WGAN-GP is less diffuse.
Local penalties ensure the Lipschitzness is somewhat satisfied all around the
selected samples, which for DRAGAN is motivated by the fact that there are narrow peaks
on the reward landscape located at the expert samples,
where it us prone to overfit (\textit{cf.} \textsc{Section}~\ref{overfitting}).
The distribution $\zeta$ used in WGAN-GP also supports data points near expert samples,
but these are not scattered all around for the sole purpose of making the whole area smooth
and escape bad basins of attraction like in DRAGAN.
In other terms, the Lipschitz-continuity constraint is applied isotropically,
from the original expert sample outwards.
By contrast, WGAN-GP's $\zeta$ only supports a few discrete directions from a given expert sample,
the lines joining said sample to all the agent-generated samples (of the mini-batch).
Intuitively, while DRAGAN smooths out the reward landscape starting from expert data points
and going in every direction from there,
WGAN-GP smooths out the reward landscape starting from expert data points
and going only in the directions that point toward agent-generated data points.
As such, one could qualify DRAGAN as \emph{isotropic} regularizer,
and WGAN-GP as \emph{directed} regularizer.

We believe that WGAN-GP outperforms DRAGAN in the setting and environments considered in this work
(\textit{cf.} \textsc{Figure}~\ref{penvariantsplots})
due to the fact that the agent benefits from having smooth reward \emph{pathways}
in the reward landscape in-between agent samples and expert samples.
Along these pathways, going from the agent sample end to the expert sample end, the
reward \emph{progressively} increases.
For the agent trying to maximize its return,
these series of gradually increasing rewards joining agent to the expert data points
are akin to an \emph{automatic curriculum} \cite{Karpathy2012-qt,OpenAI2019-vy}
assisting the reward-driven agent and
leading it towards the expert.
\textsc{Figure}~\ref{penvariantsplots} shows that WGAN-GP indeed achieves consistently
better results across every environment
but the least challenging, as seen in the \texttt{IDP} environment
(\textit{cf.}~\textsc{Table}~\ref{envtable}).
In the four considerably more challenging environments, the \emph{directed} method
allows the agent to attain overall significantly higher empirical return than its competitors.
Besides, it displays greater stability when approaching the asymptotic regime,
whereas the \emph{local} regularizers clearly suffer from instabilities,
especially DRAGAN in the results obtained in environments \texttt{Walker2d} and \texttt{HalfCheetah},
depicted in \textsc{Figure}~\ref{penvariantsplots}.
While the proposed interpretation laid out previously
corroborates the results obtained and reported in \textsc{Figure}~\ref{penvariantsplots},
it does not explain the instability issues hindering the local penalties.
We believe the jittery behavior observed in
the results obtained in environments \texttt{Walker2d} and \texttt{HalfCheetah}
(\textit{cf.}~\textsc{Figure}~\ref{penvariantsplots}) ---
once the peak performance is attained ---
is caused by $\operatorname{supp} \; \zeta$
(green areas in \textsc{Figure}~\ref{gpvariants})
not changing \emph{is size}
as the agent learns to imitate
and gets closer to the expert in $\mathcal{S} \times \mathcal{A}$.

Indeed, in DRAGAN, $\zeta$ is a stationary distribution:
it applies the regularizer on perturbations
of the expert samples,
where the additive noise's underlying sufficient statistics are constant
throughout the learning process,
and where the expert data points are distributed
according to the stationary policy $\pi_e$ and its associated state visitation distribution.
For NAGARD, the perturbations follow the same distribution,
and remain constant across the updates.
However, unlike DRAGAN, $\zeta$ is defined by adding the stationary noise
to samples from the \emph{current} agent, every update,
distributed as $\beta$ in our \emph{off}-policy setting.
Since $\beta$ is by construction non-stationary across the updates,
as a mixture of past $\pi_\theta$ updates,
$\zeta$ is non-stationary in NAGARD.
Despite $\zeta$'s having these different support and stationary traits,
the results of either local penalties are surprisingly similar.
This is due to the variance of the additive noise used in both methods
being large relative to the distance between the expert and agent samples,
at all times, in the considered environments.
As such, their $\operatorname{supp} \; \zeta$ are virtually overlapping,
which makes the two local penalties virtually equivalent,
and explains the observed similarities in-between them.

Coming back to the main point
--- \textit{``why do local penalties suffer from instabilities at the end of training?''} ---
even though the agent samples are close to the expert ones,
the local methods both apply the same large perturbation before applying the
Lipschitz-continuity penalty.
The probability mass assigned by $\zeta$ is therefore still spread similarly
over the input space, and is therefore severely decreased
in-between agent and expert samples since these are getting closer in the space.
The local methods are therefore often applying the constraint
on data points that the policy will never visit again (since it
wants to move towards the expert) and equivalently, rarely
enforces the constraint between the agent and the expert, which is where
the agent should be encouraged to go.
With this depiction, it is clearer why WGAN-GP pulls ahead.
Compared to the fixed size of $\operatorname{supp} \; \zeta$ in the local penalties,
$\zeta$ \emph{adapts} to the current needs of the agent (hence qualifying as non-stationary).
As the agent gets closer to the expert, Lipschitz-continuity is always
enforced on data points between them, which is where it potentially benefits the agent most.
The support of $\zeta$ is therefore decreasing in size as the iterations go by,
focusing the probability mass of $\zeta$ where enforcing a smooth reward landscape
matters most: where the agent should go, \textit{i.e.} in the direction of the expert data points.

Besides, considering the inherent sample selection bias \cite{Heckman1979-ui} the control agent
is subjected to, where the latter end up in $\mathcal{S} \times \mathcal{A}$ depends on
its actions, in every interaction with the dynamical system represented by its environment.
This aspect dramatically differs from the traditional \emph{non-Markovian} GAN setting ---
in which these penalties were introduced ---
where the generator's input noise is \textit{i.i.d.}-sampled.
Indeed, suffering from said sample selection bias,
an imitation agent straying from the expert demonstrations is likely to
keep on doing so until the episode is reset
(\textit{cf.} discussion in \textsc{Section}~\ref{gradpen}).
Distributions $\zeta$ whose definition involve samples generated by
the learning agent and adapt to the agent's current relative position \textit{w.r.t.}
the expert data points therefore provide valuable extra guidance
in Markovian settings.
Additionally, assuming the input also contained the \emph{phase}
--- \textit{``how far the agent/expert is in the current episode''}, $0 \leq t \leq T$ ---
(like in \cite{Peng2018-mo})
not only would the imitation task be easier,
but the benefits of the WGAN-GP penalty would be further enhanced,
as it would allow the models to exploit the temporal structure
of to the considered Markovian setting.

Finally, in reaction to the recent interest towards \textit{``zero-centered''} gradient penalties
\cite{Roth2017-sj, Mescheder2018-ck},
due to the theoretical convergence guarantees they allow for,
we have conducted a grid search on the values of
the Lipschitz constant $k$ and the regularizer importance coefficient $\lambda$,
as described in \textsc{Section}~\ref{gradpenrl}.
The results are reported in \textsc{Appendix}~\ref{gridklam}.
In short, the method performs poorly when $k=0$,
unless a very small value is used for $\lambda$.
Enforcing $0$-Lipschitzness is far too restraining for the agent to learning
anything, unless this constraint is only loosely imposed.
Conversely, a smaller $\lambda$ value yields worse results when $k=1$,
revealing the interaction between the gradient penalty hyper-parameters $k$ and $\lambda$.
In particular, we will momentarily provide comprehensive evidence along with a greater characterization
of how the choice of scaling factor $\lambda$ not only impacts the agent's performance (which is
already depicted in \textsc{Appendix}~\ref{gridklam}), but how it correlates quantitatively with
the approximate
$\mathfrak{C}$-validity displayed by the agent (\textit{cf.}~\textsc{Section}~\ref{understand}).
Unless explicitly stated otherwise, we use the WGAN-GP penalty variant, with Lipschitz constant target $k=1$,
and scaling coefficient $\lambda=10$ throughout the empirical results exhibited in both the body and appendix.

\subsection{Diagnosing $\mathfrak{C}$-validity: is the Lipschitzness premise of
the theoretical guarantees satisfied in practice?}
\label{understand}

To put things in perspective,
we first give a side-by-side rundown of how what we set out to tackle here
compares to what we have just tackled in \textsc{Section}~\ref{gradpenrl},
thereby giving a glimpse of what we set out to investigate in what follows.
In the previous section, we showed how \textit{(a)} the choice of $\zeta$
(\emph{where} do we want to encourage approximately $\mathfrak{C}$-valid behavior), and
\textit{(b)} the choice of $\lambda$
(\emph{to what degree} do we want to encourage approximately $\mathfrak{C}$-valid behavior)
both independently impact the agent's performance in terms of empirical episodic return.
In this section on the other hand,
we will show how \textit{(a)} the choice of $\zeta$, and
\textit{(b)} the choice of $\lambda$
both independently impact the agent's consistency at \emph{effectively} selecting
approximately $\mathfrak{C}$-valid actions with its learned policy $\mu_\theta$.
If we were to find a strong positive correlation between the agent's asymptotic return and
its effectively measured approximate $\mathfrak{C}$-validity rate
--- high when high, low when low, for all tested $\zeta$'s and for all tested $\lambda$'s ---
then we would have further quantitative evidence to support our work's main claim:
reward Lipschitzness is necessary to achieve high return, and higher Lipschitzness uptime correlates strongly with
higher return.
Perhaps most crucially,
we would be able to correlate high empirical episodic return with high chance of satisfying the premise of our
theoretical guarantees ($r_\varphi$'s Lipschitzness). As such, these would consequently apply in in practice too.
This would attest to the practical relevance of \textsc{Section}~\ref{theory}.

We have shown that enforcing a Lipschitz-continuity constraint
on the learned reward $r_\varphi$ (albeit indirectly via $D_\varphi$)
is instrumental in achieving expert-level performance
in off-policy generative adversarial imitation learning
(\textit{cf.} \textsc{Section}~\ref{empres1}).
We have also shown that directed regularization techniques
yield better results,
seemingly due to the better guidance they provide to the mimicking agent,
in the form of an automatic curriculum of rewards towards the expert data points
(\textit{cf.} \textsc{Section}~\ref{gradpenrl}).
Such curriculum only exists where the Lipschitz-continuity constraint
is satisfied.
Said differently, it could not exist if the constraint were not satisfied along $\mu_\theta$'s pathways
which would then involve non-smooth hurdles.
It is therefore crucially important for said constraint to be satisfied \emph{in effect} for the
state-actions pairs in the the support of the policy the agent uses in its learning update,
$\mu_\theta$, \textit{i.e.}~$\operatorname{supp} \; \mu_\theta \, \ni (s_t, a_t)$.
Still, the deterministic policy $\mu_\theta$ likely performs only approximately $\mathfrak{C}$-valid
actions as it is trained with the sole objective to maximize cumulative rewards
that represent its similarity \textit{w.r.t.} the expert $\pi_e$.
The imitation rewards corresponding to a greater degree of similarity are,
by design of the generative adversarial imitation learning framework,
situated between the agent's current position and the expert's position
on the current reward landscape.
Since this is where we apply the Lipschitzness constraint (with WGAN-GP, our baseline, as said above)
--- equivalently, since these regions are approximately $\mathfrak{C}$-valid ---
$\mu_\theta$ is likely to never select $\mathfrak{C}$-invalid actions
as it optimizes for its utility function
(\textit{cf.} \textsc{Section}~\ref{prelim}).
Conversely, in the considered setting, picking $\mathfrak{C}$-invalid actions
could in theory hinder the optimization process the policy is subject to,
as $\mu_\theta$ would \textit{a priori} venture in regions of the state-action space
that do not increase its similarity with the expert policy $\pi_e$
--- or, at the very least, for which the \emph{non-}satisfaction of the reward's Lipschitz-continuity premise
$\lVert \nabla_{s,a}^t[r_\varphi]_t \rVert _F \leq \delta$
might lead to instabilities due to
$\lVert \nabla_{s,a}^t[Q_\varphi]_t \rVert _F > \Delta_\infty$
as a direct consequence of our theoretical guarantees
(\textit{cf.}~\textsc{Section}~\ref{discussion}).
Since we do not have such a tight control over where and to what degree
the Lipschitzness constraint over the reward $r_\varphi$ is \textit{satisfied}
(hence our introduction of the notions of approximately $\mathfrak{C}$-valid samples
and probably approximately $\mathfrak{C}$-valid samples),
we instead turn to the closest surrogate over which we do have a tighter control:
where and to what degree $D_\varphi$'s constraint is \emph{enforced}.
The \textit{``where''} is controlled by the choice of $\zeta$ (determined by the gradient penalty regularization
method in use),
and the \textit{`to what degree'} by the choice of $\lambda$ scale.

Still, even in the occurrence where $D_\varphi$'s constraint is enforced
by adding $\mathfrak{R}_\varphi^\zeta (k)$ as in $\ell_\varphi^\textsc{GP}$ (\textit{cf.}~\textsc{eq}~\ref{eqgp})
at the point $(s_t,a_t)$,
the most we could say
is that $(s_t,a_t)$ is probably approximately $\mathfrak{C}$-valid,
since $(s_t,a_t) \in \operatorname{supp} \; \zeta$
--- otherwise, the gradient penalty regularizer $\mathfrak{R}_\varphi^\zeta (k)$
could never have been applied at that point in the landscape
$\mathcal{S} \times \mathcal{A}$.
In effect, enforcing the constraint at the point was enough to guarantee
that $\lVert \nabla_{s,a}^t[D_\varphi]_t \rVert _F \leq k$, and we therefore do not know whether
$(s_t,a_t)$ is approximately $\mathfrak{C}$-valid, or not.
As a direct consequence, we can \textit{a fortiori} not guarantee that
$\lVert \nabla_{s,a}^t[r_\varphi]_t \rVert _F \leq \delta$; we do not know whether
$(s_t,a_t)$ is $\mathfrak{C}$-valid, or not
--- \textit{cf.}~\textsc{Section}~\ref{indirectreg} for our discussion on indirect reward regularization,
in which we establish that $D_\varphi$'s $k$-Lipschitzness causes $r_\varphi$ to be
$\delta$-Lipschitz in practice.
On the flip side, based on the latter result about indirect Lipschitz-continuity inducement,
we can state that ensuring empirically that $\lVert \nabla_{s,a}^t[D_\varphi]_t \rVert _F \leq k$
is \emph{enough} to ensure that $\lVert \nabla_{s,a}^t[r_\varphi]_t \rVert _F \leq \delta$ is verified in practice.
In other words, showing that $(s_t,a_t)$ is approximately $\mathfrak{C}$-valid can be used as
a proxy for showing that $(s_t,a_t)$ is $\mathfrak{C}$-valid, empirically.
As such, in order to assess whether the premise of the theoretical guarantees we derived in
\textsc{Section}~\ref{theory} is satisfied in practice ($r_\varphi$'s -$\delta$-Lipschitz-continuity),
it is sufficient to assess whether the agent's actions $a_t = \mu_\theta (s_t)$
are approximately $\mathfrak{C}$-valid.
In particular, we want to know the relative impacts
the choices of $\zeta$ and the $\lambda$ in $\ell_\varphi^\textsc{GP}$
have on the propensity for an action from $\mu_\theta$ to be approximately $\mathfrak{C}$-valid.
So as to estimate how often the actions selected by the agent via $\mu_\theta$ are approximately $\mathfrak{C}$-valid,
we build an estimator that \emph{softy} approximates
$\mathds{1}_\mathfrak{C}: \mathcal{S} \times \mathcal{A} \to \{0,1\}$,
the indicator of the $\mathfrak{C}$-validity subspace over $\mathcal{S} \times \mathcal{A}$, where
$\mathds{1}_\mathfrak{C}(s_t,a_t) = 1$ when $(s_t,a_t) \in \mathfrak{C}$,
and $\mathds{1}_\mathfrak{C}(s_t,a_t) = 0$ when $(s_t,a_t) \notin \mathfrak{C}$.
Accordingly, we call our estimator \textit{soft approximate $\mathfrak{C}$-validity pseudo-indicator},
implementing a soft, $C^0$ mapping $\widehat{\mathds{1}}_\mathfrak{C}: \mathcal{S} \times \mathcal{A} \to (0,1]$,
and formally defined as,
$\forall t \in [0, T] \cap \mathbb{N},
\forall (s_t, a_t) \in \mathcal{S} \times \mathcal{A}$:
\begin{align}
\widehat{\mathds{1}}_\mathfrak{C}(s_t,a_t) \coloneqq \exp \Big(-
\max \big(0, \lVert \nabla_{s_t,a_t} \, D_\varphi (s_t,a_t) \rVert - k\big)^2\Big)
\qquad
\blacktriangleright\text{\small{\textit{soft approximate $\mathfrak{C}$-validity pseudo-indicator}}}
\label{pseudoindicator}
\end{align}
Thus, for a given pair $(s_t, a_t)$,
$\widehat{\mathds{1}}_\mathfrak{C}(s_t,a_t) = 1$
when $\lVert \nabla_{s,a}^t[D_\varphi]_t \rVert _F \leq k$
and $\widehat{\mathds{1}}_\mathfrak{C}(s_t,a_t) \to 0$
when $\lVert \nabla_{s,a}^t[D_\varphi]_t \rVert _F \gg k$.

\begin{figure}
  \center
  \begin{subfigure}[t]{0.99\textwidth}
    \center\scalebox{0.18}[0.18]{\includegraphics{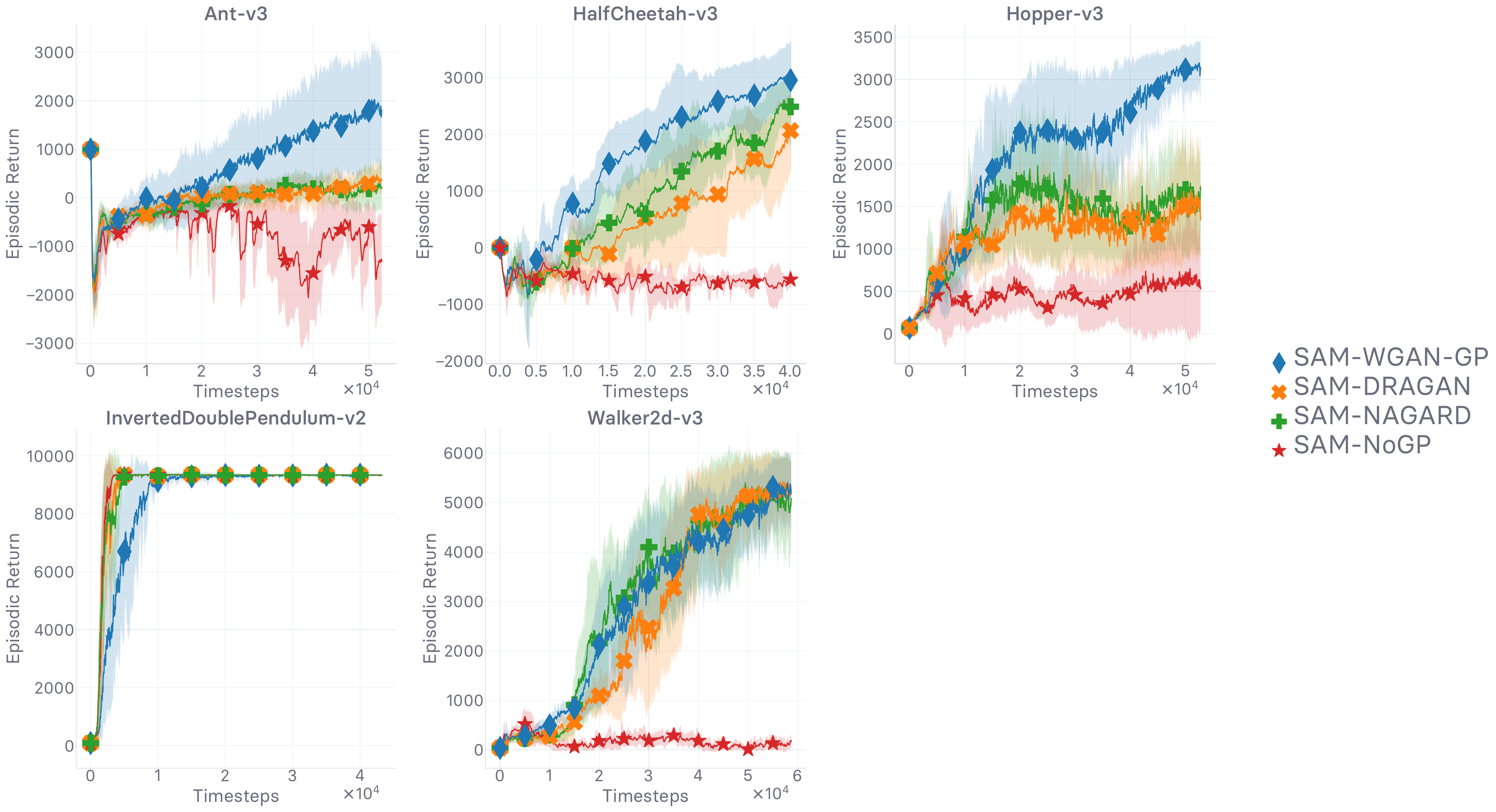}}
    \caption{Evolution of return values \textit{(higher is better)}}
    \label{pseudoindicatorempzetaret}
  \end{subfigure}
  \begin{subfigure}[t]{0.99\textwidth}
    \center\scalebox{0.18}[0.18]{\includegraphics{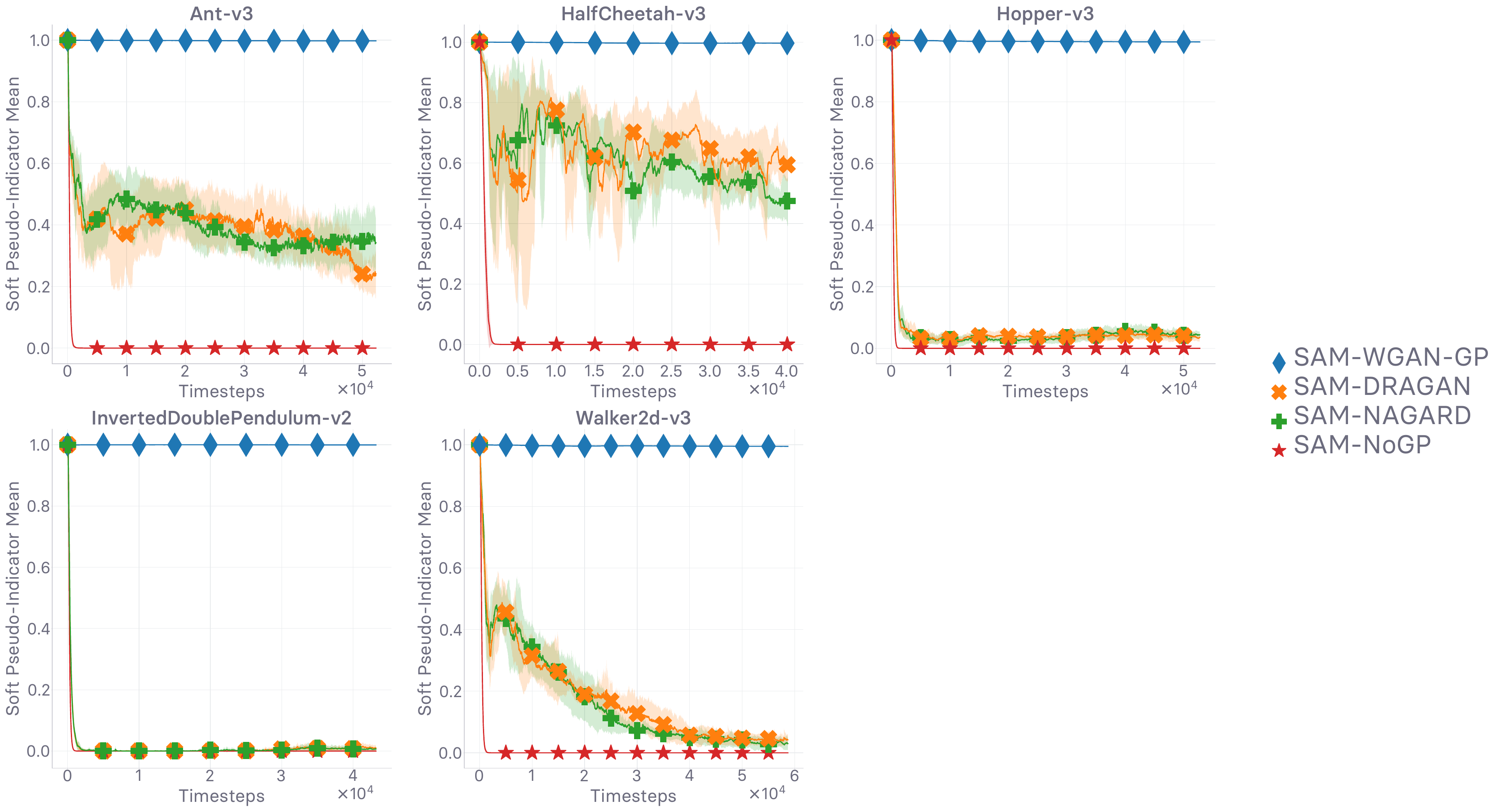}}
    \caption{Evolution of $\widehat{\mathds{1}}_\mathfrak{C}$ values
    \textit{(higher means more approx. $\mathfrak{C}$-valid)}, \textit{cf.}~\textsc{eq}~\ref{pseudoindicator}.}
  \end{subfigure}
  \begin{subfigure}[t]{0.99\textwidth}
    \center\scalebox{0.18}[0.18]{\includegraphics{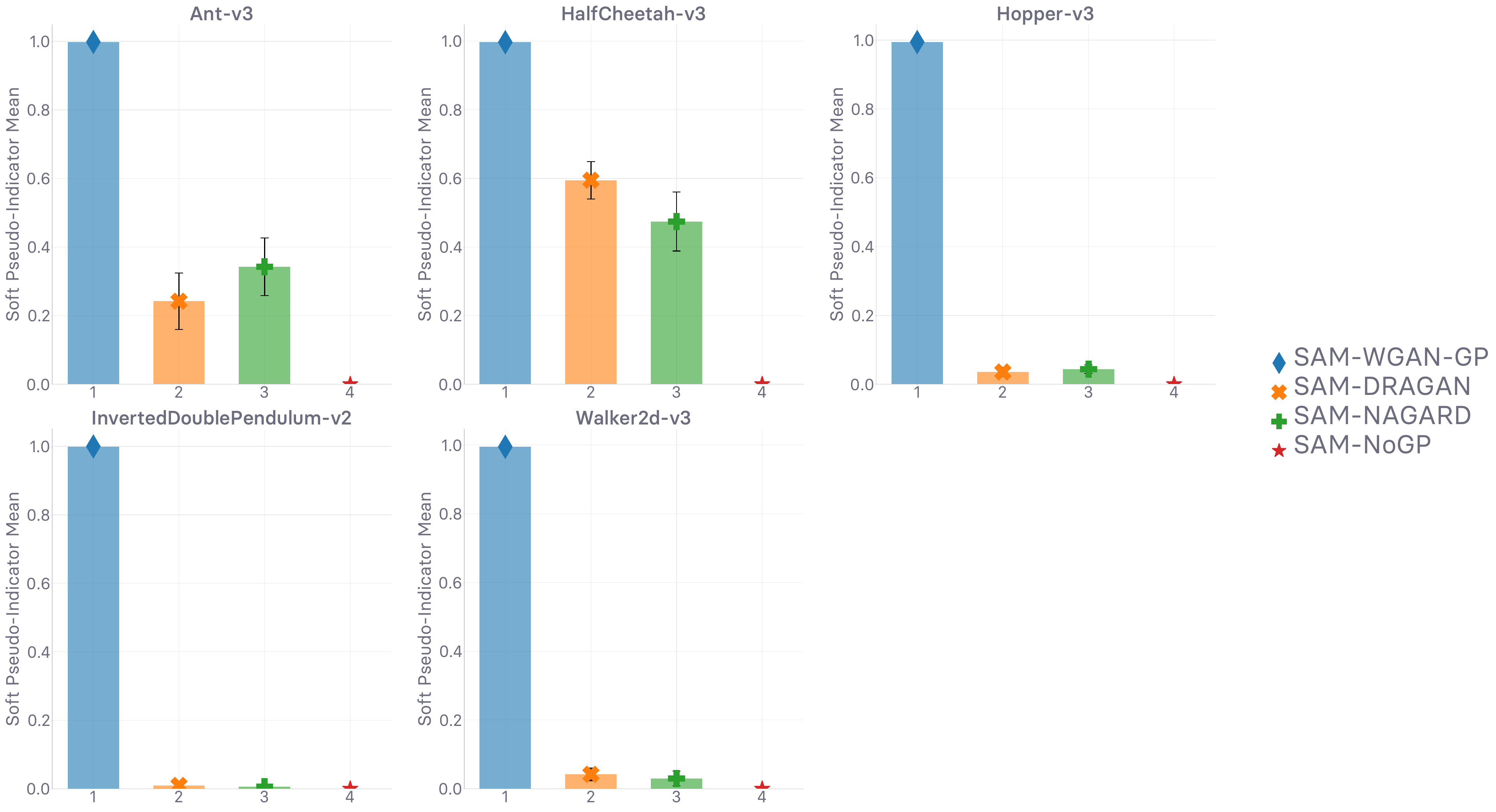}}
    \caption{Final $\widehat{\mathds{1}}_\mathfrak{C}$ values at timeout
    \textit{(higher means more approx. $\mathfrak{C}$-valid)}, \textit{cf.}~\textsc{eq}~\ref{pseudoindicator}.}
  \end{subfigure}
  \caption{
  Evaluation of several GP methods differing by their $\zeta$ distribution
  In line with how we defined it in \textsc{eq}~\ref{eqgp},
  $\zeta$ controls \textbf{\emph{``where''}} the GP constraint is enforced.
  Also, we report what happens without any GP regularization (\texttt{NoGP}).
  Explanation in text.
  Runtime is 48h.}
  \label{pseudoindicatorempzeta}
\end{figure}

\begin{figure}
  \center
  \begin{subfigure}[t]{0.99\textwidth}
    \center\scalebox{0.18}[0.18]{\includegraphics{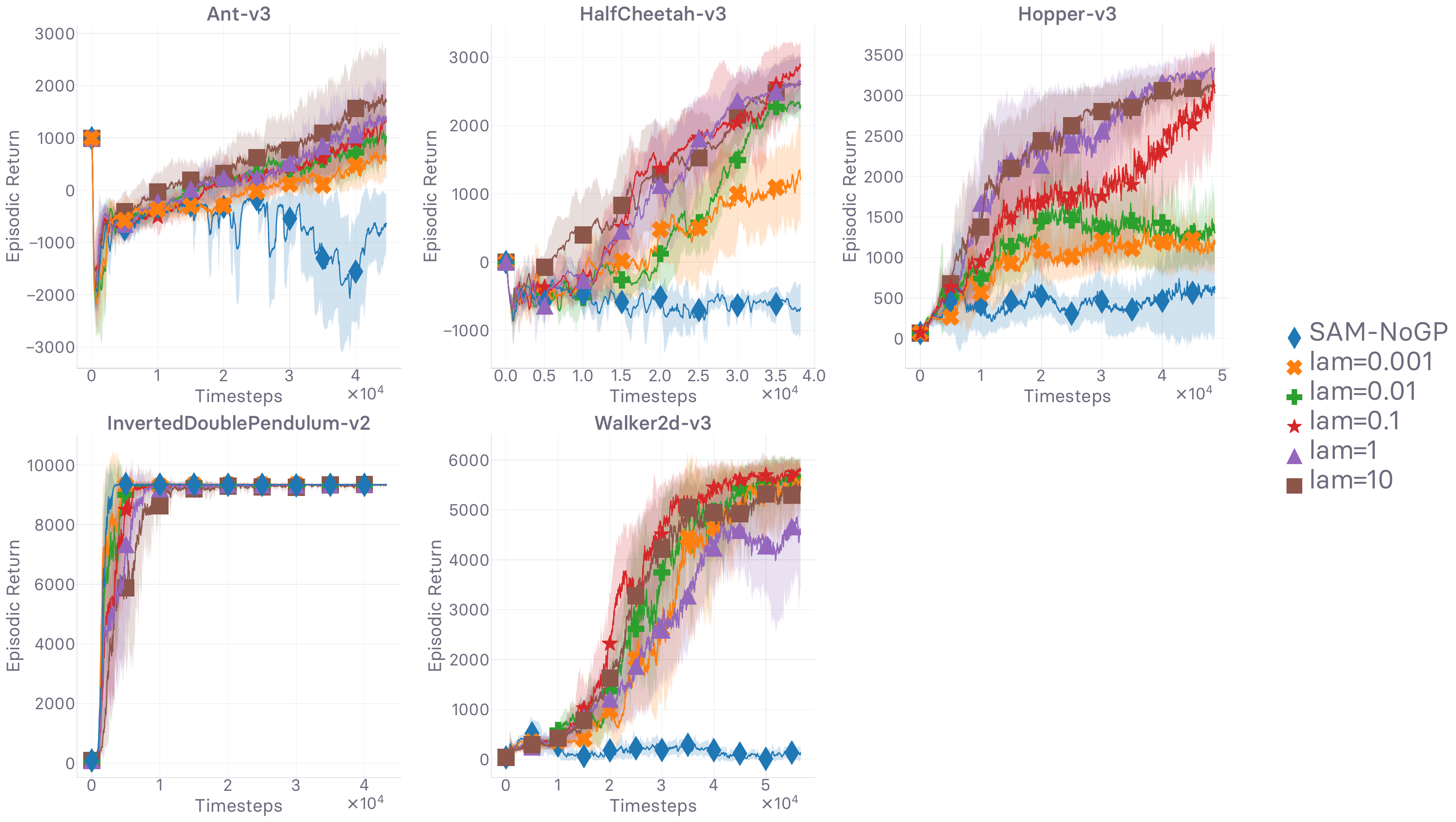}}
    \caption{Evolution of return values \textit{(higher is better)}}
    \label{pseudoindicatoremplambdaret}
  \end{subfigure}
  \begin{subfigure}[t]{0.99\textwidth}
    \center\scalebox{0.18}[0.18]{\includegraphics{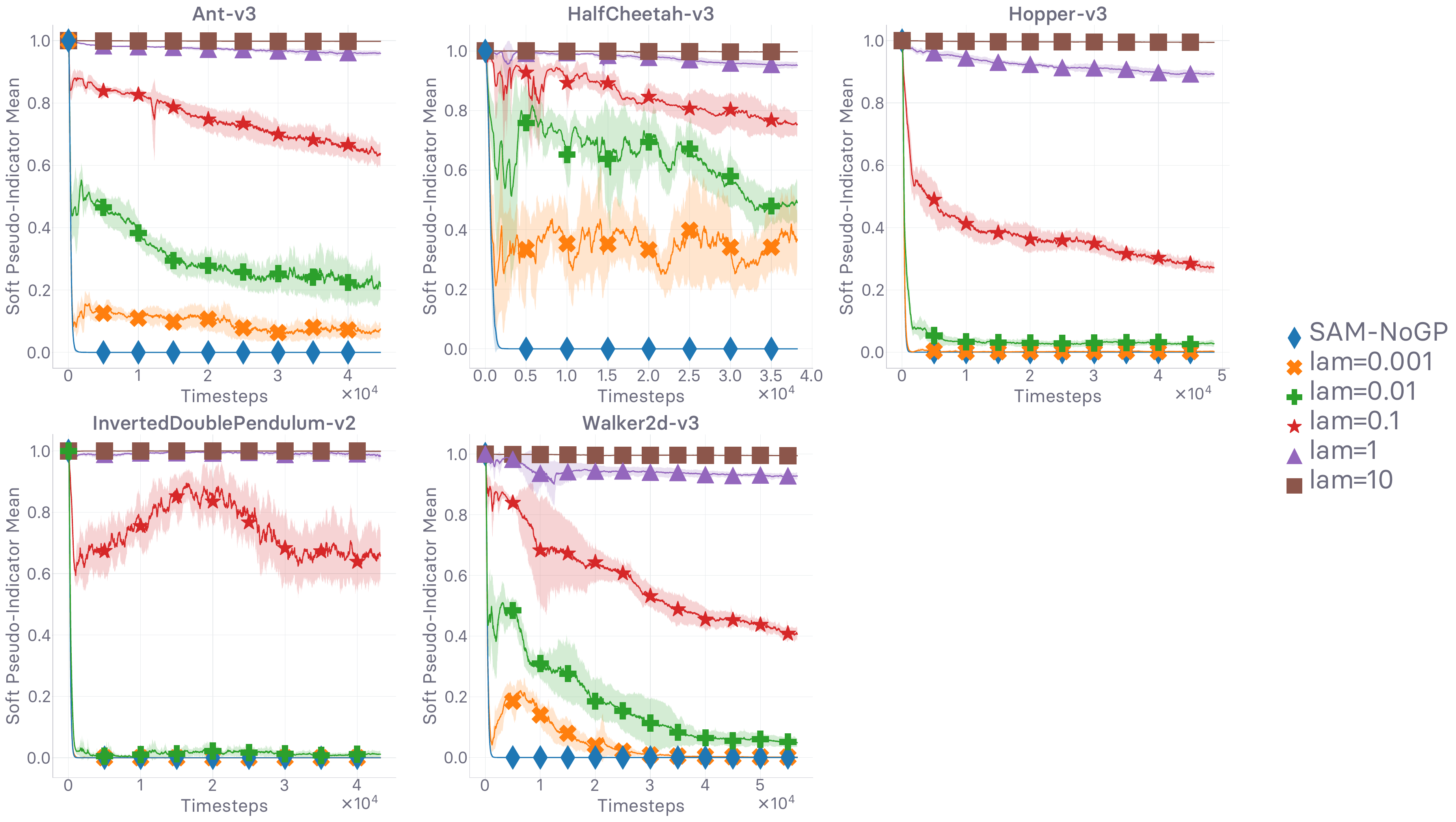}}
    \caption{Evolution of $\widehat{\mathds{1}}_\mathfrak{C}$ values
    \textit{(higher means more approx. $\mathfrak{C}$-valid)}, \textit{cf.}~\textsc{eq}~\ref{pseudoindicator}.}
  \end{subfigure}
  \begin{subfigure}[t]{0.99\textwidth}
    \center\scalebox{0.18}[0.18]{\includegraphics{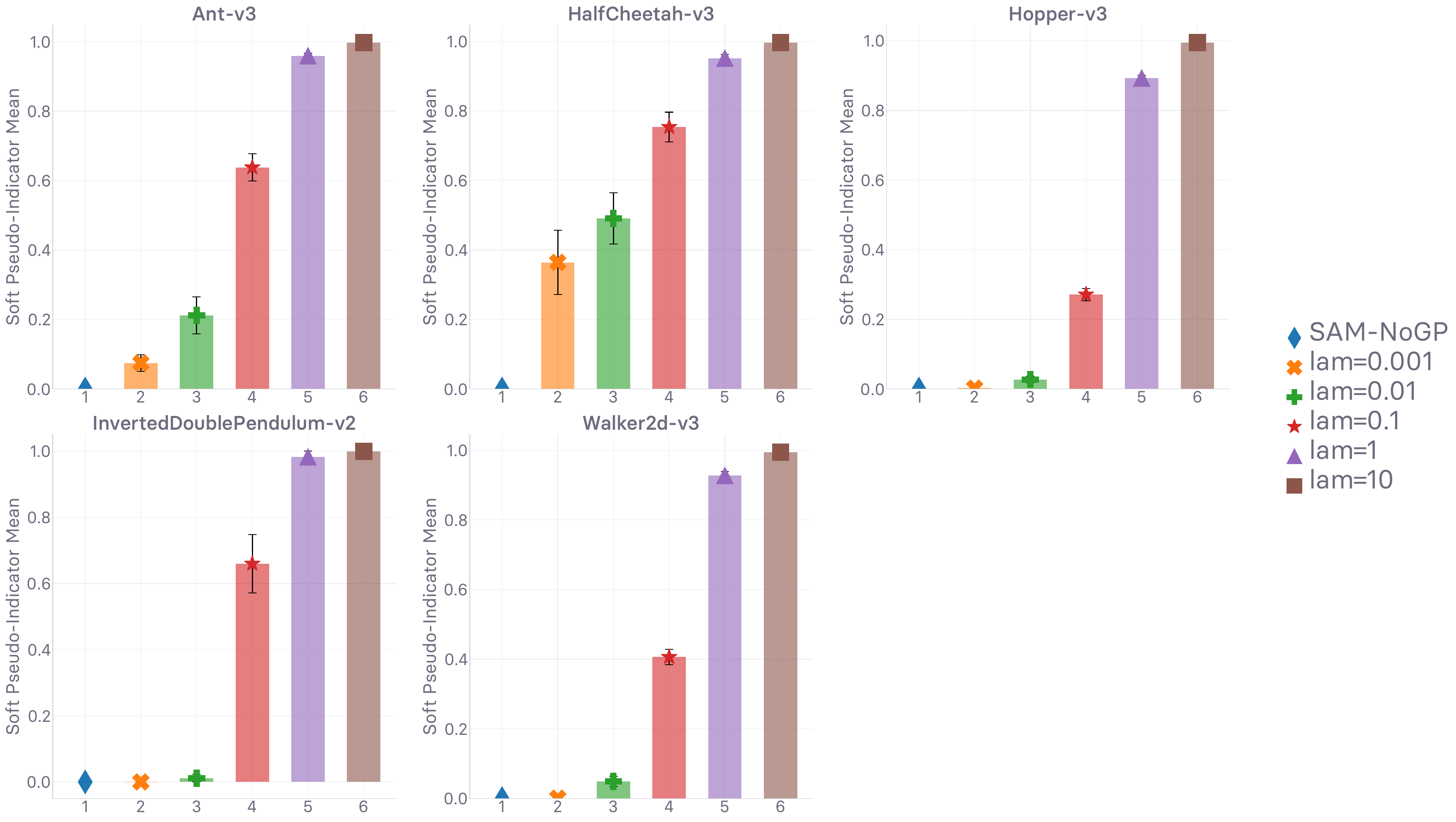}}
    \caption{Final $\widehat{\mathds{1}}_\mathfrak{C}$ values at timeout
    \textit{(higher means more approx. $\mathfrak{C}$-valid)}, \textit{cf.}~\textsc{eq}~\ref{pseudoindicator}.}
  \end{subfigure}
  \caption{
  Evaluation of several GP methods differing by their $\lambda$ scaling factor
  In line with how we defined it in \textsc{eq}~\ref{eqgp},
  $\zeta$ controls \textbf{\emph{``to what degree''}} the GP constraint is enforced.
  Also, we report what happens without any GP regularization (\texttt{NoGP}).
  Explanation in text.
  Runtime is 48h.}
  \label{pseudoindicatoremplambda}
\end{figure}

\textsc{Figures}~\ref{pseudoindicatorempzeta} and \ref{pseudoindicatoremplambda}
depict respectively the evolution of the values taken by the
soft approximate $\mathfrak{C}$-validity pseudo-indicator $\widehat{\mathds{1}}_\mathfrak{C}$
(\textit{cf.}~\textsc{eq}~\ref{pseudoindicator})
for different choices of $\zeta$ (different gradient penalty variants)
and $\lambda$ (sweep over $\mathfrak{R}_\varphi^\zeta (k)$'s scaling factor).
In \textsc{Figures}~\ref{pseudoindicatorempzeta} and \ref{pseudoindicatoremplambda},
we also share the return accumulated by the agents throughout their respective training periods,
(\textit{cf.}~\ref{pseudoindicatorempzetaret} and \ref{pseudoindicatoremplambdaret}, respectively).
In particular,
what we report in \textsc{Figures}~\ref{pseudoindicatorempzetaret} and \ref{pseudoindicatoremplambdaret}
echoes what we have already reported in \textsc{Figures}~\ref{penvariantsplots} and \ref{gridk1},
but the settings in which the agents were trained differ (ever so) slightly.
We indicate the specificities of the setting tackled in this section below, in this very paragraph.
Still, since their settings do not match perfectly,
we report their return along their
soft approximate $\mathfrak{C}$-validity pseudo-indicator $\widehat{\mathds{1}}_\mathfrak{C}$ values.
We monitor and record these values during the evaluation trials the agent periodically goes through,
in which the agent uses $\mu_\theta$ to decide what to do in a given state.
To best align with the definition of Lipschitz-continuity
(\textit{cf.}~\textsc{Definition}~\ref{lipdef}),
which is also how we designed our
soft approximate $\mathfrak{C}$-validity pseudo-indicator $\widehat{\mathds{1}}_\mathfrak{C}$,
we use one-sided gradient penalties $\mathfrak{R}_\varphi^\zeta (k)$ in the $\lambda$ sweep ---
$\max (0, \lVert \nabla_{s_t,a_t} \, D_\varphi (s_t,a_t) \rVert - k)^2$,
which \emph{purely}
encourages $\lVert \nabla_{s,a}^t[D_\varphi]_t \rVert _F \leq k$ to be satisfied
(nothing more, nothing less) ---
although we have shown the variant presents very little empirical difference with the base two-sided one
(\textit{cf.}~ablation in \textsc{Appendix}~\ref{ablationonesided}).
It is worth noting that the experiments whose results are reported in
\textsc{Figures}~\ref{pseudoindicatorempzeta} and \ref{pseudoindicatoremplambda}
carry out less iterations during the fixed allowed runtime,
due to the substantial cost entailed by computing
soft approximate $\mathfrak{C}$-validity pseudo-indicator $\widehat{\mathds{1}}_\mathfrak{C}$
at every single evaluation step, in every evaluation trial.
One could cut down that cost simply by evaluating $\widehat{\mathds{1}}_\mathfrak{C}$ less frequently,
but we decided otherwise, as we gave priority to having a finer tracking of $\widehat{\mathds{1}}_\mathfrak{C}$.
Besides, despite this slight apparent hindrance, the values of the proposed
pseudo-indicator reported in either figure seem to have reached maturity,
nearing their asymptotic regime,
in the allowed runtime.
We now go over and interpret the results reported in both figures.

In \textsc{Figure}~\ref{pseudoindicatorempzeta},
we observe that the monitored
soft approximate $\mathfrak{C}$-validity pseudo-indicator $\widehat{\mathds{1}}_\mathfrak{C}$
(\textit{cf.}~\textsc{eq}~\ref{pseudoindicator})
consistently takes values close to $1$ when using the distribution $\zeta$ advocated in WGAN-GP
to assemble the regularizer $\mathfrak{R}_\varphi^\zeta (k)$.
Conversely, \emph{not} using any gradient penalty regularizer causes the approximate $\mathfrak{C}$-validity
rate to be in the vicinity of $0$.
Albeit \textit{a priori} not surprising, it is still substantially valuable to notice that $D_\varphi$'s
$k$-Lipschitz-continuity
(and therefore $r_\varphi$'s $\delta$-Lipschitz-continuity; \textit{cf.}~\textsc{Section}~\ref{discussion})
\emph{never} happens by accident (or rather, by chance).
As for DRAGAN and NAGARD (both being non-directed gradient penalty schemes, unlike WGAN-GP;
\textit{cf.}~\textsc{Section}~\ref{gradpenrl}),
both perform similarly across the board
in terms of collected $\widehat{\mathds{1}}_\mathfrak{C}$ values.
Their recorded soft pseudo-indicator values stay around a fixed value per environment, different
for every one of them. These are within the $[0.1, 0.7]$ range, and as such,
are definitely encouraging $\lVert \nabla_{s,a}^t[D_\varphi]_t \rVert _F \leq k$ in practice,
yet are falling short of achieving the same
\textit{(a)} effective approximate $\mathfrak{C}$-validity value,
and \textit{(b)} effective approximate $\mathfrak{C}$-validity consistency
as WGAN-GP.
These phenomenona occur consistently across the spectrum of tackled environments.

In \textsc{Figure}~\ref{pseudoindicatoremplambda}, we observe the unsurprising fact that
the higher $\lambda$'s value is --- equivalently, the more we encourage the regularity property
$\lVert \nabla_{s,a}^t[D_\varphi]_t \rVert _F \leq k$ to be satisfied ---
the more $\lVert \nabla_{s,a}^t[D_\varphi]_t \rVert _F \leq k$ is satisfied in effect.
Besides confirming that gradient penalization indeed urges Lipschitzness (which we were not doubting),
the figure helps us gauge to what degree the value of $\mathfrak{R}_\varphi^\zeta (k)$'s scaling coefficient
in $\ell_\varphi^\textsc{GP}$ (\textit{cf.}~\textsc{eq}~\ref{eqgp})
affects quantitatively the satisfaction of $\lVert \nabla_{s,a}^t[D_\varphi]_t \rVert _F \leq k$
monitored via the soft proxy $\widehat{\mathds{1}}_\mathfrak{C}$.
We considered powers of $10$ for $\lambda$'s sweep, tackling the values $\lambda_i \coloneqq 10^i$,
for $i \in \{-3, -2, -1, 0, 1\}$.
The gap inbetween the $\widehat{\mathds{1}}_\mathfrak{C}$ values associated with each of these $\lambda_i$
differ per environment, but their ranking remain the same (higher $\widehat{\mathds{1}}_\mathfrak{C}$'s
for higher $i$'s).
At its lowest (\textit{i.e.}~for minimum $i$: $i=-3$) the soft pseudo-indicator values
lie more often that not near $0$. For $i = 1$, $\widehat{\mathds{1}}_\mathfrak{C}$ perfectly aligns on the $1$ value,
meaning that the value we used so far ($\lambda=10$, which corresponds to $\lambda_i$ with $i=1$) is enough
for $\mu_\theta$ to achieve a $100\%$ satisfaction rate of $\lVert \nabla_{s,a}^t[D_\varphi]_t \rVert _F \leq k$.
The case $i = 0$ is right on the edge: in some environments, the approximate $\mathfrak{C}$-validity
exactly equals $1$, while for other environments, it nears it, yet does not quite reach it.

Since we use WGAN-GP's $\zeta$ in the experiments reported in \textsc{Figure}~\ref{pseudoindicatoremplambda},
we can first conclude that picking WGAN-GP's $\zeta$ variant and $\lambda=10$ not only yields the best
empirical return (as reported and discussed in \textsc{Section}~\ref{gradpenrl}),
but also guarantees that the constraint $\lVert \nabla_{s,a}^t[D_\varphi]_t \rVert _F \leq k$
(and therefore $\lVert \nabla_{s,a}^t[r_\varphi]_t \rVert _F \leq \delta$;
\textit{cf.}~\textsc{Section}~\ref{discussion}))
is satisfied for $100\%$ of the actions performed by the agent's $\mu_\theta$ in practice.
As such, we can conclude that, in practice, the main premise of the theoretical
guarantees we have derived in \textsc{Section}~\ref{theory}
--- the reward $\delta$-Lipschitz-continuity,
$\lVert \nabla_{s,a}^t[r_\varphi]_t \rVert _F \leq \delta$ ---
is satisfied, hence making our theoretical guarantees \emph{practically} relevant and insightful.
In addition, since we showed that the learning agent's policy $\mu_\theta$ (or rather, it's companion Q-value)
is trained on a reward surrogate $r_\varphi$ that verifies
$\lVert \nabla_{s,a}^t[r_\varphi]_t \rVert _F \leq \delta$ almost $100\%$ of the time,
we have empirically proved that the agent effectively sees
virtually uninterrupted sequences of smooth rewards.
This new observation somewhat corroborates our RL-grounded interpretation of directed gradient penalization as
as the automated and adaptive creation of reward curricula (\textit{cf.}~\textsc{Section}~\ref{gradpenrl},
and particularly our schematic depiction of WGAN-GP's $\operatorname{supp} \; \zeta$ in \textsc{Figure}~\ref{wgangp}).

Despite having answered the question we asked in the title of the section (in the block right above),
interpreting the findings laid out both in this section and in the previous one side-by-side
allows us to draw another critical conclusion, substantially more meaningful than
if we were to interpret either in a vacuum.
In \textsc{Section}~\ref{gradpenrl}, we studied the impact $\zeta$ and $\lambda$
both have on the agent's performance, in terms of the empirical return in the MDP $\mathbb{M}$.
We refer here to the latter via the shorthand \textsc{Return}.
In \emph{this} section, on the other hand, we have studied the impact $\zeta$ and $\lambda$
both have on the effective approximate $\mathfrak{C}$-validity rate of the agent.
We refer here to the latter via the shorthand \textsc{Validity}.
What emerges from comparing these two sets of results is that, for every given pair $(\zeta,\lambda)$
(\emph{where} to apply the gradient penalty, and \emph{to what degree}, respectively)
in $\ell_\varphi^\textsc{GP}$ (\textit{cf.}~\textsc{eq}~\ref{eqgp}):
low \textsc{Return} co-occurs with low \textsc{Validity};
intermediate \textsc{Return} co-occurs with intermediate \textsc{Validity};
high \textsc{Return} co-occurs with high \textsc{Validity}.
Said differently, \textsc{Return} and \textsc{Validity} behave similarly under the various
pairings $(\zeta,\lambda)$ that we have considered.
Through these observations, we therefore witness a strong correlation between
\textsc{Return} and \textsc{Validity}.
Ultimately, by combining our two previous empirical analyses,
we have shown that \textsc{Validity} is a good predictor or \textsc{Return},
and \textit{vice versa}.

\textit{In fine}, compared to \textsc{Section}~\ref{empres1}, \textsc{Section}~\ref{understand} (this section)
gives a far more fine-grained diagnostic of how reward Lipschitzness relates to empirical return,
along with insights related to the practicality of our theorerical guarantees.

\subsection{Towards fulfilling the premise: a provably more robust way to further encourage Lipschitzness}
\label{purpleandres}

We introduce two new entities, $\kappa_t$
and $\tilde{r}_\varphi : \mathcal{S} \times \mathcal{A} \to \mathbb{R}$, formally defined as:
\begin{align}
\tilde{r}_\varphi (s_t,a_t)
\coloneqq \kappa_t \, r_\varphi (s_t,a_t)
\qquad
\blacktriangleright\text{\small{\textit{$\kappa_t$-preconditioned reward $\tilde{r}_\varphi$}}}
\label{precondrew}
\end{align}
$\forall t \in [0, T] \cap \mathbb{N},
\forall (s_t, a_t) \in \mathcal{S} \times \mathcal{A}$,
where $0 < \kappa_t \leq 1$, $\forall t \in [0, T] \cap \mathbb{N}$ (in any episode).

We call $\kappa_t$ a \emph{reward preconditioner} since it functionally echoes the numerical transformation
that conditions the tackled problem into a form that is more amenable to be solved via first-order optimization methods.
Since our preconditioner is a scalar,
we use the shorthand $\kappa_t$ to constrast with the usual preconditioning matricies,
denoted with capitalization.
We have the following ranking of values,
depending on the sign of the original learned synthetic reward $r_\varphi$:
$\forall t \in [0, T] \cap \mathbb{N}$ and
$\forall (s_t, a_t) \in \mathcal{S} \times \mathcal{A}$,
we have $\tilde{r}_\varphi (s_t,a_t) \leq r_\varphi (s_t,a_t)$ whenever $r_\varphi (s_t,a_t) > 0$,
and conversely,
we have $\tilde{r}_\varphi (s_t,a_t) > r_\varphi (s_t,a_t)$ whenever $r_\varphi (s_t,a_t) < 0$.

We posit that $\kappa_t$ does not depend on
(\textit{i.e.}, is constant \textit{w.r.t.})
the current state $s_t$ and action $a_t$:
\begin{align}
\dv*{\kappa_t}{s_t} = 0
\qquad \text{and} \qquad
\dv*{\kappa_t}{a_t} = 0
\qquad
\blacktriangleright\text{\small{\textit{property 1}}}
\label{purplenopresent}
\end{align}
$\forall t \in [0, T] \cap \mathbb{N},
\forall (s_t, a_t) \in \mathcal{S} \times \mathcal{A}$.
Thus, we can write
$\dv*{\tilde{r}_\varphi (s_t, a_t)}{s_t}
= \kappa_t \dv*{r_\varphi (s_t, a_t)}{s_t}
+ \dv*{\kappa_t (s_t, a_t)}{s_t} r_\varphi
= \kappa_t \dv*{r_\varphi (s_t, a_t)}{s_t}$,
and similarly
$\dv*{\tilde{r}_\varphi (s_t, a_t)}{a_t}
= \kappa_t \dv*{r_\varphi (s_t, a_t)}{a_t}$.
As such, we have
$\lVert \nabla_{s,a}^t[\tilde{r}_\varphi]_t \rVert _F
= \kappa_t \lVert \nabla_{s,a}^t[r_\varphi]_t \rVert _F$,
hence
$\lVert \nabla_{s,a}^t[\tilde{r}_\varphi]_t \rVert _F
\leq \lVert \nabla_{s,a}^t[r_\varphi]_t \rVert _F$
since $0 < \kappa_t \leq 1$, $\forall t \in [0, T] \cap \mathbb{N}$.
Applying such a preconditioner to $r_\varphi$ therefore squashes the absolute value of $r_\varphi$
and in effect \emph{shrinks} $r_\varphi$'s Lipschitz constant
(assuming here that $r_\varphi$ is $\delta$-Lipschitz, with
$\lVert \nabla_{s,a}^t[r_\varphi]_t \rVert _F \leq \delta < +\infty$)
without regard to the sign of the signal.
Formally, since $\kappa_t$ is posited constant in $s_t$ and $a_t$, we have,
$\forall t \in [0, T] \cap \mathbb{N}$
and $\forall (s_t, a_t) \in \mathcal{S} \times \mathcal{A}$:
\begin{align}
\lVert \nabla_{s,a}^t[r_\varphi]_t \rVert _F \leq \delta
\qquad \implies \qquad
\lVert \nabla_{s,a}^t[\tilde{r}_\varphi]_t \rVert _F
= \kappa_t \lVert \nabla_{s,a}^t[r_\varphi]_t \rVert _F
\leq \kappa_t \, \delta
\quad (\leq \delta)
\label{kappalip}
\end{align}
That is, if $r_\varphi$ is $\delta$-Lipschitz-continuous at $t$, then $\tilde{r}_\varphi$
is $\kappa_t \delta$-Lipschitz-continuous at $t$.
Importantly,
\textsc{eq}~\ref{kappalip} will be instrumental in proving the first stages of
our next theoretical guarantees,
in which we deal with the counterpart action-value of $\tilde{r}_\varphi$, denoted by $\widetilde{Q}_\varphi$.

Because of its \textit{``reward-squashing''} effect,
we name the method corresponding to the subtitution of $r_\varphi$ with
the preconditioned reward $\tilde{r}_\varphi$
\emph{``Pessimistic'' Reward Preconditioning Enforcing Lipschitzness}.
We dub the plug-in technique \textbf{``PURPLE''}
(it is an acronym, with minor vowel filling and letter shuffle for legibility and
easy of pronunciation).
From this point onward,
we study the effect of plugging PURPLE into SAM.
The pseudo-code of the resulting algorithm can be obtained by replacing the learned reward $r_\varphi$
in SAM's pseudo-code laid out in \textsc{Algorithm}~\ref{algosam}
with the preconditioned reward $\tilde{r}_\varphi$.

We now study how the injection of PURPLE in SAM impacts the theoretical guarantees we have previously derived in
\textsc{Section}~\ref{theory}.
Concretely, we derive the PURPLE counterparts of
\textsc{Lemma}~\ref{lemma}, \textsc{Theorem}~\ref{theorem1},
\textsc{Theorem}~\ref{theorem2}, and \textsc{Corollary}~\ref{corollary1}.
In order for us to characterize the Lipschitzness of $\widetilde{Q}_\varphi$,
we also posit that the introduced preconditioner does not depend on
(\textit{i.e.}, is constant \textit{w.r.t.})
the \emph{previously visited} (past) states and actions.
Formally:
\begin{align}
\dv*{\kappa_{t+k+1}}{s_t} = 0
\qquad \text{and} \qquad
\dv*{\kappa_{t+k+1}}{a_t} = 0
\qquad
\blacktriangleright\text{\small{\textit{property 2}}}
\label{purplenopast}
\end{align}
$\forall t \in [0, T] \cap \mathbb{N}, \forall k \in [0, T-t-1] \cap \mathbb{N},
\forall (s_t, a_t) \in \mathcal{S} \times \mathcal{A}$.
All in all, to develop the counterpart guarantees that will follow,
the preconditioner $\kappa_t$ must possess the following properties:
\begin{align}
&\dv*{\kappa_t}{s_t} = 0 \quad \text{and} \quad \dv*{\kappa_t}{a_t} = 0
\qquad
\blacktriangleright\text{\small{\textit{property 1, \textsc{eq}~\ref{purplenopresent}}}}
\nonumber \\
&\qquad
\blacktriangleright\text{\small{\textit{gave us \textsc{eq}~\ref{kappalip},
itself used in the proof (step 1) of \textsc{Theorem}~\ref{theorem1purple} (a)+(b)}}}
\nonumber \\
&\dv*{\kappa_{t+k+1}}{s_t} = 0 \quad \text{and} \quad \dv*{\kappa_{t+k+1}}{a_t} = 0
\qquad
\blacktriangleright\text{\small{\textit{property 2, \textsc{eq}~\ref{purplenopast}}}}
\nonumber \\
&\qquad
\blacktriangleright\text{\small{\textit{used in the proof of \textsc{Lemma}~\ref{lemmapurple},
itself then used to prove (step 2) \textsc{Theorem}~\ref{theorem1purple} (a)+(b)}}}
\nonumber
\end{align}
$\forall t \in [0, T] \cap \mathbb{N}, \forall k \in [0, T-t-1] \cap \mathbb{N},
\forall (s_t, a_t) \in \mathcal{S} \times \mathcal{A}$.
Note, the last two properties,
\textsc{eq}~\ref{purplenopresent} and \textsc{eq}~\ref{purplenopast},
can be condensed into,
$\forall t \in [0, T] \cap \mathbb{N}, \forall k \in [0, T-t] \cap \mathbb{N},
\forall (s_t, a_t) \in \mathcal{S} \times \mathcal{A}$:
\begin{align}
\dv*{\kappa_{t+k}}{s_t} = 0
\qquad \text{and} \qquad
\dv*{\kappa_{t+k}}{a_t} = 0
\qquad
\blacktriangleright\text{\small{\textit{property 1+2 condensed into one}}}
\end{align}
\paragraph{Property that $\kappa_t$ must have.}
\textbf{\emph{In plain English,
to get our guarantees,
we need the preconditioner to not depend on neither
current nor past states visited and actions taken by the agent.}}
Note, the property $\kappa_t \leq 1$ is only ever used in \textsc{Section}~\ref{morerobust},
and will not be leveraged anywhere else.
The developed theory will still hold if $\exists t \in [0, T] \cap \mathbb{N}$ such that $\kappa_t > 1$.

\paragraph{PURPLE in the broader algorithmic landscape.}
Setting aside the fact that $\kappa_t$ depends on a schedule indexed by the timestep $t$,
PURPLE has the effect of reducing the (policy) gradients received by the GAIL or SAM policy,
since it squashed the reward received by the agent.
This scales down the gradients traditionally designed for the policy.
The most direct adaptation of PURPLE to the GAN world would consist in scaling down
the output of the discriminator (from which the reward is directly crafted in GAIL and SAM).
The generator in a GAN is updated with gradients of the output of the discriminator \textit{w.r.t.}
its own parameters, similarly to how the actor is updated with gradients of the critic in an actor-critic.
Consequently, squashing the output of the discriminator squashes the gradients used by the generator,
which is equivalent to reducing the learning rate for the optimization of the generator
(assuming no exotic optimizer or regularizer are in use).

\begin{lemma}
\label{lemmapurple}
Let the MDP with which the agent interacts be deterministic,
with the dynamics of the environment determined
by the function $f: \mathcal{S} \times \mathcal{A} \rightarrow \mathcal{S}$.
The agent follows a deterministic policy $\mu: \mathcal{S} \rightarrow \mathcal{A}$
to map states to actions,
and receives rewards from
$r_\varphi: \mathcal{S} \times \mathcal{A} \rightarrow \mathbb{R}$
upon interaction.
The functions $f$, $\mu$ and $r_\varphi$ need be $C^0$ and differentiable
over their respective input spaces.
This property is satisfied by the usual neural network function approximators.
The ``almost-everywhere'' case can be derived from this lemma without major changes
(relevant when at least one activation function is only differentiable almost-everywhere, ReLU).
\textbf{\emph{(a)}} Under the previous assumptions,
for $k \in [0, T-t-1] \cap \mathbb{N}$ the following \textbf{(non-recursive) inequality} is verified:
\begin{align}
\lVert \nabla_{s,a}^t[\tilde{r}_\varphi]_{t+k+1} \rVert ^2_F
&\leq
\kappa_{t+k+1}^2
C_t
\, \lVert \nabla_{s,a}^{t+1}[r_\varphi]_{t+k+1} \rVert ^2_F
\end{align}
where $0 < \kappa_u \leq 1$ $\forall u \in [0, T] \cap \mathbb{N}$,
and $C_t \coloneqq A_t^2 \max(1, B_{t+1}^2)$,
$A_t$ and $B_t$ being defined as the supremum norms associated with the Jacobians of $f$
and $\mu$ respectively, with values in $\mathbb{R} \cup \{+\infty\}$:
\begin{align}
\forall t \in [0, T] \cap \mathbb{N} \text{,} \quad
\begin{cases}
A_t \coloneqq \lVert\nabla_{s,a}^t[f]_t\rVert _\infty
= \sup \big\{\lVert\nabla_{s,a}^t[f]_t\rVert _F \, : \, (s_t, a_t) \in
\mathcal{S} \times \mathcal{A} \big\} \\
B_t \coloneqq \lVert\nabla_s^t[\mu]_t\rVert _\infty
= \sup \big\{\lVert\nabla_s^t[\mu]_t\rVert _F \, : \, s_t \in
\mathcal{S} \big\}
\end{cases}
\end{align}
\textbf{\emph{(b)}} Additionally, by introducing \textbf{time-independent} upper bounds
$A, B \in \mathbb{R} \cup \{+\infty\}$
such that $\forall t \in [0, T] \cap \mathbb{N}$,
$A_t \leq A$ and $B_t \leq B$,
and $\kappa$ such that $\kappa_u \leq \kappa \leq 1$ $\forall u \in [0, T] \cap \mathbb{N}$,
the non-recursive inequality becomes:
\begin{align}
\lVert \nabla_{s,a}^t[\tilde{r}_\varphi]_{t+k+1} \rVert ^2_F
&\leq
\kappa^2
C
\, \lVert \nabla_{s,a}^{t+1}[r_\varphi]_{t+k+1} \rVert ^2_F
\end{align}
where $C \coloneqq A^2 \max(1, B^2)$ is the time-independent counterpart of $C_t$.
\end{lemma}

\emph{Proof of \textsc{Lemma}~\ref{lemmapurple}~\emph{(a)}.}
\emph{(a)} First, we take the derivative with respect to each variable separately:
\begin{align}
\nabla_s^t[\tilde{r}_\varphi]_{t+k+1}
&= \dv*{\tilde{r}_\varphi (s_{t+k+1}, a_{t+k+1})}{s_t} \\
&= \kappa_{t+k+1} \dv*{r_\varphi (s_{t+k+1}, a_{t+k+1})}{s_t}
\qquad
\blacktriangleright\text{\small{\textit{\textsc{eq}~\ref{purplenopast} (property 2), left}}} \\
&= \kappa_{t+k+1} \, \nabla_s^t[r_\varphi]_{t+k+1}
\qquad
\blacktriangleright\text{\small{\textit{repack}}}
\end{align}
\begin{align}
\nabla_a^t[\tilde{r}_\varphi]_{t+k+1}
&= \dv*{\tilde{r}_\varphi (s_{t+k+1}, a_{t+k+1})}{a_t} \\
&= \kappa_{t+k+1} \dv*{r_\varphi (s_{t+k+1}, a_{t+k+1})}{a_t}
\qquad
\blacktriangleright\text{\small{\textit{\textsc{eq}~\ref{purplenopast} (property 2), right}}} \\
&= \kappa_{t+k+1} \, \nabla_a^t[r_\varphi]_{t+k+1}
\qquad
\blacktriangleright\text{\small{\textit{repack}}}
\end{align}

By assembling the norm with respect to both input variables, we get:
\begin{align}
\lVert \nabla_{s,a}^t&[\tilde{r}_\varphi]_{t+k+1} \rVert ^2_F \nonumber \\
&= \lVert \nabla_s^t[\tilde{r}_\varphi]_{t+k+1} \rVert ^2_F
+ \lVert \nabla_a^t[\tilde{r}_\varphi]_{t+k+1} \rVert ^2_F \\
&= \kappa_{t+k+1}^2 \, \lVert \nabla_s^t[r_\varphi]_{t+k+1} \rVert ^2_F
+ \kappa_{t+k+1}^2 \, \lVert \nabla_a^t[r_\varphi]_{t+k+1} \rVert ^2_F \\
&= \kappa_{t+k+1}^2 \, \big(\lVert \nabla_s^t[r_\varphi]_{t+k+1} \rVert ^2_F
+ \lVert \nabla_a^t[r_\varphi]_{t+k+1} \rVert ^2_F\big) \\
&=
\kappa_{t+k+1}^2 \, \lVert \nabla_{s,a}^t[r_\varphi]_{t+k+1} \rVert ^2_F
\qquad
\blacktriangleright\text{\small{\textit{total norm}}}
\end{align}

As in \textsc{Lemma}~\ref{lemma}, let $A_t$, $B_t$ and $C_t$ be time-dependent quantities defined as:
\begin{align}
\forall t \in [0, T] \cap \mathbb{N} \text{,} \quad
\begin{cases}
A_t \coloneqq \lVert\nabla_{s,a}^t[f]_t\rVert _\infty
= \sup \big\{\lVert\nabla_{s,a}^t[f]_t\rVert _F \, : \, (s_t, a_t) \in
\mathcal{S} \times \mathcal{A} \big\} \\
B_t \coloneqq \lVert\nabla_s^t[\mu]_t\rVert _\infty
= \sup \big\{\lVert\nabla_s^t[\mu]_t\rVert _F \, : \, s_t \in
\mathcal{S} \big\} \\
C_t \coloneqq A_t^2 \max(1, B_{t+1}^2)
\end{cases}
\label{aandbpurple}
\end{align}

Finally, by injecting \textsc{eq}~\ref{lasteqlemma}, we directly obtain:
\begin{align}
\lVert \nabla_{s,a}^t[\tilde{r}_\varphi]_{t+k+1} \rVert ^2_F
&= \kappa_{t+k+1}^2 \, \lVert \nabla_{s,a}^t[r_\varphi]_{t+k+1} \rVert ^2_F \\
&\leq \kappa_{t+k+1}^2 \, A_t^2 \max(1, B_{t+1}^2) \, \lVert \nabla_{s,a}^{t+1}[r_\varphi]_{t+k+1} \rVert ^2_F
\qquad
\blacktriangleright\text{\small{\textit{\textsc{eq}~\ref{lasteqlemma}}}} \label{lasteqlemmapurple} \\
&= \kappa_{t+k+1}^2 \, C_t \, \lVert \nabla_{s,a}^{t+1}[r_\varphi]_{t+k+1} \rVert ^2_F
\qquad
\blacktriangleright\text{\small{$C_t$ \textit{definition}}}
\end{align}
which concludes the proof of \textsc{Lemma}~\ref{lemmapurple} \emph{(a)}. \qed

\emph{Proof of \textsc{Lemma}~\ref{lemmapurple}~\emph{(b)}.}
By introducing time-independent upper bounds $A$ and $B$ such that
$A_t \leq A$ and $B_t \leq B$
$\, \forall t \in [0, T] \cap \mathbb{N}$,
$C \coloneqq A^2 \max(1, B^2)$,
and $\kappa$ such that $\kappa_u \leq \kappa \leq 1$ $\forall u \in [0, T] \cap \mathbb{N}$,
we obtain, through \textsc{eq}~\ref{lasteqlemmapurple}:
\begin{align}
\lVert \nabla_{s,a}^t[\tilde{r}_\varphi]_{t+k+1} \rVert ^2_F
&\leq
\kappa^2 \, A^2 \max(1, B^2) \, \lVert \nabla_{s,a}^{t+1}[r_\varphi]_{t+k+1} \rVert ^2_F \\
&=
\kappa^2 \, C \, \lVert \nabla_{s,a}^{t+1}[r_\varphi]_{t+k+1} \rVert ^2_F
\end{align}
which concludes the proof of \textsc{Lemma}~\ref{lemmapurple} \emph{(b)}. \qed

\begin{theorem}[gap-dependent reward Lipschitzness]
\label{theorem1purple}
In addition to the assumptions laid out in lemma~\ref{lemmapurple},
we assume that the function $r_\varphi$ is $\delta$-Lipschitz
over $\mathcal{S} \times \mathcal{A}$.
Since $r_\varphi$ is $C^0$ and differentiable over $\mathcal{S} \times \mathcal{A}$,
this assumption can be written as
$\lVert \nabla_{s,a}^u[r_\varphi]_u \rVert _F \leq \delta$,
where $u \in [0, T] \cap \mathbb{N}$.
\textbf{\emph{(a)}} Then, under these assumptions, the following is verified:
\begin{align}
\lVert \nabla_{s,a}^t[\tilde{r}_\varphi]_{t+k} \rVert ^2_F
&\leq
\kappa_{t+k}^2 \, \delta ^2 \, \prod_{u=0}^{k-1} C_{t+u}
\end{align}
where $k \in [0, T] \cap \mathbb{N}$ and $C_v$ is defined as
in \textsc{Lemma}~\ref{lemmapurple}~\textit{(a)}, $\forall v \in [0, T] \cap \mathbb{N}$.
\textbf{\emph{(b)}} Additionally, by involving the time-independent upper bounds
introduced in \textsc{Lemma}~\ref{lemmapurple}~\textit{(b)}, we have the following:
\begin{align}
\lVert \nabla_{s,a}^t[\tilde{r}_\varphi]_{t+k} \rVert ^2_F
&\leq
\kappa^2 \, C^k \, \delta ^2
\end{align}
where $k \in [0, T] \cap \mathbb{N}$; $C$ and $\kappa$ are defined as
in \textsc{Lemma}~\ref{lemmapurple} \textit{(b)}.
\end{theorem}

\emph{Proof of \textsc{Theorem}~\ref{theorem1purple}~\emph{(a)}.}
We will prove \textsc{Theorem}~\ref{theorem1purple}~\emph{(a)} directly, not by induction
(\textsc{Lemma}~\ref{lemmapurple} proposes non-recursive inequalities,
one side containing $r_\varphi$, the other $\tilde{r}_\varphi$).
We want to prove the following \textsc{eq}~\ref{indhyp1purple}, $\forall v \in [0, T] \cap \mathbb{N}$:
\begin{align}
\lVert \nabla_{s,a}^t[\tilde{r}_\varphi]_{t+v} \rVert ^2_F
&\leq
\kappa_{t+v}^2 \, \delta ^2 \, \prod_{u=0}^{v-1} C_{t+u}
\label{indhyp1purple}
\end{align}
To do so, we will procede in two steps: \textit{(1)} prove it for $v=0$, and
\textit{(2)} prove it $\forall v \in [1, T] \cap \mathbb{N}$.

\emph{Step 1: case $v=0$.} When the gap $v=0$, \textsc{eq}~\ref{indhyp1purple} becomes
$\lVert \nabla_{s,a}^t[\tilde{r}_\varphi]_t \rVert ^2_F \leq \kappa_t^2 \, \delta ^2$,
$\forall t \in [0, T] \cap \mathbb{N}$,
which is verified by coupling
\textsc{Theorem}~\ref{theorem1purple}'s
main assumption about the $\delta$-Lipschitzness of $r_\varphi$
and the observation laid out in \textsc{eq}~\ref{kappalip}.

\emph{Step 2: case $v \in [1, T] \cap \mathbb{N}$.}
We start from the result we derived in \textsc{Lemma}~\ref{lemmapurple} \textit{(a)},
valid $\forall w \in [0, T-1] \cap \mathbb{N}$:
\begin{align}
\lVert \nabla_{s,a}^t[\tilde{r}_\varphi]_{t+w+1} \rVert ^2_F
&\leq
\kappa_{t+w+1}^2 \, C_t \, \lVert \nabla_{s,a}^{t+1}[r_\varphi]_{t+w+1} \rVert ^2_F
\qquad
\blacktriangleright\text{\small{\textit{\textsc{Lemma}~\ref{lemmapurple} (a)}}} \\
&\leq
\kappa_{t+w+1}^2 \, C_t \, \delta^2 \, \prod_{u=0}^{w-1} C_{t+1+u}
\qquad
\blacktriangleright\text{\small{\textit{\textsc{Theorem}~\ref{theorem1} (a), at $t+1$}}} \\
&=
\kappa_{t+w+1}^2 \, C_t \, \delta^2 \, \prod_{u=1}^{w} C_{t+u}
\qquad
\blacktriangleright\text{\small{\textit{index shift}}} \\
&=
\kappa_{t+w+1}^2 \, \delta^2 \, \prod_{u=0}^{w} C_{t+u}
\qquad
\blacktriangleright\text{\small{\textit{repack product}}}
\end{align}
This shows that \textsc{eq}~\ref{indhyp1purple} is verified when $v = w+1$,
$\forall w \in [0, T-1] \cap \mathbb{N}$.
\textsc{eq}~\ref{indhyp1purple} is therefore valid $\forall v \in [1, T] \cap \mathbb{N}$.

\emph{Conclusion.} We have shown that \textsc{eq}~\ref{indhyp1purple} is valid
$\forall v \in [0, T] \cap \mathbb{N}$, which concludes the proof of
\textsc{Theorem}~\ref{theorem1purple}~\emph{(a)}. \qed

\emph{Proof of \textsc{Theorem}~\ref{theorem1purple}~\emph{(b)}.}
We will prove \textsc{Theorem}~\ref{theorem1purple}~\emph{(b)} directly, not by induction
(\textsc{Lemma}~\ref{lemmapurple} proposes non-recursive inequalities,
one side containing $r_\varphi$, the other $\tilde{r}_\varphi$).
We want to prove the following \textsc{eq}~\ref{indhyp2purple}, $\forall v \in [0, T] \cap \mathbb{N}$:
\begin{align}
\lVert \nabla_{s,a}^t[\tilde{r}_\varphi]_{t+v} \rVert ^2_F
&\leq
\kappa^2 \, C^v \, \delta ^2
\label{indhyp2purple}
\end{align}
where $\kappa$ satisfies $\kappa_u \leq \kappa \leq 1$ $\forall u \in [0, T] \cap \mathbb{N}$.

To do so, we will procede in two steps: \textit{(1)} prove it for $v=0$, and
\textit{(2)} prove it $\forall v \in [1, T] \cap \mathbb{N}$.

\emph{Step 1: case $v=0$.} When the gap $v=0$, \textsc{eq}~\ref{indhyp2purple} becomes
$\lVert \nabla_{s,a}^t[\tilde{r}_\varphi]_t \rVert ^2_F \leq \kappa_t^2 \, \delta ^2
\leq \kappa^2 \, \delta ^2$,
$\forall t \in [0, T] \cap \mathbb{N}$,
which is verified by coupling
\textsc{Theorem}~\ref{theorem1purple}'s
main assumption about the $\delta$-Lipschitzness of $r_\varphi$,
the observation laid out in \textsc{eq}~\ref{kappalip},
and finally the definition of $\kappa$ (upper bound for all the $\kappa_u$'s).

\emph{Step 2: case $v \in [1, T] \cap \mathbb{N}$.}
We start from the result we derived in \textsc{Lemma}~\ref{lemmapurple} \textit{(b)},
valid $\forall w \in [0, T-1] \cap \mathbb{N}$:
\begin{align}
\lVert \nabla_{s,a}^t[\tilde{r}_\varphi]_{t+w+1} \rVert ^2_F
&\leq
\kappa^2 \, C \, \lVert \nabla_{s,a}^{t+1}[r_\varphi]_{t+w+1} \rVert ^2_F
\qquad
\blacktriangleright\text{\small{\textit{\textsc{Lemma}~\ref{lemmapurple} (b)}}} \\
&\leq
\kappa^2 \, C \, C^w \, \delta^2
\qquad
\blacktriangleright\text{\small{\textit{\textsc{Theorem}~\ref{theorem1} (b), at $t+1$}}} \\
&=
\kappa^2 \, C^{w+1} \, \delta^2
\qquad
\blacktriangleright\text{\small{\textit{repack product}}}
\end{align}
This shows that \textsc{eq}~\ref{indhyp2purple} is verified when $v = w+1$,
$\forall w \in [0, T-1] \cap \mathbb{N}$.
\textsc{eq}~\ref{indhyp2purple} is therefore valid $\forall v \in [1, T] \cap \mathbb{N}$.

\emph{Conclusion.} We have shown that \textsc{eq}~\ref{indhyp2purple} is valid
$\forall v \in [0, T] \cap \mathbb{N}$, which concludes the proof of
\textsc{Theorem}~\ref{theorem1purple}~\emph{(b)}. \qed

\begin{theorem}[state-action value Lipschitzness]
\label{theorem2purple}
We work under the assumptions laid out in
both \textsc{Lemma}~\ref{lemmapurple} and \textsc{Theorem}~\ref{theorem1purple}, and repeat
the main lines here
for \textsc{Theorem}~\ref{theorem2purple} to be self-contained:
a) The functions $f$, $\mu$ and $r_\varphi$ are $C^0$ and differentiable
over their respective input spaces,
and b) the function $r_\varphi$ is $\delta$-Lipschitz
over $\mathcal{S} \times \mathcal{A}$, i.e.
$\lVert \nabla_{s,a}^u[r_\varphi]_u \rVert _F \leq \delta$,
where $u \in [0, T] \cap \mathbb{N}$.
Then the quantity $\nabla_{s,a}^u[\widetilde{Q}_\varphi]_u$ exists
$\forall u \in [0, T] \cap \mathbb{N}$,
and verifies:
\begin{align}
\lVert \nabla_{s,a}^t[\widetilde{Q}_\varphi]_t \rVert _F
\leq
\left\{
\begin{aligned}
\kappa &\delta \, \sqrt{\frac{1 - \big( \gamma^2 C \big)^{T - t}}{1 - \gamma^2 C}},
&\qquad  &\text{if $\gamma^2 C \neq 1$} \\
\kappa &\delta\sqrt{T - t},
&\qquad &\text{if $\gamma^2 C = 1$}
\end{aligned}
\right.
\end{align}
$\forall t \in [0, T] \cap \mathbb{N}$,
where
$C \coloneqq A^2 \max(1, B^2)$, with $A$ and $B$ time-independent upper bounds of
$\lVert\nabla_{s,a}^t[f]_t\rVert _\infty$ and $\lVert\nabla_s^t[\mu]_t\rVert _\infty$
respectively
(see \textsc{eq}~\ref{aandbpurple} for definitions of the supremum norms),
and where $\kappa$ satisfies $\kappa_u \leq \kappa \leq 1$ $\forall u \in [0, T] \cap \mathbb{N}$.
\end{theorem}

\emph{Proof of \textsc{Theorem}~\ref{theorem2purple}.}
With finite horizon $T$, we have
$\widetilde{Q}_\varphi (s_t, a_t) \coloneqq \sum_{k=0}^{T-t-1} \gamma^k \, \tilde{r}_\varphi (s_{t+k}, a_{t+k})$,
$\forall t \in [0, T] \cap \mathbb{N}$,
since $f$, $\mu$, $r_\varphi$, and $\tilde{r}_\varphi$ (\textit{cf.}~\textsc{eq}~\ref{precondrew})
are all deterministic (no expectation).
Additionally, since $r_\varphi$ is assumes to be $C^0$ and differentiable
over $\mathcal{S} \times \mathcal{A}$,
$\widetilde{Q}_\varphi$ is by construction also $C^0$ and differentiable over $\mathcal{S} \times \mathcal{A}$.
Consequently, $\nabla_{s,a}^u[\widetilde{Q}_\varphi]_u$ exists, $\forall u \in [0, T] \cap \mathbb{N}$.
Since both $r_\varphi$ and $\widetilde{Q}_\varphi$ are scalar-valued (their output space is $\mathbb{R}$),
their Jacobians are the same as their gradients.
We can therefore use the linearity of the gradient operator:
$\nabla_{s,a}^t[\widetilde{Q}_\varphi]_t = \sum_{k=0}^{T-t-1} \gamma^k \, \nabla_{s,a}^t[\tilde{r}_\varphi]_{t+k}$,
$\forall t \in [0, T] \cap \mathbb{N}$.
\begin{align}
\lVert \nabla_{s,a}^t[\widetilde{Q}_\varphi]_t \rVert _F^2
&= \Bigg\lVert \sum_{k=0}^{T-t-1} \gamma^k \, \nabla_{s,a}^t[\tilde{r}_\varphi]_{t+k} \Bigg\rVert _F^2
\qquad
\blacktriangleright\text{\small{\textit{operator's linearity}}} \\
&\leq \sum_{k=0}^{T-t-1} \gamma^{2k} \, \lVert \nabla_{s,a}^t[\tilde{r}_\varphi]_{t+k} \rVert _F^2
\qquad
\blacktriangleright\text{\small{\textit{triangular inequality}}} \\
&\leq \sum_{k=0}^{T-t-1} \gamma^{2k} \, \kappa^2 \, C^k \, \delta ^2
\qquad
\blacktriangleright\text{\small{\textit{\textsc{Theorem}~\ref{theorem1purple} (b)}}} \label{theorem1purpleuse} \\
&= (\kappa \delta)^2 \sum_{k=0}^{T-t-1} \big( \gamma^2 C \big)^k
\label{geomsumpurple}
\end{align}
When $\gamma^2 C = 1$, we obtain $\lVert \nabla_{s,a}^t[\widetilde{Q}_\varphi]_t \rVert _F^2
= \delta^2 (T - t)$.
On the other hand, when $\gamma^2 C \neq 1$:
\begin{align}
\lVert \nabla_{s,a}^t[\widetilde{Q}_\varphi]_t \rVert _F^2
&\leq (\kappa \delta)^2 \, \frac{1 - \big( \gamma^2 C \big)^{T - t}}{1 - \gamma^2 C}
\qquad
\blacktriangleright\text{\small{\textit{finite sum of geometric series}}}
\end{align}
\begin{align}
\implies \quad
\lVert \nabla_{s,a}^t[\widetilde{Q}_\varphi]_t \rVert _F^2
\leq
\left\{
\begin{aligned}
& (\kappa \delta)^2 \, \frac{1 - \big( \gamma^2 C \big)^{T - t}}{1 - \gamma^2 C},
&\qquad  &\text{if $\gamma^2 C \neq 1$} \\
& (\kappa \delta)^2 (T - t),
&\qquad &\text{if $\gamma^2 C = 1$}
\end{aligned}
\right.
\end{align}
By applying $\sqrt{\cdot}$ (monotonically increasing) to the inequality,
we obtain the claimed result. \qed

Finally, we derive a corollary from \textsc{Theorem}~\ref{theorem2purple}
corresponding to the infinite-horizon regime.

\begin{corollary}[infinite-horizon regime]
\label{corollary1purple}
Under the assumptions of \textsc{Theorem}~\ref{theorem2purple},
including that $r_\varphi$ is $\delta$-Lipschitz
and that $\tilde{r}_\varphi$ is defined as in \textsc{eq}~\ref{precondrew}
over $\mathcal{S} \times \mathcal{A}$,
and assuming that $\gamma^2 C < 1$, we have, in the infinite-horizon regime:
\begin{align}
\lVert \nabla_{s,a}^t[\widetilde{Q}_\varphi]_t \rVert _F
&\leq \frac{\kappa \delta}{\sqrt{1 - \gamma^2 C}}
\end{align}
which translates into $\widetilde{Q}_\varphi$ being $\frac{\kappa \delta}{\sqrt{1 - \gamma^2 C}}$-Lipschitz
over $\mathcal{S} \times \mathcal{A}$.
\end{corollary}

\emph{Proof of \textsc{Corollary}~\ref{corollary1purple}.}
By following the proof of \textsc{Corollary}~\ref{corollary1},
using \textsc{Theorem}~\ref{theorem1purple} instead of \textsc{Theorem}~\ref{theorem1},
we arrive directly at the claimed result. \qed

\begin{remark}
\label{remarkpurple}
Say we were to write a proof analogous to the one laid out right above for \textsc{Theorem}~\ref{theorem2purple},
but using the time-\emph{dependent} version of \textsc{Theorem}~\ref{theorem1purple}
instead of the time-\emph{independent} version that we used in \textsc{eq}~\ref{theorem1purpleuse}
(version \ref{theorem1purple}~\textit{(a)} instead of \ref{theorem1purple}~\textit{(b)}).
Despite not being identifiable as a finite or infinite
sum of geometric series, the expression we would get instead of \textsc{eq}~\ref{theorem1purpleuse}
not only is a tighter bound by construction,
but it also has an interesting form:
\begin{align}
\lVert \nabla_{s,a}^t[\widetilde{Q}_\varphi]_t \rVert _F^2
&\leq \sum_{k=0}^{T-t-1}
\Bigg[
\gamma^{2k} \, \kappa_{t+k}^2 \, \delta ^2 \, \prod_{u=0}^{k-1} C_{t+u}
\Bigg]
\qquad
\blacktriangleright\text{\small{\textit{\textsc{Theorem}~\ref{theorem1purple} (a)}}}
\end{align}
Going through the first operands of the sum, and looking solely at the ``$\kappa$'' and ``$C$''
factors, we have the following:
\begin{align}
\kappa_t^2
\rightarrow \kappa_{t+1}^2 \, C_t
\rightarrow \kappa_{t+2}^2 \, C_t C_{t+1}
\rightarrow \kappa_{t+3}^2 \, C_t C_{t+1} C_{t+2}
\rightarrow \ldots
\rightarrow \kappa_{T}^2 \, C_t C_{t+1} C_{t+2} \ldots C_{T-1}
\end{align}
This observation tells us that, in the derived Lipschitz constant of $\widetilde{Q}_\varphi$,
the reward preconditioner $\kappa_t$ at time $t$ can compensate for \emph{all the past}
values $\{C_v \, | \, v<t\}$.
Intuitively, the more we wait to reduce $\kappa_t$, the more the \emph{next} $\kappa_t$'s
will need to compensate for the ``negligence'' of their predecessors.
Note, the product of $\{C_v \, | \, v<t\}$ compounds quickly.
\end{remark}

\subsection{Discussion II: implications and limitations of the theoretical guarantees}
\label{discussionpurple}

\subsubsection{Provably more robust}
\label{morerobust}

Given that, in this work, we aligned the notion of robustness of a function approximator
with the value of its Lipschitz constant (\emph{more} robust means \emph{lower} Lipschitz constant,
\textit{cf.}~\textsc{Section}~\ref{bridge}), and given that $\kappa_t$'s upper bound $\kappa$
verifies $\kappa \leq 1$ (\textit{cf.}~\textsc{Lemma}~\ref{lemmapurple}), we can write,
from the result of \textsc{Corollary}~\ref{corollary1purple}:
\begin{align}
\lVert \nabla_{s,a}^t[\widetilde{Q}_\varphi]_t \rVert _F
\leq \frac{\kappa \delta}{\sqrt{1 - \gamma^2 C}}
= \kappa \, \Delta_\infty
\coloneqq \widetilde{\Delta}_\infty
\: \leq \Delta_\infty
\label{tildeqlip}
\end{align}
where $\Delta_\infty \coloneqq \delta / \sqrt{1 - \gamma^2 C}$
is the upper bound of $Q_\varphi$'s Lipschitz constant that we derived in \textsc{Corollary}~\ref{corollary1}.
Note, all of what is written in this remark concerns the infinite-horizon regime,
but one can derive the finite-horizon counterpart trivially --- using
\textsc{Theorem}~\ref{theorem2} instead of \textsc{Corollary}~\ref{corollary1},
and \textsc{Theorem}~\ref{theorem2purple} instead of \textsc{Corollary}~\ref{corollary1purple} ---
to arrive at the same conclusion:
$\widetilde{Q}_\varphi$ has a lower derived Lipschitz constant upper bound than $Q_\varphi$
by a factor of $\kappa \leq 1$ and is therefore \emph{provably more robust} than $Q_\varphi$.
In other words, employing the simple PURPLE reward preconditioning to SAM has the effect of
making the learned Q-value provably more robust.

\subsubsection{Detached guide}

Consider the following particular form for $\kappa_t$,
$\forall t \in [0, T] \cap \mathbb{N},
\forall (s_t, a_t) \in \mathcal{S} \times \mathcal{A}$:
\begin{align}
\kappa_t \coloneqq \exp (-\alpha \, \epsilon_t)
\quad \implies \quad
\tilde{r}_\varphi (s_t,a_t)
\coloneqq \kappa_t \, r_\varphi (s_t,a_t)
\coloneqq \exp (-\alpha \, \epsilon_t) \, r_\varphi (s_t,a_t)
\label{precondrewexp}
\end{align}
where $\alpha$ is an inverse temperature hyper-parameter involved in the definition of the kernel of
the Boltzmann or Gibbs probability distribution $\kappa_t \coloneqq \exp (-\alpha \, \epsilon_t)$,
(hence $0 < \kappa_t \leq 1$), and where
$\epsilon_t \geq 0$ for now depicts an arbitrary non-negative energy function.
$\kappa_t$ is non-normalized, and as such, it is \emph{not} a probability \textit{per se}.
Nonetheless, it still echoes the propensity or tendency of the state-action pair $(s_t,a_t)$
to possess the property described by the non-negative energy $\epsilon_t$, which we define momentarily.
Low values of $\epsilon_t \geq 0$
will push the preconditioner towards the upper limit $\kappa_t \to 1$,
while high energy values will make it tend towards the lower limit $\kappa_t \to 0$ with $\kappa_t > 0$.
Equivalently, the preconditioned reward $\tilde{r}_\varphi$ will verify the approximate identity
$\tilde{r}_\varphi (s_t,a_t) \approx r_\varphi (s_t,a_t)$ whenever $\epsilon_t$
approaches zero (from above), and $\tilde{r}_\varphi (s_t,a_t) \approx 0$ whenever
the energy $\epsilon_t$ grows towards higher levels.
Under this orchestration, we need
$\dv*{\epsilon_{t+k}}{s_t} = 0$ and $\dv*{\epsilon_{t+k}}{a_t} = 0$
to be satisfied
$\forall t \in [0, T] \cap \mathbb{N}, \forall k \in [0, T-t] \cap \mathbb{N},
\forall (s_t, a_t) \in \mathcal{S} \times \mathcal{A}$
for the derived robustness guarantees to be readily applicable
(we laid out the properties $\kappa_t$ must possess in \textsc{Section}~\ref{purpleandres},
right before exposing \textsc{Lemma}~\ref{lemmapurple}).

In particular,
the soft approximate $\mathfrak{C}$-validity pseudo-indicator
(\textit{cf.}~\textsc{eq}~\ref{pseudoindicator})
is an instantiation of the $\kappa_t$ form laid out in \textsc{eq}~\ref{precondrewexp},
where $\alpha=1$ for the inverse temperature, and
$\epsilon_t = \max (0, \lVert \nabla_{s_t,a_t} \, D_\varphi (s_t,a_t) \rVert - k)^2$ for the energy.
In such an instance, $\tilde{r}_\varphi (s_t,a_t) \approx r_\varphi (s_t,a_t)$
whenever the pair $(s_t,a_t)$ is approximately $\mathfrak{C}$-valid, formally,
$\lVert \nabla_{s,a}^t[D_\varphi]_t \rVert _F \leq k$.
Conversely, in the extreme scenario where $\lVert \nabla_{s,a}^t[D_\varphi]_t \rVert _F \gg k$,
$\epsilon_t$ grows large,
$\kappa_t$ is approximately equal to $0$,
and $\tilde{r}_\varphi (s_t,a_t) \approx 0$.
As such, in effect, the agent's policy $\mu_\theta$ is punished for selecting actions that do not satisfy the
approximate $\mathfrak{C}$-validity condition above.
Besides, it is punished in accordance to how far outside the allowed
range, $[0,k]$, the norm of the Jacobian of $D_\varphi$ gets.
Nonetheless, in this particular instance,
the empirical observations we have made in \textsc{Section}~\ref{understand}
attest to the fact that, provided the right choice of $\lambda$ scaling factor and $\zeta$ distribution
(both characterizing the gradient penalization),
the approximate $\mathfrak{C}$-validity constraint $\lVert \nabla_{s,a}^t[D_\varphi]_t \rVert _F \leq k$
can easily be satisfied $100\%$ of the time by \emph{only} regularizing $D_\varphi$.
For $D_\varphi$'s $k$-Lipschitzness to be ensured,
there is therefore no need to further alter the rewards provided to the agent's policy $\mu_\theta$
through PURPLE's pessimistic reward preconditioning.
Note, however, that under such a $\epsilon_t$ formulation, we see that we clearly have
$\dv*{\epsilon_{t+k}}{s_t} \neq 0$ and $\dv*{\epsilon_{t+k}}{a_t} \neq 0$,
$\forall t \in [0, T] \cap \mathbb{N}, \forall k \in [0, T-t] \cap \mathbb{N},
\forall (s_t, a_t) \in \mathcal{S} \times \mathcal{A}$.
While this does not mean that the studied entities are not robust,
it prevents us from applying our derived results to guarantee such robustness.

Generally speaking, we will probably make the same observation whenever $\epsilon_t$ is defined from a constraint
we want to enforce on a learned function approximation, for regularization purposes.
Indeed, verifying said desideratum on the function approximator directly via the application of a regularizer seems
to always be the easiest (since most direct) solution to encourage the satisfaction of a constraint on
a \emph{differentiable} function (\textit{e.g.}~$D_\varphi$, $\mu_\theta$).
Constraints involving the Jacobian of a (\textit{a fortioni} differentiable) function of the learned system
(\textit{e.g.}~$\lVert \nabla_{s,a}^t[D_\varphi]_t \rVert _F \leq k$)
is a particular case of the general class of constraints for which \emph{direct} regularization is
\textit{a priori} prefereable to an analogous reward shaping as dictated by \textsc{eq}~\ref{precondrewexp}.
On the flip side, due to the fact that the reward --- albeit learned as a parametric function ---
is treated as an input in our computational graph,
it is not differentiated through and \emph{can} consequently
be augmented with non-differentiable nodes through the design of $\epsilon_t$.
In other words, even if it is preferable to apply regularization directly the objective of the regularized
function approximator for it to satisfy some constraint,
it might not always be possible to do so directly.
In that case, guiding the policy towards areas of the state-action landscape that satisfy said constraint
could be a surrogate solution, albeit far less preferable than acting on the targeted approximator directly.

As such, by aligning $\epsilon_t$ with said constraint, \textsc{eq}~\ref{precondrewexp} offers a way for the policy
to act in view of the satisfaction of said constraint \emph{while} enjoying the considerable advantage
of being able to treat $\epsilon_t$ as a \emph{black box}.
We will leverage this \emph{universality} in the next discussion point.

\subsubsection{Partial compensation of compounding variations}
\label{partialcompensate}

In reaction to the theoretical robustness guarantees derived in
\textsc{Theorem}~\ref{theorem2} and \textsc{Corollary}~\ref{corollary1},
we have discussed earlier in \textsc{Section}~\ref{compoundvars} that,
if the variations \emph{in space} of the policy or the dynamics are large
in the early stage of an episode (\textit{i.e.}~when $0 \leq t \ll T$),
then $\Delta_t$ (the variation bound on $Q_\varphi$) might explode.
As results, $\lVert \nabla_{s,a}^t[Q_\varphi]_t \rVert _F$ would then be unbounded,
leaving us unable to guarantee the robustness of the learned Q-value $Q_\varphi$.
The earlier large variations in either or both the policy and dynamics manifest,
the more likely these variations are to compound to unreasonably high levels.
Concretely, the degree of such compounding variations in space is entirely determined by
the operand $\gamma^2 C$ that appears in the variation bounds derived in both
\textsc{Theorem}~\ref{theorem2} and \textsc{Corollary}~\ref{corollary1}.
The exact same line of reasoning holds for the variation bounds laid out later in
\textsc{Section}~\ref{purpleandres}, in both
\textsc{Theorem}~\ref{theorem2purple} and \textsc{Corollary}~\ref{corollary1purple} respectively.
These guarantees unanimously agree on the critical role that $C$ plays in
the robustness bounds, which we here called variation bounds indifferently.
Loosely, \emph{high} values of $C$ prevent $Q_\varphi$ from enjoying the
Lipschitzness guarantees
laid out in \textsc{Section}~\ref{theory} and \textsc{Section}~\ref{purpleandres}.
As such, it is paramount to devise a way to keep $C$ in check
by somewhat controling its magnitude,
thereby preventing it from voiding our theoretical guarantees
and from adopting a brittle behavior.
We defined $C$ in \textsc{Lemma}~\ref{lemma} \textit{(b)} as
$C \coloneqq A^2 \max(1, B^2)$, where $A$ and $B$ are time-independent upper bounds of
the supremum Frobenius norms of the Jacobians of the dynamics $f$ and the policy $\mu$,
$\lVert\nabla_{s,a}^t[f]_t\rVert _\infty$ and $\lVert\nabla_s^t[\mu]_t\rVert _\infty$,
respectively
(\textit{cf.}~\textsc{eq}~\ref{aandb} for definitions of the supremum norms $\lVert \cdot \rVert _\infty$).
Simply,
$\forall t \in [0, T] \cap \mathbb{N},
\forall (s_t, a_t) \in \mathcal{S} \times \mathcal{A}$,
$\lVert\nabla_{s,a}^t[f]_t\rVert _\infty \leq A$
and $\lVert\nabla_s^t[\mu]_t\rVert _\infty \leq B$.
As such, to devise a way to limit the magnitude of $C$,
we seek ways to limit the respective magnitudes of the $A$ and $B$ majorants.
Similarly to the learned surrogate reward core $D_\varphi$, the policy $\mu_\theta$
followed by the agent (of which $\mu$ is a placeholder) is learned as a parametric
function approximator, enabling us to tame $B$ by applying a gradient penalty regularizer
\emph{directly} on the policy (exactly like we already do to ensure that $D_\varphi$
remains $k$-Lipschitz-continuous).

By contrast, we can not tame $A$ the same way (via direct regularization applied onto $f$),
due to the transition function $f$ of the world (whether real or simulated)
being a black box that we can not even query at will.
Not only is $f$ non-differentiable (the real world never is; non-trivial simulated worlds virtually never are),
but we also can \emph{not} evaluate it at \textit{any state-action pair whenever we want}.
Our desideratum then ultimately boils down to finding a way to keep $A$ in check,
since the usual candidate to enforce Lipschitzness (applying a regularizer on the Jacobian directly) ---
which is the preferable option by far for $D_\varphi$ and $\mu_\theta$ ---
is out of the question for $f$, as we have established.
Despite the fact that, by nature, we can not change $f$ in the MDP $\mathbb{M}$,
we \emph{can} change the transition function $f'$ that
effectively takes the place of $f$ in practice
and underlies the effectively observed MDP $\mathbb{M}'$ by urging the agent's policy $\mu_\theta$ to avoid
areas of the state-action landscape $\mathcal{S} \times \mathcal{A}$
that display high $\lVert\nabla_{s,a}^t[f']_t\rVert _\infty$ values.
In fact, $f'$ changes continually ($f'$ is non-stationary)
throughout the learning process as the preferences of the agent
evolve across learning episodes.
It is therefore fair to posit that we can devise a way to skew the policy towards areas
of $\mathcal{S} \times \mathcal{A}$ where
$\lVert\nabla_{s,a}^t[f']_t\rVert _\infty$ is tightly upper-bounded.
As such, we can keep $A$ in check by keeping $\lVert\nabla_{s,a}^t[f']_t\rVert _\infty$ in check in practice,
which can be approximately achieved by keeping $\lVert\nabla_{s,a}^t[f_\psi]_t\rVert _\infty$ in check,
where $f_\psi : \mathcal{S} \times \mathcal{A} \to \mathcal{S}$
is a learned functional approximation of the effective dynamics $f'$.

\textit{In fine},
we urge the constraint $\lVert\nabla_{s,a}^t[f]_t\rVert _\infty \leq A$ to be satisfied
by encouraging $\mu_\theta$ to avoid areas where $\lVert\nabla_{s,a}^t[f_\psi]_t\rVert _\infty$
is high, which itself can be relaxed into $\lVert\nabla_{s,a}^t[f_\psi]_t\rVert _F$.
Note, even if $f_\psi$ is differentiable, regularizing it via gradient penalization does \emph{not}
have any effect on the value of $\lVert\nabla_{s,a}^t[f']_t\rVert _\infty$,
since the agent does \emph{not} interact with $f_\psi$, but with $f'$.
For our line of reasoning to hold, we want $\lVert\nabla_{s,a}^t[f_\psi]_t\rVert _F$ to be
a high-fidelity depiction of $\lVert\nabla_{s,a}^t[f']_t\rVert _\infty$.

We maintain the parametric model $f_\psi$
because it allows us to approximate the norm of the Jacobian of the dynamics
wherever we want, whenever we want.
In order for $\mu_\theta$ to avoid areas where $\lVert\nabla_{s,a}^t[f_\psi]_t\rVert _F$ is high,
we leverage the universal preconditioner form exhibited in \textsc{eq}~\ref{precondrewexp}.
Concretely, we reward the agent \emph{less} for
\emph{not} navigating areas of $\mathcal{S} \times \mathcal{A}$ that satisfy the constraint
$\lVert\nabla_{s,a}^t[f_\psi]_t\rVert _F \leq \tau$.
The Lipschitz constant $\tau$ we want to enforce onto $f_\psi$ is a hyper-parameter that must be tuned,
like $k$ for $D_\varphi$.
We push $\mu_\theta$ towards areas where $\lVert\nabla_{s,a}^t[f_\psi]_t\rVert _F \leq \tau$
(where $f_\psi$ is $\tau$-Lipschitz-continuous, thereby also satisfying the premise of the guarantees)
by defining the energy function $\epsilon_t^\psi$ in the
\emph{model-based} preconditioner $\kappa_t^\psi$ as a one-sided gradient penalty, as follows:
\begin{align}
\tilde{r}_\varphi^\psi (s_t,a_t)
\coloneqq \kappa_t^\psi \, r_\varphi (s_t,a_t)
\quad \text{where} \quad
\begin{cases}
\kappa_t^\psi &\coloneqq \max\big(\kappa_\text{min}, \exp \big(-\alpha \, \epsilon_t^\psi\big)\big)
\quad \text{with} \\
\epsilon_t^\psi &\coloneqq
\max \big(0, \lVert\nabla_{s,a}^t[f_\psi]_t\rVert _F - \tau\big)^2 \Big/ \sigma_\textsc{on}^\psi
\end{cases}
\label{precondrewf1}
\end{align}
\begin{align}
\iff \qquad
\tilde{r}_\varphi^\psi(s_t,a_t)
\coloneqq \max\bigg(\kappa_\text{min}, \exp \bigg(- \frac{\alpha}{\sigma_\textsc{on}^\psi}
\max \big(0, \lVert\nabla_{s,a}^t[f_\psi]_t\rVert _F - \tau\big)^2\bigg)\bigg)
\, r_\varphi (s_t,a_t)
\label{precondrewf2}
\end{align}
$\forall t \in [0, T] \cap \mathbb{N},
\forall (s_t, a_t) \in \mathcal{S} \times \mathcal{A}$,
where
$\sigma_\textsc{on}^\psi$ denotes an online, running estimate of
the standard deviation of
$\max (0, \lVert \nabla_{s_t,a_t} \, f_\psi (s_t,a_t) \rVert _F - \tau)^2$.
For completeness, we remind here that we used the same online normalization technique
in our RED experiments
(\textit{cf.} \textsc{Section}~\ref{empres1}),
inspired by the discussion laid out in in \cite{Burda2018-vl}
on the importance of such normalization technique
when the reward is grounded on a prediction loss.
Considering the edge cases, and omitting here the clipping to $\kappa_\text{min}$,
when $\epsilon_t^\psi$ is close to zero,
$\kappa_t^\psi$ is approximately equal to $1$,
\textit{i.e.} $\tilde{r}_\varphi (s_t,a_t) \approx r_\varphi (s_t,a_t)$
(\textit{cf.}~\textsc{eq}~\ref{precondrewf1}, \ref{precondrewf2}).
Conversely, in the extreme scenario where $\epsilon_t^\psi$ is very large
(\textit{i.e.}~$\lVert\nabla_{s,a}^t[f_\psi]_t\rVert _F \gg \tau$),
$\kappa_t^\psi$ is approximately equal to $0$,
and $\tilde{r}_\varphi (s_t,a_t) \approx 0$.

Looking at the model-based instantiation of PURPLE laid out in \textsc{eq}~\ref{precondrewf1},
and specifically of the form exhibited in \textsc{eq}~\ref{tildeqlip},
we see that the energy $\epsilon_t^\psi$ depends on the current state $s_t$ and action $a_t$.
Indeed, from the definitions of $\epsilon_t^\psi$ and $\kappa_t^\psi$, we immediately see that
$\dv*{\epsilon_{t+k}^\psi}{s_t} \neq 0$ and $\dv*{\epsilon_{t+k}^\psi}{a_t} \neq 0$,
which direclty leads to
$\dv*{\kappa_{t+k}^\psi}{s_t} \neq 0$ and $\dv*{\kappa_{t+k}^\psi}{a_t} \neq 0$,
$\forall t \in [0, T] \cap \mathbb{N}, \forall k \in [0, T-t] \cap \mathbb{N},
\forall (s_t, a_t) \in \mathcal{S} \times \mathcal{A}$.
As such,
the crafted preconditioner does not satisfy the eligibily conditions
for the derived theoretical guarantees to be applicable,
which were represented in condensed form in \textsc{Section}~\ref{purpleandres},
right before exposing \textsc{Lemma}~\ref{lemmapurple}.
If we had used the \emph{supremum} Frobenius norm
$\lVert\nabla_{s,a}^t[f_\psi]_t\rVert _\infty$
to formulate $\epsilon_t^\psi$ instead of relaxing it to
$\lVert\nabla_{s,a}^t[f_\psi]_t\rVert _F$, its non-supremum counterpart,
$\epsilon_t^\psi$ would \emph{not} depend on $s_t$ and $a_t$ (or any visited state or picked action),
and our robustness guarantees would be readily applicable.
Still, such a supremum Frobenius norm is intractable in practice.
In order for us to be able to evaluate the developed prototype empirically,
we resorted to the obvious tractable relaxation consisting in simply dropping the supremum altogether
for this diagnostics-oriented case.

Now that we have laid out how the pessimistic model-based preconditioner $\kappa_t^\psi$ impacts the reward
received by the agent artificially upon interaction, we consider how this preconditioning affects the
Lipschitz constant of $\widetilde{Q}_\varphi$ in the infinite-horizon setting,
denoted by $\widetilde{\Delta}_\infty$ (\textit{cf.}~\textsc{eq}~\ref{tildeqlip}).
As $\lVert\nabla_{s,a}^t[f]_t\rVert _\infty$ grows larger,
its upper-bound $A$ grows larger.
Assuming $B$ (upper-bounding $\lVert\nabla_s^t[\mu]_t\rVert _\infty$) remains unaffected and remains constant,
larger values of $A$ cause larger values of $C \coloneqq A^2 \max(1, B^2)$,
which in turn push the denominator of the Lipschitz constant
$\widetilde{\Delta}_\infty^\psi \coloneqq \kappa_t^\psi \delta / \sqrt{1 - \gamma^2 C}$
towards $0$ from above,
exposing $\widetilde{\Delta}_\infty^\psi$ to diverge to $+\infty$.
Without preconditioning
($\kappa_t^\psi = 1$),
the task of compensating for such a low-valued denominator would be left to $\delta$ alone,
and picking $\delta \approx 0$ would be the only way to maintain the robustness bound from diverging.
With preconditioning however, we can also try to prevent it from diverging with
the preconditioner $\kappa_t^\psi$, whose value can be set far more finely (\emph{per} timestep).
Specifically, with the $\kappa_t^\psi$ formulation laid out in \textsc{eq}~\ref{precondrewf1} and \ref{precondrewf2},
and assuming $\lVert\nabla_{s,a}^t[f_\psi]_t\rVert _F$ approximates
$\lVert\nabla_{s,a}^t[f]_t\rVert _\infty$ well
--- \textit{i.e.} $\lVert\nabla_{s,a}^t[f_\psi]_t\rVert _F$
mirrors the behavior of $\lVert\nabla_{s,a}^t[f]_t\rVert _\infty$,
we hold an analogous line of reasoning for the \emph{numerator} of $\widetilde{\Delta}_\infty^\psi$.
As $\lVert\nabla_{s,a}^t[f]_t\rVert _\infty$ grows larger,
$\lVert\nabla_{s,a}^t[f_\psi]_t\rVert _F$ grows larger
(with we can translate into $\lVert\nabla_{s,a}^t[f_\psi]_t\rVert _F \gg \tau$),
which consequently pushes the preconditioner $\kappa_t^\psi$
towards $0$ from above.
As such, the premise \textit{``$\lVert\nabla_{s,a}^t[f]_t\rVert _\infty$ grows larger''}
pushes both the numerator and denominator of $\widetilde{\Delta}_\infty^\psi$ towards $0$ from above,
taming the quotient in effect.
Nonetheless, note, we can not \emph{eliminate} the influence of $\lVert\nabla_{s,a}^t[f]_t\rVert _\infty$ on the bound.
Still, the \emph{partial compensation} of the detrimental impact
of $\lVert\nabla_{s,a}^t[f]_t\rVert _\infty$ on $\widetilde{\Delta}_\infty^\psi$ ---
that we were able to secure by proposing the model-based pessimistic reward preconditioning $\kappa_t^\psi$
(\textit{cf.}~\textsc{eq}~\ref{precondrewf1}, \ref{precondrewf2}) ---
can be tuned extensively in practice
to achieve the desired level of compensation.
We used $\kappa_\text{min} = 0.7$, $\alpha = 1$, and $\tau \in \{6,7\}$ in the experiments
we conducted to showcase how the proposed model-based reward preconditioning laid out above can help us
achieve our robustness desideratum.

\begin{figure}
  \center
  \begin{subfigure}[t]{0.33\textwidth}
    \center\scalebox{0.18}[0.18]{\includegraphics{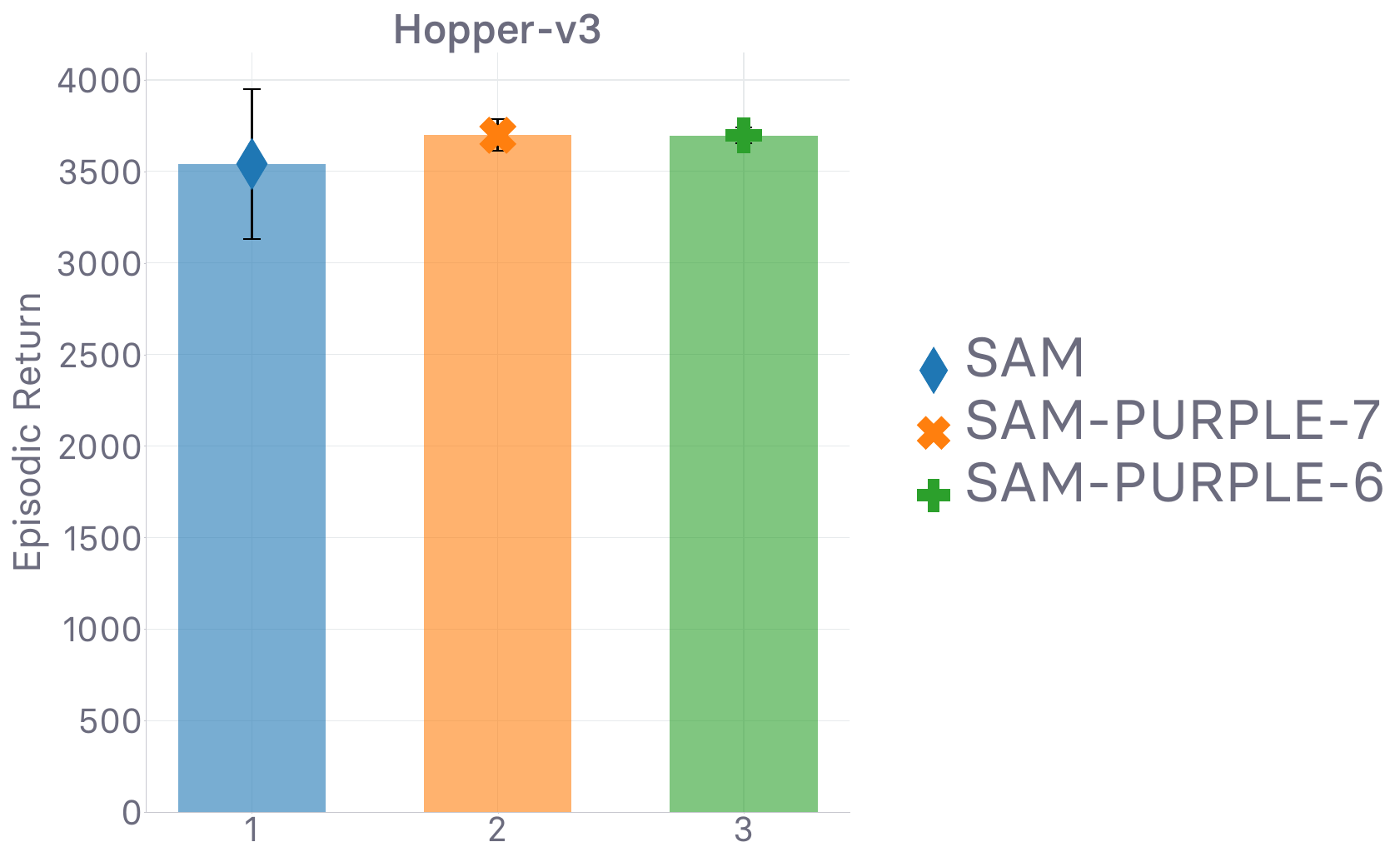}}
    \caption{Return values \textit{(higher is better)}}
  \end{subfigure}
  \begin{subfigure}[t]{0.33\textwidth}
    \center\scalebox{0.18}[0.18]{\includegraphics{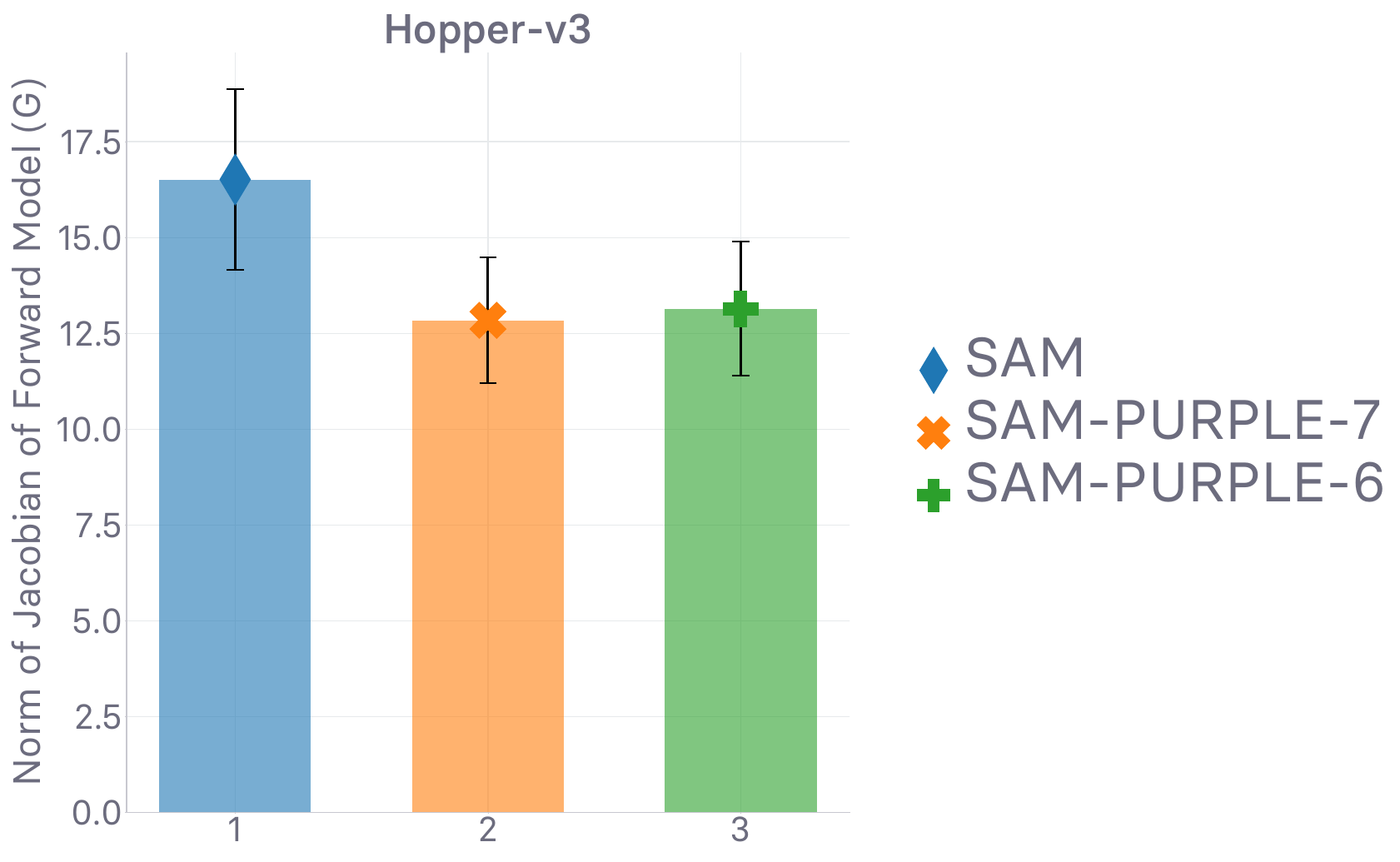}}
    \caption{$G$ values \textit{(lower is better)}}
  \end{subfigure}
  \begin{subfigure}[t]{0.33\textwidth}
    \center\scalebox{0.18}[0.18]{\includegraphics{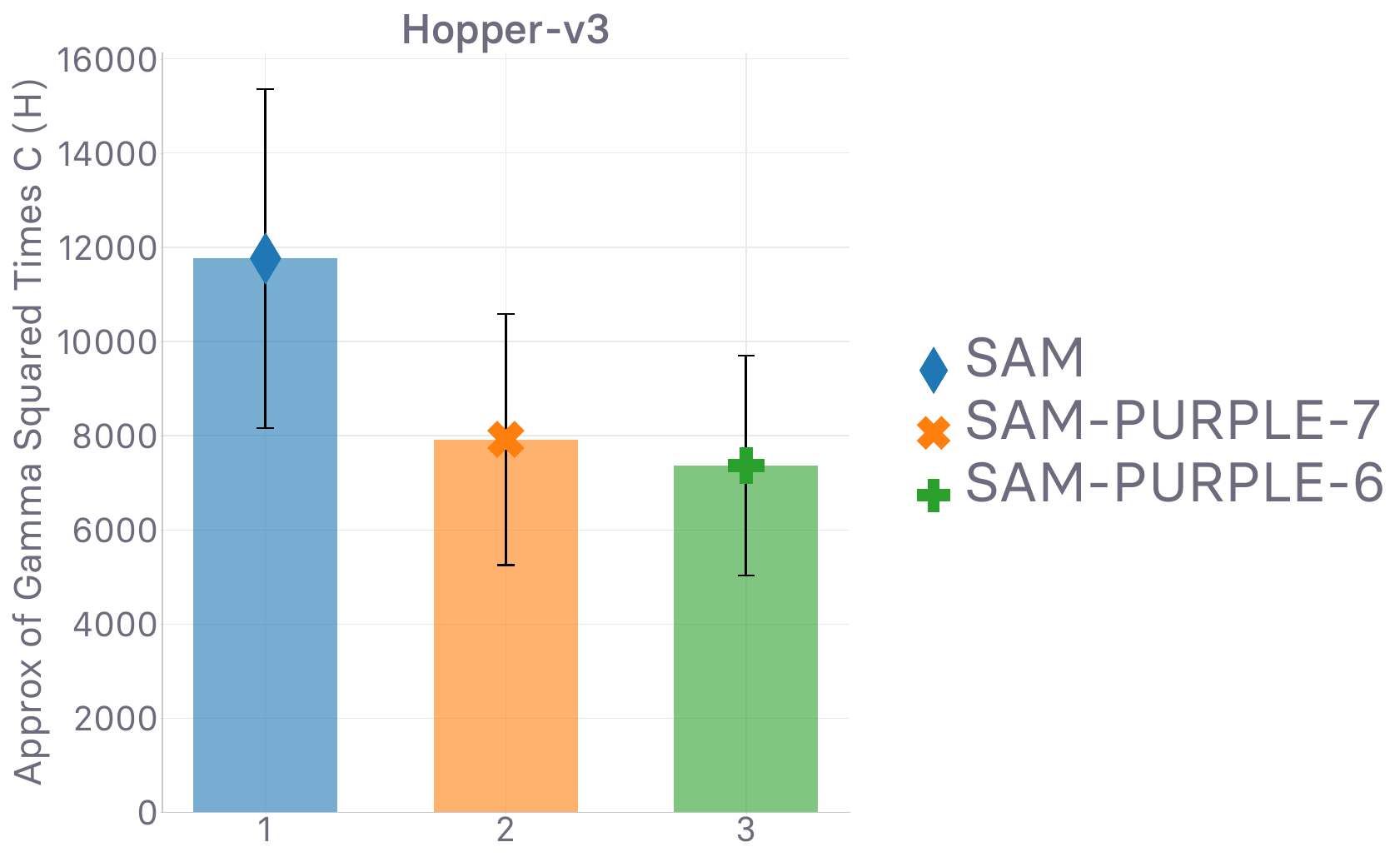}}
    \caption{$H$ values \textit{(lower is better)}}
  \end{subfigure}
  \caption{Empirical evaluation of
  \textit{(a)} the empirical return,
  \textit{(b)} the norm of the Jacobian of the forward model $f_\psi$ defined by
  $G \coloneqq \lVert\nabla_{s,a}^t[f_\psi]_t\rVert _F$, and
  \textit{(c)} the approximation of $\gamma^2 C$ defined by
  $H \coloneqq \gamma^2 \lVert\nabla_{s,a}^t[f_\psi]_t\rVert _F^2 \max (1, \lVert\nabla_s^t[\mu_\theta]_t\rVert _F^2)$.
  \textit{SAM-PURPLE-7} and \textit{SAM-PURPLE-6} are two instantiations of SAM
  (\textit{cf.} \textsc{Algorithm}~\ref{algosam}),
  augmented with
  the \emph{model-based} instantiation of PURPLE
  whose template is laid out in
  \textsc{eq}~\ref{precondrewf1} and \ref{precondrewf2},
  with $\tau=7$ and $\tau=6$ respectively.
  We indicate how to read the plots (whether \textit{lower} or \textit{higher} is better) in the caption of
  each column.
  Despite displaying overlapping return curves, note how \emph{tighter} the standard deviation envelope is
  for PURPLE runs.
  Runtime is 96 hours.}
  \label{purplepsi}
\end{figure}

Since we aim to \textit{showcase} its potential benefits, as opposed to convince the reader to plug this
preconditioning method in every future architecture, we conducted \textit{illustrative} experiments
only in the \texttt{Hopper} environment (neither the easiest, nor the hardest among the ones
considered, \textit{cf.}~\textsc{Table}~\ref{envtable}).
Note, when it comes to $D_\varphi$'s gradient penalty regularization,
we use the default $\zeta$ and $\lambda$ (\textit{cf.}~\textsc{Section}~\ref{gradpenrl}):
the directed $\zeta$ distribution of WGAN-GP, with $\lambda=10$ as scaling factor.
Since the evaluated policy is penalized for navigating areas of $\mathcal{S} \times \mathcal{A}$ where
$\lVert\nabla_{s,a}^t[f_\psi]_t\rVert _F > \tau$, we monitor
$G \coloneqq \lVert\nabla_{s,a}^t[f_\psi]_t\rVert _F$.
We expect to observe \emph{lower} values of $G$ when using the studied preconditioning.
In order to grasp the extent to which variations can compound in the system,
and therefore highlight the need for mechanims allowing the main method to contain such compounding of variations
(like the proposed one),
we also monitor an approximation of $\gamma^2 C$, relaxed as
$H \coloneqq \gamma^2 \lVert\nabla_{s,a}^t[f_\psi]_t\rVert _F^2 \max (1, \lVert\nabla_s^t[\mu_\theta]_t\rVert _F^2)$.
We expect to see the same ranking of methods in the plots depicting $G$ and $H$ respectively.
These are all reported in \textsc{Figure}~\ref{purplepsi}.

Note, the steep surge in overall computational cost caused by the evaluation of the monitored metrics
($G$ and $H$) and expecially $\kappa_t^\psi$ lowered the number of iterations our agent could do in the allowed runtime.
As such, we increased said runtime from the usual 0.5-day or 2-day duration to a 4-day duration (or 96 hours)
Such runs are more costly to orchestrate,
hence the sparser array of experiments to offset the steeper cost in compute.
In \textsc{Figure}~\ref{purplepsi},
we observe that, at evaluation time,
the model-based PURPLE instantiation in \textsc{eq}~\ref{precondrewf1} and \ref{precondrewf2}
indeed enables the agent to achieve \emph{lower} values of $G$ and $H$,
with the \emph{same} episodic return.
Said differently, it seems that the agent --- with preconditioning, compared to the one without ---
achieves the same proficiency, with the same convergence speed, while making decisions
that are \emph{safer} in terms of incurred variations of the approximate dynamics $f_\psi$.
\textbf{\emph{
So, even if the preconditioner is not needed to reach a higher return (or reach it faster) \textit{per se},
we have showcased that the studied model-based reward preconditioning can increase the robustness of the main method
by augmenting it with the means to tame \textit{a priori} untamable entities in the system (here, the dynamics).
}}
Still, the studied model-based instantiation of PURPLE is set back by several drawbacks.
\textit{a)} We need to maintain a forward model $f_\psi$ that approximates the effective transition function $f'$.
\textit{b)} To be estimated, $\kappa_t^\psi$ requires explicit calls to an automatic differentiation library,
making its frequent computation (every time a mini-batch is sampled from the replay buffer) extremely expensive overall.
\textit{c)} The threshold $\tau$ (to be enforced as Lipschitz constant for $f_\psi$) must be set such that
\emph{not every} decision made by the agent is penalized, while making sure it is still strict enough in that respect.
Besides, we observed in practice that the range of values taken by $\lVert\nabla_{s,a}^t[f_\psi]_t\rVert _F$
varies greatly across environments.
As such, $\tau$ must be tuned carefully per environment, making the overall process
tedious and computationally expensive.
In effect, this brings us back to the original issues of reward shaping \cite{Ng1999-lv},
that adversarial IL \cite{Ho2016-bv} circumvented.

\subsubsection{Total compensation of compounding variations}

Inspired by the insight laid out in \textsc{Remark}~\ref{remarkpurple},
we derive theoretical guarantees that characterize the robustness of $\widetilde{Q}_\varphi$
when using a preconditioner defined as follows:
\begin{align}
\kappa_{t+k} \coloneqq
\frac{1}{\sqrt{\prod_{u=0}^{k-1} C_{t+u}}}
\qquad &\text{where (\textit{cf.}~\textsc{eq}~\ref{aandbpurple})} \:
\forall v \in [0, T-1], \nonumber \\
&\quad C_v \coloneqq \lVert\nabla_{s,a}^v[f]_v\rVert _\infty^2
\max \big(1, \lVert\nabla_s^{v+1}[\mu]_{v+1}\rVert _\infty^2\big)
\label{complexkappa}
\end{align}
$\forall t \in [0, T] \cap \mathbb{N}$, and
$\forall k \in [0, T-t] \cap \mathbb{N}$.
Since the norms involved in $C_v$ are \emph{supremum} ones,
the preconditioner $\kappa_t$ verifies
$\dv*{\kappa_{t+k}}{s_t} = 0$ and
$\dv*{\kappa_{t+k}}{a_t} = 0$,
$\forall t \in [0, T] \cap \mathbb{N},
\forall k \in [0, T-t] \cap \mathbb{N},
\forall (s_t, a_t) \in \mathcal{S} \times \mathcal{A}$.
The reward preconditioner therefore verifies the properties
one must satisfy for the derived robustness guarantees to be applicable
(\textit{cf.}~\textsc{Section}~\ref{purpleandres}).
Again, note, the property $\kappa_t \leq 1$ is only ever used in \textsc{Section}~\ref{morerobust},
and has not been leveraged anywhere else.
Given that the developed theory still holds if $\exists t \in [0, T] \cap \mathbb{N}$ such that $\kappa_t > 1$,
the fact that the preconditioner defined in \textsc{eq}~\ref{complexkappa} does not necessarily lie in
the $(0,1]$ interval is not an issue \textit{a priori}.
Still, in practice,
it will virtually always be below $1$.

We now derive the associated counterparts of
\textsc{Theorem}~\ref{theorem2purple} and \textsc{Corollary}~\ref{corollary1purple}.

\begin{theorem}[state-action value Lipschitzness]
\label{theorem2purplecomplex}
We work under the assumptions laid out in
both \textsc{Lemma}~\ref{lemmapurple} and \textsc{Theorem}~\ref{theorem1purple}, and repeat
the main lines here
for \textsc{Theorem}~\ref{theorem2purplecomplex} to be self-contained:
a) The functions $f$, $\mu$ and $r_\varphi$ are $C^0$ and differentiable
over their respective input spaces,
and b) the function $r_\varphi$ is $\delta$-Lipschitz
over $\mathcal{S} \times \mathcal{A}$, i.e.
$\lVert \nabla_{s,a}^u[r_\varphi]_u \rVert _F \leq \delta$,
where $u \in [0, T] \cap \mathbb{N}$.
Then the quantity $\nabla_{s,a}^u[\widetilde{Q}_\varphi]_u$ exists
$\forall u \in [0, T] \cap \mathbb{N}$.
Assuming in addition that the reward preconditioner used on $r_\varphi$ to obtain $\tilde{r}_\varphi$
is defined according to \textsc{eq}~\ref{complexkappa},
the action-value $\widetilde{Q}_\varphi$ verifies:
\begin{align}
\lVert \nabla_{s,a}^t[\widetilde{Q}_\varphi]_t \rVert _F
\leq
\delta \, \sqrt{\frac{1 - \gamma^{2(T - t)}}{1 - \gamma^2}}
\end{align}
$\forall t \in [0, T] \cap \mathbb{N}$.
Note, the bound now only depends on $\delta$, $\gamma$, and $T-t$, the \textit{``remaining time in the episode''}.
\end{theorem}

\emph{Proof of \textsc{Theorem}~\ref{theorem2purplecomplex}.}
The reward preconditioner used to assemble $\tilde{r}_\varphi$ from $r_\varphi$
is defined according to \textsc{eq}~\ref{complexkappa}.
As carried out in \textsc{Remark}~\ref{remarkpurple},
we start the proof of \textsc{Theorem}~\ref{theorem2purplecomplex} analogously
to the one laid out for \textsc{Theorem}~\ref{theorem2purple},
but using the time-\emph{dependent} version of \textsc{Theorem}~\ref{theorem1purple}
instead of the time-\emph{independent} version that we used in \textsc{eq}~\ref{theorem1purpleuse}
(version \ref{theorem1purple}~\textit{(a)} instead of \ref{theorem1purple}~\textit{(b)}).
Our starting point then aligns with the crux of \textsc{Remark}~\ref{remarkpurple}.
As such, we have:
\begin{align}
\lVert \nabla_{s,a}^t[\widetilde{Q}_\varphi]_t \rVert _F^2
&\leq \sum_{k=0}^{T-t-1}
\Bigg[
\gamma^{2k} \, \kappa_{t+k}^2 \, \delta ^2 \, \prod_{u=0}^{k-1} C_{t+u}
\Bigg]
\qquad
\blacktriangleright\text{\small{\textit{\textsc{Theorem}~\ref{theorem1purple} (a)}}} \\
&= \sum_{k=0}^{T-t-1}
\Bigg[
\gamma^{2k} \, \frac{1}{\prod_{u=0}^{k-1} C_{t+u}} \, \delta ^2 \, \prod_{u=0}^{k-1} C_{t+u}
\Bigg]
\qquad
\blacktriangleright\text{\small{\textit{\textsc{eq}~\ref{complexkappa}}}} \\
&= \delta ^2 \, \sum_{k=0}^{T-t-1} (\gamma^2)^k
\label{beforegeom}
\end{align}
Since we defined $\gamma$ to be within the interval $[0,1)$ in \textsc{Section}~\ref{prelim}, we trivially have
$\gamma^2 < 1$, hence $\gamma^2 \neq 1$ and:
\begin{align}
\lVert \nabla_{s,a}^t[\widetilde{Q}_\varphi]_t \rVert _F^2
&\leq \delta ^2 \, \frac{1 - \gamma^{2(T - t)}}{1 - \gamma^2}
\qquad
\blacktriangleright\text{\small{\textit{finite sum of geometric series}}}
\end{align}
By applying $\sqrt{\cdot}$ (monotonically increasing) to the inequality,
we obtain the claimed result. \qed

Finally, we derive a corollary from \textsc{Theorem}~\ref{theorem2purplecomplex}
corresponding to the infinite-horizon regime.

\begin{corollary}[infinite-horizon regime]
\label{corollary1purplecomplex}
Under the assumptions of \textsc{Theorem}~\ref{theorem2purplecomplex},
including that $r_\varphi$ is $\delta$-Lipschitz
and that $\tilde{r}_\varphi$ is defined as in \textsc{eq}~\ref{precondrew}
over $\mathcal{S} \times \mathcal{A}$,
we have, in the infinite-horizon regime:
\begin{align}
\lVert \nabla_{s,a}^t[\widetilde{Q}_\varphi]_t \rVert _F
&\leq \frac{\delta}{\sqrt{1 - \gamma^2}}
\end{align}
which translates into $\widetilde{Q}_\varphi$ being $\frac{\delta}{\sqrt{1 - \gamma^2}}$-Lipschitz
over $\mathcal{S} \times \mathcal{A}$.
\end{corollary}

\emph{Proof of \textsc{Corollary}~\ref{corollary1purplecomplex}.}
As we adapt the proof of \textsc{Theorem}~\ref{theorem2purplecomplex} to
the infinite-horizon regime, \textsc{eq}~\ref{beforegeom} becomes
\begin{align}
\lVert \nabla_{s,a}^t[\widetilde{Q}_\varphi]_t \rVert _F^2
\leq \delta ^2 \, \sum_{k=0}^{+\infty} (\gamma^2)^k
= \frac{\delta ^2}{1 - \gamma^2}
\qquad
\blacktriangleright\text{\small{\textit{infinite sum of geometric series}}}
\end{align}
since we defined $\gamma$ to be within the interval $[0,1)$ in \textsc{Section}~\ref{prelim},
\textit{i.e.} $\gamma^2 < 1$.
We then apply $\sqrt{\cdot}$ to the inequality. \qed

In these theoretical guarantees, we have shown that by carefully crafting PURPLE's reward preconditioner
according to \textsc{eq}~\ref{complexkappa}, we obtain upper-bounds $\widehat{\Delta}_\infty$
on the Lipschitz constant of the resulting
action-value $\widetilde{Q}_\varphi$ that are \emph{independent} of $C_v$,
$\forall v \in [0, T-1]$ --- where
$C_v \coloneqq \lVert\nabla_{s,a}^v[f]_v\rVert _\infty^2
\max \big(1, \lVert\nabla_s^{v+1}[\mu]_{v+1}\rVert _\infty^2\big)$
(\textit{cf.}~\textsc{eq}~\ref{complexkappa}).
In other words, we have shown that such preconditioner design allows us to \emph{totally} compensate
for the compounding variations
\textit{a)} first tackled in the discussion led in \textsc{Section}~\ref{compoundvars}, and
\textit{b)} then addressed \emph{only partially} by the model-based reward preconditioning discussed profusely
in \textsc{Section}~\ref{partialcompensate} (of which we showcase the applicability in practice).
Echoing what motivated the emergence of \textsc{Remark}~\ref{remarkpurple} in the first place,
the form adopted by the reward preconditioning (\textit{cf.}~\textsc{eq}~\ref{complexkappa})
that allowed us to derive the robustness guarantees of
\textsc{Theorem}~\ref{theorem2purplecomplex} and
\textsc{Corollary}~\ref{corollary1purplecomplex} enjoys an insightful and intuitive \emph{interpretation}.
Going through the elements of the series described by
the preconditioner of \textsc{eq}~\ref{complexkappa},
$(\kappa_{t+k})_k$,
$\forall t \in [0, T] \cap \mathbb{N}$, and
$\forall k \in [0, T-t] \cap \mathbb{N}$,
we have the following sequence of consecutive preconditioning values:
\begin{align}
&\kappa_{t+k}\big\rvert_{k=0} = \kappa_t \coloneqq 1
\; \rightarrow \;
\kappa_{t+k}\big\rvert_{k=1} = \kappa_{t+1} \coloneqq \frac{1}{\sqrt{C_t}}
\; \rightarrow \;
\kappa_{t+k}\big\rvert_{k=2} = \kappa_{t+2} \coloneqq \frac{1}{\sqrt{C_t C_{t+1}}} \nonumber \\
&\; \rightarrow \;
\kappa_{t+k}\big\rvert_{k=3} = \kappa_{t+3} \coloneqq \frac{1}{\sqrt{C_t C_{t+1} C_{t+2}}}
\; \rightarrow \;
\ldots
\; \rightarrow \;
\kappa_{t+k}\big\rvert_{k=T-t} = \kappa_{T} \coloneqq \frac{1}{\sqrt{C_t C_{t+1} C_{t+2} \ldots C_{T-1}}}
\end{align}
We observe that, when purposely defined as such,
the reward preconditioner $\kappa_{t+k}$
at a given stage $t+k$ compensates for the $C_v$'s
of all the \emph{previous} timesteps --- backwards from $t+k-1$ to $t$, where $\widetilde{Q}_\varphi$'s
Lipschitz constant is characterized.
In order to prevent the upper-bound on $\lVert \nabla_{s,a}^t[\widetilde{Q}_\varphi]_t \rVert _F$
to be burdened by incipient, potentially prone to compound, variation of
$C_v \coloneqq \lVert\nabla_{s,a}^v[f]_v\rVert _\infty^2
\max \big(1, \lVert\nabla_s^{v+1}[\mu]_{v+1}\rVert _\infty^2\big)$,
the preconditioner can \emph{actively anticipate} said incipient compounding variations to compound further
within the time remaining in the episode by preemptively squashing the \emph{current} surrogate reward at $t+k$
based on how much $C_v$'s variations have accumulated since $t$ until $t+k-1$.
The proposed interpretation of the studied preconditioner aligns with our intuitive desideratum:
\textit{``if you want to fend off from compounding of variations that threaten the stability of
your action-value, make the latter more robust as soon as you see, from past metrics
--- here, monitored $C_v$ values --- that said
variations might actually compound soon''}.

Despite appealing in principle thanks to its salient interpretation,
and justified by theoretical guarantees,
we did not experiment with the proposed preconditioner in practice.
Indeed, considering how we have shown in \textsc{Section}~\ref{partialcompensate}
that the values in effect taken by $C_v \coloneqq \lVert\nabla_{s,a}^v[f]_v\rVert _\infty^2
\max \big(1, \lVert\nabla_s^{v+1}[\mu]_{v+1}\rVert _\infty^2\big)$ do not seem to affect
the agent's return in practice,
we do not expect the interpretable preconditioner tackled in this discussion
to bring anything \emph{practically} in the considered environments.
Using a gradient penalty constraint to induce local Lipschitz-continuity of the
function at the core of the reward function is, in a sense, \emph{all you need}
to achieve peak expert performance in the considered off-policy
generative adversarial imitation learning setting.
Still, we believe the design and study of methods able to actively tune their level of robustness
--- aligned in this work with the concept of spatial, local Lipschitz-continuity ---
depending on the choices (or more pessimistically, on the \emph{mistakes}) made by the agent
to be an interesting avenue of future work.
Besides,
by augmenting the reward-less MDP $\mathbb{M}$ (from which we first stripped the environmental reward)
with our adversarially learned reward, preconditioned in line with \textsc{eq}~\ref{complexkappa},
the resulting MDP has a \emph{memory},
since the reward $\tilde{r}_\varphi$ depends on entities
($C_v$'s)
from previous timesteps in the episode.
In effect, due to such a reward preconditioning formulation, the Markov property is not satisfied anymore as,
given the present, the future now \emph{does} depend on the past.
We believe the observations made and results derived in this work
could pave the way to further investigations aiming to decipher known methods and ultimately pinpoint
the \emph{most minimal} setup for it to still do well.

\section{Conclusion}

In this work, we conducted an in-depth study of the stability problems
incurred by off-policy generative adversarial imitation learning.
Our contributions closely follow the line of reasoning, and are as follows.
\textbf{\textit{(1)}}
We characterized the various inherent hindrances the approach suffers from,
in particular how learned parametric rewards affect the learned parametric state-action value.
\textbf{\textit{(2)}}
We showed that enforcing a local Lipschitz-continuity
constraint on the discriminator network used to formulate the imitation surrogate reward
is a \textit{sine qua non} condition for the approach
to empirically achieve expert performance
in challenging continuous control problems, within a number of timesteps that still enable
us to call the method sample-efficient.
\textbf{\textit{(3)}}
In line with the first and second steps,
we derived theoretical guarantees that characterize
the Lipschitzness of the Q-function
when the reward is assumed $\delta$-Lipschitz-continuous.
\textbf{\emph{Note, the reported theoretical results are valid for any reward
satisfying the condition, nothing is specific to imitation.}}
\textbf{\textit{(4)}}
We propose a new RL-grounded interpretation of the usual
GAN gradient penalty regularizers --- differing by
\emph{where} they induce Lipschitzness --- along with an explanation
as to \textit{(a)} why they all have such a positive impact on stability,
but also \textit{(b)} how to make sense of the empirical gap between them.
\textbf{\textit{(5)}}
We show that, in effect,
the consistent satisfaction of the Lipschitzness constraint on the reward
is a strong predictor of how well the mimicking agent performs empirically.
\textbf{\textit{(6)}}
Finally, we introduce a pessimistic reward preconditioning technique which
\textit{(a)} makes the base method it is plugged into provably more robust,
and \textit{(b)} is accordingly backed by several theoretical guarantees.
\textbf{\emph{As in (3), these guarantees
are not not specific to imitation and have a wide range of applicability.}}
We give an illustrative example of how the technique can help further increasing
the robustness of the method it is plugged into empirically.

\bibliography{bibliography}

\clearpage

\appendix

\section{Hyper-parameters}
\label{hps}

The function approximators used in every learned module
are two-layer multi-layer perceptrons, but the widths of their respective layers
differ.
We use layers of sizes 100-100 for the discriminator (from which the reward is formulated),
300-200 for the actor, and 400-300 for the critic, as they achieved the best overall
result across the environments of the suite in our early experiments.
Unless specifically stated otherwise,
the discriminator network uses spectral normalization \cite{Miyato2018-wc}
at every layer,
while the actor and critic networks both use layer normalization \cite{Ba2016-bs}
at every layer.
Every neural network is initialized via orthogonal initialization \cite{Saxe2013-rm}.
Each network has its own optimizer (\textit{cf.} \textsc{Section}~\ref{bridge}
for a complete description of the optimization problems the networks
of parameter $\varphi$, $\omega$, and $\theta$ are involved in,
along with the loss they optimize).
We use \textsc{Adam} \cite{Kingma2014-op} for each of them,
with respective learning rates reported in \textsc{Table}~\ref{hptable},
while the other parameters of the optimizer are left to the default
PyTorch \cite{Paszke2019-zf} values.
In practice, we replace the squared error loss involved in the loss optimized by the critic
(\textit{cf.} \textsc{eq}~\ref{omegaloss})
by the Huber loss, as is commonly done in temporal-difference learning
with function approximation and target networks \cite{Mnih2013-rb,Mnih2015-iy}.
As for the activations functions used in the neural networks,
we used ReLU non-linearities in both the actor and critic,
and used Leaky-ReLU \cite{Maas2013-dw} non-linearities with a leak of $0.1$ in the discriminator.
We used an \emph{online} version of batch normalization
(described earlier in \textsc{Section}~\ref{empres1})
to standardize the actor and critic observations before they are fed to them.
We do not use any learning rate scheduler, for any module.

\begin{table}[H]
\centering
\begin{tabular}{l|c}
\hline
Hyper-parameter&Selected Value \\
\hline\hline
\rowcolor{MyLightGray} Training steps per iteration&$2$ \\
Evaluation steps per iteration&$10$ \\
\rowcolor{MyLightGray} Evaluation frequency&$10$ \\
\hline
Actor learning rate&$2.5 \times 10^{-4}$ \\
\rowcolor{MyLightGray} Critic learning rate&$2.5 \times 10^{-4}$ \\
Actor clip norm&$40$ \\
\rowcolor{MyLightGray} Critic weight decay scale&$0$ \\
\hline
Rollout length&$2$ \\
\rowcolor{MyLightGray} Effective batch size&$1024$ \\
Discount factor $\gamma$&0.99 \\
\rowcolor{MyLightGray} Replay buffer $\mathcal{R}$ size&$100000$ \\
Exploration (\textit{cf.} \textsc{Section}~\ref{bridge})& $\sigma_a=0.2$, $\sigma_b=0.2$ \\
\rowcolor{MyLightGray} Param. noise update frequency&$50$ \\
Target update Polyak scale $\tau$&$0.005$ \\
\rowcolor{MyLightGray} Multi-step lookahead $n$&$10$ \\
\hline
Target smoothing - noise $\sigma$ \cite{Fujimoto2018-pe}&$0.2$ \\
\rowcolor{MyLightGray} Target smoothing - noise clip \cite{Fujimoto2018-pe}&$0.5$ \\
Actor update delay \cite{Fujimoto2018-pe}&$2$ \\
\hline
\rowcolor{MyLightGray} Reward training steps per iteration&$1$ \\
Agent training steps per iteration&$1$ \\
\rowcolor{MyLightGray} Discriminator learning rate&$5.0 \times 10^{-4}$ \\
Entropy regularization scale&$0.001$ \\
\rowcolor{MyLightGray} Positive label-smoothing&Real labels $\sim \operatorname{unif}(0.7,1.2)$ \\
\hline
Positive-Unlabeled \cite{Xu2019-uo} - coeff. $\eta$&$0.25$ \\
\hline
\end{tabular}
\caption{Hyper-parameters used in this work.
Unless explicitly stated otherwise, every method uses these.
The \emph{``effective''} batch size corresponds to the size of the mini-batch
\emph{aggregated} across parallel workers of the distributed architecture.
In our case, every worker --- of the grand total of $n=16$ workers ---
samples a mini-batch of size $64$ from its (individual) replay buffer,
resulting in an effective batch size of $64 \times 16 = 1024$.}
\label{hptable}
\end{table}

\section{Sequential Decision Making Under Uncertainty In Non-Stationary Markov Decision Processes}
\label{nsmdps}
In \textsc{Section}~\ref{prelim}, we have defined $\mathbb{M}$ as a stationary MDP,
in line with a vast majority of works in RL.
Note, a stochastic process or a distribution is commonly said \emph{stationary} if
it remains unchanged when shifted in time.
While the stationarity assumption allows for the derivation of various theoretical guarantees
and is overall easier to deal with analytically,
it fails to explain the inner workings of complex realistic simulations,
and \textit{a fortiori} the real world.
One critical challenge incurred when modeling the world as a non-stationarity MDP is the
unavailability of convergence guarantees for standard practical RL methods.
Crucially, assuming stationarity in the dynamics $p$
is necessary for the Markov property to hold,
which is required for the convergence of Q-learning \cite{Watkins1989-ir}
algorithms \cite{Abdallah2016-pd}
like DQN \cite{Mnih2013-rb,Mnih2015-iy}.
As such, designing methods yielding agents that are robust against the non-stationarities
naturally occurring in their realistic environments
is a challenging yet timely milestone.
Methods equipping models against unforeseen changes in the data distribution,
a phenomenon qualified as \emph{concept drift} \cite{Schlimmer1986-nj},
are surveyed in \cite{Gama2014-wv} who dedicate the study to the supervised case.
In RL, a analysis of non-stationarity issues inherent to the Q-learning loss optimization
under function approximation \cite{Sutton1999-ii}
proposes qualitative and quantitative diagnostics along
with a new replay sampling method to alleviate the isolated weaknesses \cite{Fu2019-kb}.
Non-stationarities are characterized by how they manifest in time.
A distribution is \emph{switching} if abrupt changes, called \emph{change points},
occur while remaining stationarity
in-between, making it in effect piece-wise stationary
\cite{Da_Silva2006-gj,Jaksch2010-bf,Garivier2011-ax,Abdallah2016-pd,Gajane2018-ee,
Padakandla2019-mb,Auer2019-xz}.
The change points are either given by an oracle or discovered via change point detection
techniques.
Once exhibited, one can employ stationary methods individually on each segment.
A distribution is \emph{drifting} if it gradually changes at an unknown rate
\cite{Besbes2014-hm,Anava2016-qk,Luo2018-ib,Ortner2019-pw,Chen2019-gt,
Cheung2019-yq,Cheung2019-cu,Russac2019-da}.
The change can occur continually or as a slow transition between stationary plateaus,
making it considerably more difficult to deal with, theoretically and empirically.
In a non-stationary MDP,
the non-stationarities can manifest in the dynamics $p$
\cite{Nilim2005-rq,Da_Silva2006-gj,Xu2007-iq,Lim2013-ml,Abdallah2016-pd},
in the reward process $r$
\cite{Even-dar2005-rg,Dick2014-br},
or in both conjointly \cite{Yu2009-xp,Yu2009-yo,Abbasi-Yadkori2013-vd,Gajane2018-ee,
Padakandla2019-mb,Yu2019-sc,Lecarpentier2019-av}.
The adversarial bilevel optimization problem
--- guiding the adaptive tuning of the reward
for every policy update ---
present in this work
is reminiscent of the stream of research pioneered by \cite{Auer1995-mm}
in which the reward is generated by an omniscient \emph{adversary},
either arbitrarily or adaptively with potentially malevolent drive
\cite{Yu2009-xp,Yu2009-yo,Lim2013-ml,Gajane2018-ee,Yu2019-sc}.
Non-stationary environments are almost exclusively tackled from a theoretical perspective
in the literature (\textit{cf.} previous references in this section).
Specifically, in the \emph{drifting} case, the non-stationarities are traditionally dealt with
via the use of sliding windows.
The accompanying (dynamic) regret analyses all rely on strict assumptions.
In the switching case, one needs to know the number of occurring switches beforehand,
while in the drifting case, the change variation need be upper-bounded.
Specifically, \cite{Besbes2014-hm,Cheung2019-yq} assume the total change to be
upper-bounded by some preset variation budget,
while \cite{Cheung2019-cu} assumes the variations are uniformly bounded in time.
\cite{Ortner2019-pw} assumes that the \textit{incremental}
variation (as opposed to \textit{total} in \cite{Besbes2014-hm,Cheung2019-yq}) is upper-bounded
by a \textit{per-change} threshold.
Finally, in the same vein, \cite{Lecarpentier2019-av} posits \emph{regular evolution},
by making the assumption that both the transition and reward functions are Lipschitz-continuous
\textit{w.r.t.} time.

\section{Adaptive Policy Update based on Gradient Similarities}
\label{cossim}

\begin{figure}[H]
  \center
  \begin{subfigure}[t]{0.49\textwidth}
    \center\scalebox{0.16}[0.16]{\includegraphics{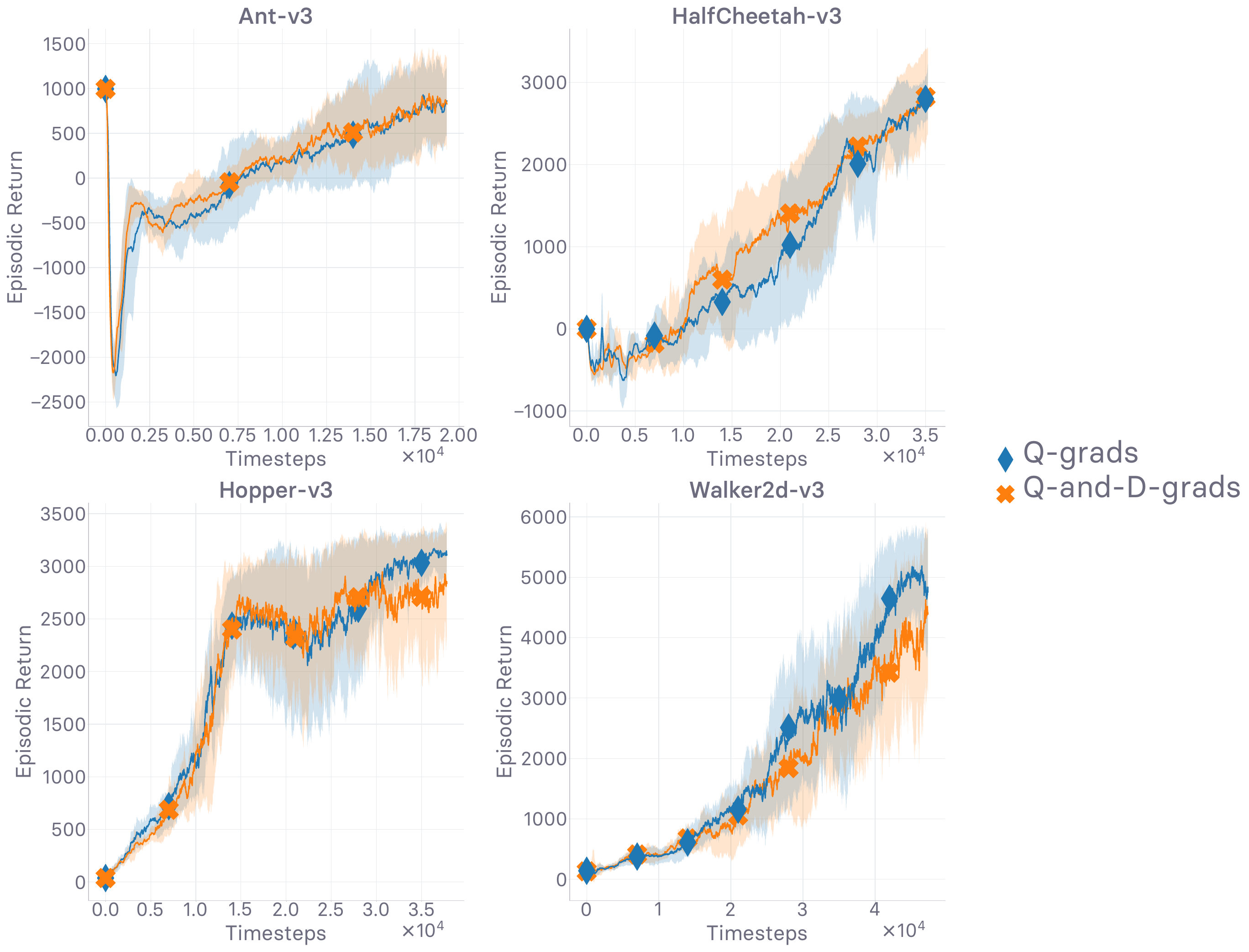}}
    \caption{Evolution of return values \textit{(higher is better)}}
  \end{subfigure}
  \begin{subfigure}[t]{0.49\textwidth}
    \center\scalebox{0.16}[0.16]{\includegraphics{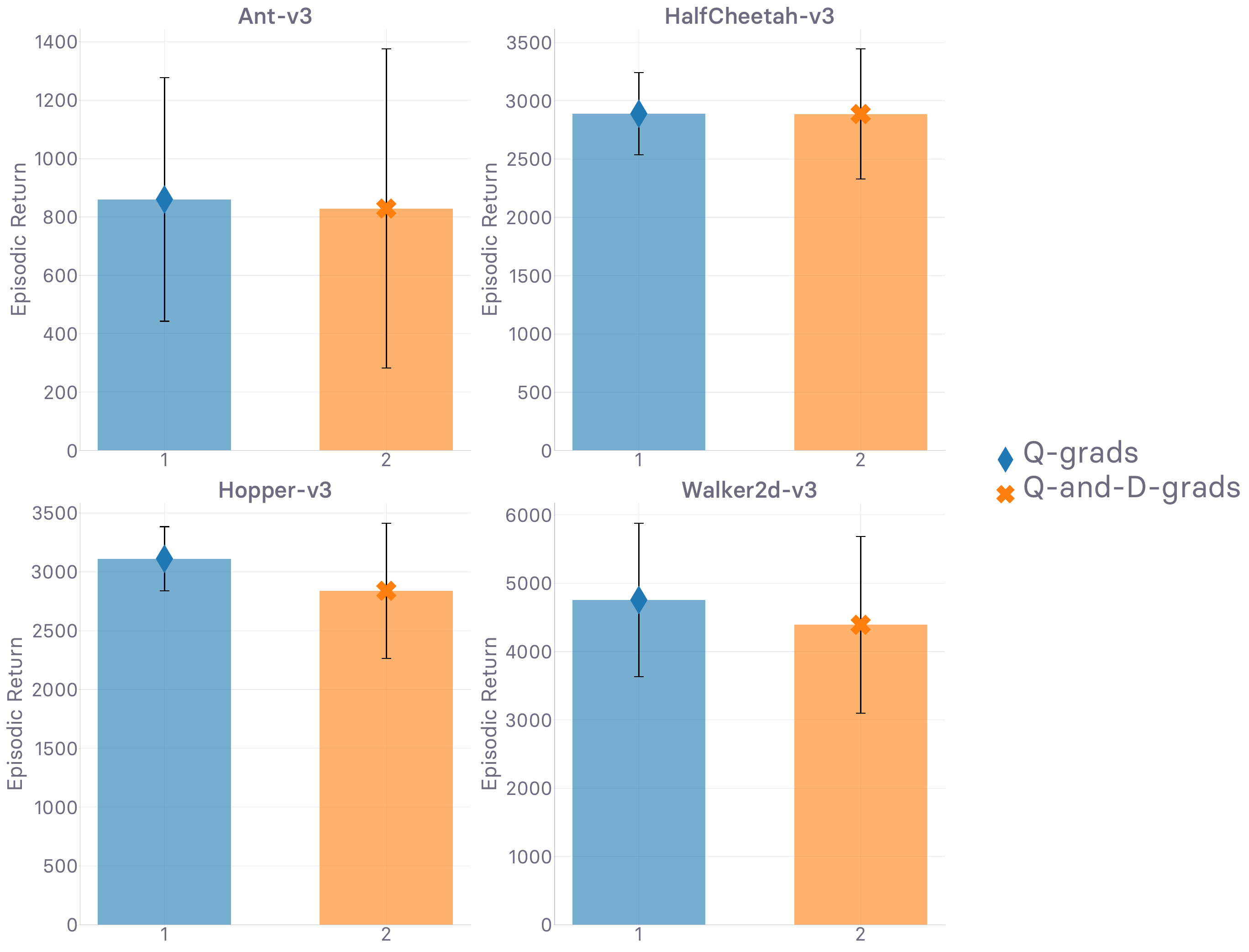}}
    \caption{Final return values at timeout \textit{(higher is better)}}
  \end{subfigure}
  \caption{
  Comparison of the gradient used to update the policy in this work,
  involving the gradient of the state-action value,
  against an adaptive hybrid method involving \emph{also}
  the gradient of the discriminator, and combining both gradients
  based on their cosine similarity.
  Runtime is 12 hours.}
  \label{cosimplots}
\end{figure}

\section{Clipped Double-Q Learning and Target Policy Smoothing}
\label{ablationtd3}

\begin{figure}[H]
  \center
  \begin{subfigure}[t]{0.99\textwidth}
    \center\scalebox{0.18}[0.18]{\includegraphics{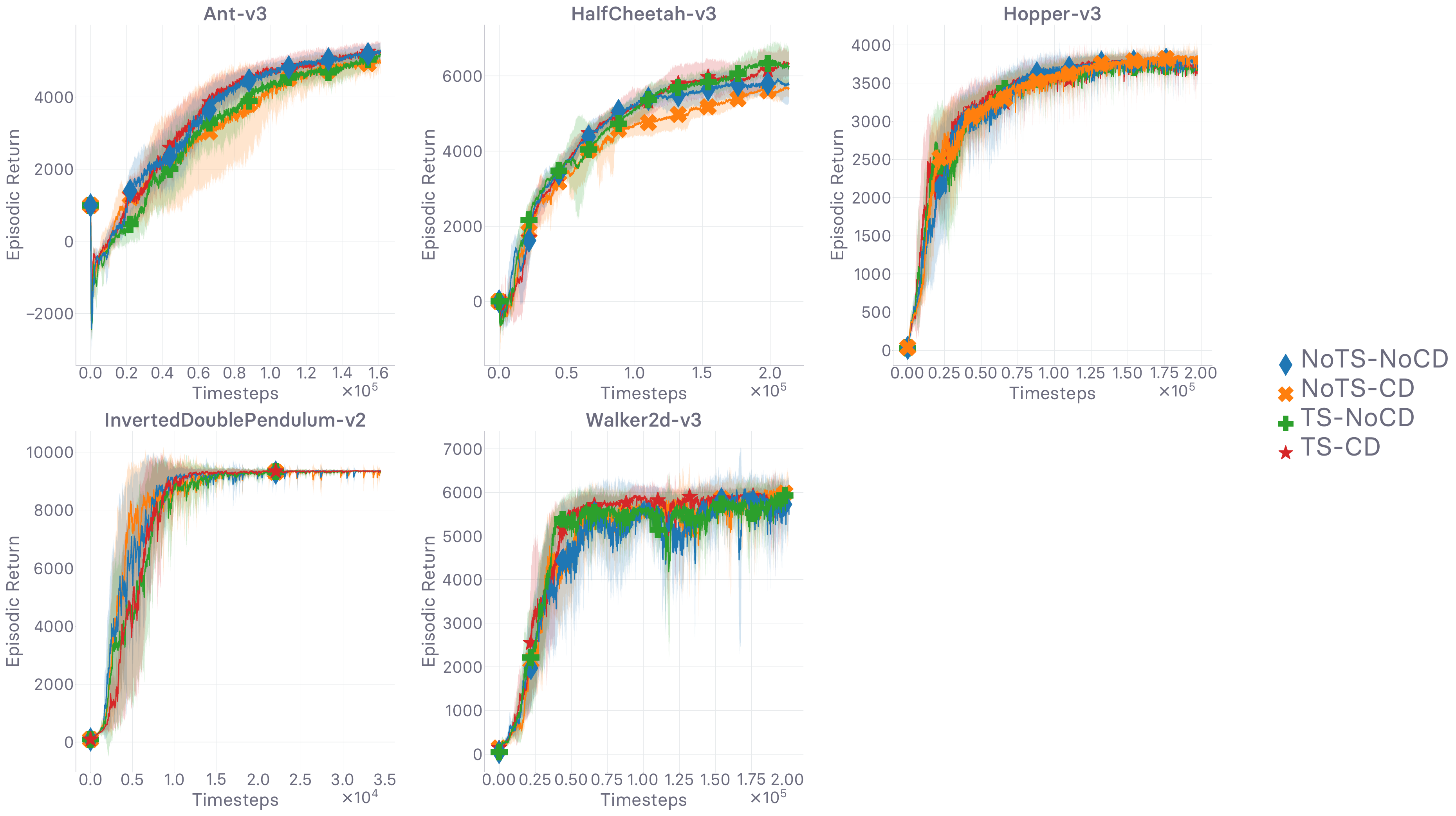}}
    \caption{Evolution of return values \textit{(higher is better)}}
  \end{subfigure}
  \begin{subfigure}[t]{0.99\textwidth}
    \center\scalebox{0.18}[0.18]{\includegraphics{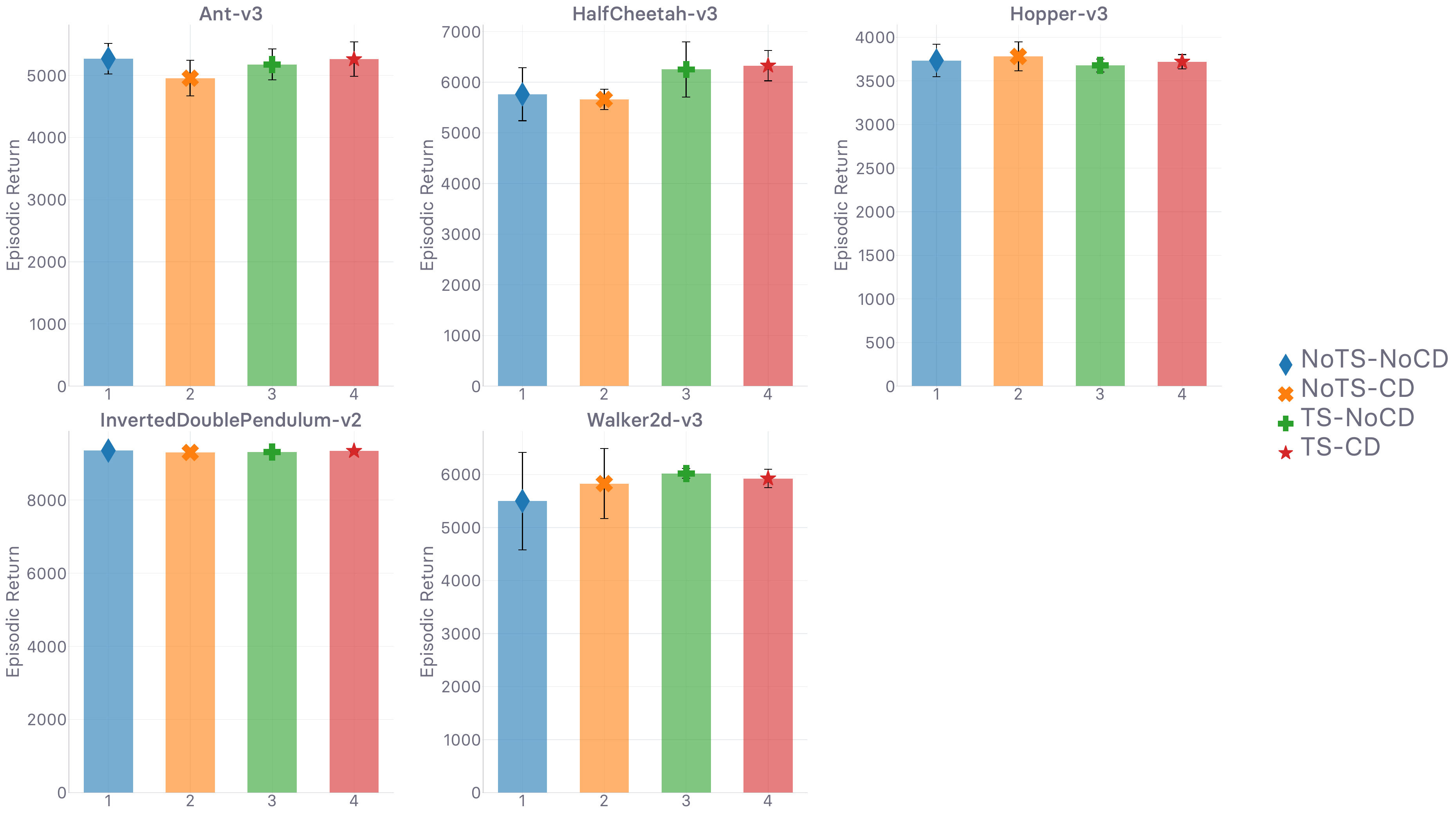}}
    \caption{Final return values at timeout \textit{(higher is better)}}
  \end{subfigure}
  \caption{
  Ablation study on the use of the clipped double Q-Learning (CD)
  and target smoothing (TS) techniques,
  both from \cite{Fujimoto2018-pe},
  \emph{with} gradient penalty regularization \cite{Gulrajani2017-mr}.
  Runtime is 48 hours}
\end{figure}

\section{Gradient Penalty}

\subsection{One-sided Gradient Penalty}
\label{ablationonesided}

\begin{figure}[H]
  \center
  \begin{subfigure}[t]{0.99\textwidth}
    \center\scalebox{0.18}[0.18]{\includegraphics{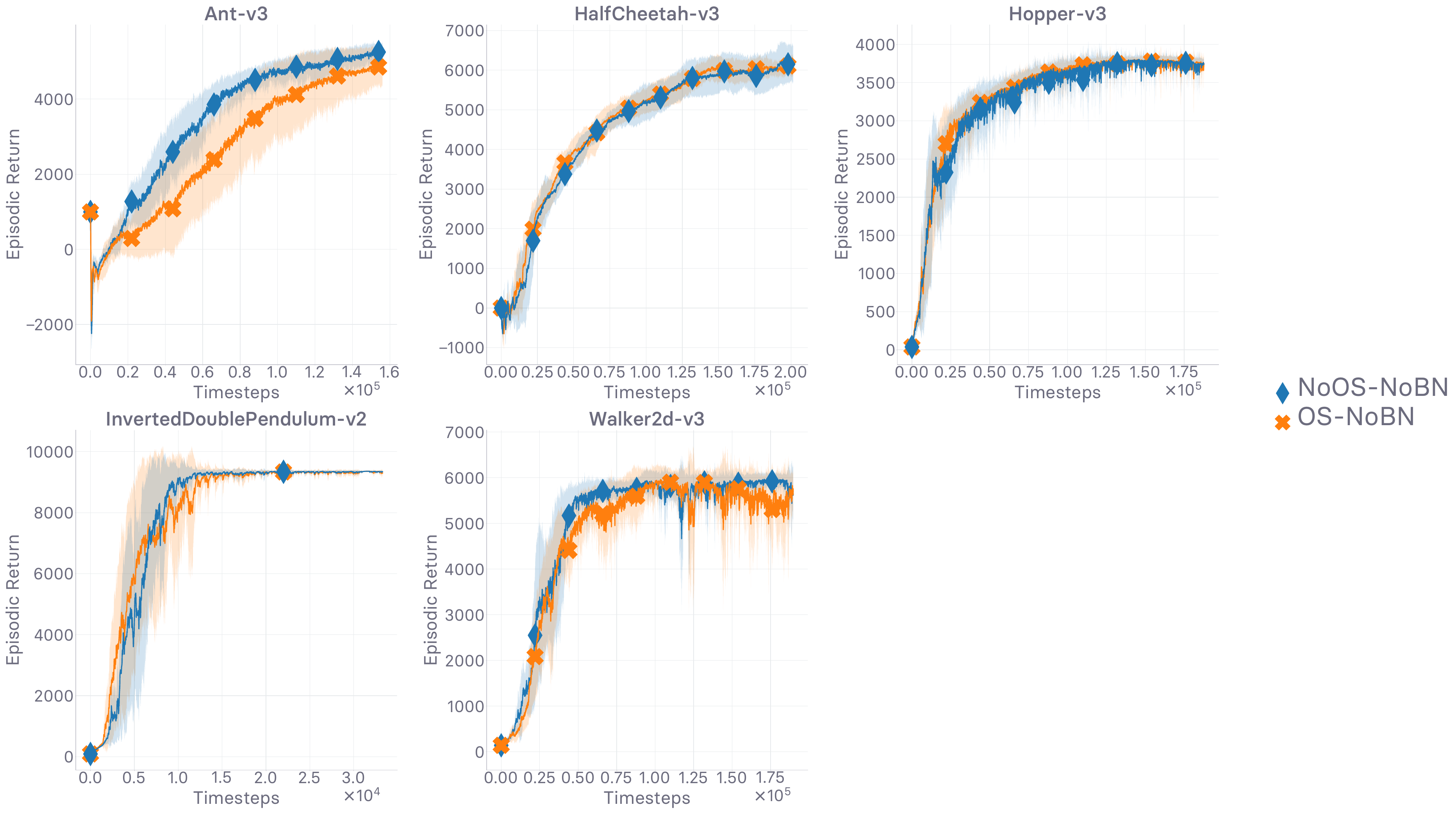}}
    \caption{Evolution of return values \textit{(higher is better)}}
  \end{subfigure}
  \begin{subfigure}[t]{0.99\textwidth}
    \center\scalebox{0.18}[0.18]{\includegraphics{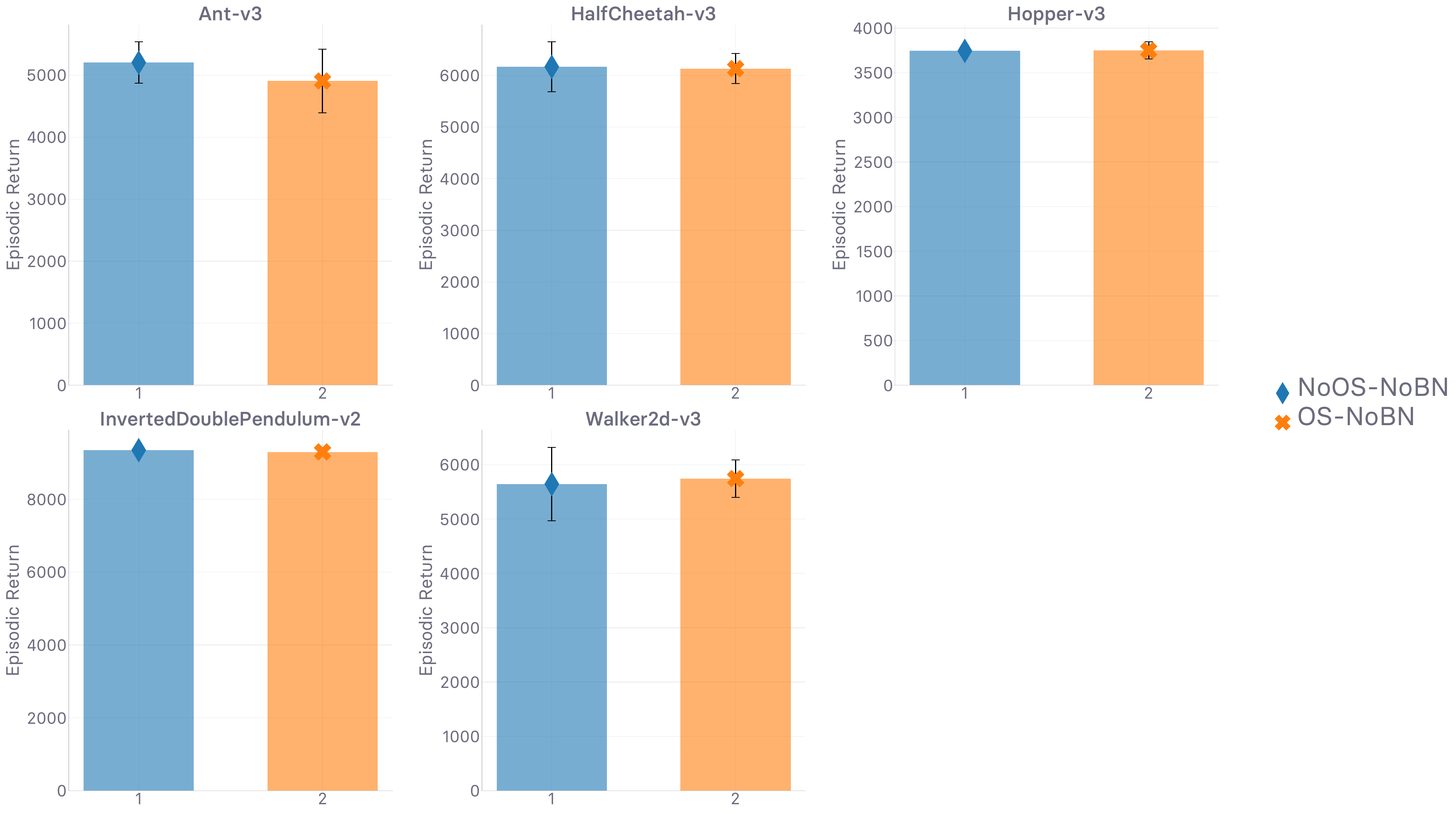}}
    \caption{Final return values at timeout \textit{(higher is better)}}
  \end{subfigure}
  \caption{
  Ablation study on the use of the one-sided (OS) penalty variant \cite{Gulrajani2017-mr}.
  Runtime is 48 hours}
\end{figure}

\subsection{Online Batch Normalization in Discriminator}
\label{ablationbn}

\begin{figure}[H]
  \center
  \begin{subfigure}[t]{0.99\textwidth}
    \center\scalebox{0.18}[0.18]{\includegraphics{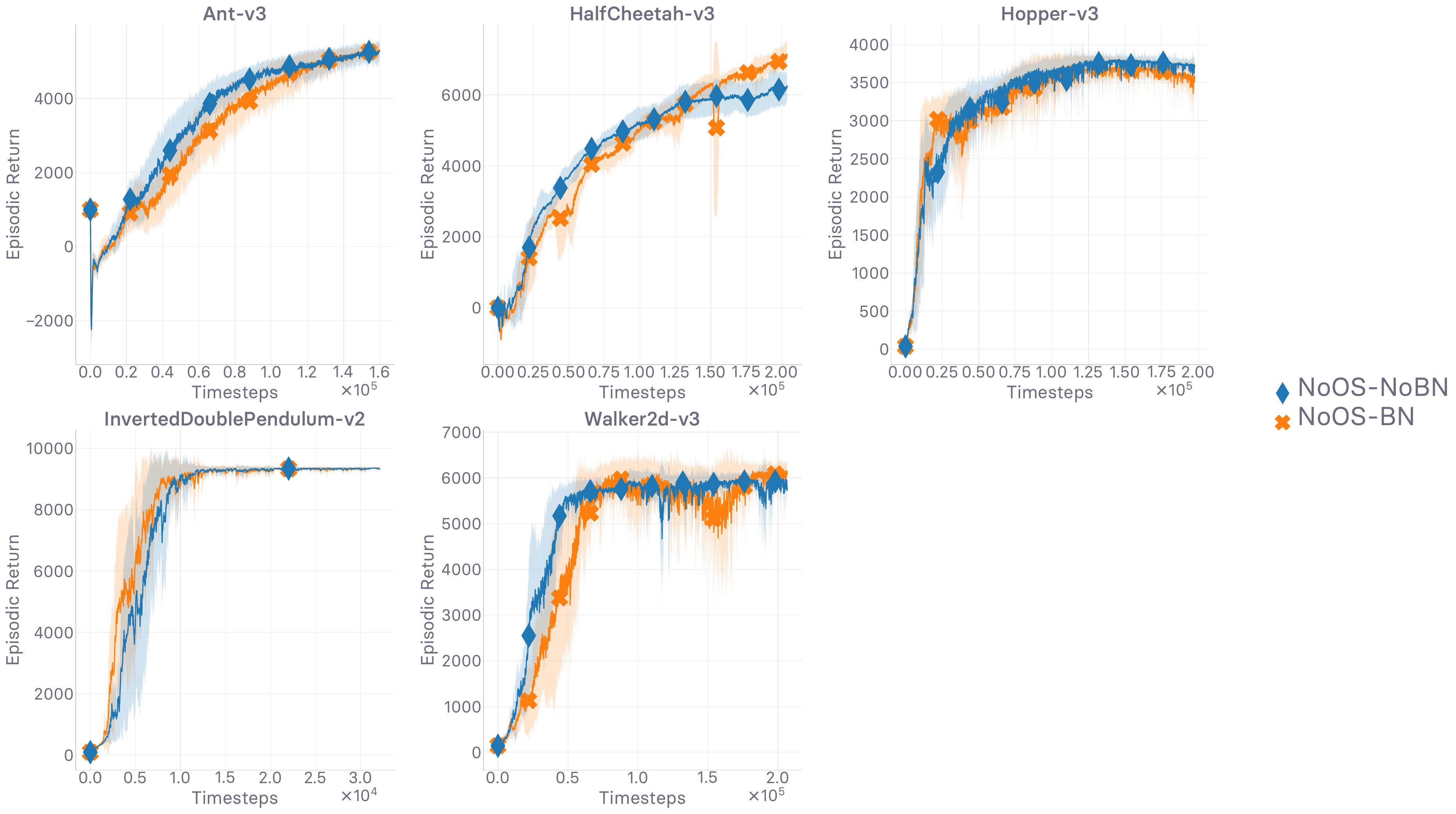}}
    \caption{Evolution of return values \textit{(higher is better)}}
  \end{subfigure}
  \begin{subfigure}[t]{0.99\textwidth}
    \center\scalebox{0.18}[0.18]{\includegraphics{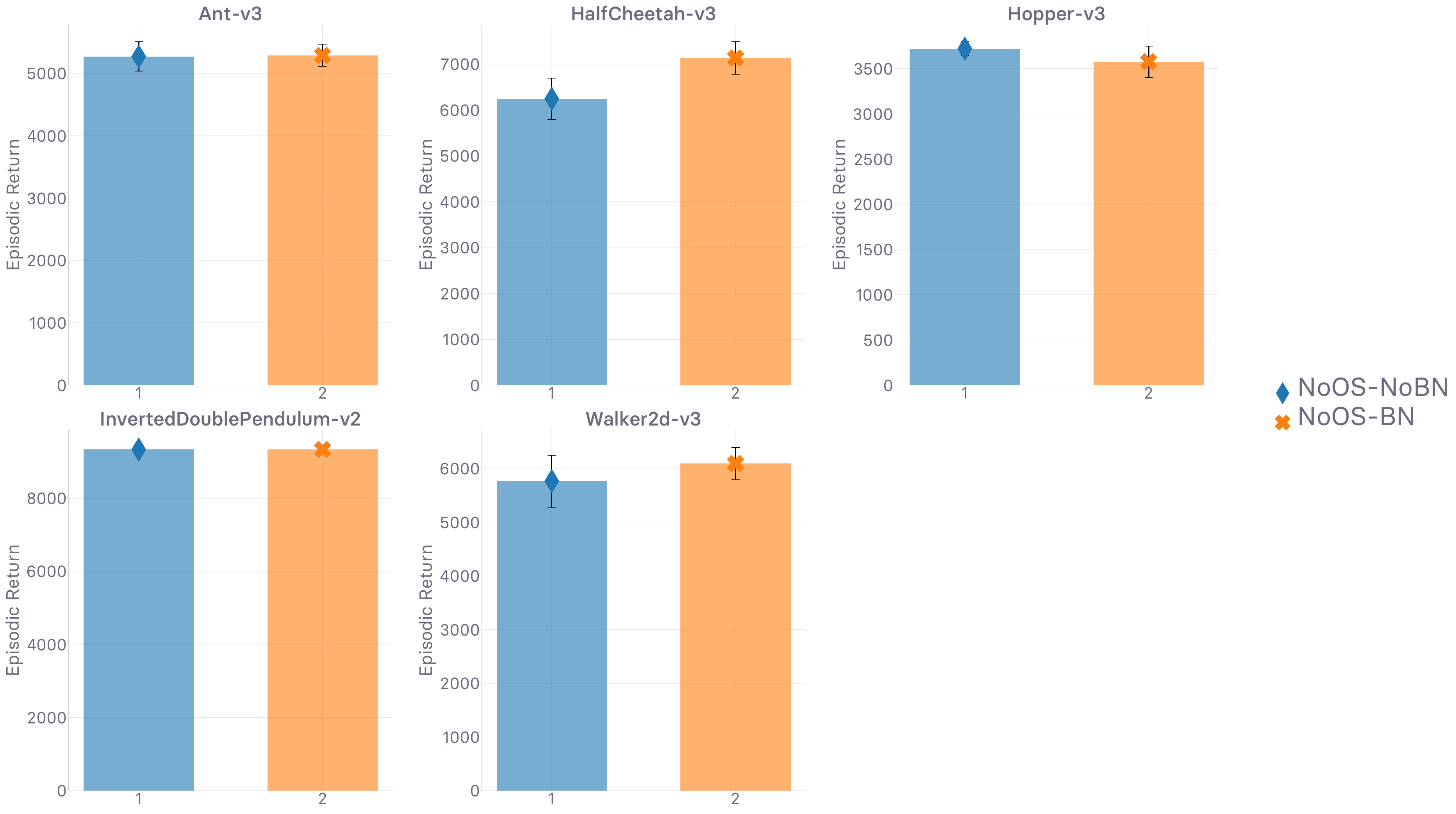}}
    \caption{Final return values at timeout \textit{(higher is better)}}
  \end{subfigure}
  \caption{
  Ablation study on the use of online batch normalization (BN) in the
  discriminator for its impact on the gradient penalization \cite{Gulrajani2017-mr}.
  Runtime is 48 hours}
\end{figure}

\subsection{Target $k$ and Coefficient $\lambda$ Grid Search}
\label{gridklam}

\begin{figure}[H]
  \center
  \begin{subfigure}[t]{0.49\textwidth}
    \center\scalebox{0.16}[0.16]{\includegraphics{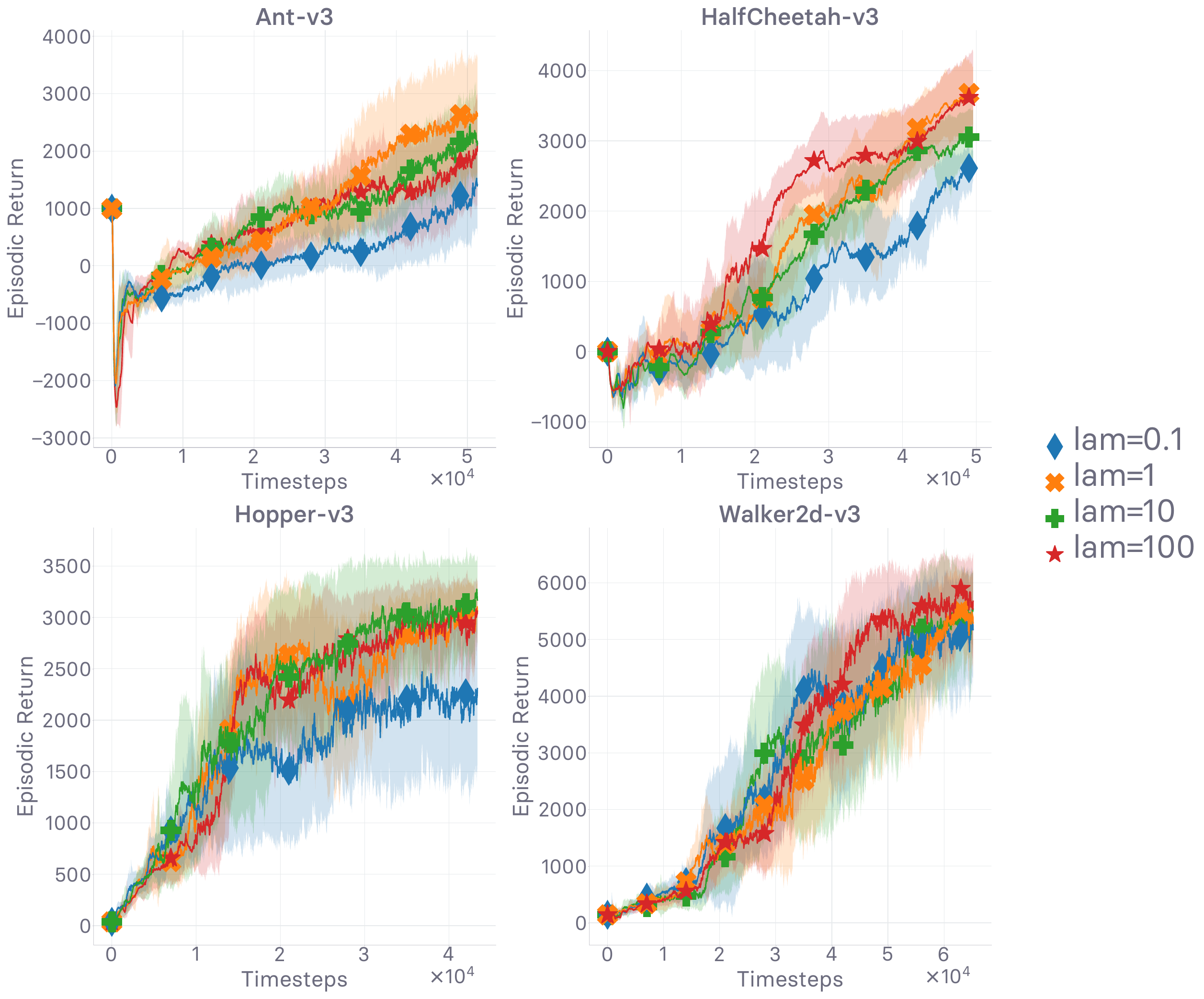}}
    \caption{Evolution of return values \textit{(higher is better)}}
  \end{subfigure}
  \begin{subfigure}[t]{0.49\textwidth}
    \center\scalebox{0.16}[0.16]{\includegraphics{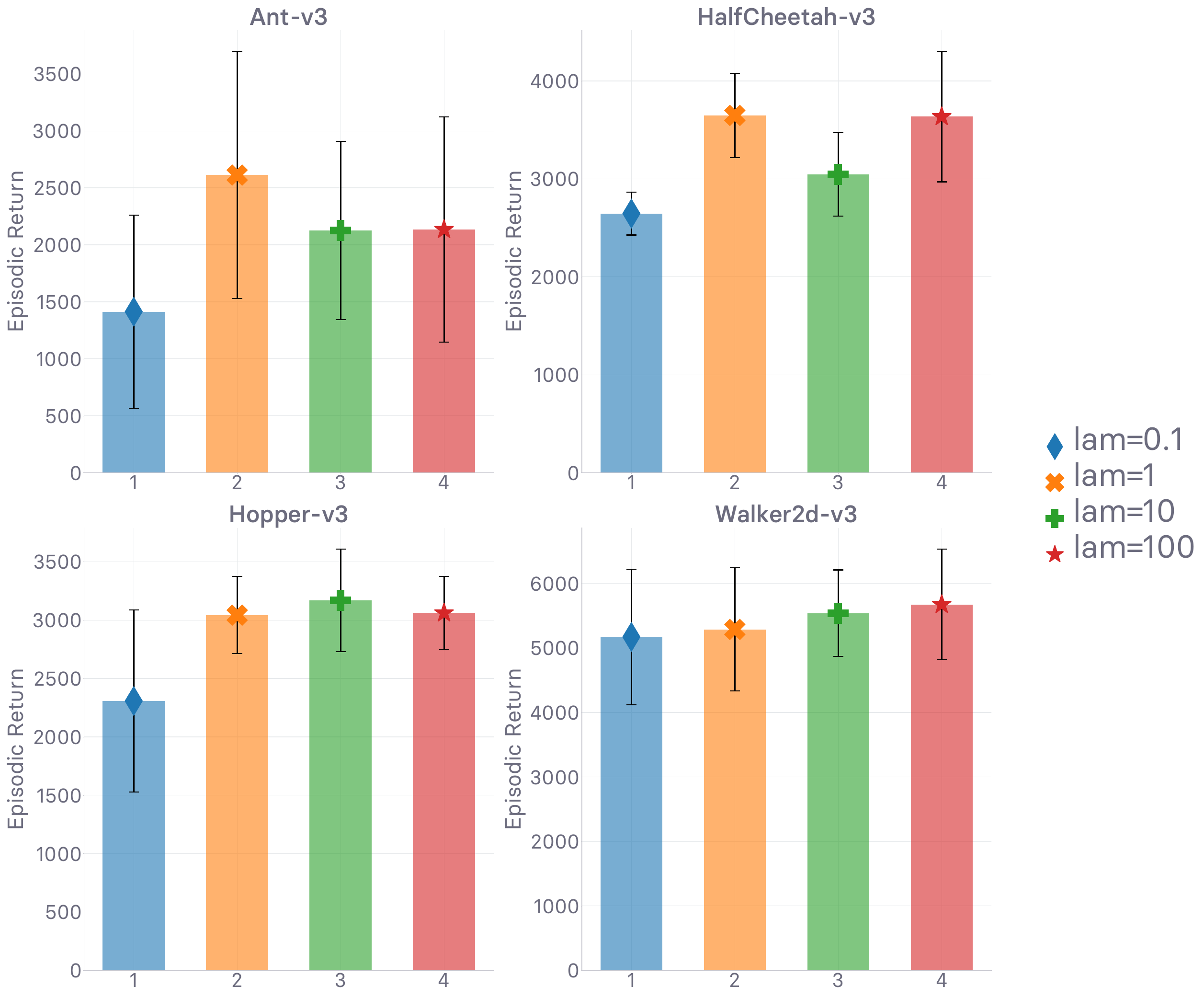}}
    \caption{Final return values at timeout \textit{(higher is better)}}
  \end{subfigure}
  \caption{
  Grid search over the hyper-parameter $\lambda$ when $k=1$.
  Runtime is 12 hours.}
  \label{gridk1}
\end{figure}

\begin{figure}[H]
  \center
  \begin{subfigure}[t]{0.49\textwidth}
    \center\scalebox{0.16}[0.16]{\includegraphics{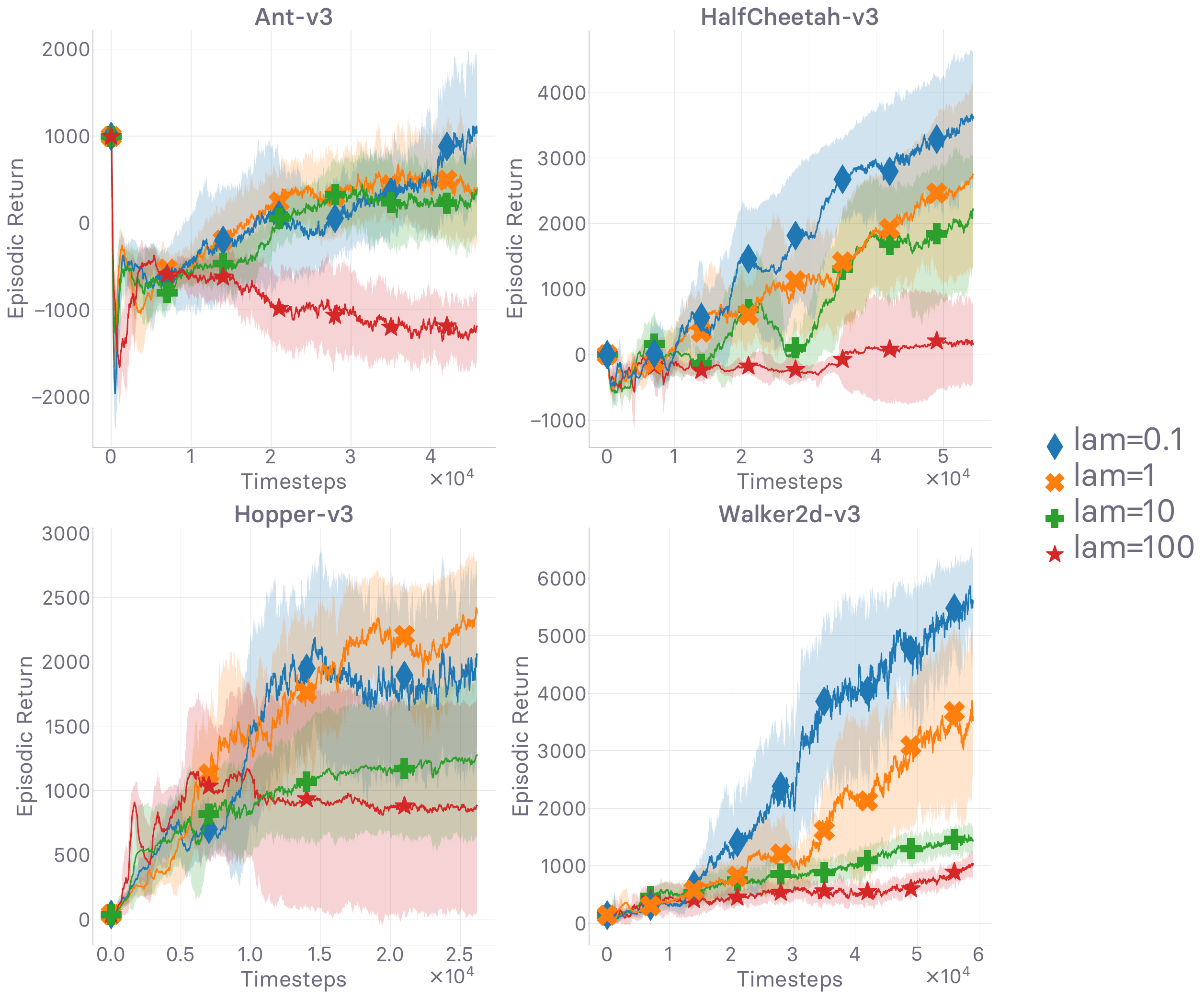}}
    \caption{Evolution of return values \textit{(higher is better)}}
  \end{subfigure}
  \begin{subfigure}[t]{0.49\textwidth}
    \center\scalebox{0.16}[0.16]{\includegraphics{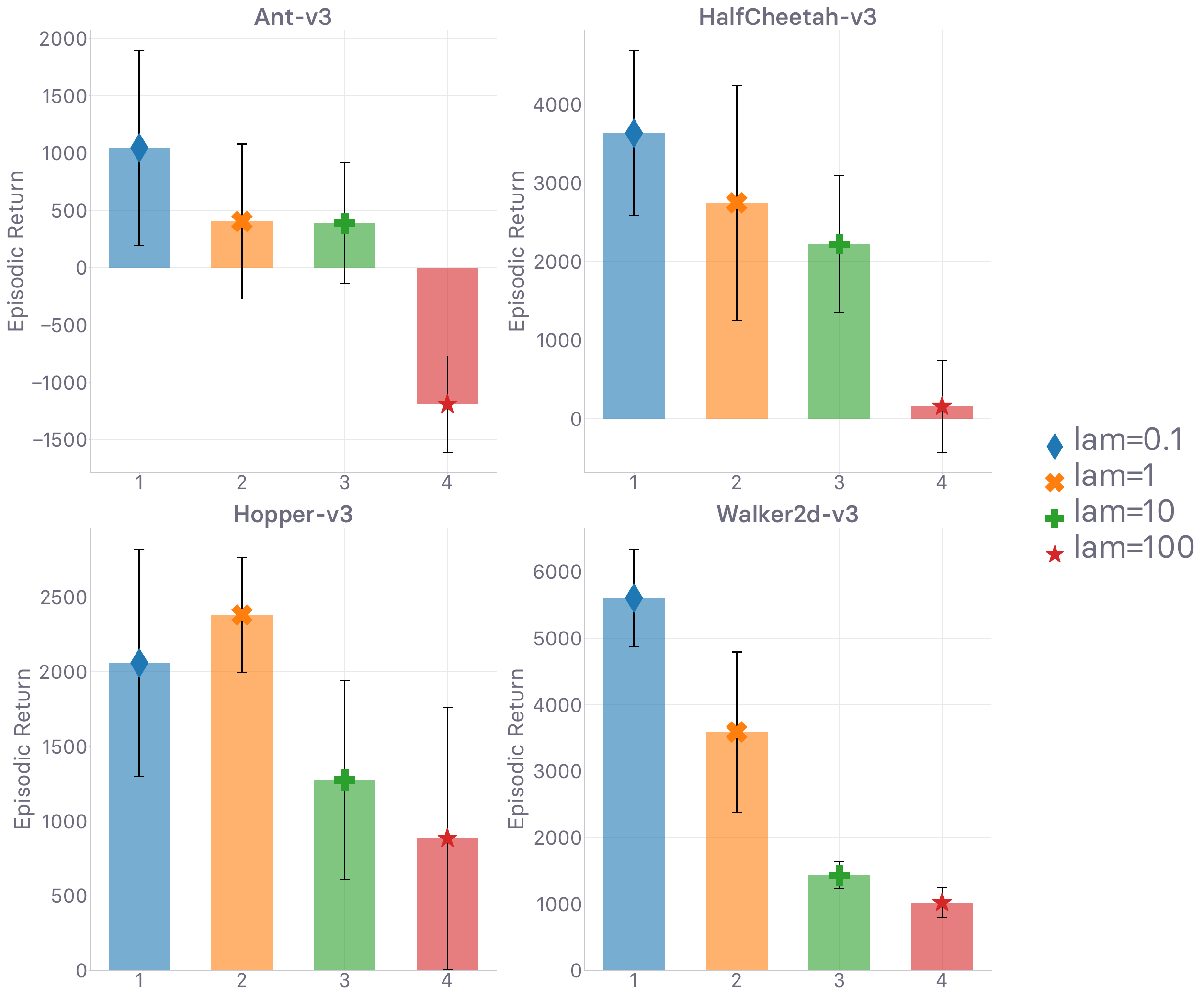}}
    \caption{Final return values at timeout \textit{(higher is better)}}
  \end{subfigure}
  \caption{
  Grid search over the hyper-parameter $\lambda$ when $k=0$.
  Runtime is 12 hours.}
\end{figure}

\section{Reward Formulation}
\label{ablationreward}

\begin{figure}[H]
  \center
  \begin{subfigure}[t]{0.49\textwidth}
    \center\scalebox{0.16}[0.16]{\includegraphics{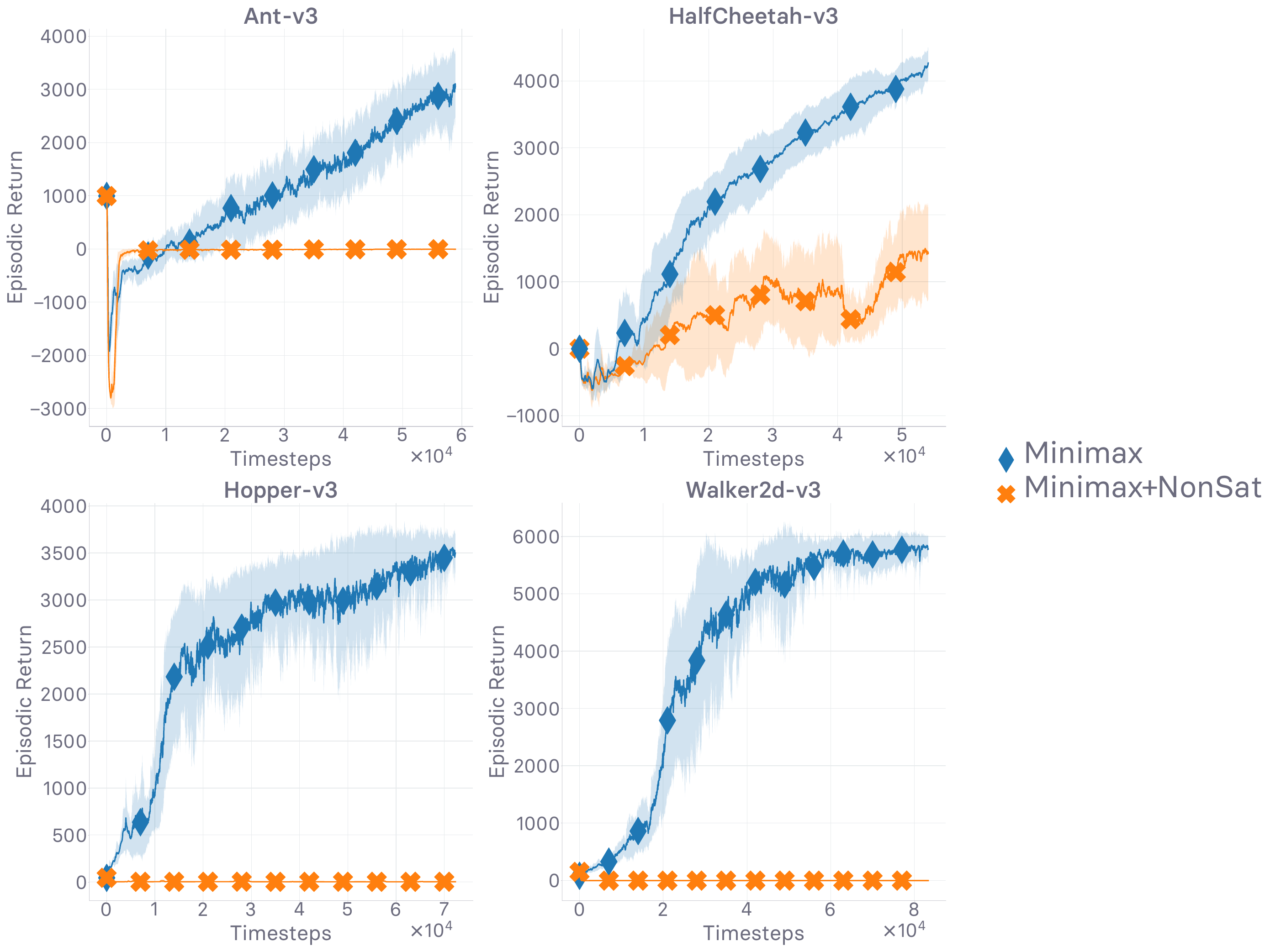}}
    \caption{Evolution of return values \textit{(higher is better)}}
  \end{subfigure}
  \begin{subfigure}[t]{0.49\textwidth}
    \center\scalebox{0.16}[0.16]{\includegraphics{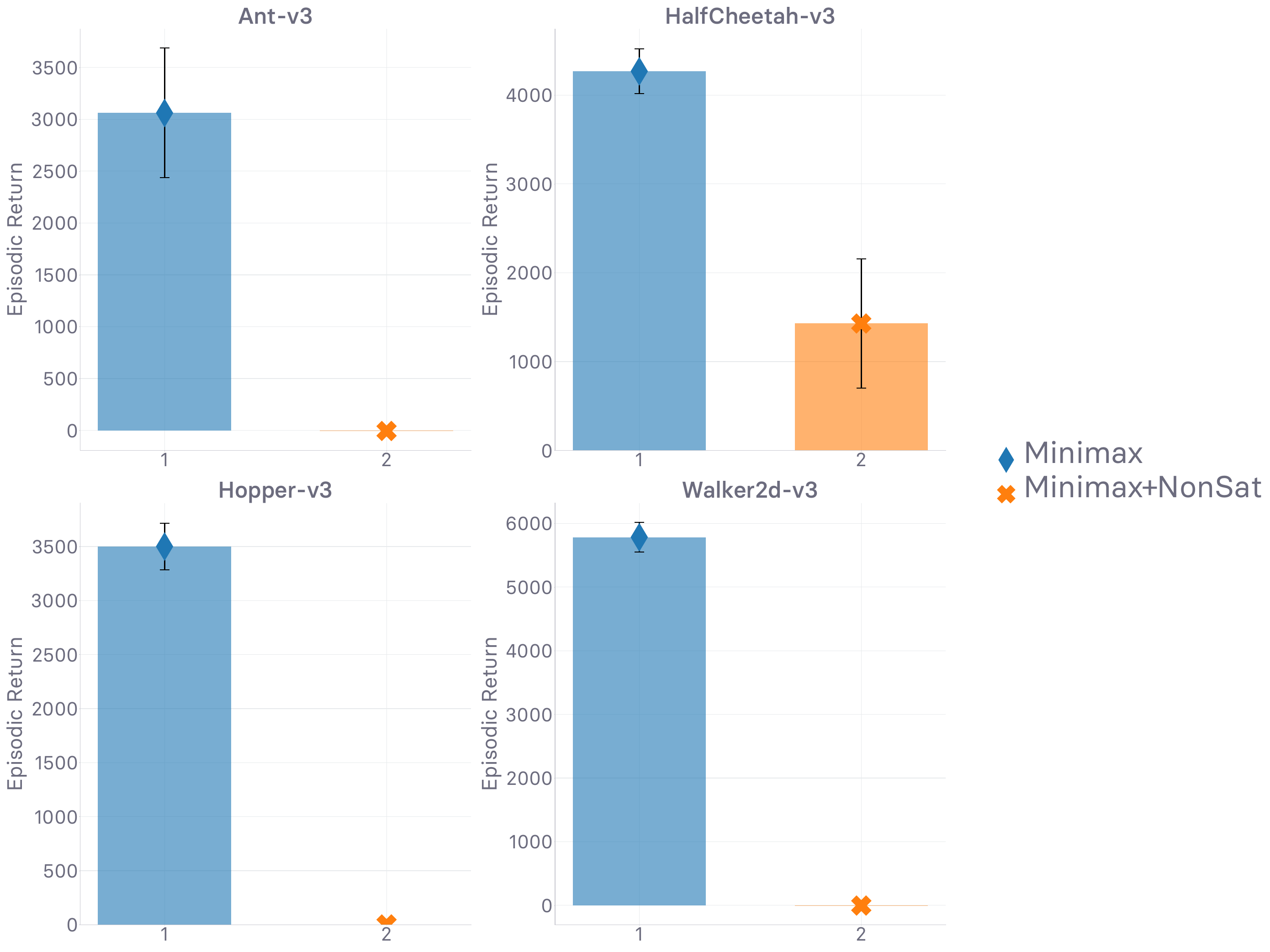}}
    \caption{Final return values at timeout \textit{(higher is better)}}
  \end{subfigure}
  \caption{
  Comparison of two ways to define the surrogate imitation reward $r_\varphi$
  from the discriminator $D_\varphi$.
  \textit{``Minimax''} corresponds to
  $r_\varphi^\textsc{mm} \coloneqq -\log(1-D_\varphi)$,
  while \textit{``Minimax + Non-Saturating''} denotes the use of
  $r_\varphi^\textsc{ns} \coloneqq -\log(1-D_\varphi) + \log(D_\varphi)$,
  as described in \textsc{Section}~\ref{bridge}.
  Runtime is 12 hours.}
\end{figure}

\section{Discount Factor}
\label{ablationdiscount}

\begin{figure}[H]
  \center
  \begin{subfigure}[t]{0.49\textwidth}
    \center\scalebox{0.16}[0.16]{\includegraphics{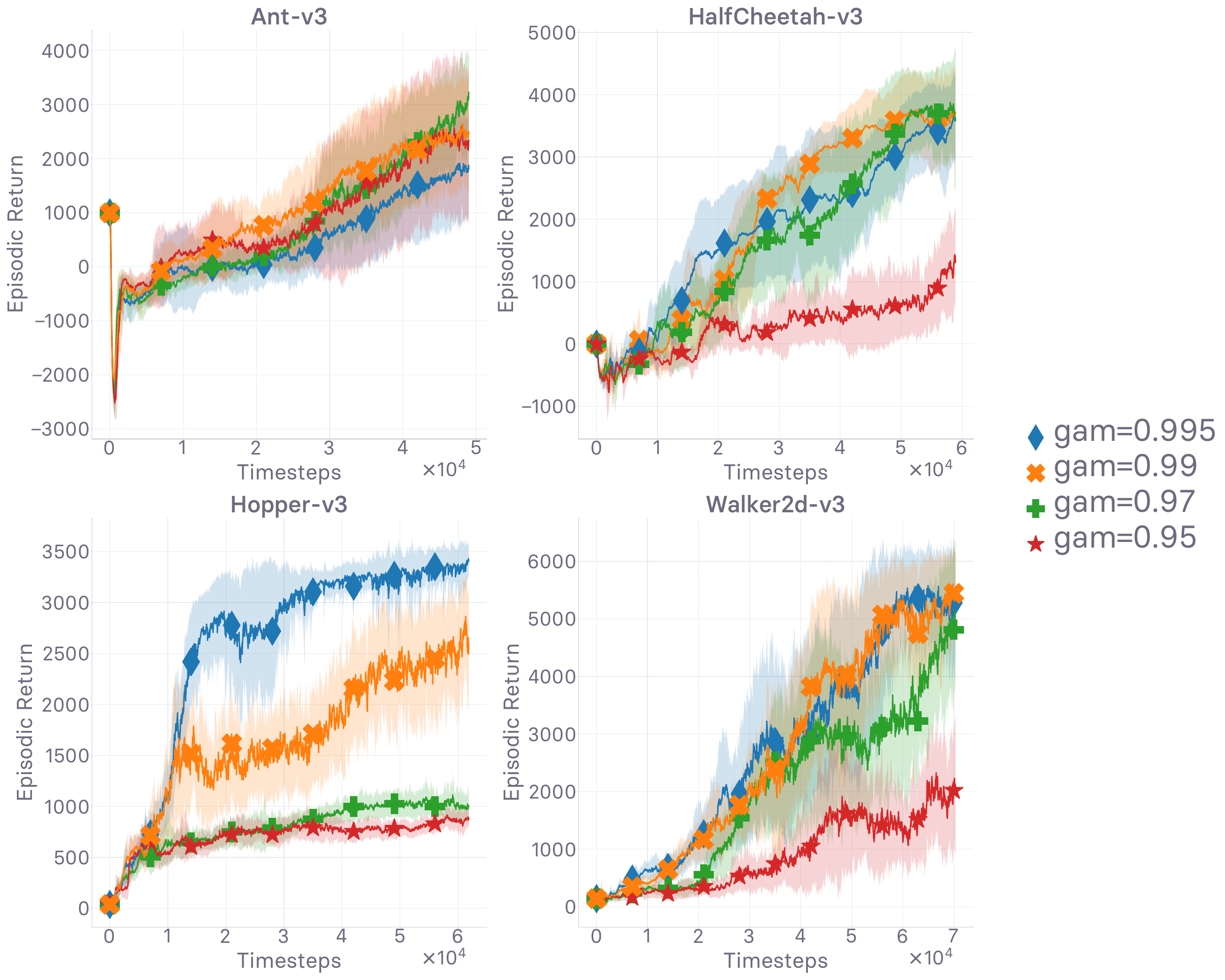}}
    \caption{Evolution of return values \textit{(higher is better)}}
  \end{subfigure}
  \begin{subfigure}[t]{0.49\textwidth}
    \center\scalebox{0.16}[0.16]{\includegraphics{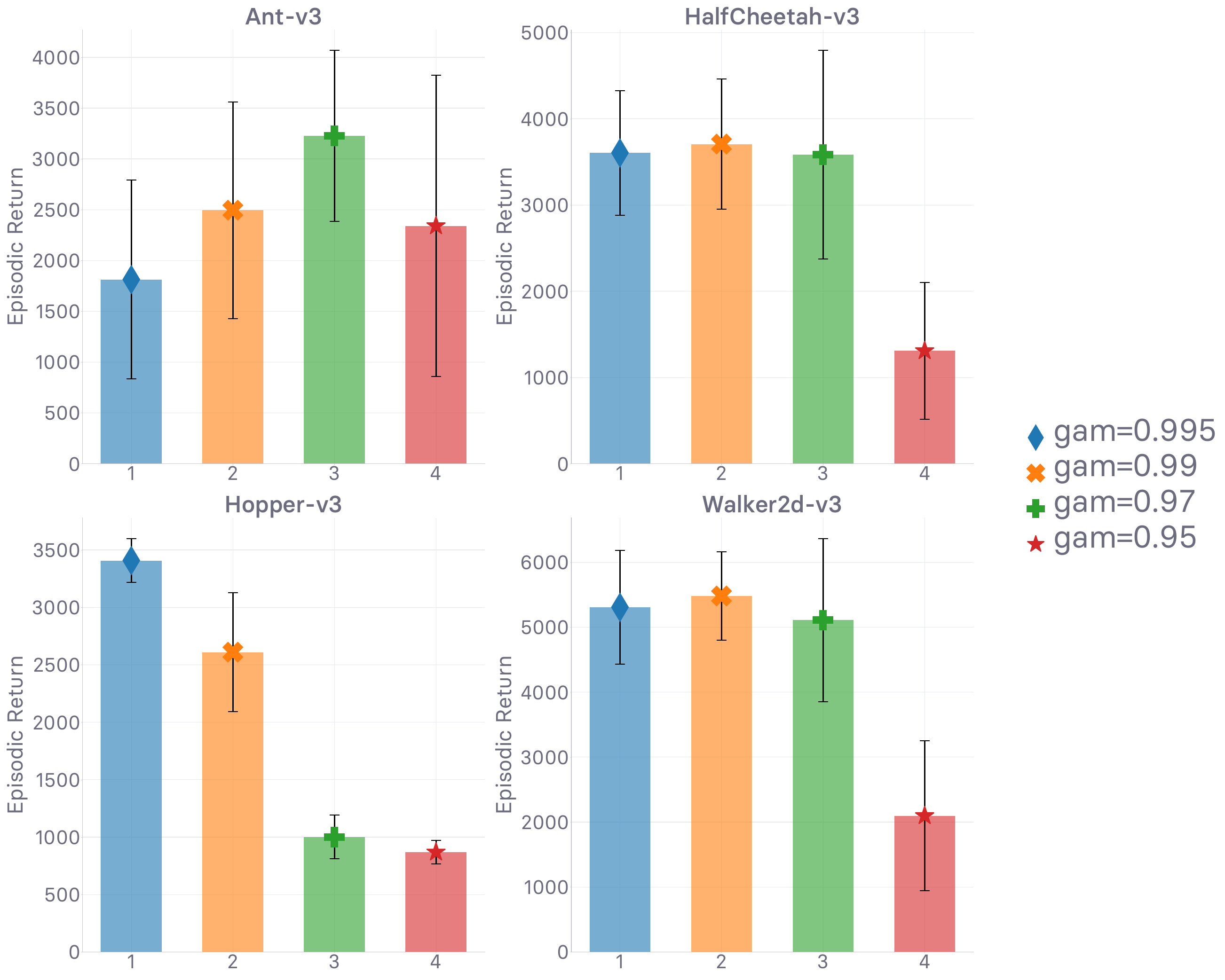}}
    \caption{Final return values at timeout \textit{(higher is better)}}
  \end{subfigure}
  \caption{
  Grid search over the discount factor $\gamma$.
  Runtime is 12 hours.}
\end{figure}

\section{Return Normalization}
\label{ablationretnorm}

\begin{figure}[H]
  \center
  \begin{subfigure}[t]{0.49\textwidth}
    \center\scalebox{0.16}[0.16]{\includegraphics{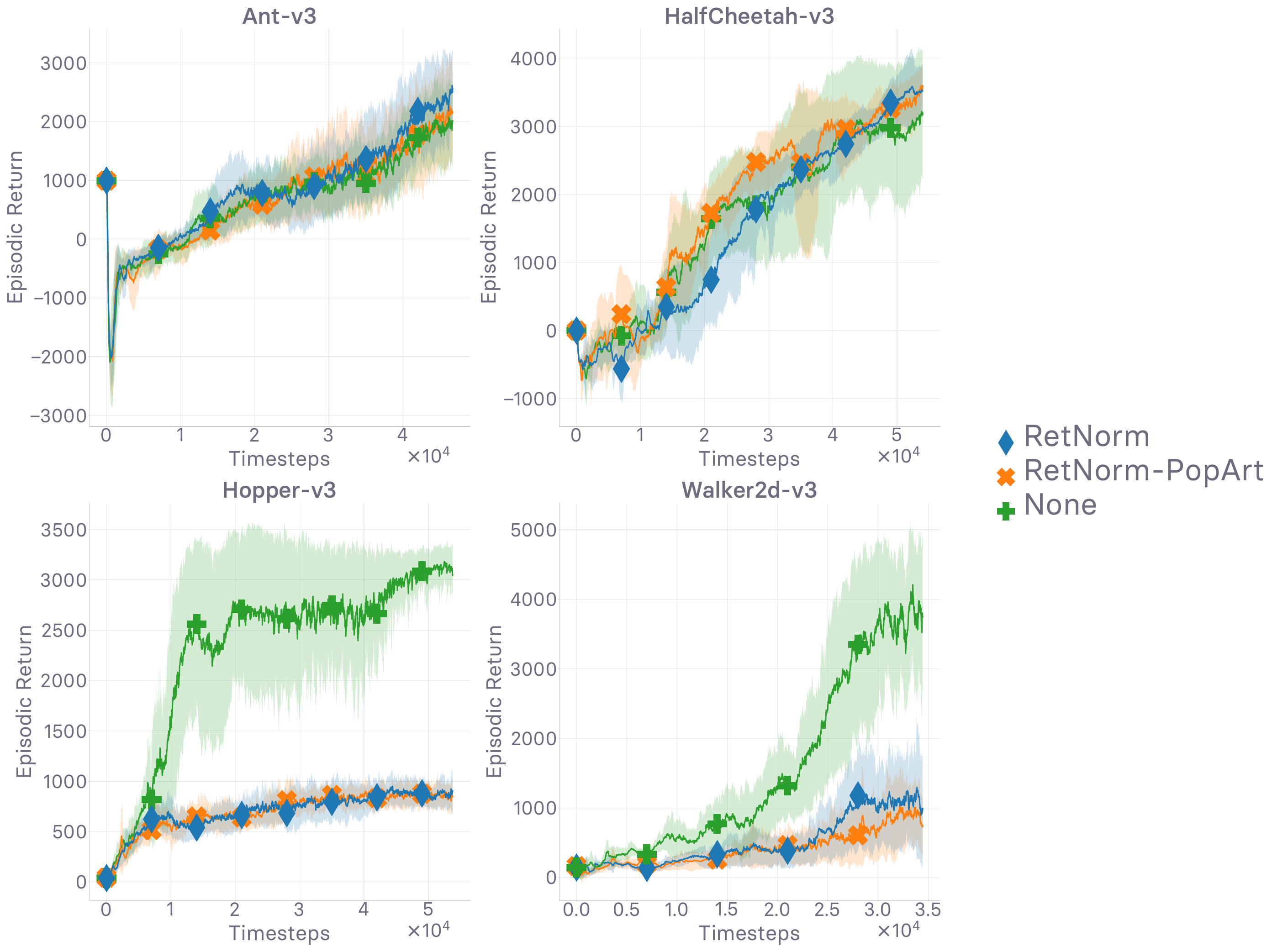}}
    \caption{Evolution of return values \textit{(higher is better)}}
  \end{subfigure}
  \begin{subfigure}[t]{0.49\textwidth}
    \center\scalebox{0.16}[0.16]{\includegraphics{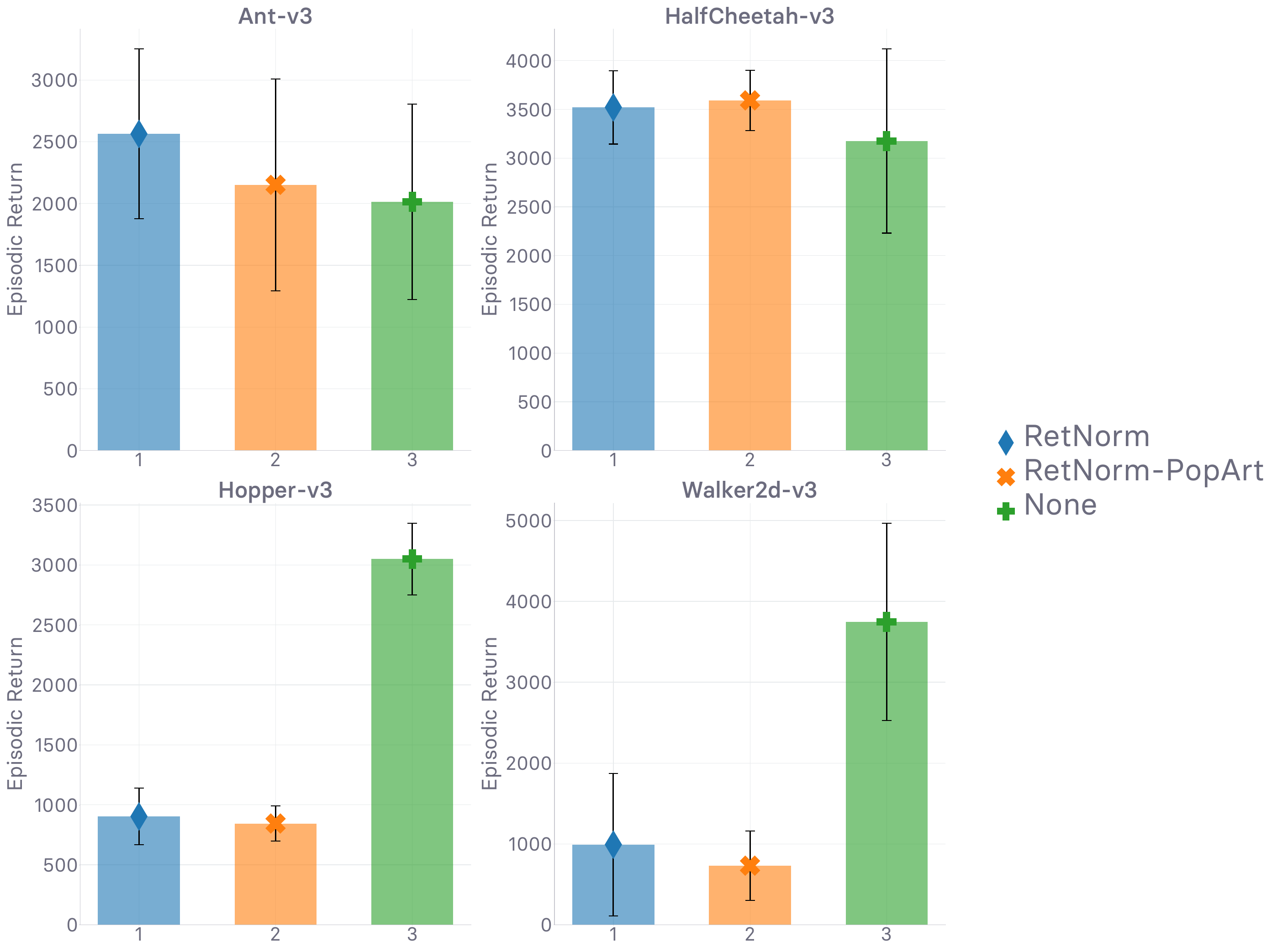}}
    \caption{Final return values at timeout \textit{(higher is better)}}
  \end{subfigure}
  \caption{
  Ablation study on return normalization and \textsc{Pop-Art} \cite{Van_Hasselt2016-bh}.
  Runtime is 12 hours.}
\end{figure}

\section{Exploration}
\label{ablationexplo}

\begin{figure}[H]
  \center
  \begin{subfigure}[t]{0.49\textwidth}
    \center\scalebox{0.16}[0.16]{\includegraphics{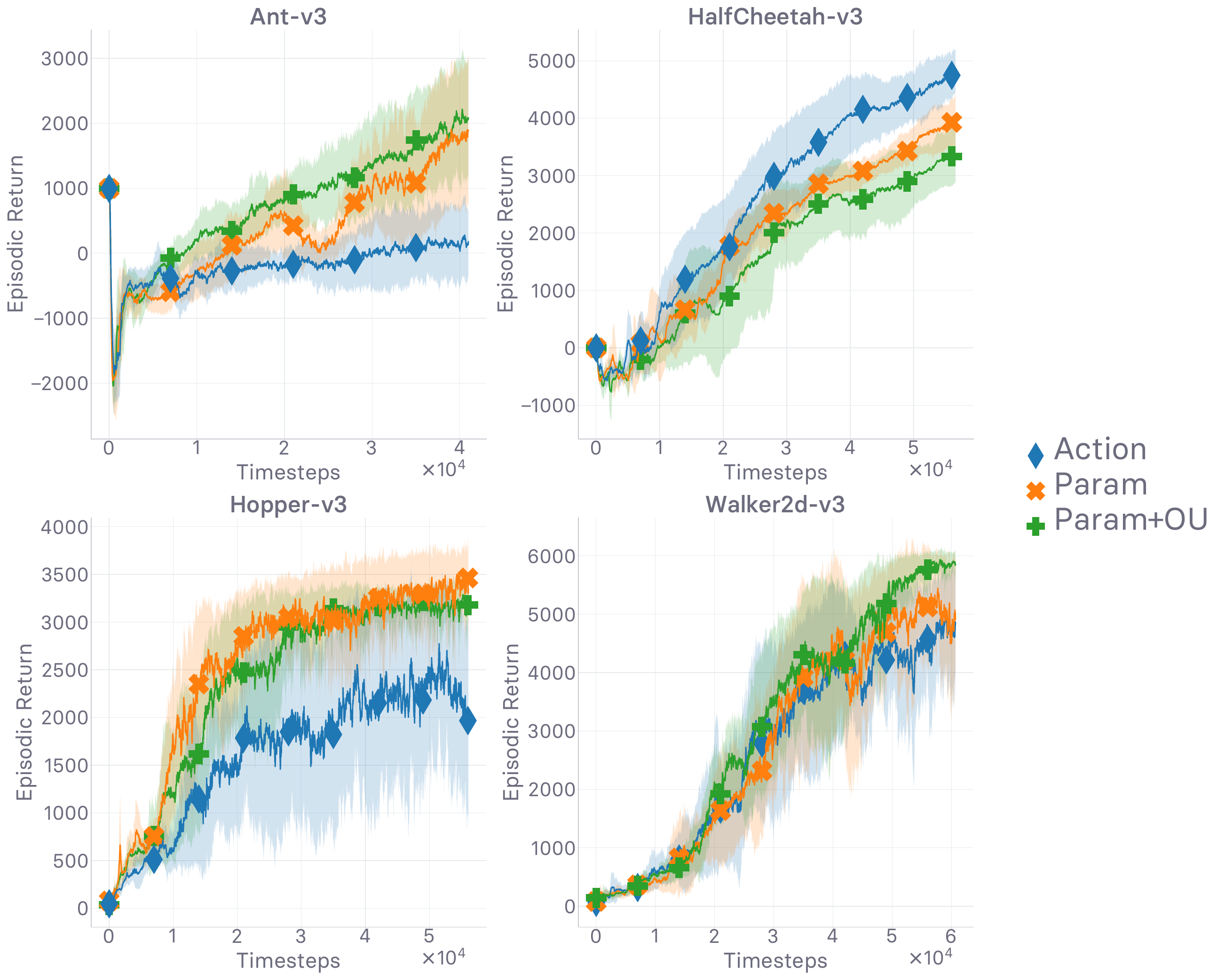}}
    \caption{Evolution of return values \textit{(higher is better)}}
  \end{subfigure}
  \begin{subfigure}[t]{0.49\textwidth}
    \center\scalebox{0.16}[0.16]{\includegraphics{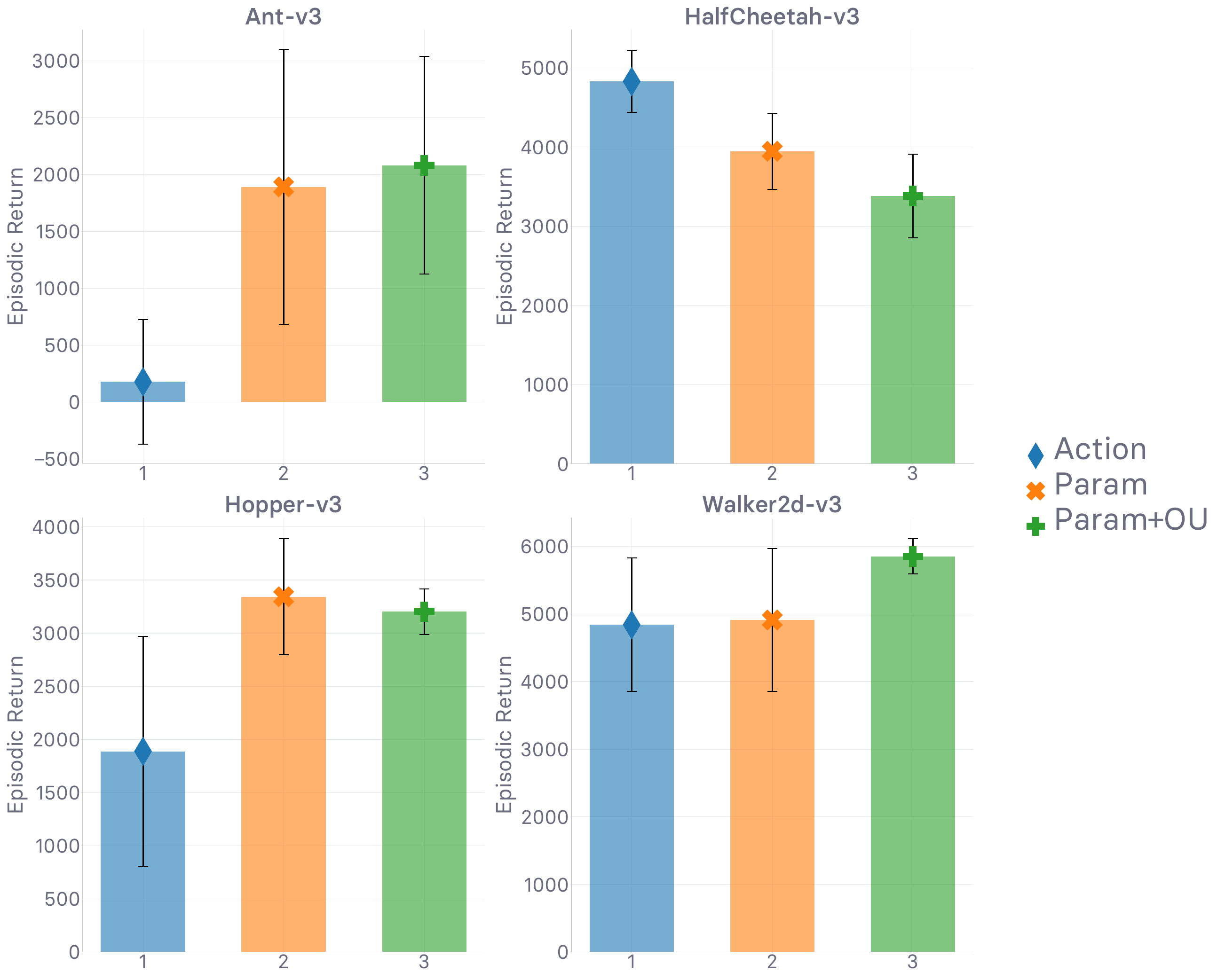}}
    \caption{Final return values at timeout \textit{(higher is better)}}
  \end{subfigure}
  \caption{
  Evaluation of the considered method under several exploration strategies.
  \textit{``Action''} corresponds to defining $\pi_\theta$
  by directly applying additive Gaussian noise to the \emph{action} returned by $\mu_\theta$.
  As such,
  $\pi_\theta(\cdot, s_t) = \mu_\theta(s_t) + \epsilon$,
  where $\epsilon \sim \mathcal{N}(0,\sigma)$,
  with $\sigma=0.2$.
  \textit{``Param''} denotes the application of additive noise in the network \emph{parameters}
  directly, and
  \textit{``Param + OU''} corresponds to the additional application of temporally correlated
  noise, generated sequentially by a Ornstein-Uhlenbeck process, on the action
  (\textit{cf.} \textsc{Section}~\ref{bridge}
  for a description of these two last approaches,
  and \textsc{Table}~\ref{hptable} for the associated hyper-parameters).
  Despite the absence of a clear winner,
  we use the combination of parameter noise and temporally correlated action noise
  in every experiment reported in this work, as it seems to yield the best results.
  Runtime is 12 hours.}
\end{figure}

\end{document}